\documentclass[10pt,twocolumn,letterpaper]{article}
\usepackage[preprint]{iccv}      %

\definecolor{iccvblue}{rgb}{0.21,0.49,0.74}
\usepackage[pagebackref,breaklinks,colorlinks,allcolors=iccvblue]{hyperref}

\usepackage{preamble}
\usepackage{macros}
\usepackage{booktabs}
\usepackage{enumitem}

\usepackage{nicematrix}
\usepackage{color, colortbl}
\definecolor{TableRowHighlight}{rgb}{0.92, 0.92, 0.92}

\usepackage{makecell}

\def\paperID{8277} %
\def\confName{ICCV}
\def\confYear{2025}

\newcommand*\samethanks[1][\value{footnote}]{\footnotemark[#1]}

\title{Graph-Based Captioning: \\
Enhancing Visual Descriptions by Interconnecting Region Captions}
\author{
\begin{tabular}{ccc}
\normalfont{Yu-Guan Hsieh}$^1$\thanks{Work done at Apple.}
& \normalfont{Cheng-Yu Hsieh}$^2$\samethanks[1]
& \normalfont{Shih-Ying Yeh}$^3$\samethanks[1]\\
\small{\texttt{yuguan@spellbrush.com}}
&
\small{\texttt{cydhsieh@cs.washington.edu}}
&
\small{\texttt{kblueleaf@gapp.nthu.edu.tw}}
\end{tabular}
\\[1.2em]
\normalfont{Louis Béthune}$^4$
\hspace{2em}
\normalfont{Hadi Pouransari}$^4$
\hspace{2em}
\normalfont{Pavan Kumar Anasosalu Vasu}$^4$
\\
\small{\texttt{\{l\_bethune, mpouransari, panasosaluvasu\}@apple.com}}
\\[0.6em]
\normalfont{Chun-Liang Li}$^4$
\hspace{2em}
\normalfont{Ranjay Krishna}$^{2}$\samethanks[1]
\hspace{2em}
\normalfont{Oncel Tuzel}$^4$
\hspace{2em}
\normalfont{Marco Cuturi}$^4$
\\
\small{\texttt{ranjay@cs.washington.edu}},~~ \small{\texttt{\{chunliang\_li, otuzel, cuturi\}@apple.com}}
\\[.8em]
$^1$ Spellbrush \hspace{1em}
$^2$ University of Washington \hspace{1em}
$^3$ National Tsing Hua University \hspace{1em}
$^4$ Apple \\[1.2em]
Code: \url{https://github.com/apple/ml-gbc}\\[0.2em]
Dataset: \url{https://huggingface.co/graph-based-captions} \
}

\begin{document}

\maketitle

\begin{abstract}
Humans describe complex scenes with compositionality, using simple text descriptions enriched with links and relationships. While vision-language research has aimed to develop models with compositional understanding capabilities, this is not reflected yet in existing datasets which, for the most part, still use plain text to describe images. In this work, we propose a new annotation strategy, graph-based captioning (GBC) that describes an image using a labeled graph structure, with nodes of various types.
The nodes in GBC are created through a two-stage process: first, identifying and describing entity nodes; second, linking these nodes by highlighting \textit{compositions} and \textit{relations} among them.
Since \textit{all} GBC nodes hold plain text descriptions, GBC retains the flexibility found in natural language, but can also encode hierarchical information in its edges.
We demonstrate that GBC can be produced automatically, using off-the-shelf multimodal LLMs and object detection models, by building a new dataset GBC10M that gathers GBC annotations for about 10M images of the CC12M dataset.
Through CLIP training on GBC10M, we show that leveraging GBC nodes' annotations---particularly those in composition and relation nodes---significantly boosts the model's performance across various benchmarks compared to when other annotations are used.
To further explore the opportunities provided by GBC, we also investigate the use of GBC as middleware for text-to-image generation, and show the extra benefits of incorporating the graph structure in this task.

\end{abstract}

\section{Introduction}
\label{sec:introduction}

The availability of huge paired image/caption datasets has revolutionized our ability to develop advanced multimodal models, enabling a range of tasks such as efficient text-to-image synthesis, text-guided image manipulation, and fine-grained image understanding through multimodal large language models~\citep{radford2021learning,li2021align,alayrac2022flamingo,podell2024sdxl,liu2023visual,brooks2023instructpix2pix}. %
The quality and granularity of these datasets plays, therefore, a crucial role. While quality can be addressed by filtering out data~\cite{schuhmann2022laionb,fang2024data,goyal2024scaling} or with a simple recaptioning strategy~\cite{lai2024veclip,doveh2023dense,nguyen2023improving,fan2023improving},
there is ample interest in the community to provide more detailed, fine-grained information for each image ~\cite{liu2023visual,BetkerImprovingIG,chen2024pixartalpha,esser2024scaling}.
To obtain better annotations, \textbf{we draw inspiration from  compositionality}, a fundamental characteristic of human perception that is reflected in the natural language used to describe our surroundings~\cite{cresswell1973logics,fodor1988connectionism,janssen1997compositionality,hupkes2020compositionality,bottou2014machine,chomsky1965some}. 
Compositionality plays an especially important role when examining larger images found in the wild, which have a rich coarse-to-fine, hierarchical structure, commonly represented by a scene graph~\cite{johnson2015image}.
While scene graphs have been successfully applied to image retrieval~\cite{schuster2015generating,johnson2015image}, generation~\cite{mishra2024scene,farshad2023scenegenie}, and pre-training~\cite{herzig2023incorporating,huang2024structure}, the scale of scene graph dataset is typically small.
For instance, Visual Genome~\cite{krishna2017visual} only contains around $100$k images.

\begin{figure*}[t]
    \centering
    \includegraphics[width=0.95\textwidth]{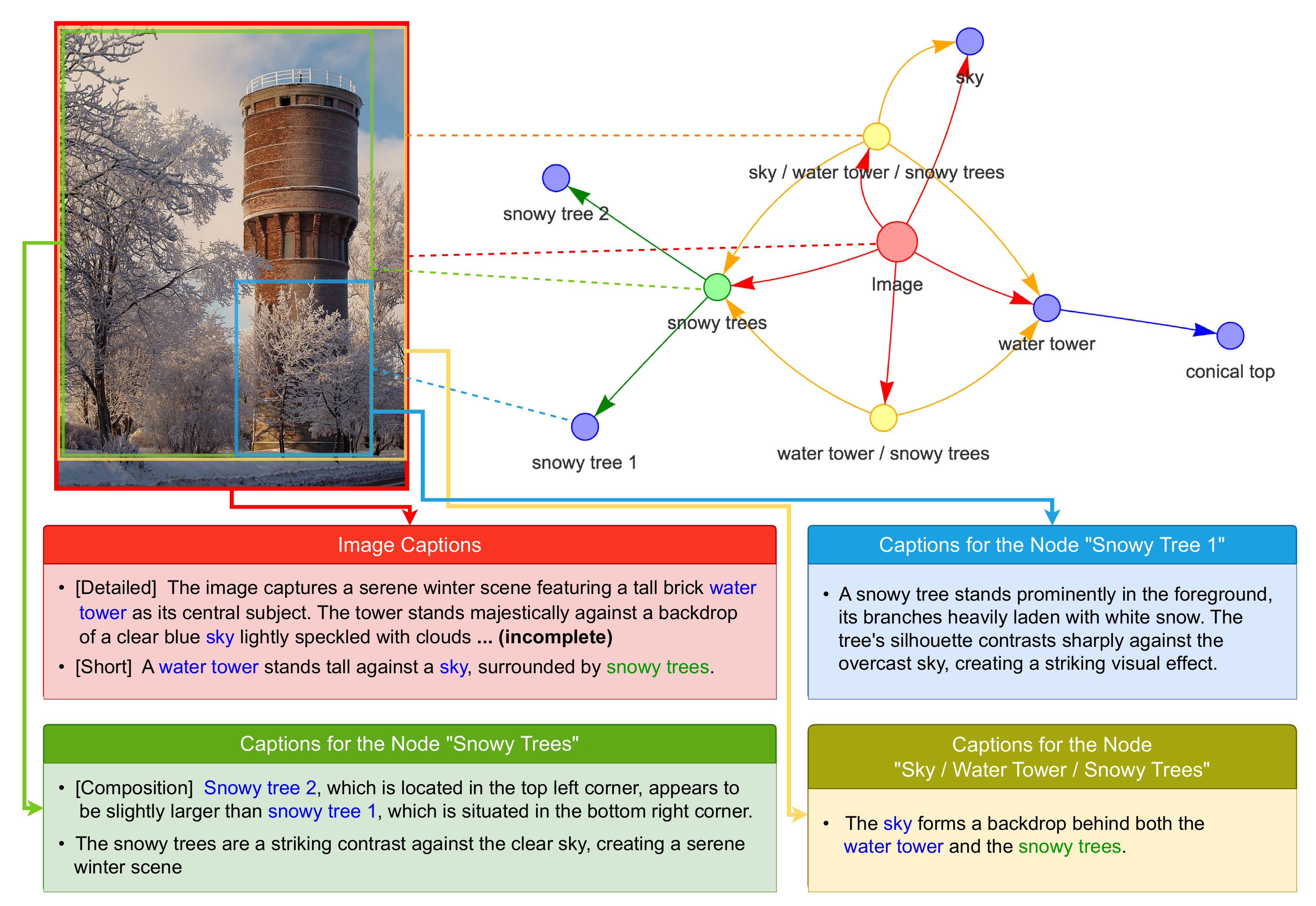}
    \caption{An illustration of our proposed graph-based captions. The image node, entity nodes, composition nodes, and relation nodes are respectively colored in red, blue, green, and yellow.
    The color texts in the captions correspond to the labels of the outgoing edges, which are summarized as node labels in the figure.
    More examples are provided in \cref{apx:dataset-examples}.
    }
    \label{fig:GBC-ex-main}
\vspace{-0.5em}
\end{figure*}

\paragraph{Contributions\afterhead}
To overcome the limitations of existing datasets and annotation formats that either struggle to represent the hierarchical nature of scenes or are of small size and lack flexibility in their description, this paper makes a series of contributions as summarized below.

1. \textbf{We propose graph-based captioning (GBC)}, a new vision-language data format that captions images with a graph-based structure akin to scene graphs while retaining the flexibility and intuitiveness of plain text description.
GBC contains four types of nodes: (1) an image node with captions of the entire image, (2) entity nodes that contain descriptions of individual objects, (3) composition nodes that link objects in the images of the same type,
and (4) relation nodes that describe the spatial (``the tree is to the left of the tower'') or semantic (``The branch is covered in snow'') relationships between objects of different types.
Importantly, encoding non-hierarchical relations as nodes rather than edges enables us to capture relationships involving more than two objects
(\S~\ref{subsec:GBCdesc}).

2. \textbf{We design a workflow to produce GBC annotations at scale.}
Our approach combines a multimodal large language model (MLLM) with an open-vocabulary detection model.
Initially, the MLLM generates both short and detailed captions for the entire image, which are used to identify entities.
The detection model is then applied to locate bounding boxes for each identified entity.
This process is recursively applied to create a GBC for each detected proposal.
Finally, the MLLM is prompted to generate composition and relation captions that link multiple entity nodes together (\S~\ref{subsec:dataset-construct}).

3. \textbf{We create large-scale GBC dataset containing $10$ million images with $\approx 534$ words per image using the aforementioned workflow}.
While ours is the first vision-language dataset that contains structured captions, a few recent datasets contain dense annotations,
and only \cite{wang2023all} has a scale that is similar to ours.
Our dataset is released under the CC BY-NC 4.0 license (\S~\ref{subsec:gbc12M}).

4. \textbf{We demonstrate the benefit of GBC via CLIP training experiments.}
Concretely, we show that the diversity of captions found in GBC nodes improves CLIP model performance across image-to-text retrieval, text-to-image retrieval, compositionality, and semantic segmentation tasks, while retaining comparative performance on zero-shot ImageNet classification.
Remarkably, we observe that composition and relation nodes, which can only be obtained through the GBC workflow, boost performance. Moreover, we perform ablation on the influence of annotation format on retrieval performance using a set-aside test set from GBC. 
In this case, we see that an architecture that is tailored to the GBC format provides comparable or even better performance than that obtained when describing an image with detailed captions, suggesting that GBC can be a promising alternative to traditional image captioning formats (\S~\ref{sec:exp}).

5. \textbf{We demonstrate the benefit of GBC as middleware for text-to-image generation.}
By breaking down the text-to-image generation process into the subtasks of text-to-GBC and GBC-to-image, we provide users with a powerful intermediary for fine-grained image manipulation.
The text-to-GBC subtask is handled by a lightweight language model, while the GBC-to-image subtask can be completed through a training-free approach.
Importantly, we show that the inclusion of graph information significantly enhances the performance of a baseline method that otherwise performs poorly when only bounding box information is used (\S~\ref{sec:exp-t2i}).

\section{Related works}
\label{sec:preliminaries}

\begin{figure*}[t]
    \centering
    \includegraphics[width=0.9\textwidth]{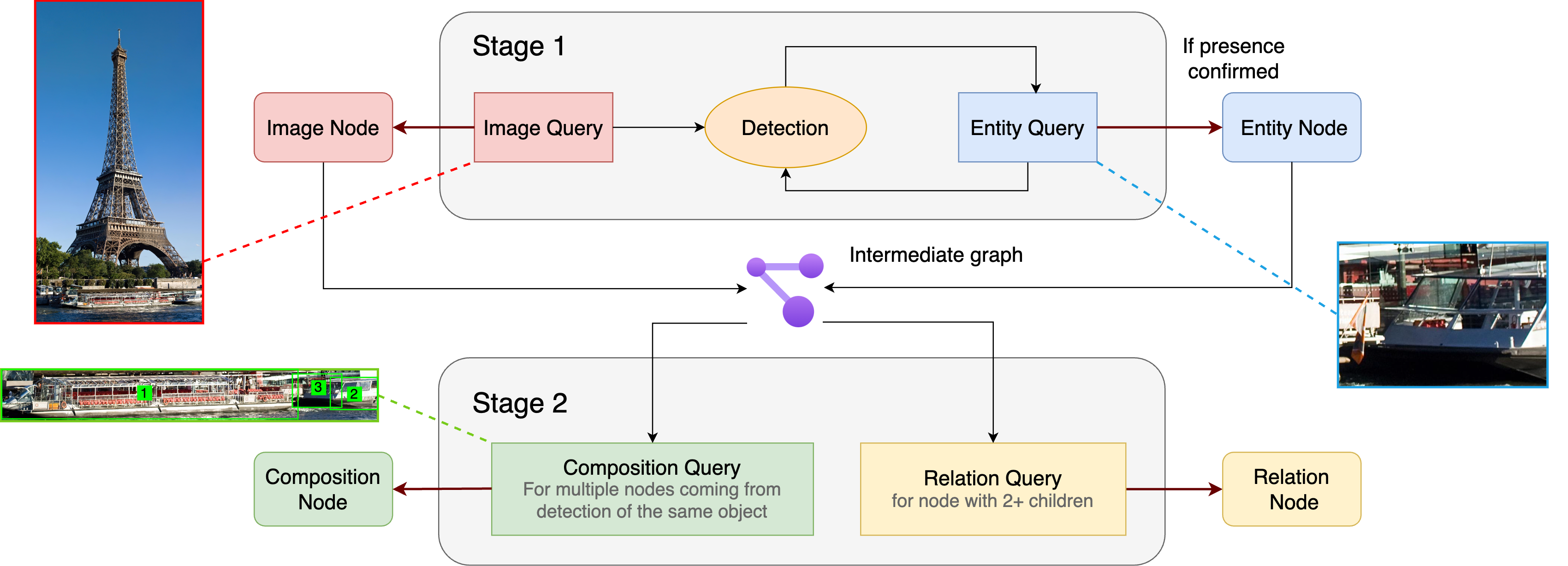}
    \caption{Our image annotation process involves four types of queries that are performed in two separate stages, with the detection model serves to single out the regions that are used for different queries.}
    \label{fig:dataset-query}
    \vspace{-0.5em}
\end{figure*}

In this section, we discuss related works on vision-language datasets. We refer the readers to \cref{apx:related} for works that are specific to CLIP~\citep{radford2021learning} and text-to-image models.

\vspace{0.3em}
\textbf{Vision-language datasets\afterhead}\quad
First vision-language datasets were manually built using human annotations, such as Flickr30k~\citep{young2014image}, COCO~\citep{lin2014microsoft} and Visual Genome~\citep{krishna2017visual}. This yielded annotations of high quality, but unfortunately of short length, and in limited amounts (with no dataset containing more than 130k images).
Several studies have then demonstrated the benefits of using larger scale datasets obtained by crawling the web, such as YFCC100M~\citep{thomee2016yfcc100m}, RedCaps~\citep{desai2021redcaps}, or Wikipedia-based image-text dataset (WIT)~\citep{srinivasan2021wit}. The quality of these data became a concern when it was noticed that in some situations the caption was only loosely related (or not related at all) with the image, which can be detrimental to the overall performance~\citep{santurkar2022caption}. This motivated researchers to use automatic filtering procedures to select higher-quality data samples, like in Localized Narratives~\cite{pont2020connecting} or Conceptual Captions (CC3M)~\cite{sharma2018conceptual}, and its successor CC12M~\citep{changpinyo2021cc12m}. These efforts have reached billion scale with LAION-5B~\citep{schuhmann2022laionb}, and %
LAION-CAT~\citep{radenovic2023filtering}. 
In a similar vein, Meta-CLIP~\citep{xu2023demystifying} reproduces the processing of the seminal CLIP paper~\citep{radford2021learning} on a subset of the Common Crawl dataset, SemDeDup~\cite{abbas2023semdedup} relies on the embeddings provided by foundation models to filter data and remove duplicates, while DFN~\citep{fang2024data} uses filtering networks trained on high quality data to extract subsets of Common Crawl.  

\vspace{0.3em}
\textbf{VL datasets with dense captioning\afterhead}\quad
It was noticed recently that using entirely generated captions from raw images, as in DAC~\citep{doveh2023dense} and AS-1B~\cite{wang2023all}, could improve results over filtering approaches.
These datasets are characterized by their long and detailed captions that describe every element within a scene. 
Complementing these efforts, \citet{urbanek2023picture} introduced DCI, a dataset featuring similarly dense annotations but curated by humans and on a smaller scale.
Alternatively, DOCCI~\citep{onoe2024docci} focuses on a set of only 15k high quality, high resolution, paired image-captions, manually selected and annotated by one of the authors,
with typical caption length of more than $135$ words. In the ImageInWords~\citep{garg2024imageinwords} dataset, captions are iteratively improved by humans, on top of previously human or machine annotated captions, yielding 9K densely captioned images. %

\section{Improving image annotations with GBC}
\label{sec:dataset}

We introduce in this section our new captioning format to represent an image, explain how we can use any off-the-shelf \ac{MLLM} and open-vocabulary detection model to obtain such captions, and briefly describe the two datasets GBC1M, and GBC10M that we construct following the proposed workflow.
Additional details about the data preparation process and the datasets can be found in the \cref{apx:dataset-construct,apx:dataset-info}.

\subsection{Representing an image with GBC}\label{subsec:GBCdesc}

To encode the structured information contained in an image, we propose to represent each image as a \ac{DAG}, denoted as $\graph=(\vertices,\edges)$.
Each node of the graph $\vertex\in\vertices$ is associated with a bounding box.
Starting with the root node, which corresponds to the entire image (image node), other nodes can either hold a set of objects (composition node and relation node), or a single object in the image (entity node).
Moreover, to benefit from the expressive power of natural language descriptions and to ensure smooth integration of our annotations into the existing ecosystems of methods that rely primarily on image-text pairs, we label each node $\vertex$ with a set of captions $\vv[\captions][\vertex] = \{\capn_1, \ldots, \capn_{\vv[\ncap][\vertex]}\}$.

The edges, on the other hand, are used to encode the \emph{hierarchy} between the nodes.
More specifically, there is an edge $\edge\in\edges$ from $\vertexalt$ to $\vertex$ only if the content associated to $\vertex$ is part of the content associated to $\vertexalt$.
This relation is also reflected by the edge label $\ve[\labeling][\edge]$ which should appear in the captions of the source node $\vertexalt$ and be able to represent the object(s) associated to the target node.\footnote{Ideally, we would also like to distinguish between multiple appearances of the same text in a caption. However, this is not explicitly handled by our current dataset construction workflow so we omit it here.}

An examplar GBC, generated automatically through our workflow is provided in \cref{fig:GBC-ex-main}. It should be noted that the only manual addition in that graph comes from the "title label" of each node, which is obtained by taking the union of labels found in its incoming edges. Compared to the standard scene graph annotation, the use of node captions provides flexibility to describe complex concepts, while the underlying graph still captures the inherent structure of the image.
Our dataset, whose construction is detailed in \cref{subsec:dataset-construct} next, includes several different types of captions tailored to the structure of the DAG. At the root image node, we provide both detailed and short captions to cater to varying levels of granularity.
Captions at composition nodes and relation nodes explicitly describe the arrangement and interaction of multiple objects, while the captions at the entity nodes provide detailed description of a single object.

\subsection{GBC dataset construction workflow}
\label{subsec:dataset-construct}

We show how to produce GBC annotations automatically, using any pre-trained \ac{MLLM} and open-vocabulary detection model. This
results in a workflow that is comparable, in compute time and complexity, to that of other widespread recaptioning approaches.
At a high level, we use a \ac{MLLM} model to provide captions and identify potential entity nodes, followed by a detection model to provide bounding box coordinates for these entities.

\vspace{0.25em}
\textbf{Data annotation\afterhead}\quad
Our overall process to annotate a single image is shown in \cref{fig:dataset-query}.
To account for the different types of nodes, we design four query templates as listed below:

\begin{itemize}
    \item \textbf{Image query:} We ask the model to provide detailed caption for the image, identify prominent elements, and summarize the long caption with a concise one that contains all these elements.
    The identified elements are then passed to the detection model to obtain the bounding boxes.
    \item \textbf{Entity query:} For each bounding box, we crop out the region and ask the model whether a specific object appears in the cropped image. Moreover, we also ask the model to describe the object and identify prominent elements of the object when it is present. The identified elements are again passed to detection models for detection.
    \item \textbf{Composition query:} In the case where multiple bounding boxes are returned for a single type of object, we ask the model to describe the composition of these objects with an annotated image.
    \item \textbf{Relation query:} For image or entity nodes with more than two children, we ask the model to describe the relations between its children.
\end{itemize}

Provided that there is no guarantee that all the detected objects would end up as a node in the graph---consider the case where the \ac{MLLM} says that the object is not present or just fails to reply in the correct format---we split the entire process into two stages, and we only perform composition queries and relation queries \textit{after} discovering all the entity nodes.
Finally, to improve efficiency and to reduce redundant information, we train two dedicated classifier on top of Jina Embeddings~\cite{gunther2023jina} to decide whether a piece of text is suitable for object detection and whether two texts can represent the same object in an image. The former is applied to every identified element while the later results in \emph{merging} of nodes when a new query targets a region that has already been queried with similar texts.

\subsection{GBC1M and GBC10M}
\label{subsec:gbc12M}

\begin{table}[t]
    \centering
    \begin{tabular}{@{\hskip 0.6em}l@{\hskip 1.2em}|@{\hskip 0.6em}cc@{\hskip 0.6em}}
    \toprule
         &  GBC1M & GBC10M\\
    \midrule
        \# Images
        & 1,013,592 & 10,138,757 \\
        \# Vertices / Image
        & 12.12 & 12.24
        \\
        \# Edges / Image
        & 22.28 & 21.81
        \\
        \# Captions / Image
        
        & 17.40 & 17.67
        \\
        \# Words / Image
        & 593.14 & 533.98
        \\
        Average Graph Diameter
        & 4.55 & 4.41
        \\
    \bottomrule
    \end{tabular}
    \vspace{0.75em}
    \caption{Key statistics of the GBC1M and GBC10M datasets. We report number of images, average number of vertices, edges, captions, and words per image, and average graph diameter.}
    \label{tab:gbc-dataset-stat}
\vspace{-1em}
\end{table}

Following the process outlined in \cref{subsec:dataset-construct}, we annotate the CC12M dataset~\cite{changpinyo2021cc12m} with graph-based captions using LLaVA-1.6~\cite{liu2023visual,liu2024llavanext} as the MLLM and Yolo-World~\cite{Cheng2024YOLOWorld} as the open-vocabulary detection model.
Specifically, we construct two sets of annotations: GBC1M for a subset of around 1M of images, with all the queries performed with the Yi-34B version of LLaVA-1.6, and GBC10M for a subset of around 10M of images, with LLaVA-1.6 Yi-34B for image and composition queries, and LLaVA-1.6 Mistral-7B for entity and relation queries.\footnote{Our larger dataset does not cover the entire CC12M both because some images were no longer accessible at the time we accessed the images, and because we discard images for which the MLLM model's reply to the image query does not comply with the prescribed format.}

We provide statistics of the above two datasets in \cref{tab:gbc-dataset-stat}.
We note that these two datasets have very similar per-image statistics, with the number of words being the only exception, as LLaVA-1.6 Yi-34B tends to provide longer descriptions than LLaVA-1.6 Mistral-7B.
Moreover, our datasets use an average number of around 500 words to describe each image.
This is comparable to other dataset with rich annotations such as DCI (1111 words/img)~\cite{urbanek2023picture} and DOCCI (136 words/img)~\cite{onoe2024docci}.

\section{CLIP training with GBC}
\label{sec:exp}
We present in this section a comprehensive set of experiments to compare different image annotation schemes from a CLIP training perspective.
We first show that compared to existing annotation schemes, GBC annotations can bring improvements on a range of benchmarks across classification, retrieval, and dense prediction task.
Then, we demonstrate how GBC allows one to encode denser, more descriptive textual information to better represent images on retrieval tasks.
Additional experimental details and further ablations are respectively provided in \cref{apx:exp-setup,apx:exp-add}.

\vspace{-0.3em}
\subsection{Annotation formats}
\label{subsec:annotation}
We outline below the different types of image annotations that are considered in our experiments, each providing different opportunities to leverage information from the image.

\textbf{Short caption\afterhead}\quad
Each image is paired with a short caption, as in common image-text datasets.

\textbf{Long caption\afterhead}\quad
One can improve image description using a longer caption.
The long captions that we use in our experiments are of 110 words on average, as compared to short captions, of only 28 words on average. We extend the context length of text encoders in CLIP models from 77 to 512 for this setup.

\textbf{Region captions\afterhead}\quad
Alternatively, more captions can be provided for an image, especially those that describe a specific region of the image.
While this format includes all region captions, it does not include the relational information between region descriptions found in GBC.

\textbf{Graph-based captions\afterhead}\quad
Finally, we consider the GBC format as proposed in \cref{subsec:GBCdesc}.
The GBC format includes region captions, but also provides additional information, stored in relation and composition nodes.
With this in mind, we explore three different ways to leverage GBC annotations:
    \begin{itemize}[leftmargin=*,itemsep=.01cm,topsep=0cm,parsep=2pt]
        \item A direct way to leverage GBC is to treat captions for all nodes in the graph as positive texts for the image, i.e. as what we do for region captions. 
        We refer to such method as \textbf{GBC-captions}.
        \item Another strategy is to traverse from the root image node through the graph and concatenate the captions at the visited vertices into a single long caption.
        We then train a CLIP model with 512 context length in the standard fashion. We refer to this method as \textbf{GBC-concat}.

        \item To fully benefit from the graph information, when available, we introduce additional cross-attention layers to leverage the graph topology (see \cref{apx:SAHA} for details).
        This allows us to encode the entire graph into a text embedding that also contains the information about the graph structure.
        We refer to this method as \textbf{GBC-graph}.
    \end{itemize}

As GBC annotation encapsulates all existing \textit{short}, \textit{long}, and \textit{region} caption formats, we are able to instantiate all the setups by using only a subset of annotation available in our curated GBC10M dataset.
Specifically, taking only the short or detailed caption at the root image node creates the \textit{short} and \textit{long} caption setup, respectively.
To mimic the \textit{region} caption setup, we drop the relation and composition captions from GBC annotations.
Additionally, to isolate the impact of using long captions and the additional captions uncovered by GBC, we omit long captions from both the region caption setup and the three GBC setups.
By adapting GBC into these configurations, we ensure consistent text annotation quality across different methods.

\begin{table}[t]
    \centering
    \resizebox{\columnwidth}{!}
    {
    \begin{NiceTabular}{l|cc|cc|c|c|c}[colortbl-like]
    \toprule
        \multirow{2}{*}[-0.25em]{Annotation}
        &
        \multicolumn{2}{c|}{Flickr-1k}
        &
        \multicolumn{2}{c|}{MSCOCO-5k}
        &
        \multirow{2}{*}[-0.25em]{ImageNet}
        &
        \multirow{2}{*}[-0.25em]{SugarCrepe}
        &
        \multirow{2}{*}[-0.25em]{ADE20K}
        \\
        \cmidrule(lr){2-3}
        \cmidrule(lr){4-5}
        & T2I
        & I2T
        & T2I
        & I2T
        & 
        &
        & %
        \\
    \midrule
        CC12M & 46.4 & 64.6 & 25.0 & 39.4 & 39.2 & 72.9  & 41.7 \\
    \midrule
        Short & 56.3 & 73.2 & 30.7 & 46.7 & 38.8 & 76.0 & 42.0\\
        Long & 56.4 & 75.2 & 31.8 & \underline{50.1} & \underline{39.6} & \textbf{77.0} & 42.8 \\
        Region & \underline{58.3} & 76.6 & 31.5 & 49.1 & 38.5 & 75.6 & 43.5 \\[0.05em]
        \vphantom{\rule{0pt}{1em}}%
        \rowcolor{TableRowHighlight}GBC-captions & \textbf{60.6} & \textbf{79.3} & \textbf{34.1} & \textbf{51.9} &  \textbf{40.8} & \underline{76.7} & \textbf{45.0} \\
        \rowcolor{TableRowHighlight} GBC-concat & 56.1 & 76.0 & 31.4 & 48.5 & 39.0 & 75.7 & 42.1 \\
        \rowcolor{TableRowHighlight} GBC-graph  & 58.0 & \underline{76.9}  & \underline{31.9} & 49.2 & 38.4 & 74.4 & \underline{43.8} \\
    \bottomrule
    \end{NiceTabular}
    }
    \vspace{0.5em}
    \caption{Comparative performance on various existing benchmarks when trained using different annotation schemes.
    For retrieval tasks we report Recall@1, and for ADE20K we report the mIOU.
    As a baseline, we also report performance of a model trained on the same set of images using original CC12M captions.
    The highest scores for each task are highlighted in bold, and the second-highest scores are underlined.}
    \label{tab:standard-eval}
\vspace{-1.5em}
\end{table}

\subsection{Experimental setup}
\label{subsec:exp-setup}

We perform CLIP training on our GBC10M dataset, while leaving out 10,151 samples as the test set.
Following common practice, we use the CLIP score computed by a pre-trained CLIP model~\cite{fang2024data} to filter our training set, discarding the 5\% of captions with the lowest scores for each type.
In addition, we retain the original CC12M captions associated with each image.
Specifically, in all setups, both the original caption and the short synthetic caption are used as positive texts for the image during training.
This prevents the severe distribution shifts that could occur from using only long or region captions when evaluating on standard benchmarks.

\textbf{Objective\afterhead}\quad
To pair an image with multiple captions in training CLIP models, we adopt a multiple-positive contrastive loss in the spirit of LaCLIP~\cite{fan2023improving} and DAC~\cite{doveh2023dense}.
Briefly speaking, compared to standard CLIP objective, the multiple-positive loss sums over the loss on each positive captions of an image while all the captions from the images in the same batch are used in the normalization term.

\textbf{Model and hyperparmeters.}\quad
We use the standard CLIP ViT-B/16 model, with the only difference of longer context length of text encoder for long caption and GBC-concat, and a replacement of the vanilla transformer block by our dedicated attention block in text encoder for GBC-graph.
We fix the global batch size (\ie number of images in each batch) to 4,096 for all the methods. The models are trained for 45,000 steps with AdamW and cosine scheduler at a learning rate of $10^{-3}$. This roughly correspond to 20 epochs of training. We evaluate at the EMA checkpoint at epoch 10, as we observe that further training provides little to no improvement in performance across the benchmarks.

\vspace{-0.3em}
\subsection{Evaluations on existing benchmarks}
\label{subsec:standard-benchmark}

We compare the CLIP models derived from different annotation schemes on an array of evaluation benchmarks, including: \textbf{Flickr-1k}~\cite{plummer2015flickr30k} and \textbf{MSCOCO-5k}~\cite{lin2014microsoft} for zero-shot retrieval, \textbf{ImageNet}~\cite{russakovsky2015imagenet} for zero-shot classification, \textbf{SugarCrepe}~\cite{hsieh2024sugarcrepe} for compositional understanding evaluation, and \textbf{ADE20k}~\cite{zhou2019semantic} for semantic segmentation that measures models' dense prediction performances.
Table~\ref{tab:standard-eval} illustrates our results, from which we draw the following two key insights.

\textbf{GBC annotation leads to clear performance gains by encoding relational information.}\quad
Table~\ref{tab:standard-eval} demonstrates that training models with more detailed textual information, such as long captions or region captions, consistently enhances downstream performance, particularly in retrieval tasks and dense prediction.
However, the most significant improvements are seen with GBC-captions, which augment traditional region captions with relational and compositional descriptions, yielding about a $5\%$ recall increase for retrieval tasks and a $3\%$ boost in mIOU in segmentation task compared to using only short captions.
Given that the GBC workflow is uniquely positioned to provide such relational information, this demonstrates that GBC captures valuable insights not present in conventional captions.

\textbf{How the captions are used matters\afterhead}\quad
Compared to GBC-captions, the improvements achieved by GBC-concat and GBC-graph on these benchmarks are of a smaller margin.
This indicates that the way GBC annotations is used significantly impacts performance.
Specifically, this worse performance is likely due to a mismatch between training and evaluation.
For instance, the graph information that would benefit GBC-graph most is not provided in any of these benchmarks.
We address this discrepancy below.

\vspace{-0.3em}
\subsection{Evaluation on GBC test set}
\label{subsec:gbc-test}

\begin{table}[t]
    \centering
    \resizebox{\columnwidth}{!}
    {
    \begin{NiceTabular}{l|cc|cc|cc|cc|cc}[colortbl-like]
    \toprule
        \multicolumn{1}{c}{
        \multirow{2}{*}[-0.24em]{Annotation}}
        &
        \multicolumn{2}{c}{Short}
        &
        \multicolumn{2}{c}{Long}
        &
        \multicolumn{2}{c}{GBC-captions}
        &
        \multicolumn{2}{c}{GBC-concat}
        &
        \multicolumn{2}{c}{GBC-graph}
        \\[-0.1em]
        \cmidrule(lr){2-3}
        \cmidrule(lr){4-5}
        \cmidrule(lr){6-7}
        \cmidrule(lr){8-9}
        \cmidrule(lr){10-11}
        \multicolumn{1}{c}{}
        & T2I
        & I2T
        & T2I
        & I2T
        & T2I
        & I2T
        & T2I
        & I2T
        & T2I
        & I2T
        \\
    \midrule
        Short & 85.8 & 86.2 & 85.0 & 87.2 &  57.3 & 37.0 & \textbf{87.4} & \textbf{88.2} & - & -\\
        Long & 86.4  & 87.5  & \textbf{95.4} & \textbf{95.7} & 44.3 & 33.1 & 90.5 & 91.4 & - & - \\
        Region & 85.3 & 86.1 & 85.5 & 88.2 & \textbf{91.5} & 79.3 & 89.5 & \textbf{90.0}  & - & -   \\[0.05em]
        \vphantom{\rule{0pt}{0.95em}}%
        \rowcolor{TableRowHighlight} GBC-captions &  86.8 & 87.6  & 87.2 & 89.6 & \textbf{91.3} & 80.9 & 90.1 & \textbf{91.0} & - & - \\
        \rowcolor{TableRowHighlight} GBC-concat & 86.1 & 86.5 & 92.7 & 93.5 & 57.5 & 37.9 & \textbf{94.6} & \textbf{94.9} & - & -\\
        \rowcolor{TableRowHighlight} GBC-graph & 84.8 & 85.7 & 85.5 & 88.0 & 90.8 & 79.8 & 89.6 & 90.5 &  \underline{\textbf{95.9}} & \underline{\textbf{96.1}} \\[-0.1em]
    \bottomrule
    \end{NiceTabular}
    }
    \vspace{0.5em}
    \caption{Image and text retrieval performance on GBC test set when trained and evaluated using different types of annotations (Rows: models trained from different annotations; Columns: Evaluations on different annotations).
    For each trained model,
    we highlight the best evaluation performance in T2I and I2T retrievals in bold. 
    For the GBC-captions column, we include all the captions from our graph by default except in cases where training is done solely on the region annotations. In this case, excluding relation and composition captions, as done during training, results in better performance.}
    \label{tab:gbc-eval}
\vspace{-1em}
\end{table}

To assess the effectiveness of different annotation formats, we provide the model with these annotations at \emph{test time}.
Annotations that better describe the images should, ideally, result in better retrieval performance when they are used.
Specifically, we use our own test set and consider performing retrieval with the various types of annotations presented in \cref{subsec:annotation}.
Note, however, that when using region captions or GBC-captions, no single text embedding can naturally encompass all the relevant information.
To address this limitation, we perform retrieval based on the average CLIP score between the image embedding and the text embeddings of the provided captions in this setup.
We report our results in \cref{tab:gbc-eval}, where the rows correspond to the annotations at training time and the columns correspond to the annotation at test time.
Unsurprisingly, we see a strong tendency that when a model is trained with a certain annotation format, it performs the best when we use the same format for retrieval.
Among the few exceptions, we note that models trained to pair with shorter captions may have better performance when concatenation of short captions is provided at test time. This leads us to the following two observations.

\textbf{Denser textual information benefits retrieval.}\quad
The table clearly shows that training with richer annotations—such as long captions, GBC-concat, or GBC-graph—enhances retrieval performance.
This improvement suggests that these methods provide a more effective representation of the images.
Specifically, GBC-graph yields the best performance, indicating that the proposed GBC format consists in a viable alternative to the commonly used detailed captions.

\textbf{Simple augmentation during training does not allow to exploit additional information when available\afterhead}\quad
Our observations from \cref{subsec:standard-benchmark} show that treating all captions as independent positives yields the best performance on existing benchmarks.
However, there is no evidence that this method could harness the richer information from multiple captions when they are provided together in test time.
Indeed, whether we use average CLIP score or concatenation, the retrieval performance of these methods significantly lags behind those methods that are trained directly with captions that individually encompass rich information.

\section{Text-to-image generation with GBC}
\label{sec:exp-t2i}
While current image generation models mostly rely on natural language instructions to create images, 
the inherent ambiguity in natural language can limit their ability to produce content that aligns precisely with the user’s intent.
Additionally, natural languages may lack the expressive power needed to convey all the nuances present in an image.
To overcome these challenges, 
a number of works have introduced middleware between text and image to enable more fine-grained control over the generated content~\cite{tipo2024yeh,yang2024mastering,lian2024llmgrounded,omost,feng2024ranni,nie2024BlobGEN}.
In this section, we complement this line of work by showing how the graph information in GBC can further enhance the image generation process.
Specifically, we explain how GBC can be used as middleware for text-to-image generation by dividing the process into two distinct tasks: text-to-GBC and GBC-to-image.
For simplicity, we discard all relation nodes and composition captions, and assume there is only one caption per node.
Additional details and experimental results are reported in \cref{apx:algo,apx:exp-setup-t2i,apx:exp-add-t2i}.

\begin{figure}[t]
    \centering
    \includegraphics[width=0.48\textwidth]{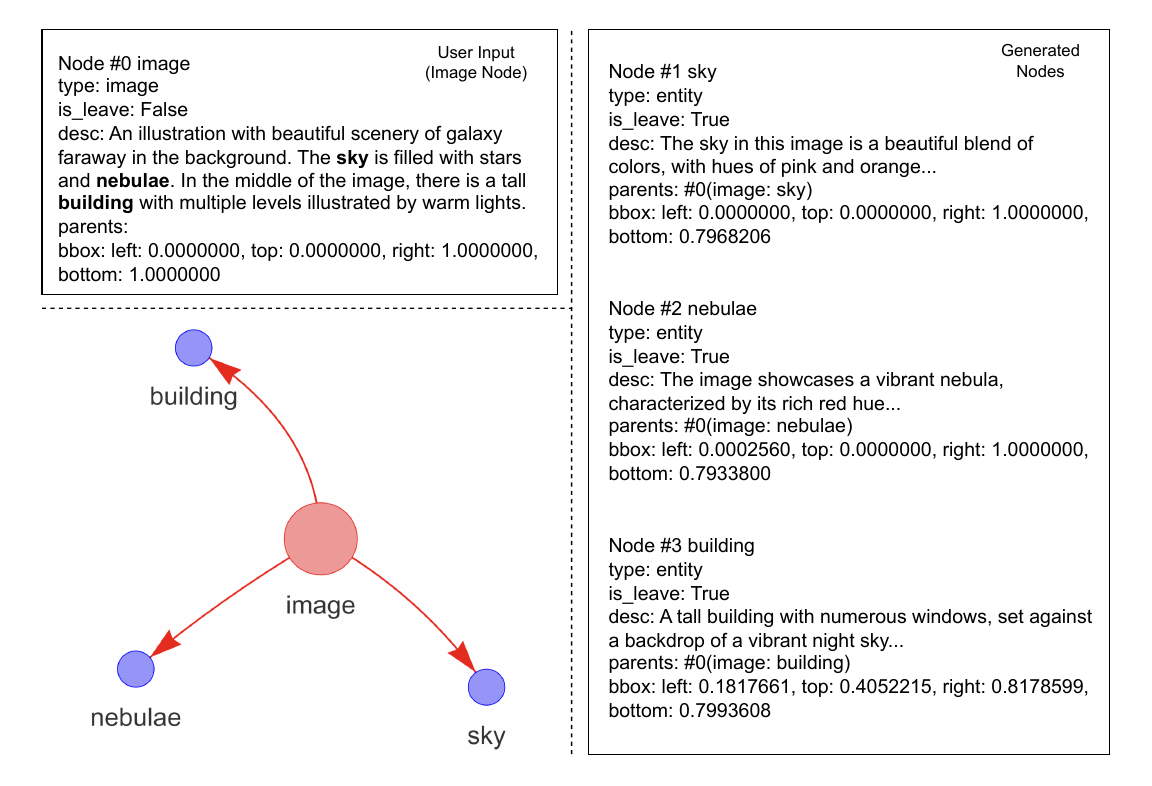}
    \caption{An example of generated graph from our 200M prompt generation model.}
    \label{fig:Text-to-GBC}
\vspace{-0.5em}
\end{figure}

\textbf{Text-to-GBC\afterhead}\quad
To generate GBC descriptions from a user-provided prompt, we encode GBC in natural languages 
and train a small language model of 200M parameters to generate the entire GBC graph from the short prompt contained in the image node. We find that as we are focusing on a very specific task, this already gives satisfying results. We show an example of the generated graph in \ref{fig:Text-to-GBC}. The small size of the additional model ensures that we add minimum overhead to the entire image generation process.

\textbf{GBC-to-image\afterhead}\quad
Our image generation experiments are based on SDXL~\cite{podell2024sdxl}, a latent text-conditional diffusion model. We consider various methods that generate images using different subsets of information from GBC and demonstrate that providing graph information helps generate images that better align with user intent.

\begin{figure*}[t]
    \centering
    \includegraphics[width=0.95\textwidth]{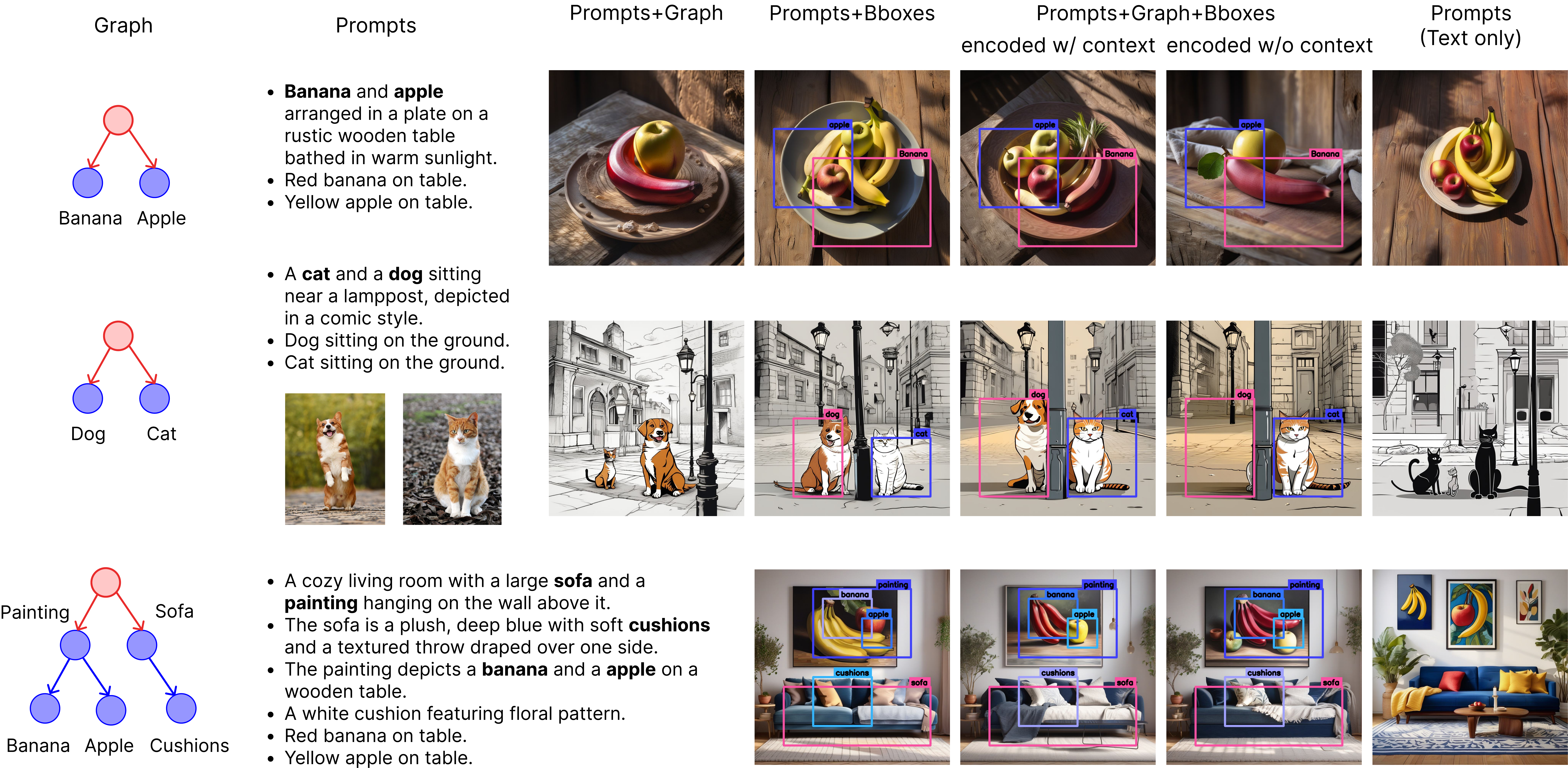}
    \caption{
    Images generated using GBC prompts with different algorithms.
    Some algorithms use only a strict subset of GBC information.
    We note that although more advanced methods for generating images from region prompts exist, our goal here is to highlight how incorporating additional graph information can enhance a simple, training-free approach that might otherwise perform poorly when only bounding box information is exploited.
    Image prompts are provided for the second example using IP adapter~\cite{ye2023ip-adapter}.
    The method that only leverages prompts and graph does not work for the third example as the depth of the corresponding graph is greater than 1.
    }
    \label{fig:GBC-diff-main}
\vspace{0em}
\end{figure*}

\begin{enumerate}
\item \textbf{Text only:} For this, we simply generate images by concatenating prompts from different nodes in a BFS order (same as GBC-concat for CLIP training).
\item \textbf{Text and bounding boxes:}
Methods for incorporating bounding boxes in image generation include training-free alternatives such as MultiDiffusion-type approaches~\cite{bar2023multidiffusion,yang2024mastering,lian2024llmgrounded}, BoxDiff~\cite{xie2023boxdiff}, and DenseDiffusion~\cite{kim2023dense}, as well as training-based alternatives like GLIGEN~\cite{li2023gligen}.
In this work, we implement a simple training-free baseline that encodes each text prompt independently and manipulates cross-attention masks so that each image patch only attends to a prompt when its corresponding bounding box contains that image patch~\cite{RegionalPrompter}.
This approach is more efficient than previous methods but often underperforms, as the image generation process may disregard additional prompts even though we attend to them in cross-attention.
\item \textbf{Text, bounding boxes, and graph:}
We find that the baseline approach presented in the previous bullet point can be drastically improved when graph information is available.
For this, we implement the following modifications of the algorithm:
\begin{itemize}
    \item 
    \textit{Improved cross-attention mask:}
    The graph structure introduces a natural hierarchy among prompts.
    To ensure that image patches focus on the most fine-grained descriptions, we design the mask so that a patch attends to a prompt only if its bounding box contains the patch and none of its descendant prompts do so.
    \item 
    \textit{Confining each node’s content to its bounding box:}
    When a bounding box is provided for a node associated with a label $\labeling$ (i.e., $\labeling$ is the label of one of the edges pointing to the node), we want to restrict $\labeling$ from appearing in regions outside the bounding box.
    This is achieved by masking out $\labeling$ from the parent prompt that points to the node using the edge labeled $\labeling$ in cross-attention and by adding $\labeling$ to negative prompts for patches outside the bounding box in question.
    
    \item 
    \textit{Encoding prompts with parent context (optional):} In some cases, encoding each prompt with additional contextual information can be beneficial. To achieve this, we concatenate each prompt with its parent prompt before passing it through the text encoder and then mask out the parent prompt in cross-attention to reduce its influence.
    \end{itemize}
\item
\textbf{Text and graph:} We note that the algorithm described above can be extended to support arbitrary segmentation masks for each node.
Additionally, segmentation masks for each object in the description can be derived from cross-attention scores, provided that these objects are accurately generated by the model.
Using this approach, we implement a method similar to the one proposed by \citet{ge2023expressive} that generates images based solely on text and graph information.
We find that this method performs well for star graphs (i.e., where every entity prompt has global prompt as its parent), but can be hardly generalized to more complex graph topologies.
\end{enumerate}

We compare the four approaches via qualitative examples in \cref{fig:GBC-diff-main}.
As expected, the method incorporating all the information from GBC performs the best.
In particular, we can generate a red banana and a yellow apple without encountering attribute binding issues when we leverage the underlying graph structure (first and forth image). 
However, such attribute binding issues can arise when the global and individual prompts are encoded together (third image) or when they are attended to simultaneously in cross-attention (second image). 
On the contrary, in the cat and dog example, encoding individual prompts with their parent prompts proves beneficial, resulting in a higher success rate for generating both animals in the image. 
Notably, even with the base global prompt, SDXL struggles to generate both a cat and a dog simultaneously.
The success rate of the approach that relies solely on text and graph is hence affected, as it assumes the base model can generate all key objects described.
In the same example, we demonstrate that image prompts can be provided for individual nodes if the underlying model can handle them. 
Finally, our last example illustrates that our approach can extend beyond star graphs, generating images based on more complex GBC prompts, confirming the advantages of incorporating graph structures in GBC.

\section{Conclusion}
\label{sec:conlcusion}

We propose graph-based captioning (GBC) as a new image-text annotation format and curated the GBC1M and GBC10M datasets using modern MLLMs and detection models.
Building on CLIP training, we introduce several baseline methods to leverage these datasets, demonstrating that models trained with GBC annotations achieve superior performance across various benchmarks compared to those trained with traditional annotation formats.
For text-to-image generation, we show that using GBC as middleware enables finer control in image generation due to the rich information provided by the graph structure.
In summary, our work demonstrates that GBC provides a versatile and powerful foundation for developing more advanced vision-language models across various applications, enhancing both image-text representation learning and text-to-image generation.

\newpage

{
    \small
    \bibliographystyle{ieeenat_fullname}
    \bibliography{references}
}

\newpage
\onecolumn
\appendix
\noindent\rule{\textwidth}{1pt}
\begin{center}
\vspace{3pt}
{\Large Appendix}
\vspace{-3pt}
\end{center}
\noindent\rule{\textwidth}{1pt}

\section*{Table of Contents}
\startcontents[sections]
\printcontents[sections]{l}{1}{\setcounter{tocdepth}{2}}
\newpage

\section{Related works, limitations, and societal impact}
\label{apx:related}
This appendix delves deeper into the broader context of our study, examines additional related works, discusses the limitations of our methodologies, and explores its potential societal impacts.

\subsection{Additional related works}

In this section, we include works related to CLIP training and text-to-image generation.

\vspace{0.25em}
\textbf{CLIP with recaptioning\afterhead}\quad
CLIP~\citep{radford2021learning} is a seminal vision-language model that utilizes text and image encoders to generate joint latent representations.
While there is an extensive body of literature on CLIP training---ranging from modifications in the objective~\citep{yu2022coca, li2022blip}, data augmentation techniques~\cite{li2023scaling, fini2023improved}, to training procedures~\cite{zhai2023sigmoid,sun2023eva}---it is impossible to cover all developments comprehensively here.
Among these, particularly relevant to our work is the recent trend that highlights the benefits of enhancing caption quality through dedicated models.
For instance, VeCLIP~\citep{lai2024veclip} enriches image alt-text with outputs from LLaVA, while similar recaptioning strategies have also been explored by~\citet{nguyen2023improving,doveh2023dense}, and \citet{mobileclip2024}. 
On the other hand, LaCLIP~\citep{fan2023improving} employs LLaMA~\citep{touvron2023llama} to rewrite captions.
Going further, SynthCLIP~\cite{hammoud2024synthclip} leverages a dataset with entirely generated captions and images for CLIP training.

\vspace{0.25em}
\textbf{CLIP with additional annotations\afterhead}\quad
There has been a plethora of research on training CLIP models with diverse annotations such as long captions, region captions, and scene graphs.
As for long captions, DreamLIP~\cite{zheng2024dreamlip} proposes to sample sub-captions from the long description to construct multiple positive pairs, while Long-CLIP~\citep{zhang2024longclip} addresses CLIP's 77-token limitation by modifying the positional encoding to accommodate longer text sequences during fine-tuning. 
Meanwhile, region annotations with varying granularity have been considered by works including GLIP~\citep{li2022grounded}, X-VLM~\citep{zeng2022multi}, and RegionCLIP~\citep{zhong2022regionclip}.
Their objectives match features of image crops to their specific descriptions.
Efforts that aim to improve CLIP training with the help of scene graphs include CLIP-SGVL~\cite{herzig2023incorporating} and Structure-CLIP~\cite{huang2024structure}.
The former integrates scene graphs to define additional objective for image encoder, while the later uses scene graphs to guide the generation of negative captions, and to enrich the text encoder with additional contextual information.

\vspace{0.25em}
\textbf{Image generation with layout conditioning\afterhead}\quad
Numerous works have studied image generation with layout conditioning, with the conditioning frequently represented as a single prompt with grounding information or bounding boxes along with their respective annotations on top of a global prompt. 
Early methods primarily focused on training generation models directly with such conditions \cite{zhao2019image}, and some even employed scene graph conditioning and only used layouts as intermediaries for generation \cite{farshad2023scenegenie,johnson2018image}.
More recent works, such as \cite{zheng2023layoutdiffusion}, have effectively integrated layout information into modern image generation models using similar strategies.
However, the advent of text-to-image models trained on large-scale paired image/caption datasets has rendered training-from-scratch approaches less necessary.
Instead, training additional adapters~\cite{li2023gligen,nie2024BlobGEN} or controlnets~\cite{zhang2023adding,zhao2024uni} is often sufficient to incorporate layout information into image generation.

Another thread of work focuses on training-free approaches.
MultiDiffusion~\cite{bar2023multidiffusion} proposes to run the diffusion model sampling process in parallel for different regions and combine them on the fly.
This method ensures that each region contained the desired information when individual prompt is simple enough, and has been further refined in LLM-grounded diffusion~\cite{lian2024llmgrounded} and
RPG-DiffusionMaster~\cite{yang2024mastering}.
The downside of these approaches is the need for running the sampling multiple times, making the total cost proportional to the number of annotated regions in the image.
Another set of approaches manipulates cross-attention.
\citet{chen2024training} distinguish between forward and backward guidance methods. Forward approaches involve direct intervention in cross-attention scores, while backward approaches update intermediate states at each sampling step to minimize a specific energy function.
They found that the backward approach generally outperforms the forward approach.
However, it is also more computationally expensive due to the additional gradient steps involved.
The baseline methods we implement can be seen as variants of the forward approach.
More advanced modulations to improve forward approaches have been proposed by \citet{kim2023dense} for image generation based on segmentation mask information.
On the other hand, BoxDiff~\cite{xie2023boxdiff} can be considered a variant of the backward approach.

\vspace{0.25em}
\textbf{Text-to-image generation with additional middleware\afterhead}\quad
It has become increasingly common to transform user-provided prompts into more detailed prompts before feeding them into text-to-image models~\cite{BetkerImprovingIG,tipo2024yeh}.
To enhance the model's ability to generate images with complex compositions, other works have focused on incorporating grounding information into user prompts.
This can be achieved through various means, such as region annotations~\cite{lian2024llmgrounded,yang2024mastering}, semantic panels~\cite{feng2024ranni}, dense blob representations~\cite{nie2024BlobGEN}, or \textit{Canvas}~\cite{omost}.
Additionally, \citet{gu2024kaleido} proposed conditioning the final generation on both the initial prompt and the generated middleware, demonstrating promising results.

\subsection{Limitations and perspectives}

We discuss below the limitations of our works from three different perspectives, the procedure and format, the datasets, and the experiments.
These limitations also naturally point to several future directions that are to be explored. %

\subsubsection{Limitation concerning the GBC procedure and format}

While GBC remains a versatile high-level annotation format that in principle applies to any image, its design is inherently tied to the coarse-to-fine and compositional nature of natural images.
This design orientation means that GBC is not necessarily the most suitable for certain types of images such as scientific imagery, homogeneous patterns, or abstract art.
Specifically, scientific imagery often requires annotations that convey precise, quantifiable data rather than relational or descriptive text.
This limitation highlights the need for tailored approaches to different visual content categories to address their unique characteristics.

\subsubsection{Limitation concerning the GBC datasets}

Our datasets are curated with the help of LLaVA and Yolo-World, and hence inherit their limitations.
This includes but is not limited to, the bias and hallucination from LLaVA captioning, incorrectly identified objects from Yolo-World, and the inability of Yolo-World to recognize certain object category (see \cref{apx:dataset-examples} for concrete examples).
Moreover, our approach mainly distinguishes between objects of the same type via composition nodes.
Yet, we believe that there is a more effective strategy than merely assigning numbers to these objects.

\subsubsection{Limitation concerning our experiments}

Our experiments on CLIP training and retrieval tasks are of relatively small scale and do not fully uncover the potential of the GBC annotation format, especially as we believe that what GBC offers may not be fully represented in traditional benchmarks.
Our text-to-image experiments focus on training-free approaches for image generation, which is inherently limited by the underlying model's capabilities.
This limitation is evident in the model's difficulty generating multiple objects accurately, even with simple prompts, and the attribute binding issues that arise within the text encoder.
A training-based approach would likely mitigate these challenges.
Lastly, we did not investigate the application of GBC for training MLLMs, which could be a promising avenue for future exploration.

\subsection{Societal impact}

Our paper introduces the GBC datasets and procedure, both aimed at advancing the development of multimodal models.
Specifically, the structured approach of GBC, designed to provide detailed descriptions, may help overcome representational biases inherent in existing captioning pipelines, offering more accurate descriptions of images.
The potential benefits of these advancements extend across a range of applications, such as assistive technologies and scientific research.
However, alongside these benefits, there are challenges including the potential spread of misinformation and concerns about privacy.
A comprehensive discussion of these broader societal impacts, both positive and negative, extends beyond the immediate focus of our methodological study.

\section{Dataset construction}
\label{apx:dataset-construct}
In this appendix, we provide all the missing details about our dataset construction process that are not mentioned in \cref{subsec:dataset-construct,subsec:gbc12M}.

\subsection{Query templates}

To make the \ac{MLLM} models fulfill the tasks described in \cref{subsec:dataset-construct}, we perform COT prompting~\cite{wei2022chain} with few-shot examples.
The four templates for our queries are shown in \cref{fig:image-query-template-1} to \ref{fig:relation-query-template}.
We make the following remarks concerning the design of our prompts.

\vspace{0.25em}
\textbf{Prompt structure\afterhead}\quad
We craft these prompts with the help of ChatGPT, which results in prompts that might be more complicated than necessary.
Meanwhile, we did notice that the inclusion of few-shot examples is crucial for the model to adhere to the required output formats.
Given that using always the same few-shot examples might significantly bias the model's output, it could be beneficial to randomly retrieve examples from a diverse pool for each query, but we did not pursue this exploration.

\vspace{0.25em}
\textbf{[Single] and [Multiple] annotations\afterhead}\quad
Since a detection model could output multiple candidate bounding boxes for an input text, we ask the \ac{MLLM} to annotate each identified element with either [single] or [multiple].
We then proceed with slightly different algorithms in the two cases, to encourage the selection of either only one, or multiple bounding boxes.
In particular, we use respectively an NMS threshold of $0.05$ and $0.2$ for objects labeled with [single] and [multiple].
However, these labels do not necessarily dictate the final count of bounding boxes; multiple boxes may still be selected for items labeled [single], and vice versa.

\vspace{0.25em}
\textbf{Dynamically filled-in elements\afterhead}\quad
To ensure that the response of the \ac{MLLM} is relevant, the prompts are dynamic and reflect the content of the current image (the image query being the only exception).
Such information comes from previous queries and can be naturally retrieved for different queries.
The only nonobvious part is the \emph{hard coded hints} for composition queries, which we explain below.

\vspace{0.25em}
\textbf{Hard coded hints for composition queries.}\quad
After numerous attempts, we observe that LLaVA-1.6 struggles with accurately describing the composition of multiple objects in a scene, even when these objects are annotated with bounding boxes.
To overcome this limitation, we guide the models with hints generated programmatically using a set of predefined rules. Specifically, we begin by constructing a Euclidean minimum spanning tree based on the centers of the bounding boxes. We then select a random node as the root and perform a \ac{DFS} on the tree. During this search, we interleave descriptions of the edges, which detail the geometric relations between two objects based on the positions of their bounding boxes, with node descriptions. These node descriptions are added when an object is located at a particular extremity of the composition, such as the rightmost or top-left position.

\subsection{Text classifiers}

Both of our text classifiers are trained for binary classification using logistic loss. To determine whether a piece of text is suitable for object detection, we utilize a single linear layer added on top of the Jina Embedding.\footnote{\url{https://huggingface.co/jinaai/jina-embeddings-v2-small-en} ~~Accessed: 2024-05-01} For the task of assessing whether two texts can represent the same object, we concatenate their Jina embeddings and process them through an MLP. This MLP includes layer normalization, a hidden layer that expands the input dimensionality by a factor of four, followed by SiLU activation and the final linear layer.
The dataset for the training of our text classifiers are prepared with the help of ChatGPT.

\subsection{Other details for data annotation}

We incorporate LLaVA-1.6 into our pipeline using \texttt{llama.cpp}.\footnote{\url{https://github.com/ggerganov/llama.cpp} ~~Accessed: 2024-05-01}
Moreover, to speed up the annotation process, 
we utilize models quantized at different precision levels: the vision encoders at 6-bit precision, the LLM component of LLaVA-1.6 Mistral-7B at 5-bit precision, and the LLM component of LLaVA-1.6 Yi-34B at 3-bit precision.%
\footnote{
\url{https://huggingface.co/cmp-nct/llava-1.6-gguf} ~~Accessed: 2024-05-01}
We use the default hyperparameters for inference except for a temperature of 0.1 and context window of size of 5952 (note that LLaVA-1.6 can use up to 2880 image tokens).
We discard any responses that do not comply with our required format.

As for the object detection model, we use YOLO-Worldv2-X trained with input resolution of $640\times 640$.%
\footnote{\url{https://huggingface.co/wondervictor/YOLO-World/blob/main/yolo_world_v2_x_obj365v1_goldg_cc3mlite_pretrain-8698fbfa.pth} ~~Accessed: 2024-05-01}
We set the confidence threshold to 0.05 and retain a maximum of six bounding boxes for each input text, selecting those with the highest confidence scores. We exclude any region whose size is smaller than 5,000. To prevent repetitive descriptions of the same element, we keep only those bounding boxes that occupy less than 80\% of the current image region for detections arising from entity queries. Regarding node merging, we consider two bounding boxes to be overlapping if their intersection occupies more than 85\% of the area of each bounding box involved.

\begin{figure}[p]

\begin{tcolorbox}[colback=red!5!white,colframe=red!50!black,title=Query Template for Image Query]

\setlength{\parskip}{6pt}
\renewcommand{\baselinestretch}{1.2}\selectfont

\vspace{0.4em}
\textbf{System message}\\[-.7em]

\begin{scriptsize}
As an AI visual assistant, your role is to conduct an in-depth analysis of images and articulate a detailed account of the visible elements. You must then distill this information into a precise and concise caption that accurately reflects the content of the image.

Step-by-Step Process:

Detailed Caption:\\
- Conduct a thorough examination of the image to note all elements present, including main subjects, minor objects, background details, and any text.\\
- Prepare a detailed caption that accounts for all these elements, emphasizing the whole objects within the scene.

Top-Level Element Identification:\\
- Identify and format concrete objects: Begin by identifying concrete objects within the image that are detectable by object recognition models. Each identified object should be formatted as [object\_name][node\_type] where [node\_type] is either [single] or [multiple]:\\
\hphantom{1} - [single]: Applied to items that appear only once in the image, represented as a unique entity within its context, such as a [cat][single] or a [chair][single]. This category is used regardless of the object's size or location in the frame and is intended for items that are not repeated elsewhere in the image. For example, a [stop sign][single] on a street corner or a [tree][single] in a field.\\
\hphantom{1}  - [multiple]: Applied to items that are present more that once within the image, emphasizing their plurality. Examples include [dogs][multiple] playing in a park, [chairs][multiple] in a café, [park benches][multiple] along a pathway, [girls][multiple] on a street, [pillows][multiple] on a couch, [paintings][multiple] on a wall, and [lights][multiple] across a ceiling.\\
- Entire objects only: When identifying elements within an image, only include objects that stand alone as the main subjects. Avoid breaking down the top-level objects into smaller components.\\
- Grouping similar items: When general items, such as houses, trees, players, or people, appear multiple times in the image, they should be grouped together under a single [multiple] label rather than described separately. This approach applies even if these items might have been described individually in the detailed caption.\\
- No abstraction: Do not include abstract qualities like colors (blue, red, white), patterns, or expressions.\\
- No numbering: Do not use any number to label objects. Just use [houses][multiple].\\
- No directional description: Do not use positional terms for individual elements. Instead, group similar items under a single [multiple] label, like [cowboys][multiple].

Concise Formatted Caption:\\
- Use the identified elements to construct a concise formatted caption. Use brackets to denote each identified object, following the [object\_name][node\_type] format. The object name should only appear in the bracket.\\
- Restrict the number of elements mentioned in the concise caption to avoid overcrowding and ensure clarity. Prioritize the inclusion of key elements that define the scene or the subject's essence.\\
- The concise caption should contain at most two sentences.

Example Adjustments:\\
- Character attributes: When analyzing an image featuring a person with distinctive attributes such as armor or tattoos, focus on the person as a whole rather than the individual attributes. The correct annotation would be [person][single], encompassing all aspects of the person appearance without breaking them down into separate elements.\\
- Architectural features: In the case of architectural elements, avoid itemizing components like the roof, windows, or door if they contribute to the overall structure of a building. For a singular building in the image, use [house][single]. If the image depicts a series of buildings, such as a row of houses with varying designs, annotate them collectively as [houses][multiple], regardless of their individual features.\\
- Groups of similar objects: For scenes containing groups of similar objects or individuals, such as girls playing in a park, group them under a single [multiple] label. Even if the individuals are engaged in different activities or have distinct appearances, they should be annotated as [girls][multiple] to emphasize their collective presence within the scene. Similarly, even if multiple dogs or chairs have different colors, they should be labeled as [dogs][multiple] and [chairs][multiples].

\end{scriptsize}
\end{tcolorbox}
\caption{The system prompt used for image query (first half).}
\label{fig:image-query-template-1}
\end{figure}

\begin{figure}[p]%
\newpage
\begin{tcolorbox}[colback=red!5!white,colframe=red!50!black,parbox=false,title=Query Template for Image Query]

\setlength{\parskip}{6pt}
\renewcommand{\baselinestretch}{1.2}\selectfont

\textbf{System message (continued)}\\[-.7em]

\begin{scriptsize}

Example Captions:

For an image featuring multiple elements like a logo:

Detailed Caption: A design showcasing a prominent grey 'N' at the top, with three smaller NEO Business Bank logos directly below it, two colored squares positioned to the bottom left, and a line of text to the bottom right detailing the availability of various file formats for the design.
Top-Level Element Identification:\\
- ['N'][single]\\
- [Logos][multiple]\\
- [Squares][multiple]\\
- [Text][single]\\
Concise Formatted Caption: A design showcasing grey ['N'][single] positioned over NEO Business Bank [logos][multiple], accompanied by colored [squares][multiple] and [text][single] at the bottom.

For an illustration of a zebra:

Detailed Caption: An animated zebra stands upright on two legs, waving in a welcoming manner, next to a wooden signpost at the beginning of a dirt path. This path leads to a quaint wooden cabin with a thatched straw roof, surrounded by a simple wooden fence. In the background, there's another similar cabin. The scene is completed by a clear sky overhead and multiple trees dotting the landscape, contributing to the lush greenery.
Contextual Considerations: The zebra's legs are part of its overall form and should not be listed separately.\\
Top-Level Element Identification:\\
- [Zebra][single]\\
- [Signpost][single]\\
- [Dirt path][single]\\
- [Cabins][multiple]\\
- [Trees][multiple]\\
- [Sky][single]\\
Concise Formatted Caption: An animated [zebra][single] waves next to a wooden [signpost][single] on a [dirt path][single] that leads towards wooden [cabins][multiple], with [trees][multiple] enhancing the lush greenery under a clear [sky][single].

For a photo of two men on street:

Detailed Caption: A photo of two men standing side by side on a city street. The man on the left has long hair and is wearing a beige blazer over a white shirt with black trousers. He is smiling and looking directly at the camera. The man on the right has short hair and is dressed in a gray blazer over a black shirt with gray trousers. He also smiles at the camera. They are standing on a sidewalk lined with shops and buildings, suggesting they are in a commercial or urban area. The lighting suggests it might be late afternoon or early evening.
Contextual Considerations: The two men, despite their distinct appearances and attire, should be grouped together under a single label since they both fall under the category of "men".\\
Top-Level Element Identification:\\
- [Two men][multiple]\\
- [Sidewalk][single]\\
- [Shops][multiple]\\
- [Buildings][multiple]\\
- [City street][single]\\
Concise Formatted Caption: [Two men][multiple] stand side by side on a [sidewalk][single] along a [city street][single], lined with [shops][multiple] and [buildings][multiple], each dressed in coordinated blazers and trousers.

\end{scriptsize}

\tcbline

\textbf{User message:}

\begin{scriptsize}
Following the instruction, please provide a detailed caption and a concise formatted caption for the given image. Note that it is crucial for you to put elements that can be detected and should be further described in picture in brackets as [object\_name][node\_type] in the concise formatted caption.

\end{scriptsize}

\end{tcolorbox}
\caption{The system and user prompts used for image query (second half).}
\label{fig:image-query-template-2}
\end{figure}

\begin{figure}[p]

\begin{tcolorbox}[colback=blue!5!white,colframe=blue!50!black,title=Query Template for Entity Query]

\setlength{\parskip}{6pt}
\renewcommand{\baselinestretch}{1.2}\selectfont

\vspace{0.4em}
\textbf{System message}\\[-.7em]

\begin{scriptsize}
Your task is to perform an in-depth analysis of a cropped image focusing on a requested object, like a "house". The process involves a step-by-step evaluation to identify the object's presence, describe its features, craft concise captions, and assess any prominent objects.

Process Overview:

Verify Object Presence:\\
- Examine the image to determine if the specified object, or any instance of it, is present.\\
- State the presence with "Object Present: Yes" or "Object Present: No".

Provide Appropriate Caption (If Object Is Present):\\
- Provide a detailed description of the object, focusing solely on its features without reference to other elements in the image.\\
- The description should contain at most 50 words.

Assessment of Prominent Objects:\\
- Evaluate the described features to determine if any stand out for further description and are detectable by an object detection model. This is crucial for complex objects such as 'man', 'woman', 'family', 'couple', 'cat', or 'house', where components or distinctive attributes are significant. For example, when analyzing 'woman', consider elements like dress [single], skirt [single], or hair [single] as prominent features. For simpler objects like 'cup' or 'chair', additional descriptions may not be needed.

Identification of Prominent Features (If Applicable):\\
- If there are prominent features identified, list and format these features for potential detection by an object detection model.\\
- Ensure these features are parts or components of the main object and not the entire object itself.\\
- Use [single] for unique, standalone items, and [multiple] for features present more than once, such as roof [single] or windows [multiple].\\
- Group similar items under a single [multiple] label rather than describing them separately, even if individual descriptions were provided in the detailed caption. For example, multiple distinct windows in a house should be labeled as windows [multiple] rather than individually enumerated.\\
- For groups like families or couples, identify members separately (e.g., man [single], woman [single]) rather than as a collective unit. This contrasts with grouping similar inanimate objects (e.g., windows [multiple]), where individual distinction isn't necessary.\\
- Consistency with the caption: Ensure that the features identified as [single] or [multiple] are also mentioned in the caption.

Example Responses:

Example 1: Object Not Present

Object Presence: No\\
Detailed Caption: N/A\\
Prominent Features: N/A\\
Identification of Prominent Features: N/A

Example 2: Object Present Without Prominent Features (requested object: "cup")

Object Presence: Yes\\
Detailed Caption: A simple ceramic cup on a wooden table. The cup has a smooth, unadorned surface and a standard curved handle on one side.\\
Prominent Features: No\\
Identification of Prominent Features: N/A

\end{scriptsize}

\end{tcolorbox}
\caption{The system prompt used for entity query (first half).}
\label{fig:entity-query-template-1}
\end{figure}

\begin{figure}[p]

\begin{tcolorbox}[colback=blue!5!white,colframe=blue!50!black,title=Query Template for Entity Query]

\setlength{\parskip}{6pt}
\renewcommand{\baselinestretch}{1.2}\selectfont

\vspace{0.4em}
\textbf{System message (continued)}\\[-.7em]

\begin{footnotesize} 

Example 3: Object Present With Prominent Features (requested object: "family")

Object Presence: Yes\\
Detailed Caption: A family of four is captured enjoying a sunny day in the park. The father, in casual attire, is engrossed in setting up a picnic spot, while the mother, donned in a summer dress, is laying out a feast on a blanket. Nearby, two children, a boy and a girl, engage in playful antics; the boy is kicking a football with fervor, and the girl, adorned in a light frock, is gleefully chasing bubbles.\\
Prominent Features: Yes\\
Identification of Prominent Features:\\
- Father: [single]\\
- Mother: [single]\\
- Boy: [single]\\
- Girl: [single]

Example 4: Object Present With Prominent Features (requested object: "car")

Object Presence: Yes\\
Detailed Caption: A vintage car in pristine condition, with shiny chrome bumpers and classic spoke wheels. The car's body is painted in a vibrant red, and the leather interior is visible through the clear windows. A unique hood ornament adorns the front, adding to the car's elegance.\\
Prominent Features: Yes\\
Identification of Prominent Features:\\
- Chrome bumpers: [single]\\
- Wheels: [multiple] \\
- Hood ornament: [single]

\end{footnotesize}

\tcbline

\textbf{User message:}

\begin{footnotesize}
Please assess the image focusing on '\{\}'. Start by confirming its presence with 'Object Present: Yes' or 'Object Present: No'. If present, describe its key features in a detailed caption with at most 50 words. Then, evaluate if any aspects stand out for further emphasis, stating 'Prominent Features: Yes' or 'No' while preferring "Yes". If yes, list a few notable features in brackets, applying [single] or [multiple] as appropriate. Importantly, do not include '\{\}' in features. Instead, you should break it down.

\end{footnotesize}

\tcbline

\begin{minipage}{0.75\linewidth}
\textbf{Example filled-in elements}\\[-0.7em]

\begin{footnotesize}
\begin{itemize}
    \item boat
    \item boat
\end{itemize}
\end{footnotesize}
\end{minipage}
\begin{minipage}{0.2\linewidth}
\includegraphics[width=\textwidth]{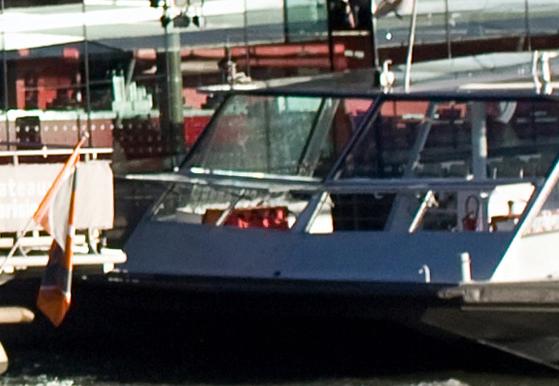}
\end{minipage}

\end{tcolorbox}
\caption{The system and user prompts used for entity query (second half).
The placeholders `\{\}' are dynamically filled with the name of the object, i.e., the label of the associated incoming edge.}
\label{fig:entity-query-template-2}
\end{figure}

\begin{figure}[p]

\begin{tcolorbox}[colback=green!5!white,colframe=green!50!black,title=Query Template for Composition Query]

\setlength{\parskip}{6pt}
\renewcommand{\baselinestretch}{1.2}\selectfont

\vspace{0.4em}
\textbf{System message}\\[-.7em]

\begin{scriptsize}
Your role is to analyze images containing objects within pre-labeled bounding boxes and describe the compositional arrangement of these objects based on provided hints. You will then provide general descriptions that apply to all the objects collectively.

Input Image Format Explanation:\\
- The image will feature objects of interest, each enclosed within a bounding box.\\
- Each bounding box will be numbered centrally to uniquely identify it.\\
- The objects will be similar in nature (e.g., all dogs) and positioned within a scene.

Utilizing Hints for Analyzing Composition:\\
- Begin by reviewing the hints provided regarding the spatial arrangement of the objects.\\
- These hints may specify the relative positions of objects (e.g., "Object 3 is in the top right corner").\\
- Use the hints to guide your description of how the objects relate to each other within their bounding boxes.

Output Format:\\
- Composition Description: Start with "Composition:" followed by a description informed by the hints and using the bounding box numbers. This description should elucidate the spatial arrangement of the objects as per the hints.\\
- General Descriptions: Provide observations that apply to all objects within the specified group, excluding unrelated elements or background details. Preface this section with "General descriptions:".

Additional Guidelines:\\
- Describe the spatial arrangement of objects without inferring spatial relations from the sequence of numbers.\\
- Utilize clear spatial language to articulate the composition.\\
- The description should reflect the actual visual composition, not the order of numbers in the bounding boxes.

Examples:

Example for 3 Dogs in Bounding Boxes:

Query Prompt: "Please describe the composition of the 3 dogs in the bounding boxes, followed by some general descriptions that apply to all dogs."

System Response:

Composition: Dog 3 is in front, with dog 2 to the left and dog 1 to the right.\\
General descriptions:\\
- The three dogs are aligned in a row on the grass.\\
- They share similar sizes and features, suggesting they may be from the same breed.

Additional Examples:

For 5 Flowers in a Garden Bed in Bounding Boxes:\\
Composition: Flower 4 takes a central position, flanked by flower 2 and flower 3 on either side, while flower 1 and flower 5 bookend the arrangement at the outer edges.\\
General descriptions:\\
- Each flower is in full bloom, indicating a peak growing season.

For 2 Cats in a Window in Bounding Boxes:\\
Composition: Cat 1 is positioned on the left side of the window sill, while cat 2 is curled up on the right.\\
General descriptions:\\
- Both cats are basking in the sunlight coming through the window.\\
- Their relaxed postures suggest a shared sense of comfort and tranquility.

\end{scriptsize}

\end{tcolorbox}
\caption{The system prompt used for composition query.}
\label{fig:composition-query-template-1}
\end{figure}

\begin{figure}[p]

\begin{tcolorbox}[colback=green!5!white,colframe=green!50!black,title=Query Template for Composition Query]

\setlength{\parskip}{6pt}
\renewcommand{\baselinestretch}{1.2}\selectfont

\textbf{User message:}

\begin{small}
Please describe the composition of the \{\} in the bounding boxes, followed by some general descriptions that apply to all \{\}. The composition should include \{\} and be based on the following hints (do not mention hints or bounding boxes in the response).\\
\{\}

\end{small}

\tcbline

\textbf{Example filled-in elements}

\begin{small}
\begin{itemize}
    \item boats
    \item boats
    \item boat 2, boat 3, boat 1
    \item
    - boat 2 is on the right side of the composition\\
    - boat 2 is to the right of boat 3\\
    - boat 3 is to the right of boat 1\\
    - boat 1 is on the left side of the composition
\end{itemize}
\end{small} 

\vspace{0.3em}
\includegraphics[width=\textwidth]{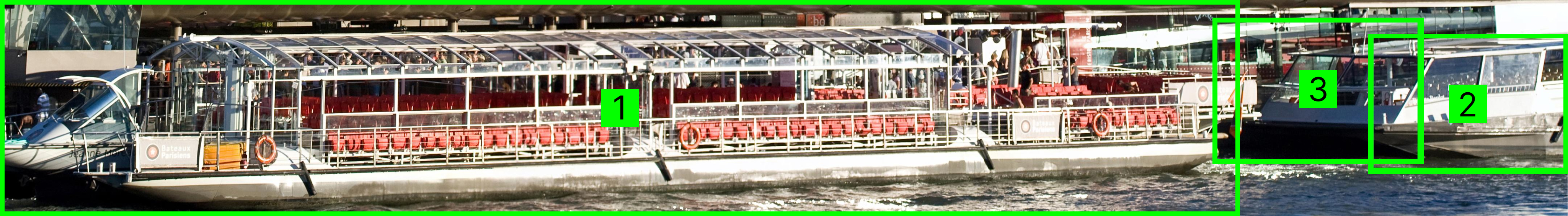}

\end{tcolorbox}
\caption{The user prompt used for composition query. The placeholders `\{\}' are dynamically filled with the name of the object, the labels of the out edges (in the form of the name of the object plus number), and hard coded hints that are obtained using the positions of the bounding boxes.}
\label{fig:composition-query-template-2}
\end{figure}

\begin{figure}[p]

\begin{tcolorbox}[colback=yellow!5!white,colframe=yellow!50!black,title=Query Template for Relation Query]

\setlength{\parskip}{5pt}
\renewcommand{\baselinestretch}{1.2}\selectfont

\vspace{0.4em}
\textbf{System message}\\[-.7em]

\begin{scriptsize}
Your role involves analyzing the spatial and direct interactions between pre-identified elements within an image, described through annotations like [beach], [turquoise waters], [people], [shoreline], [lush vegetation]. Your task is to objectively describe how these elements are related or positioned relative to each other within the scene.

Steps for Identifying Objective Relations:\\
1. Review Annotated Elements: Start by examining the list of annotated elements. Understand the nature of each element as it is presented in the image.\\
2. Identify Spatial Positions: Determine the spatial positioning of these elements in relation to each other. Focus on direct relationships such as touching, overlapping, or proximity without implying any further interpretation.\\
3. Describe Direct Interactions: Look for and describe any direct interactions between elements, such as one element supporting another, blocking, or leading into another.\\
4. Format Relations List: Provide your findings as a list of bullet points. Each bullet should detail a direct and observable relationship between two or more elements, using their annotated identifiers for reference.

Example Relations Based on Annotated Elements:

For elements: [beach], [turquoise waters], [people], [shoreline], [lush vegetation], you might reply:

- The [people] are standing on the [beach], with the [lush vegetation] to their left.\\
- [Turquoise waters] lap against the [beach] at the [shoreline], with [people] scattered along its edge.\\
- [Lush vegetation] flanks the left side of the [beach], providing a natural border.\\
- The [shoreline] separates the [beach] from the [turquoise waters].\\
- To the right of the [lush vegetation], the [beach] stretches towards the [turquoise waters].

For another set of elements: [eagle], [snake], [wings], you might reply:

- The [eagle] has its [wings] spread above the [snake].\\
- The [snake] is positioned below the [eagle].\\
- The [eagle]'s claws are near or in contact with the [snake].

Guidelines for Reporting Relations:\\
1. Ensure descriptions are based solely on visible or directly inferable relationships without adding interpretations or assumptions.\\
2. Maintain clarity and precision in articulating the spatial and interactional dynamics between elements.\\
3. Stick to objective descriptions that reflect the physical and observable aspects of the elements' relationships.\\
4. Only answer the list of bullet points without anything before or after.\\
5. Do not include any bullet point with 1 or even 0 elements.

- Visible Relationships Only: Report relationships that are clearly depicted in the image. If no clear relationships are visible, state "No visible relationships."\\
- Objective Descriptions: Keep descriptions factual and based solely on what can be seen in the image.\\
- Avoid Assumptions: Do not infer or assume any relationships that aren't clearly shown in the image.\\
- Bullet Point Format: Present each observable relationship as a separate bullet point, avoiding any descriptive text not related to the direct relationships.\\
- No Relation Inference: Refrain from implying relationships or positions that are not explicitly shown. If elements are simply present without any discernible interaction, it is acceptable to say "Elements are present without visible interaction."\\
- Avoid Single Element Points: Do not include bullet points that mention only one element or have no elements at all. Each bullet point must reference the relationship between two or more elements.

\end{scriptsize}

\tcbline

\setlength{\parskip}{6pt}
\renewcommand{\baselinestretch}{1.2}\selectfont

\textbf{User message:}

\begin{footnotesize}
Please infer and list of at most \{\} relations between \{\} in this images.
\end{footnotesize}

\tcbline

\setlength{\parskip}{6pt}
\renewcommand{\baselinestretch}{1.2}\selectfont

\begin{minipage}{0.88\linewidth}
\textbf{Example filled-in elements:}\\[-1em]

\begin{footnotesize}
\begin{itemize}
    \item 4
    \item eiffel tower, sky, trees, boat, water
\end{itemize}
\end{footnotesize}
\end{minipage}
\begin{minipage}{0.08\linewidth}
\includegraphics[width=0.85\textwidth]{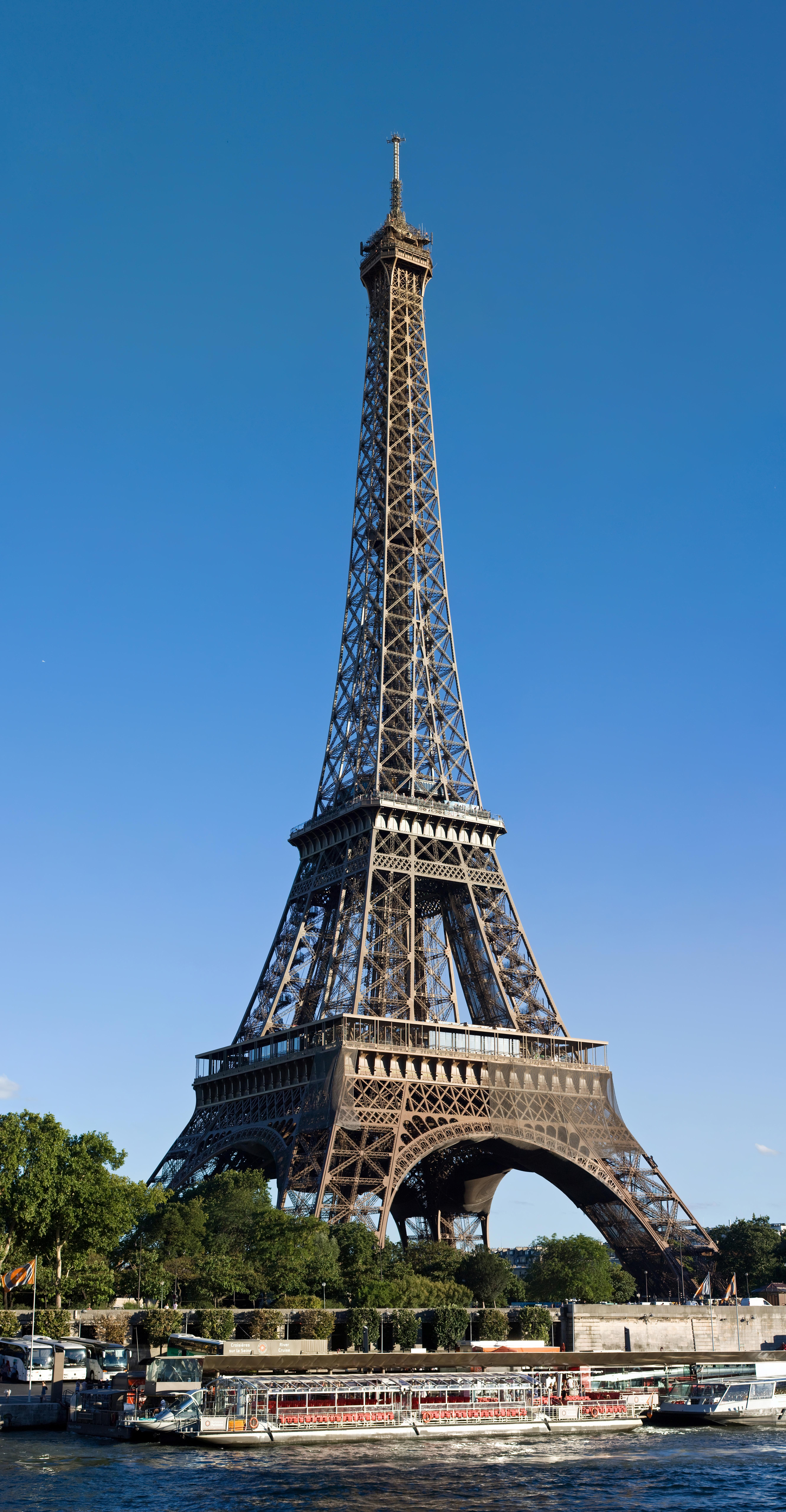}
\end{minipage}

\end{tcolorbox}
\caption{The system and user prompts used for relation query. The placeholders `\{\}' are dynamically filled with a random number between $2$ and the number of involved objects, and the names of these objects.}
\label{fig:relation-query-template}
\end{figure}

\clearpage

\subsection{Computation cost}

We list below the major computation cost of our data preparation process.
\begin{itemize}
    \item GBC1M: With our processing pipeline, it takes an average of around 3 minutes to annotate each image on an A100 80G when all the queries are performed with LLaVA-1.6 Yi-34B.
     As a result, annotating 1 million images took us around 6 days with 300 A100 80Gs.
    \item GBC10M: 
    The average annotation time per image on an A100 80G is improved to 1 minute when relation and entity queries are performed with LLaVA-1.6 Mistral-7B.
    This process is about twice slower on a V100 32G.
    In this regard, our GBC12M dataset was compiled in roughly 6 days using 500 A100 80Gs and 1,000 V100 32Gs.
\end{itemize}

We also compute CLIP score for each caption using the DFN-5B model.\footnote{\url{https://huggingface.co/apple/DFN5B-CLIP-ViT-H-14-378} ~~Accessed: 2024-05-01}
This computation takes around 3 hours for every 10,000 images on a V100 32G.

\section{Dataset information}
\label{apx:dataset-info}
In this appendix, we provide information about dataset release,
dataset statistics, %
and visualizations of a few examples from our GBC10M dataset. %

\subsection{Data release and licensing}
\label{apx:dataset-release}

\disclose{Our datasets are available at \url{https://huggingface.co/graph-based-captions}, released under the CC BY-NC 4.0 license.}{Our datasets have been released under the CC BY-NC 4.0 license.}
Following CC12M, we include URLs to images along with captions generated through our GBC procedure, all stored in JSON lines format. 
Comprehensive documentation including a dataset card and croissant metadata is also provided in the data repository.

\paragraph{Personal identifiable information and offensive content\afterhead}
Our dataset comprises only captions generated by MLLM models (LLaVA 1.6 Yi-34B and LLaVA 1.6 Mistral-7B), which were trained on carefully curated data.
The images, sourced from CC12M, are generally free from offensive content.
In particular, CC12M is the result of a filtering operation involving \textit{adult content detection} on images and their captions. 
While CC12M images may include human faces, we do not host the images directly; only the URLs are provided. 
Additionally, we conduct toxicity check with Detoxify~\cite{Detoxify} on a subset of examples in GBC dataset and find no harmful contents.
While it was not possible to manually examine all the samples produced by GBC pipeline, we believe that the protective measures of the source dataset and model are sufficient to avoid both harmful content, and privacy leakages.

\subsection{Dataset statistics}
\label{apx:dataset-stat}

In this section, we provide statistical insights into the GBC1M and GBC10M datasets. In particular, we zoom in on the statistics at image, vertex, edge, and caption levels, and present distributions of several key metrics including for example caption length, region size, and CLIP score.
Since most of these metrics exhibit long-tailed distributions, we often group excessively large values into a single histogram bin for better visualization.

\subsubsection{Image and graph statistics}
\label{apx:image-graph-stat}

\begin{figure}[p]
    \centering
    \includegraphics[width=0.42\textwidth]{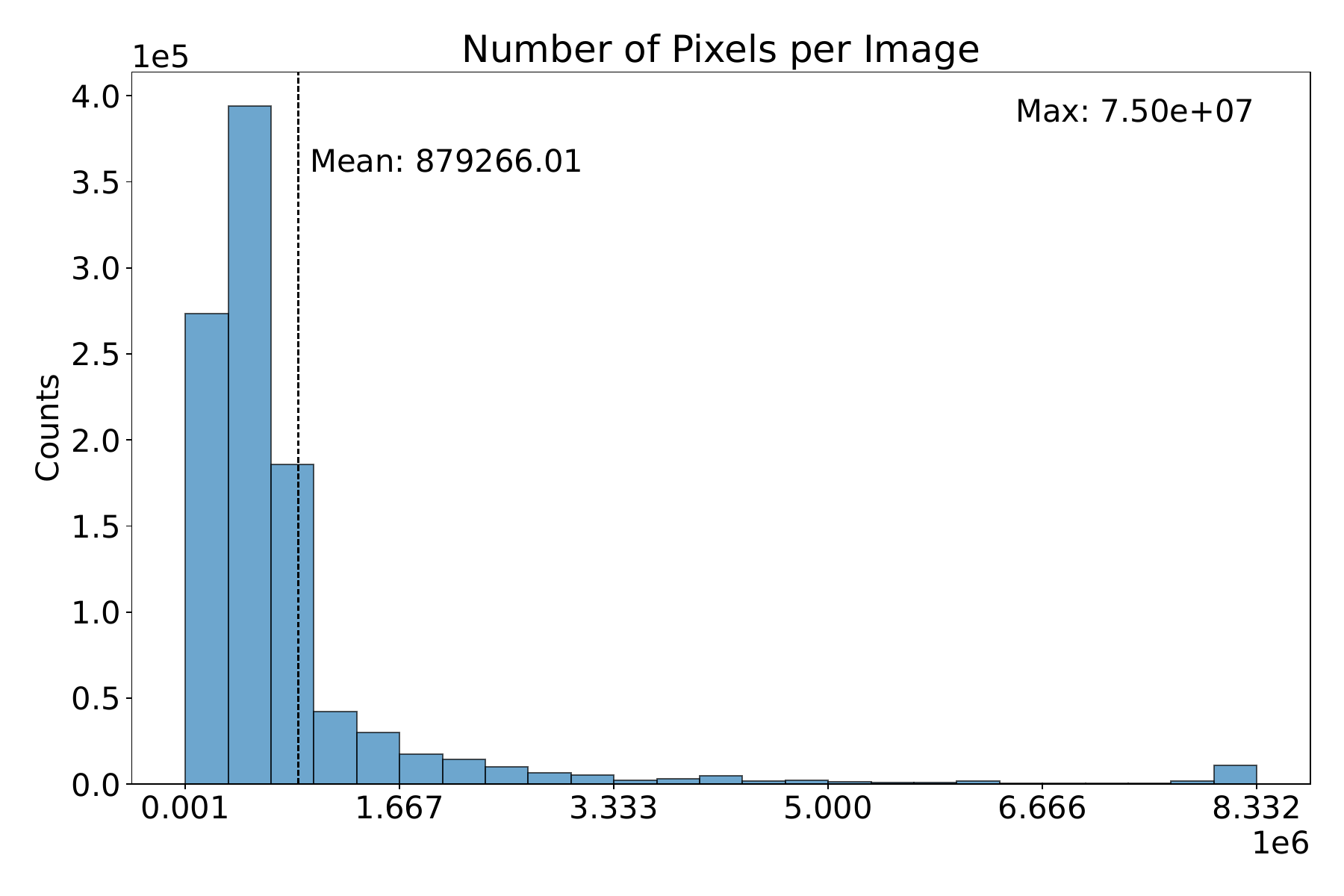}
    \includegraphics[width=0.42\textwidth]{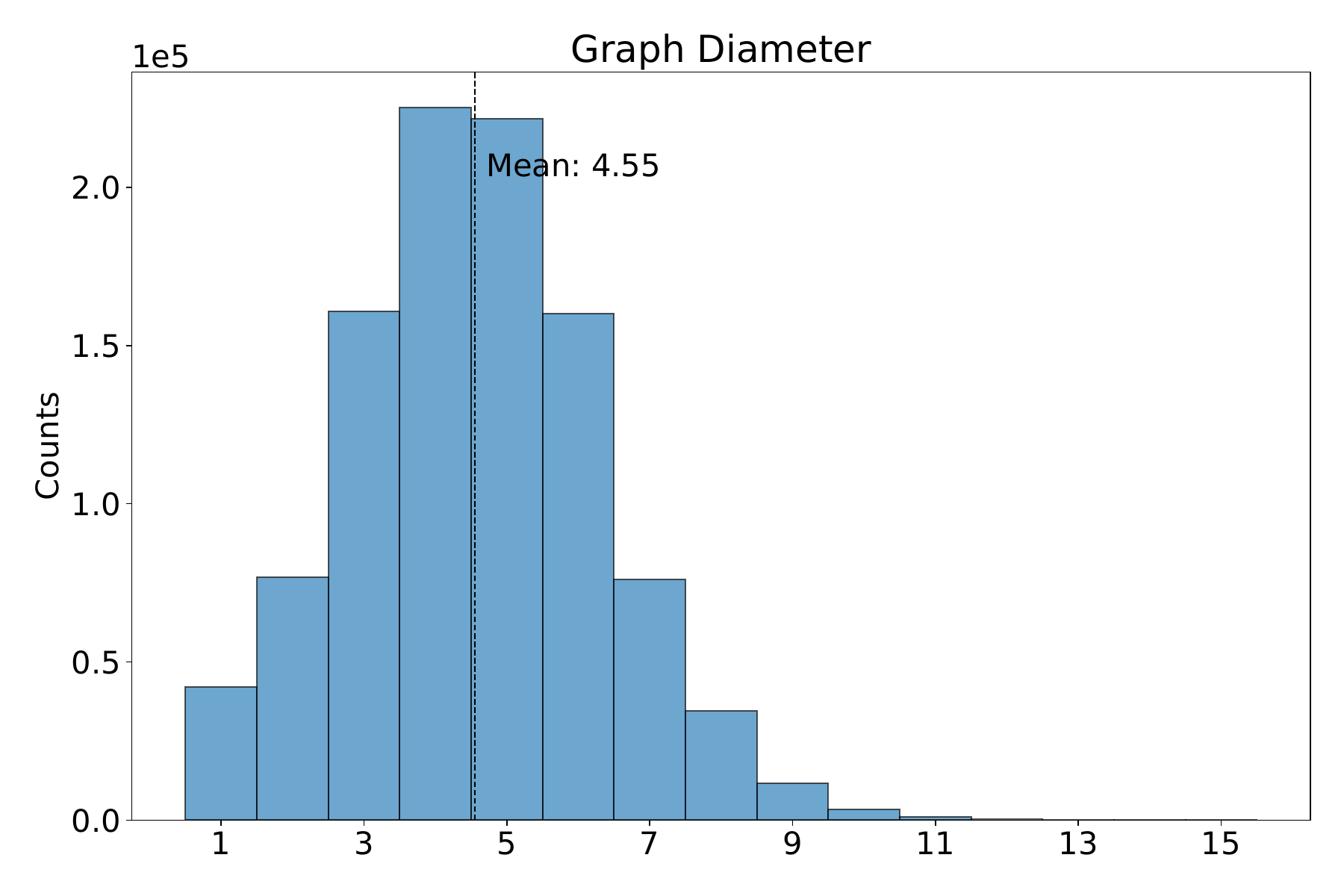}
    \includegraphics[width=0.42\textwidth]{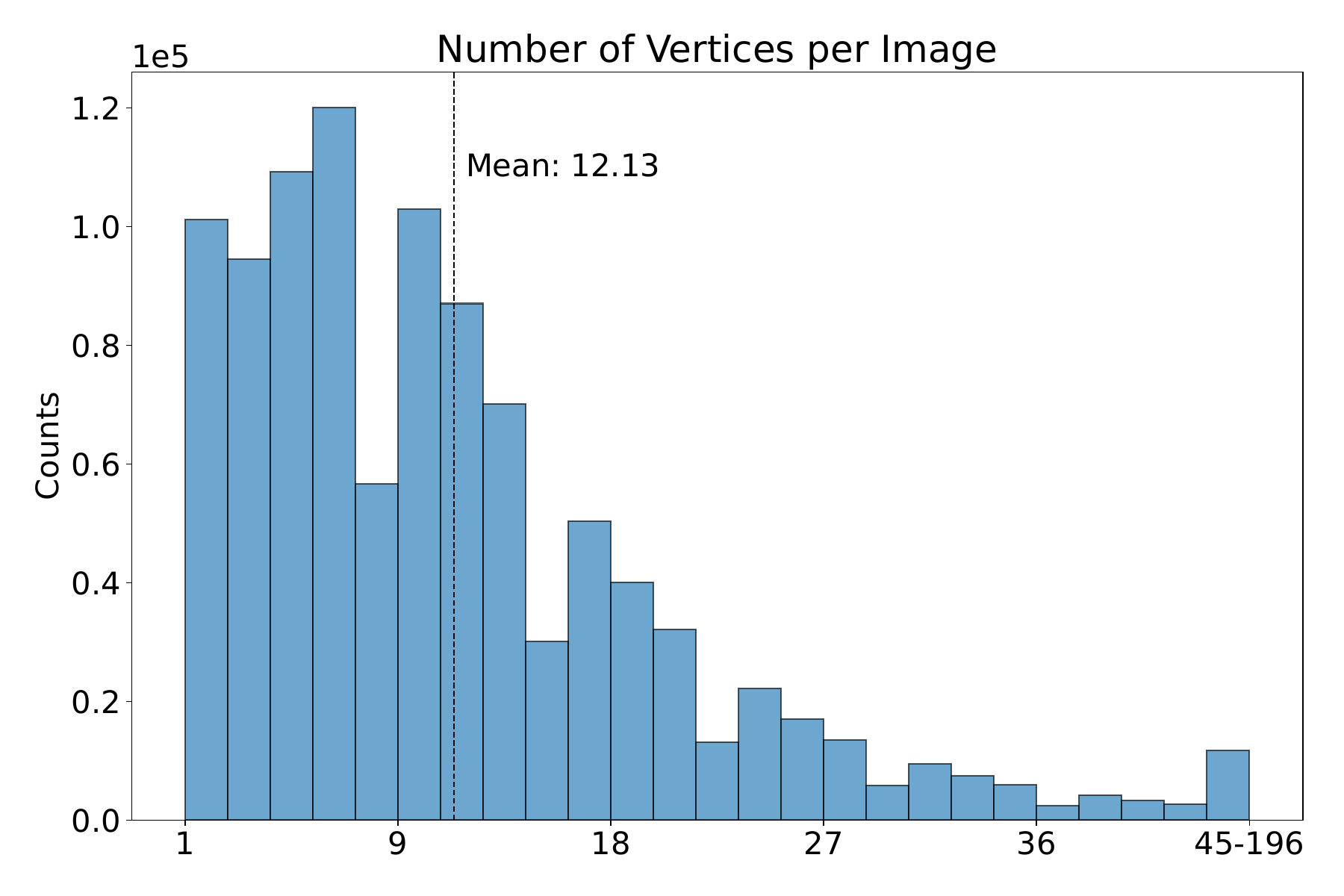}
    \includegraphics[width=0.42\textwidth]{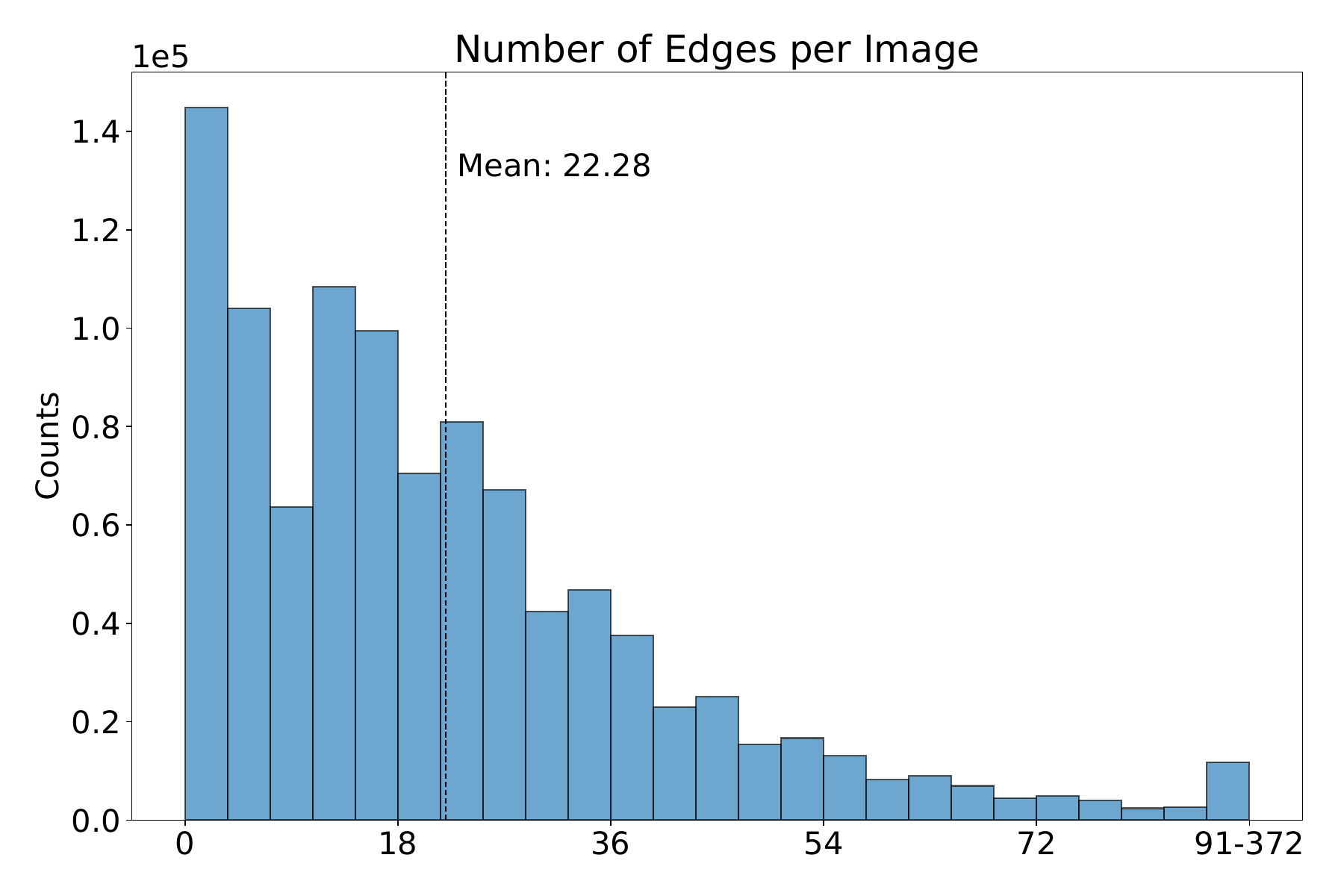}
    \caption{Distributions of metrics at image and graph level in the GBC1M Dataset.}
    \label{fig:gbc1m-image-stat}
\end{figure}

\begin{figure}
    \centering
    \includegraphics[width=0.42\textwidth]{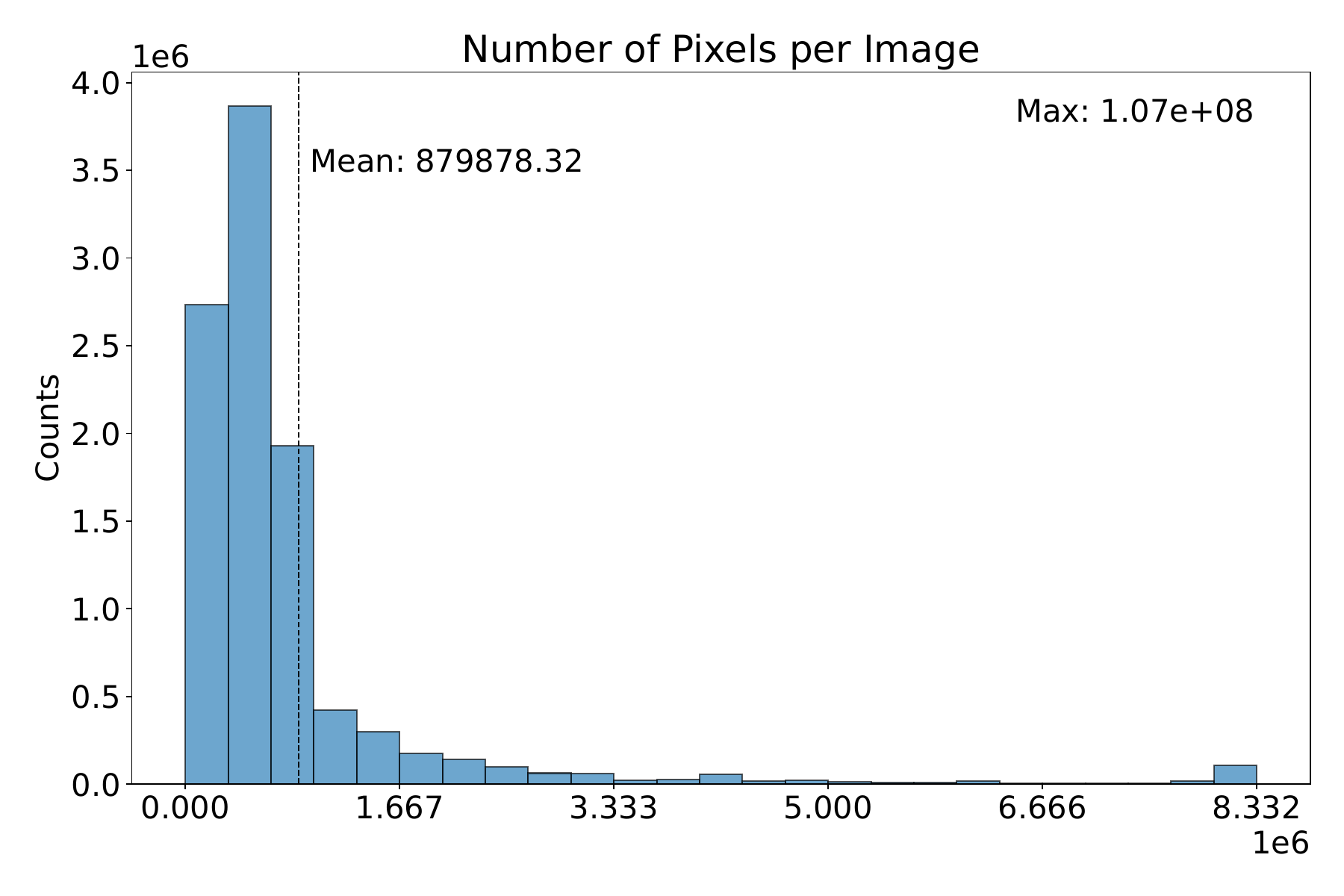}
    \includegraphics[width=0.42\textwidth]{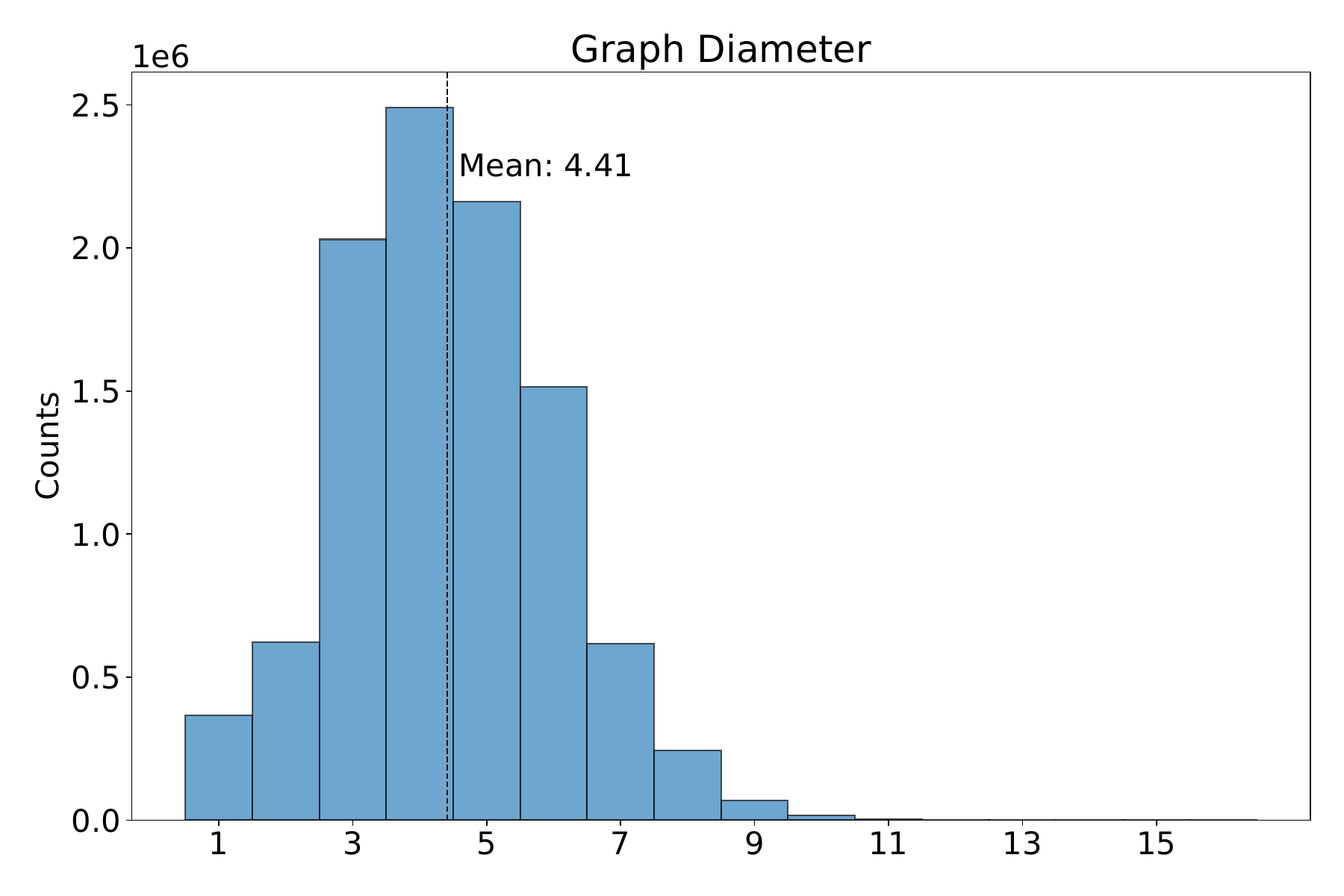}
    \includegraphics[width=0.42\textwidth]{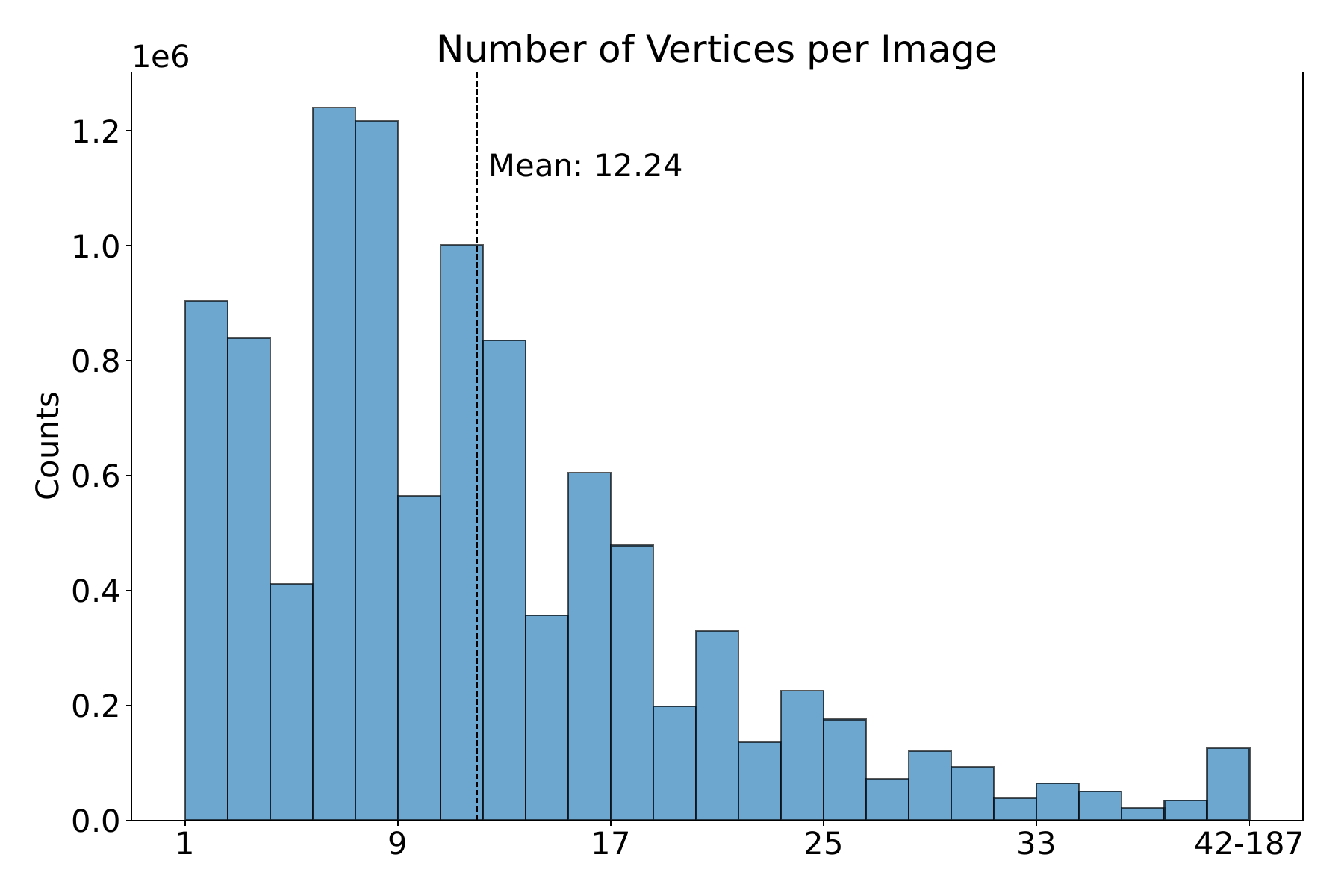}
    \includegraphics[width=0.42\textwidth]{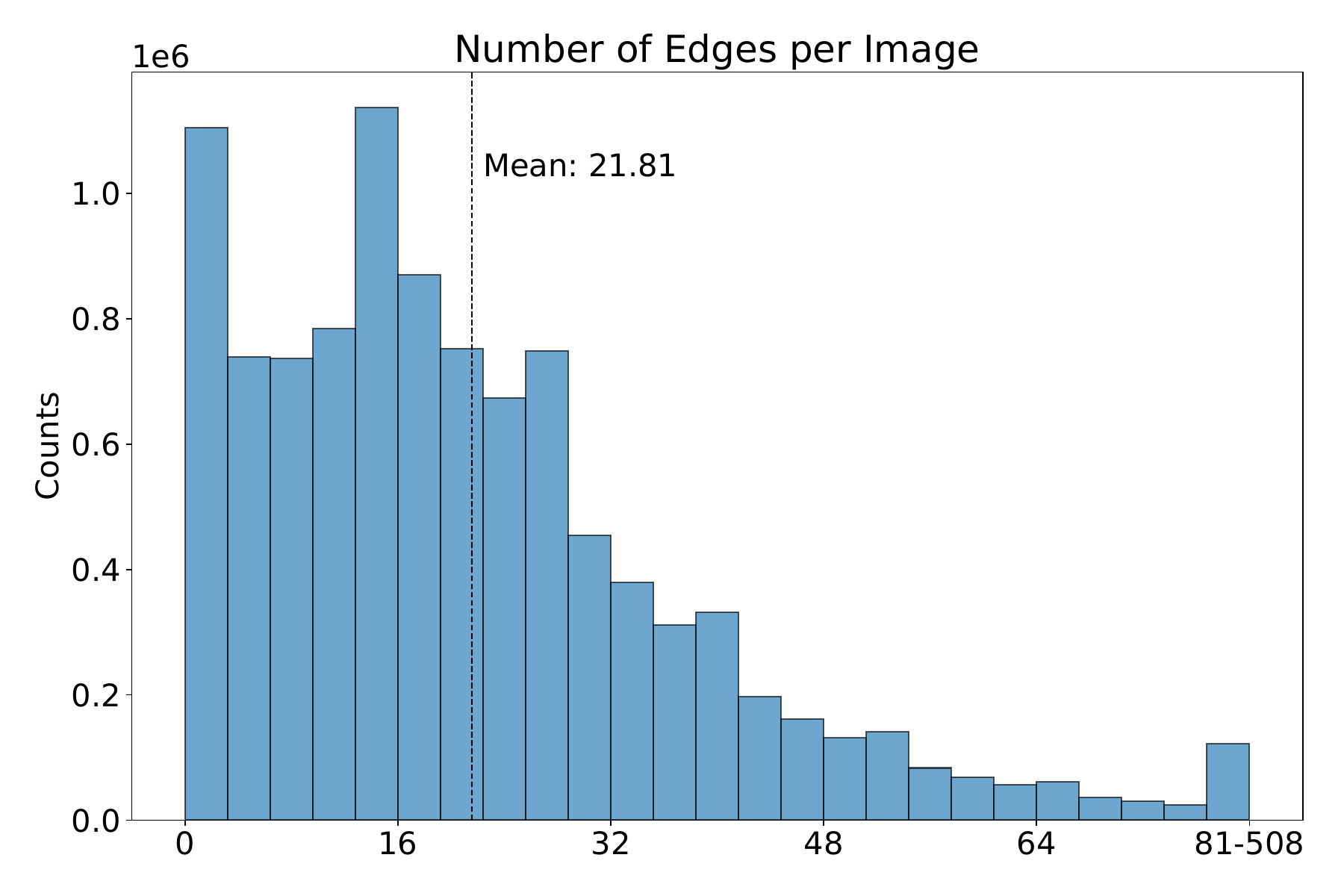}
    \caption{Distributions of metrics at image and graph level in the GBC10M Dataset.}
    \label{fig:gbc10m-image-stat}
\end{figure}

We first look at the sizes of the images and of the annotation graphs, \ie the numbers of vertices and edges in these graphs and their diameters (which is measured as the length of the longest path in a directed graph).
The distributions of these metrics are shown in \cref{fig:gbc1m-image-stat,fig:gbc10m-image-stat}.
We see that the image size has a very long-tailed distribution, with the majority of images having around $786\times 786$ pixels.
Conversely, the distributions of graph diameters are more similar to that of a Poisson or a binomial distribution, with most of the graphs having a diameter between 3 and 6.
Finally, as one could expect, the numbers of vertices and edges share quite similar distributions.

While we expect the size of a graph to reflect the inherent complexity of an image, we acknowledge that our annotations are influenced by the biases of the used models.
In particular, we observe that our annotation process tends to yield larger graph for natural images compared to other types of images such as artworks or graphic designs.


\clearpage

\subsubsection{Vertex statistics}

\begin{table}[t]
    \centering
    \renewcommand{\arraystretch}{1.25}
    \setlength\tabcolsep{1.4em}
    \begin{tabular}{@{\hskip 1.2em}l|@{\hskip 1.2em}cc@{\hskip 1.2em}}
    \toprule
    Dataset & \# Images
    & \# Regions / Image
    \\
    \midrule
    COCO~\cite{lin2014microsoft}
    & 123,000 & 7
    \\
    Visual Genome \cite{krishna2017visual}
    &  108, 249 & 42
    \\
    Objects365 \cite{shao2019objects365}
    & 638,000 & 16
    \\
    Open Images \cite{kuznetsova2020open}
    & 1.7M & 8
    \\
    BigDetection \cite{cai2022bigdetection}
    & 3.5M & 10
    \\
    SA-1B \cite{kirillov2023segment}
    & 11M & 100 
    \\
    AS-1B \cite{wang2023all}
    & 11M & 110
    \\
    DCI \cite{doveh2023dense}
    & 7,805 & 40
    \\
    GBC1M (ours)
    & 1.1M & 11
    \\
    GBC10M (ours)
    & 10.1M & 11
    \\
    \bottomrule
    \end{tabular}
    \vspace{0.75em}
    \caption{Comparison of number of regions per image among several vision-language datasets with region-based annotations.
    We use the statistics reported in the original paper although some datasets, such as COCO and Open Images have been updated after their initial release.
    Moreover, for Open Images we report the number for the training set with bounding box annotation \cite[Tab. 5]{kuznetsova2020open}.
    For DCI, we compute the average number of regions per image ourselves using their open-sourced dataset with 7,805 images as this number is not reported in \cite{doveh2023dense}.
    }
    \label{tab:dataset-stat-regions}
\vspace{-1em}
\end{table}

We have shown previously that our datasets contain an average of 12 vertices per graph.
This translates to 11 regions per image after excluding the root node that represents the entire image.
We compare this number with several other vision-language datasets with region-based annotations in \cref{tab:dataset-stat-regions}.
As one can see, this number aligns well with many of these datasets, 
particularly those used for detection, such as COCO and Object365.
However, it lags behind compared to Visual Genome and more recent datasets with dense annotations, such as AS-1B and DCI.
We believe this discrepancy can be attributed to both the top-down design of our annotation process, which tends to overlook less significant components of the images, and the limitations of the detection model used. 
Notably, both AS-1B and DCI utilize Segment Anything~\cite{kirillov2023segment} to identify regions of interest. Segment Anything is trained on SA-1B, which has much denser annotations compared to the object detection datasets used for training Yolo-World.

\begin{figure}[p]
    \centering
    \includegraphics[width=0.42\textwidth]{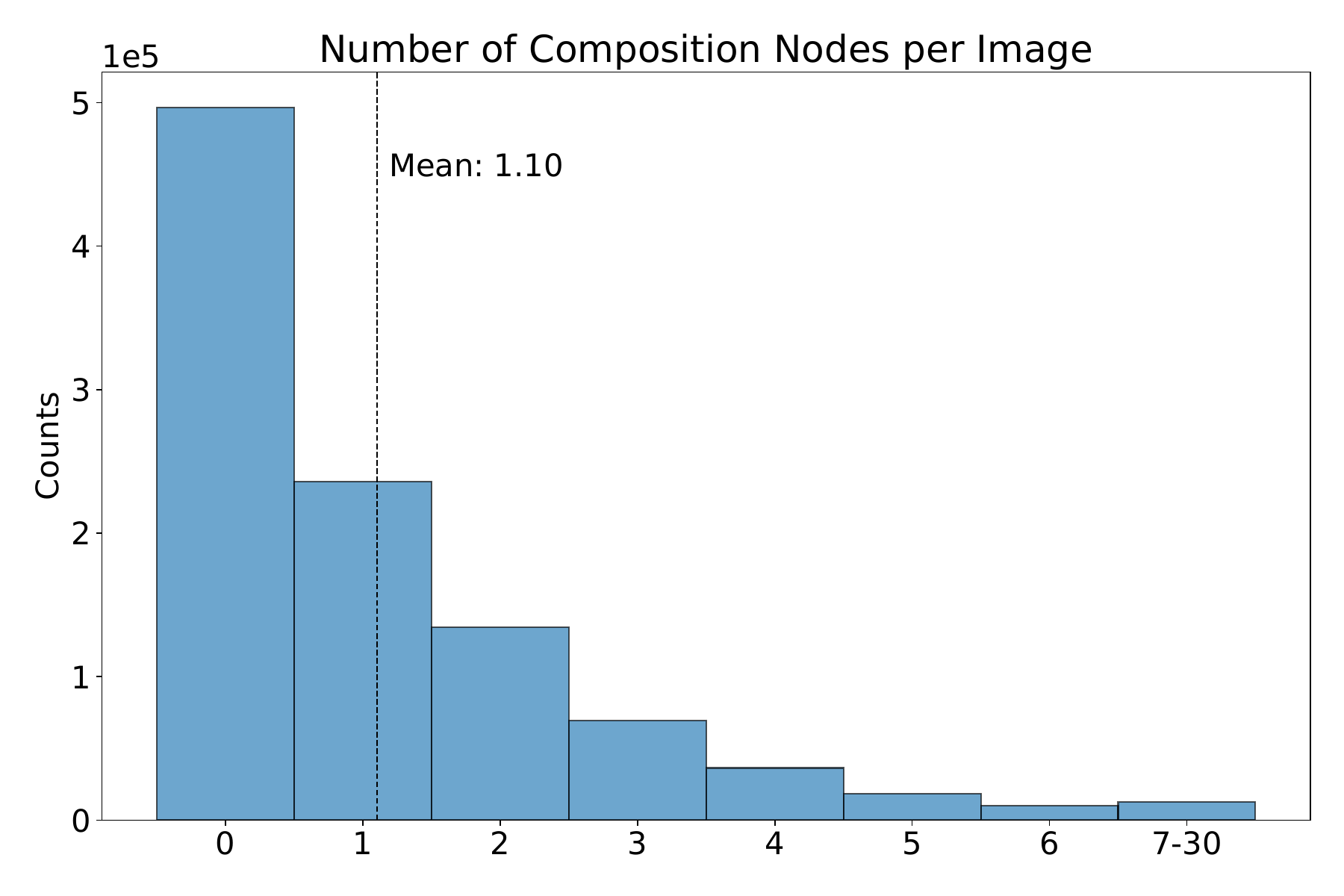}
    \includegraphics[width=0.42\textwidth]{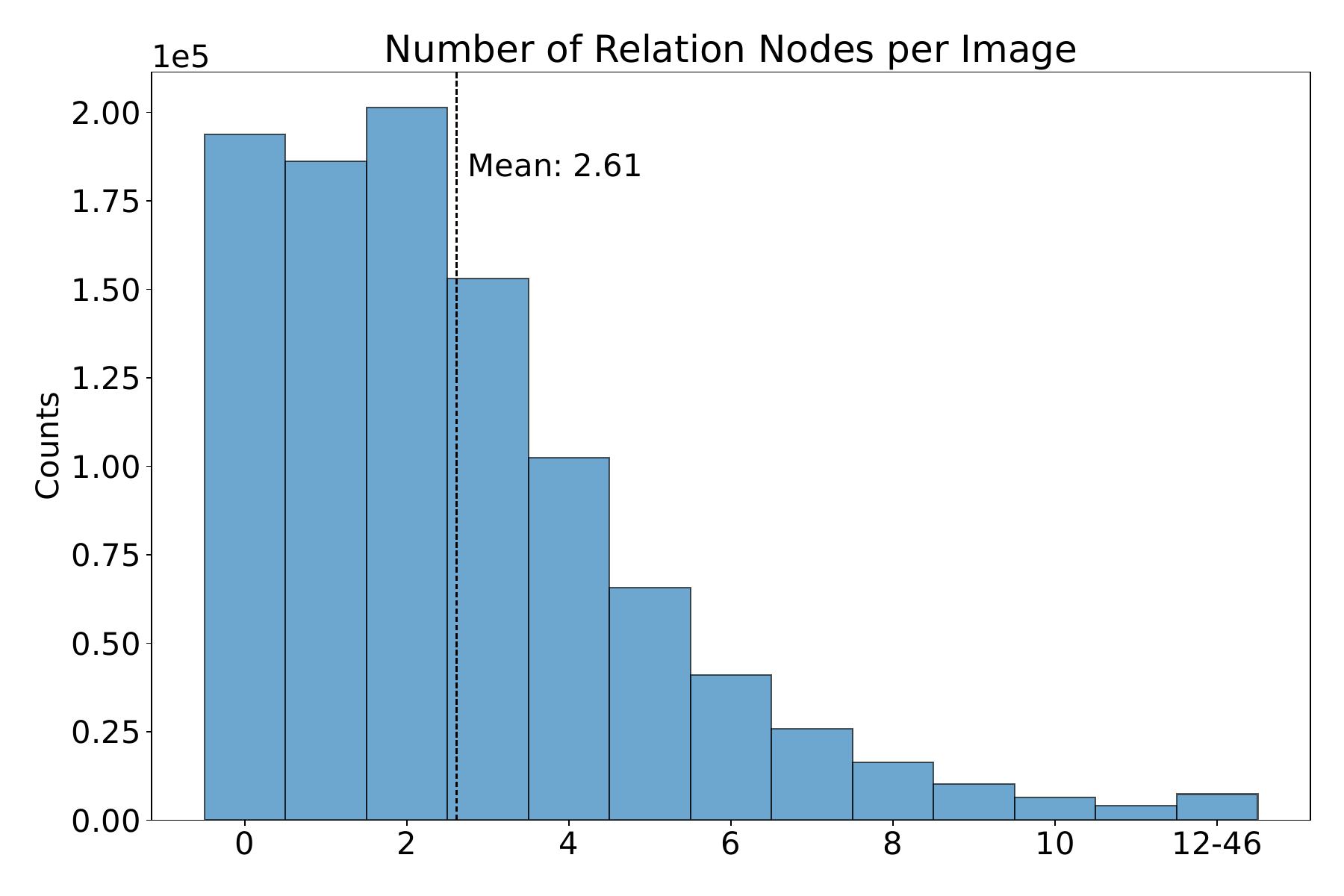}
    \includegraphics[width=0.42\textwidth]{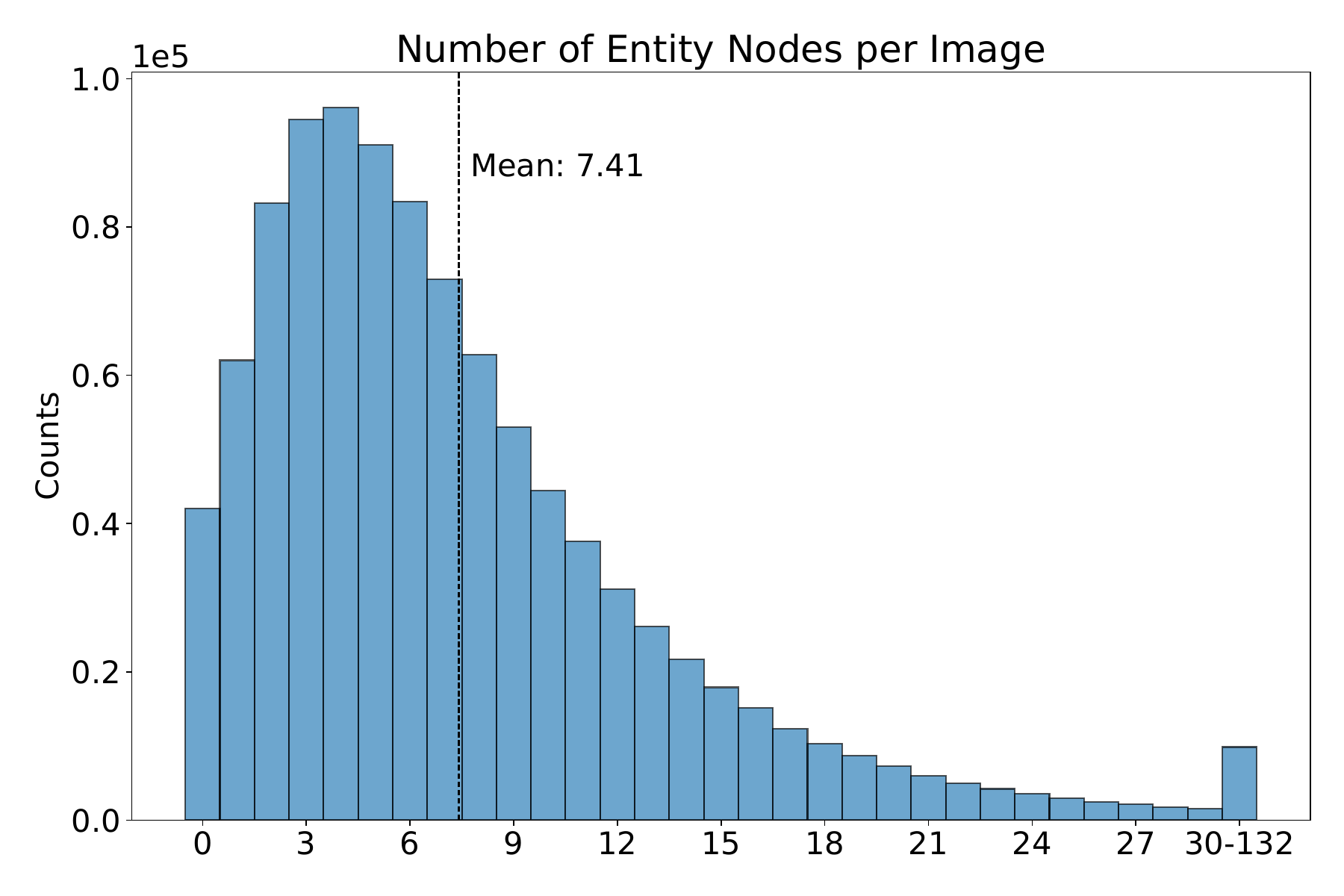}
    \includegraphics[width=0.42\textwidth]{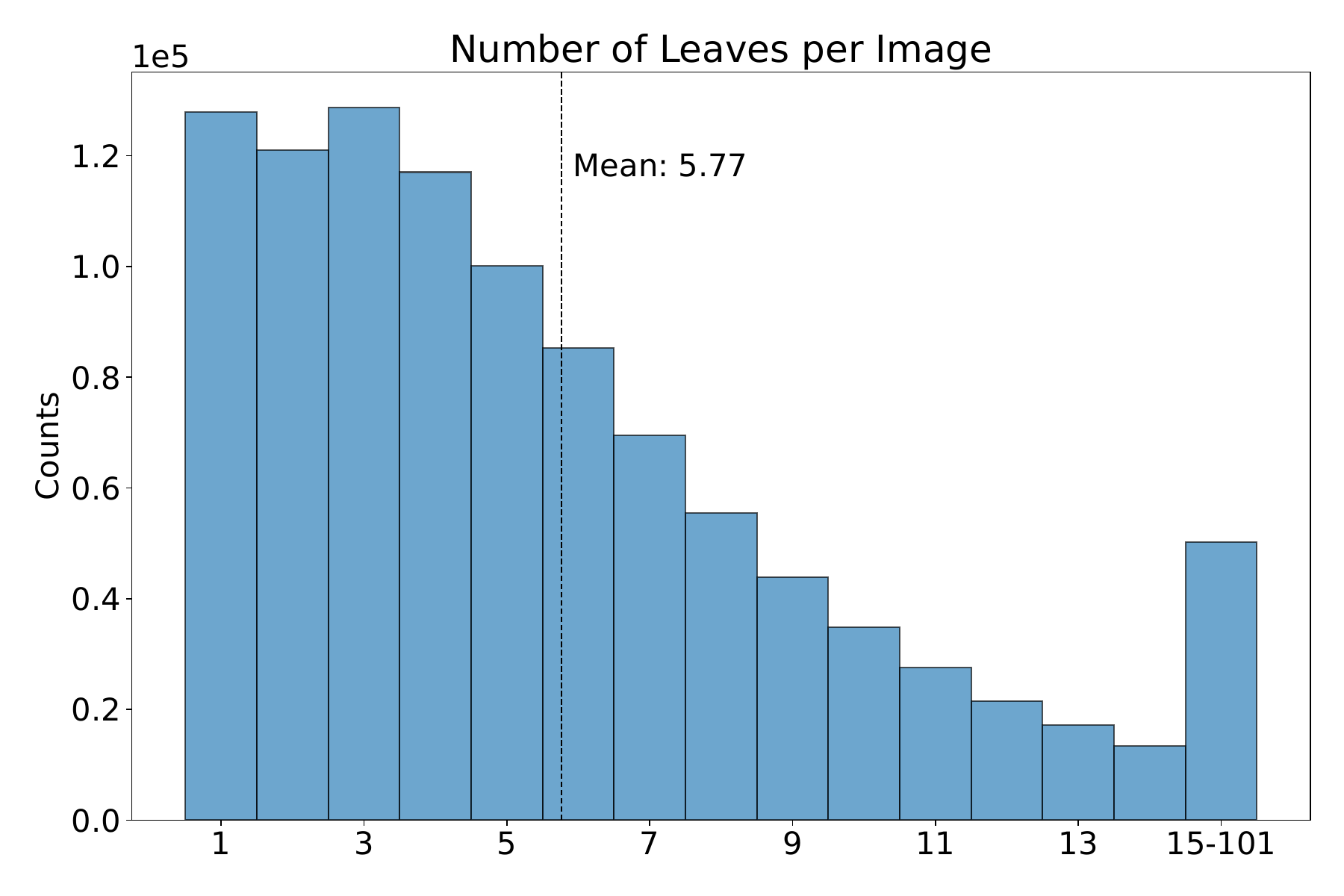}
    \caption{Distributions of vertex numbers across different types of vertices in the GBC1M Dataset.}
    \label{fig:gbc1m-vertex-number-stat}
\end{figure}

\begin{figure}
    \centering
    \includegraphics[width=0.42\textwidth]{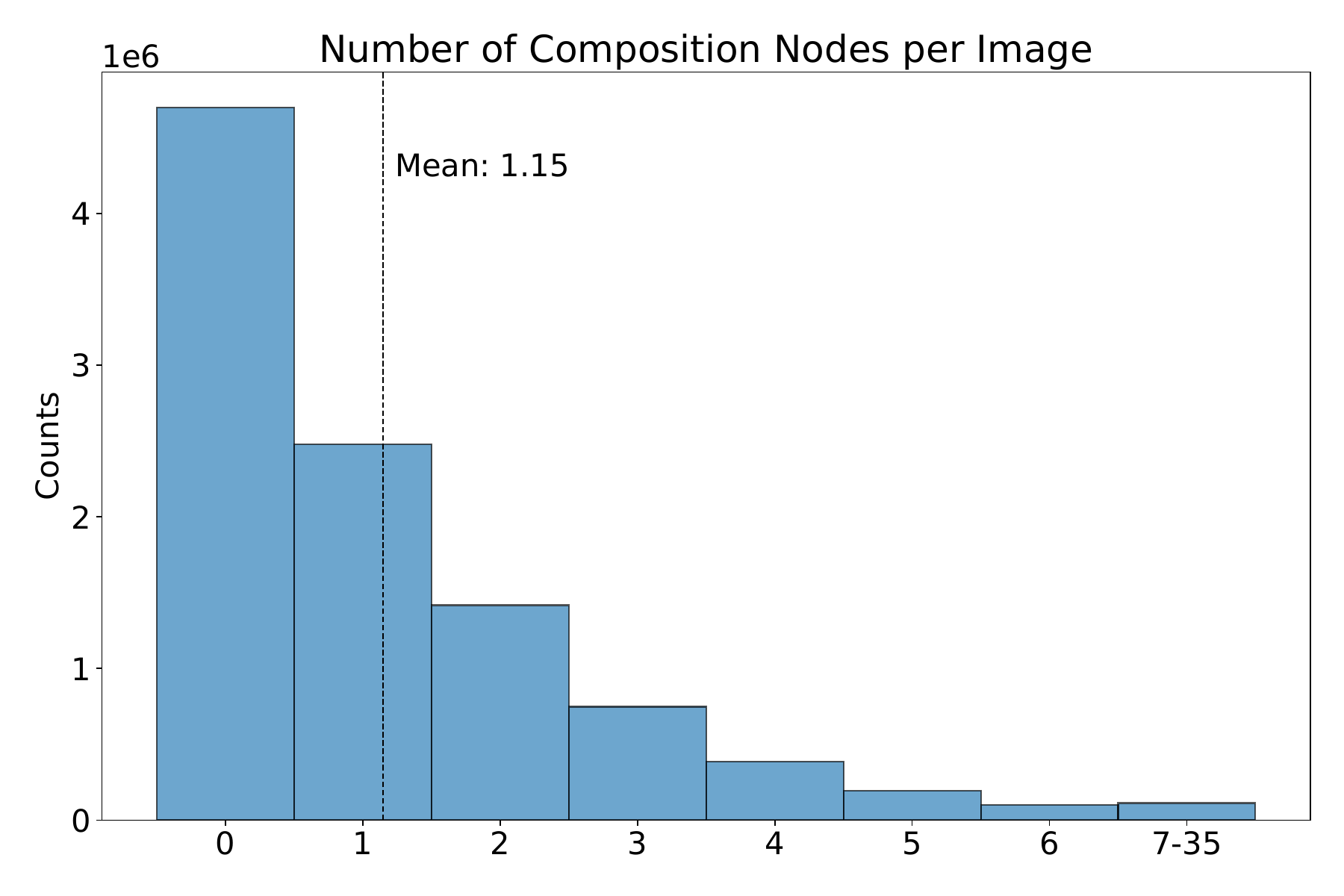}
    \includegraphics[width=0.42\textwidth]{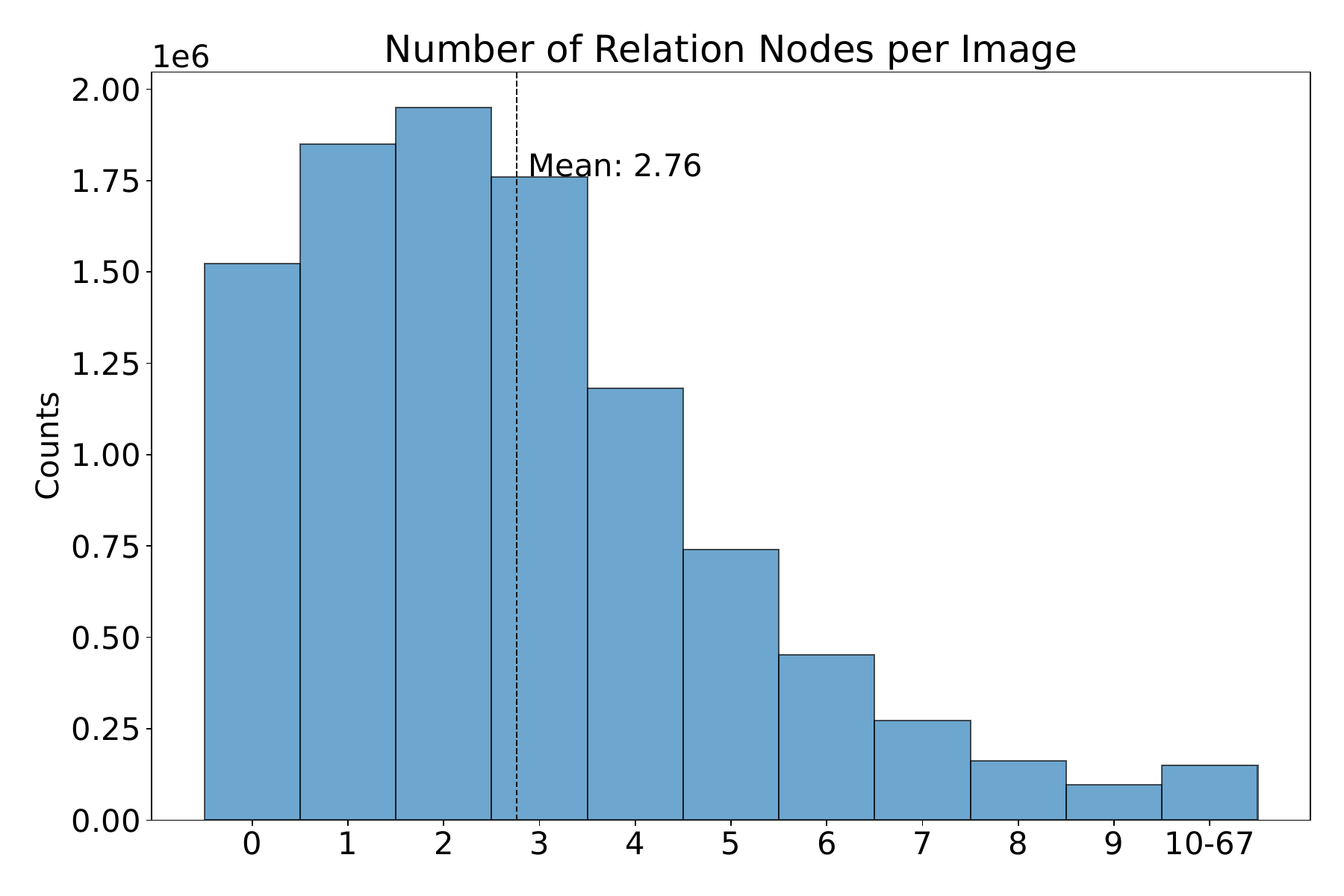}
    \includegraphics[width=0.42\textwidth]{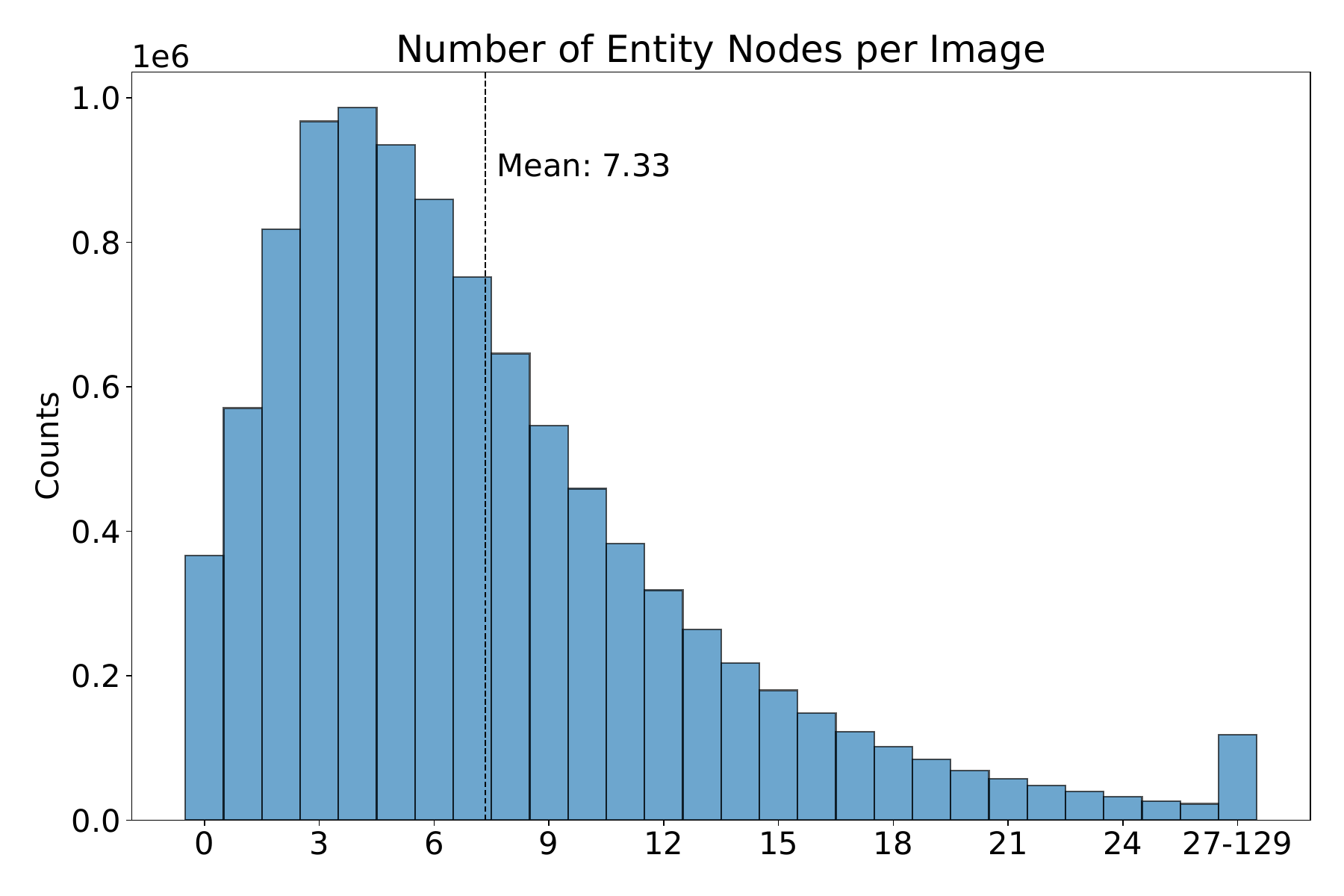}
    \includegraphics[width=0.42\textwidth]{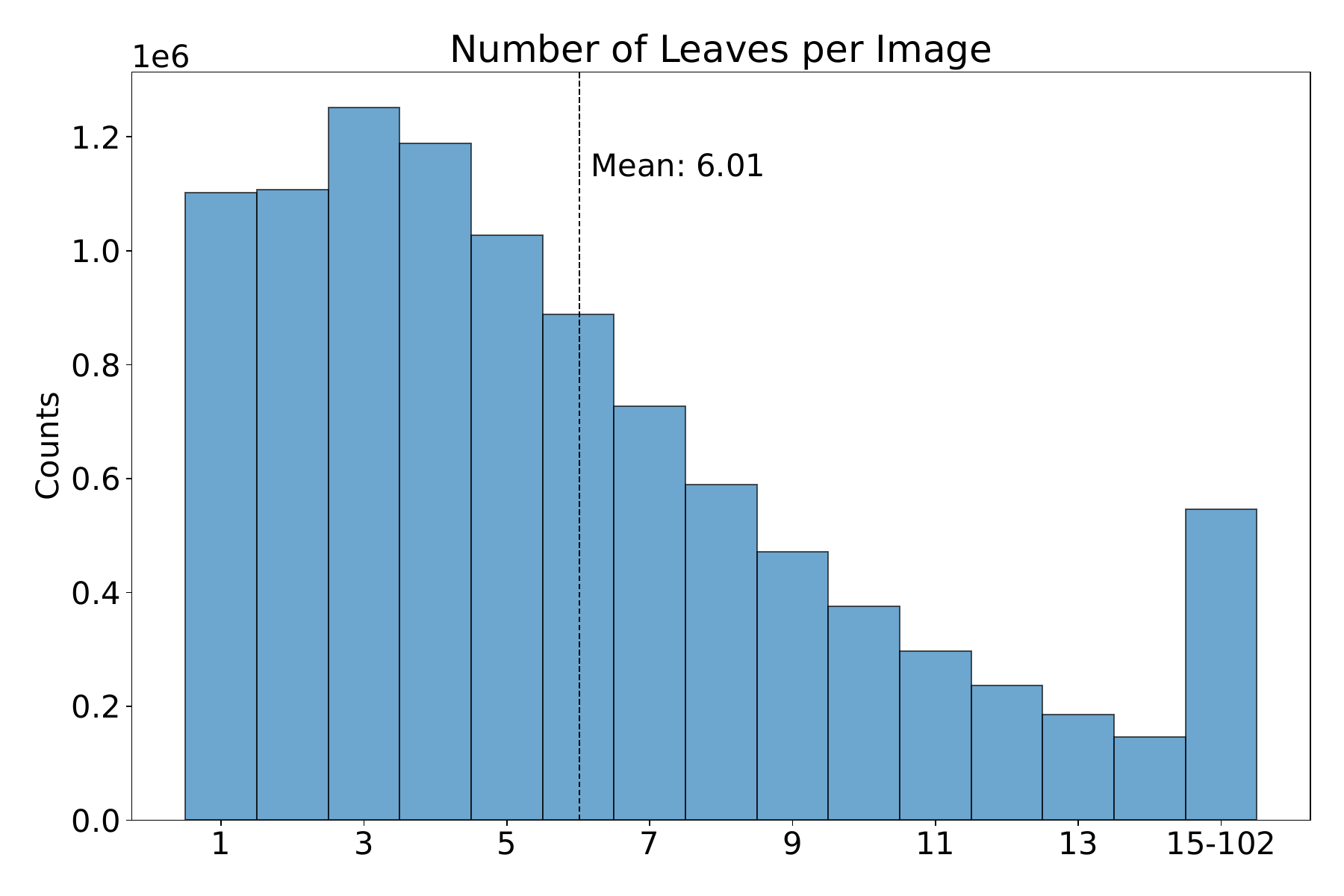}
    \caption{Distributions of vertex numbers across different types of vertices in the GBC10M Dataset.}
    \label{fig:gbc10m-vertex-number-stat}
\end{figure}

We next examine how this number is distributed across the different types of nodes that are present in our graphs.
For this, we plot the distributions of the numbers of composition nodes, relation nodes, entity nodes, and leaves (\ie the nodes without any children) in \cref{fig:gbc1m-vertex-number-stat,fig:gbc10m-vertex-number-stat}. 
As seen in the figures, a large number of vertices are entity nodes, which focus on describing a single object. 
In spite of this, we still have an average number of 4 vertices per graph that are dedicated to describing the composition or relationships between multiple objects.

\begin{figure}[tp]
    \centering
    \includegraphics[width=0.36\textwidth]{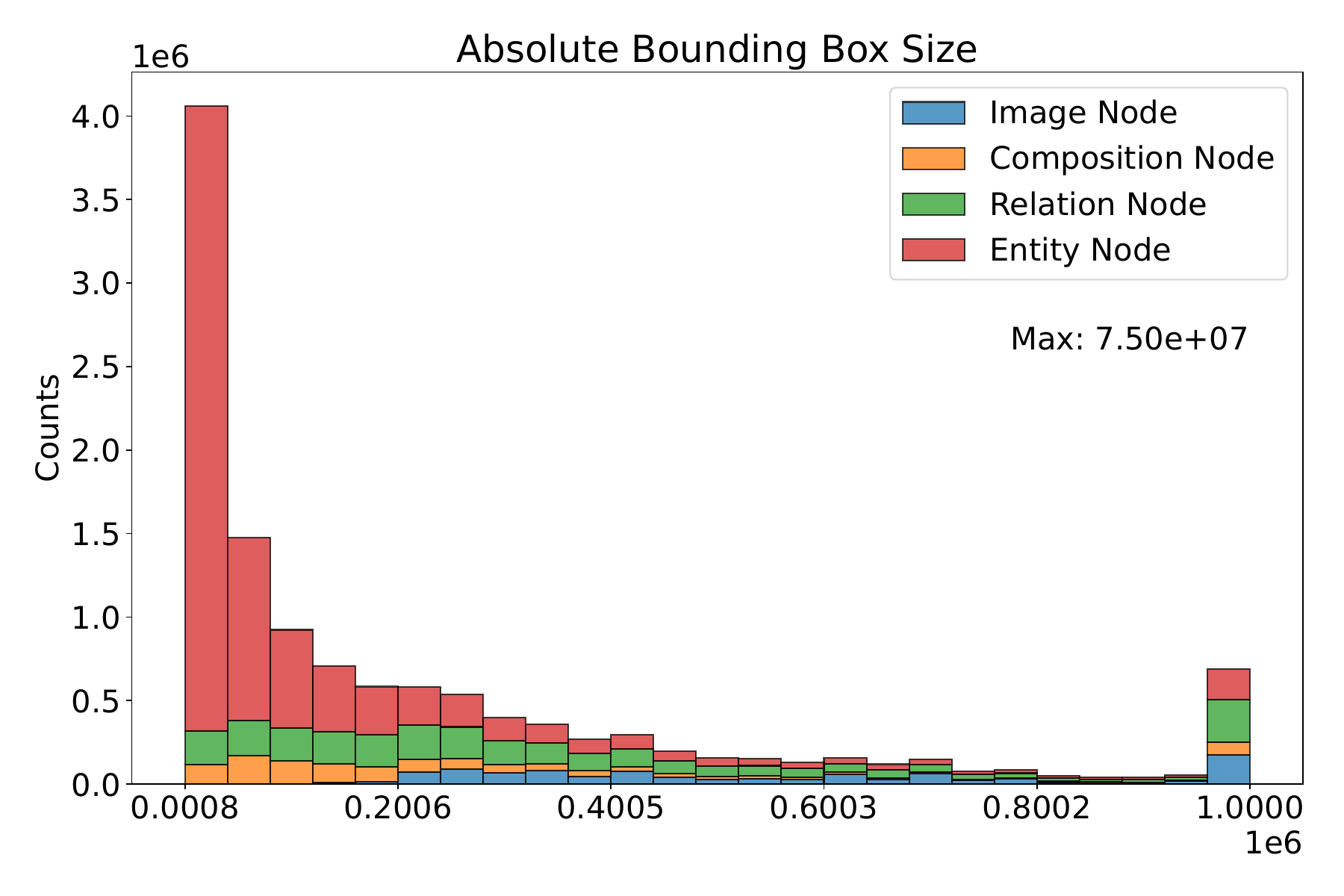}
    \hspace{2em}
    \includegraphics[width=0.36\textwidth]{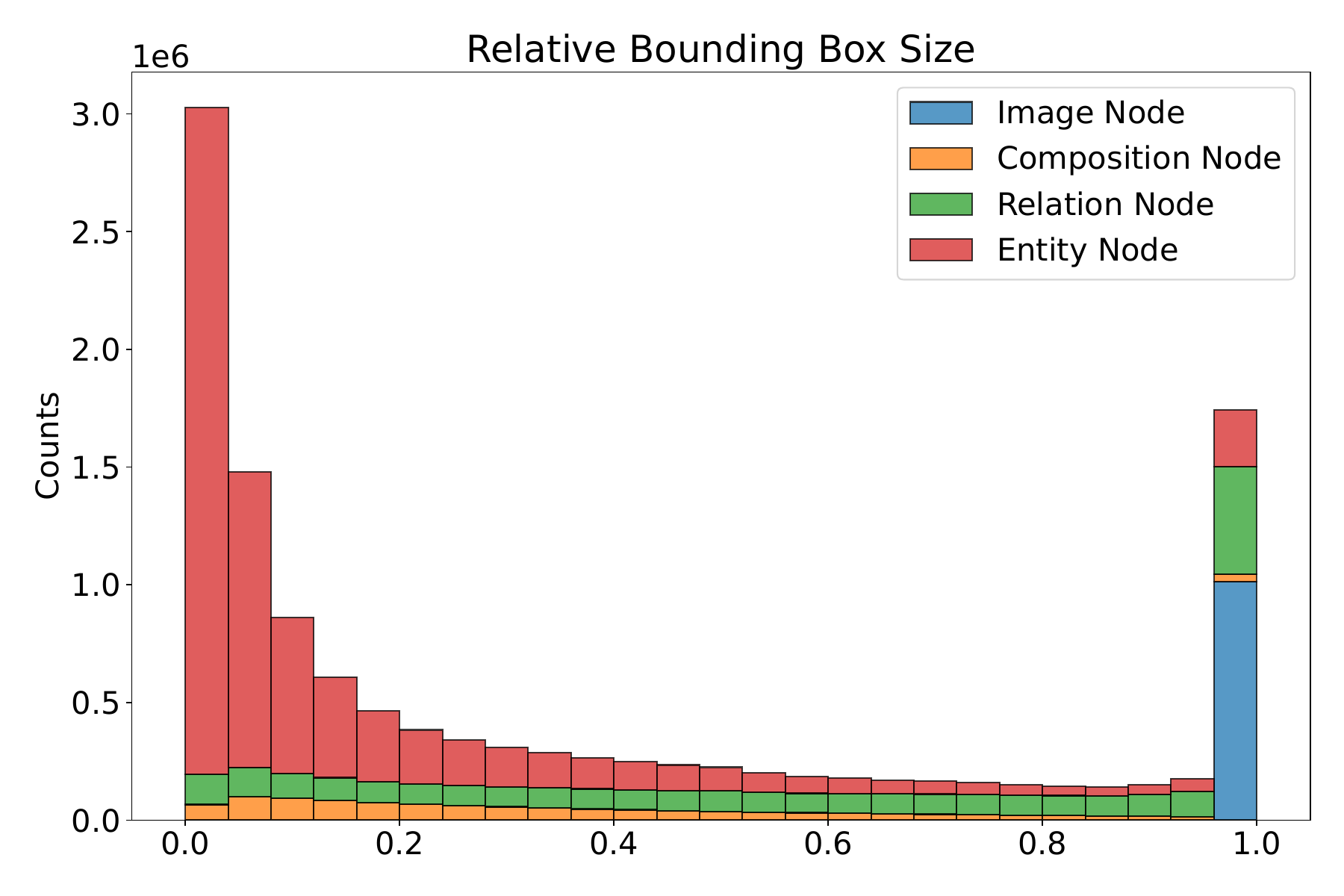}
    \caption{Distribution of bounding box sizes in the GBC1M Dataset. We show both the absolute size (number of pixels) and the relative size (normalized by image size).}
    \label{fig:gbc1m-region-size-stat}
\end{figure}

\begin{figure}
    \centering
    \includegraphics[width=0.36\textwidth]{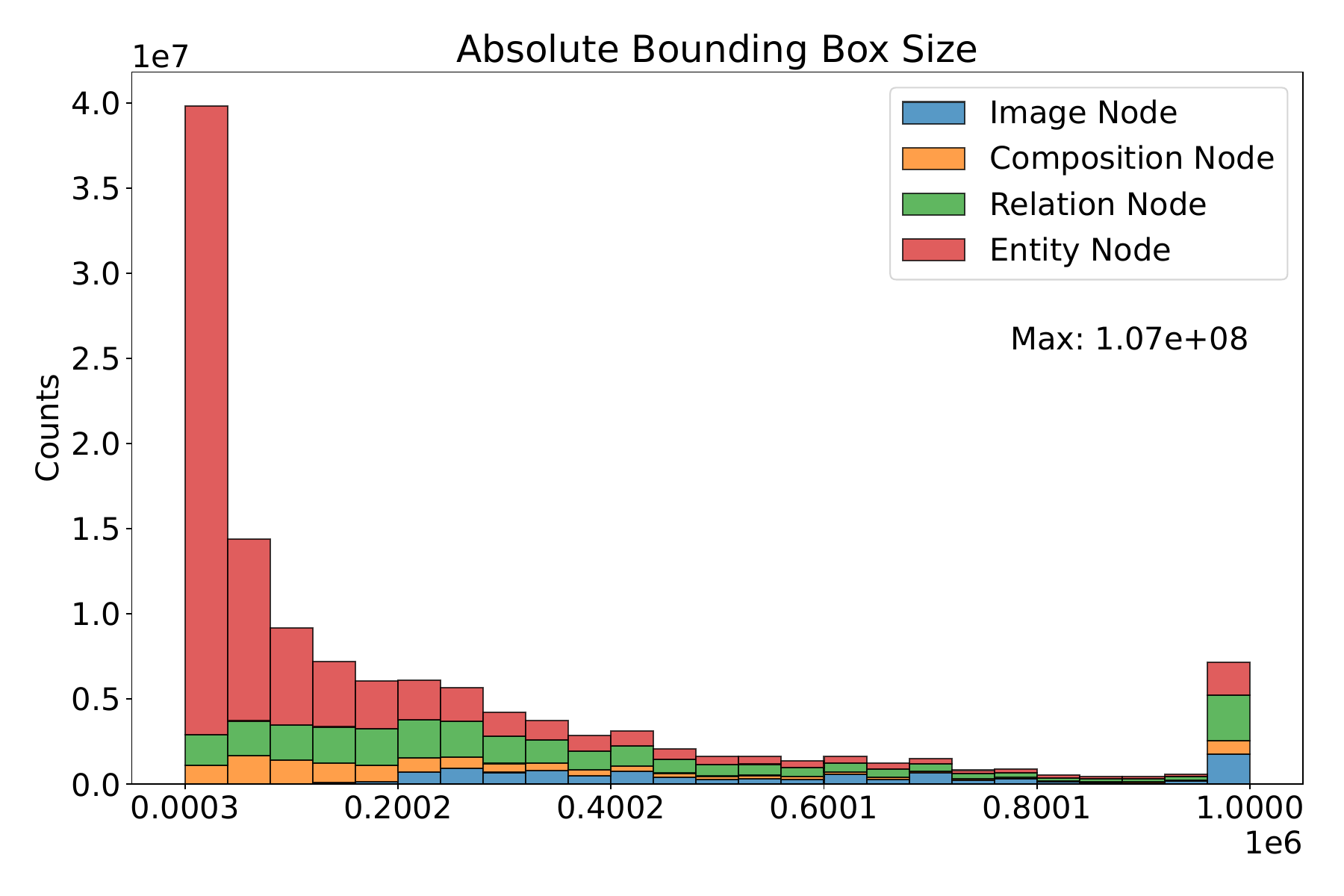}
    \hspace{2em}
    \includegraphics[width=0.36\textwidth]{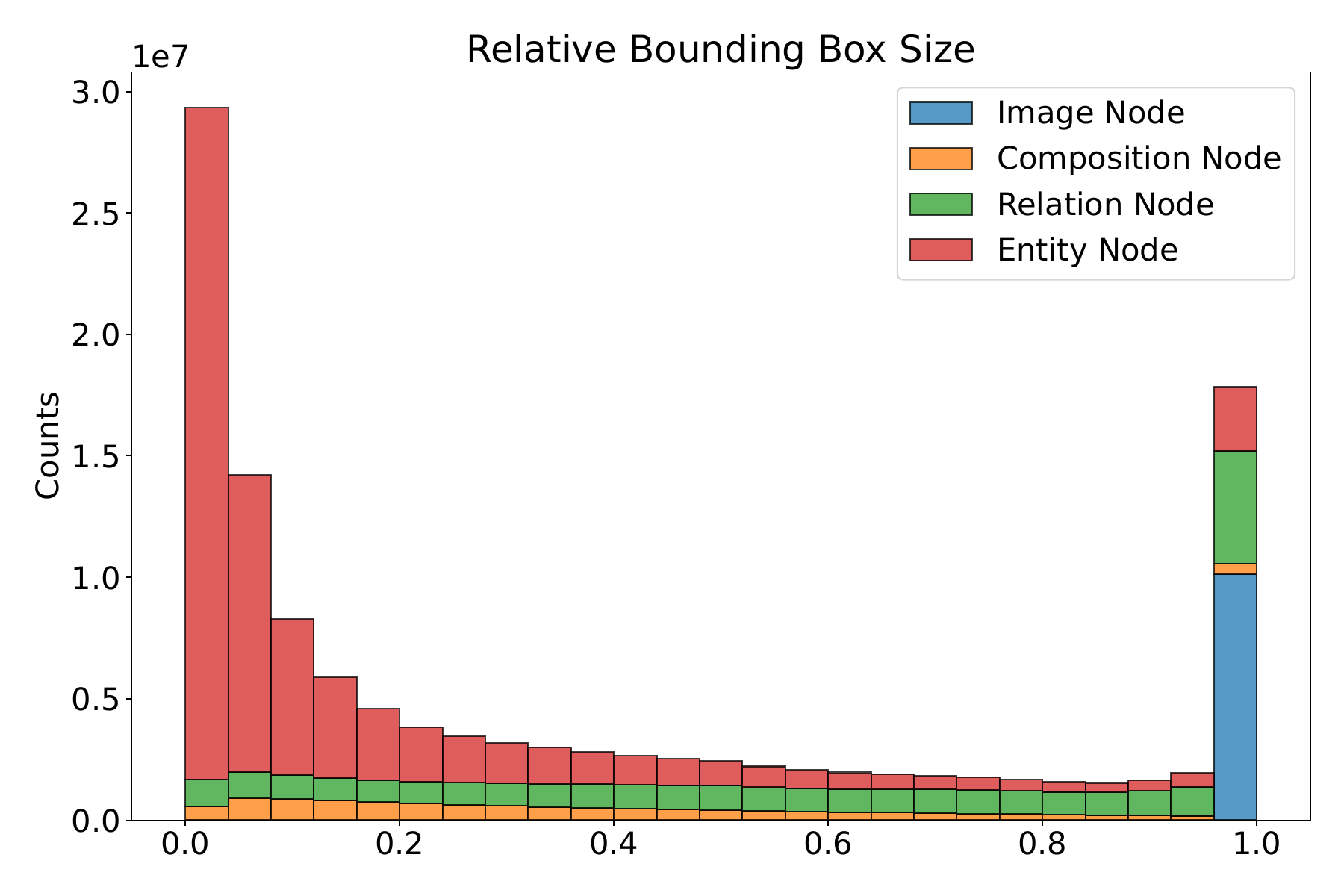}
    \caption{Distribution of bounding box sizes in the GBC10M Dataset. We show both the absolute size (number of pixels) and the relative size (normalized by image size).}
    \label{fig:gbc10m-region-size-stat}
\end{figure}

To complete our investigation, we visualize the distributions of the sizes of the vertices' bounding boxes in \cref{fig:gbc1m-region-size-stat,fig:gbc10m-region-size-stat}.
We note that most of the regions have small relative size (smaller than $0.1$).
This is also observed in other datasets such as Visual Genome~\cite[Fig. 15]{krishna2017visual} and Open Images~\cite[Fig. 20]{kuznetsova2020open}.
Relation nodes, whose bounding boxes are defined as the minimum bounding box containing the union of all the involved objects' bounding boxes, have sizes that spread more uniformly across different ratios.
We also observe a large number of entity nodes with bounding boxes that have a relative size close to 1.
This likely corresponds to background objects that spans across the entire image, such as ``sky'' or ``grass''.

\subsubsection{Edge statistics}

\begin{figure}[tp]
    \centering
    \begin{subfigure}{0.36\textwidth}
        \centering
        \includegraphics[width=\textwidth]{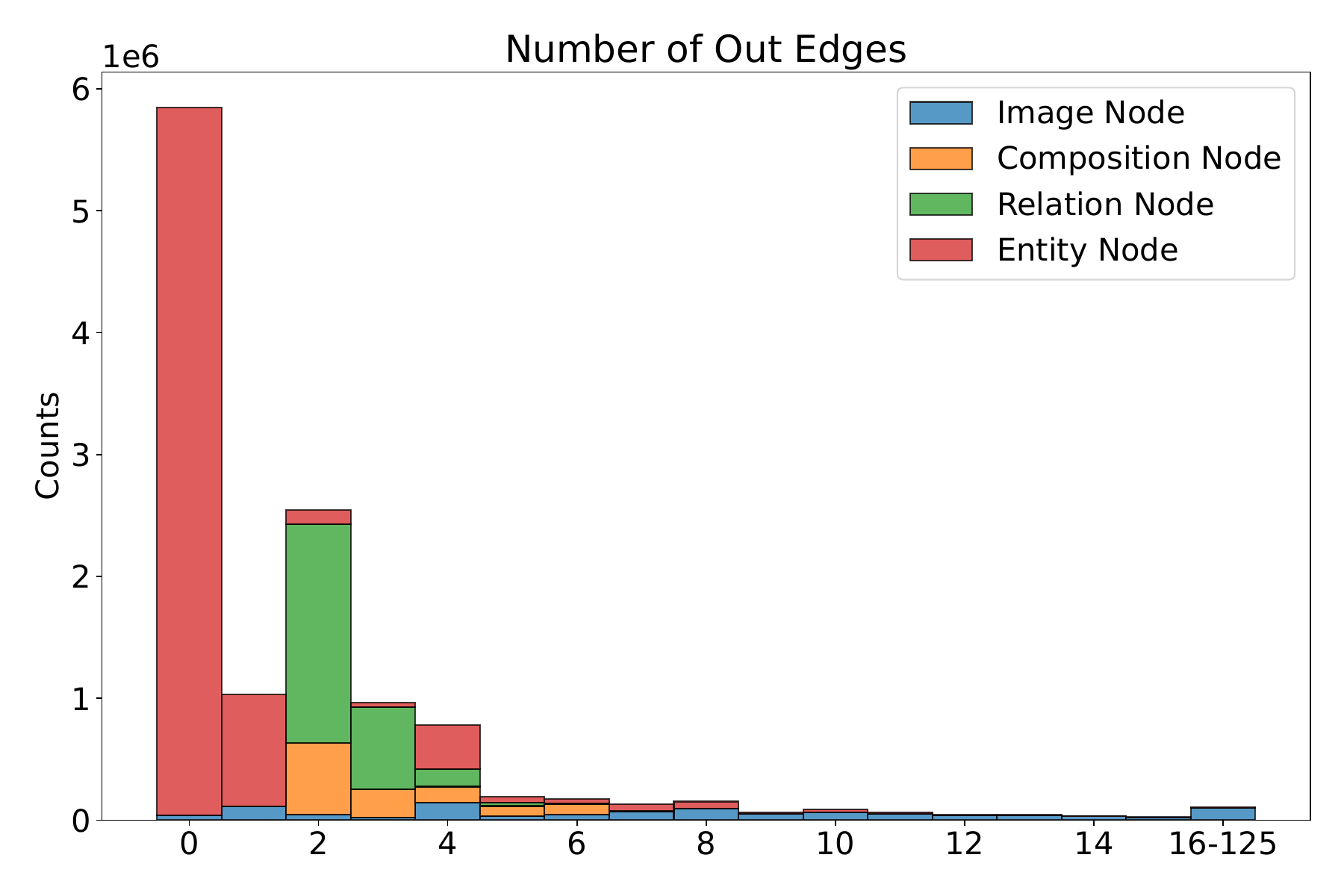}
        \caption{GBC1M}
    \end{subfigure}
    \hspace{2em}
    \begin{subfigure}{0.36\textwidth}
        \centering
        \includegraphics[width=\textwidth]{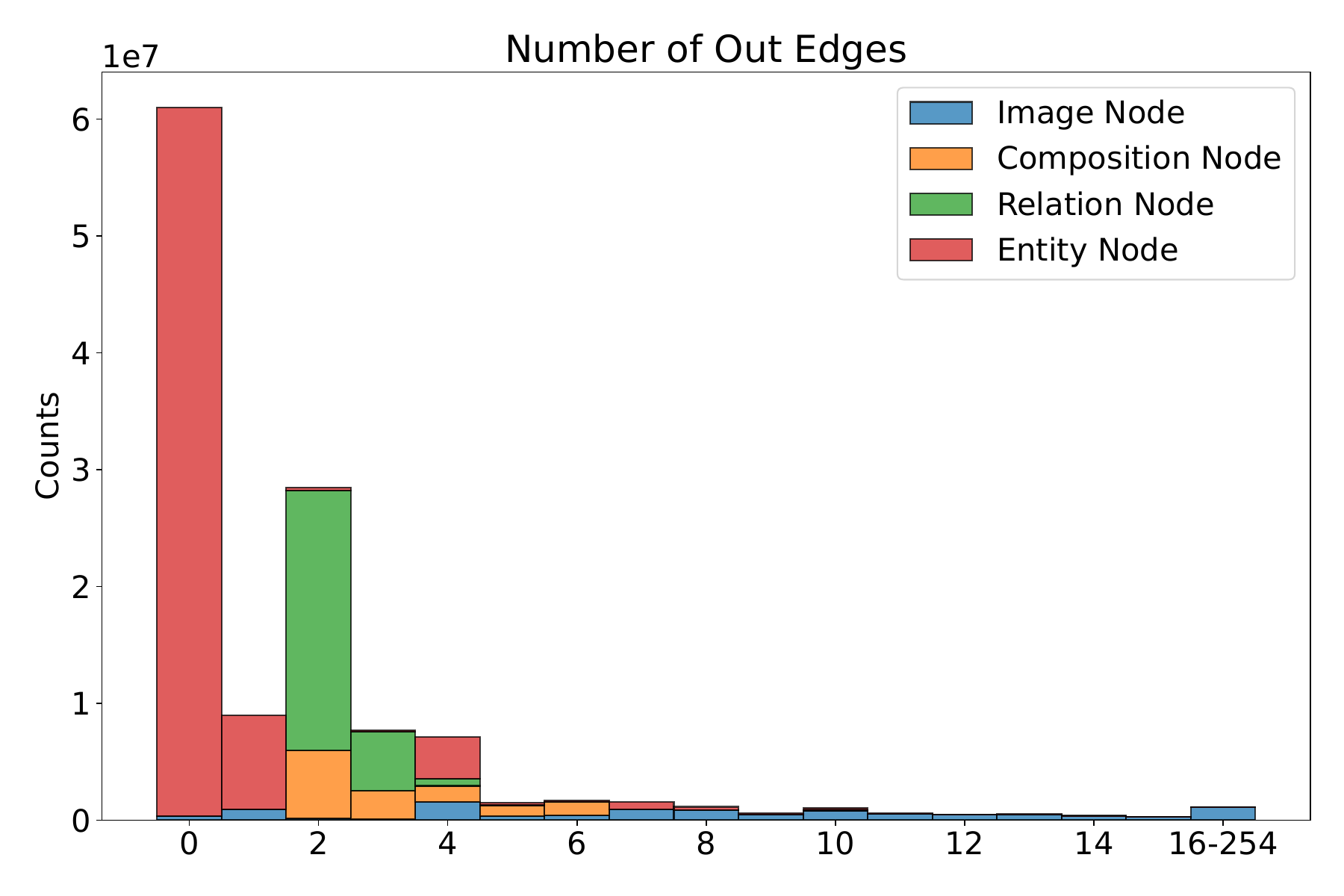}
        \caption{GBC10M}
    \end{subfigure}
    \caption{Distributions of number of outgoing edges across different types of vertices in the GBC1M (left) and GBC10M (right) datasets.}
    \label{fig:gbc1m-10m-out-edge-number-stat}
\end{figure}

Our datasets feature an average of 22 edges per graph.
We analyze the origins of these edges in \cref{fig:gbc1m-10m-out-edge-number-stat}, which shows their distributions across different types of source vertices.
The figure indicates that the image node is responsible for a large proportion of these edges, suggesting that many of the entities that we identify directly come from the image caption.
This is natural provided that an image often contains many objects, while it is less common to need further decomposition of a single object for detailed description.
Besides this, these figures also indicate the number of entities that are involved in our composition and relation descriptions.
Notably, we see that most of these descriptions only contain 2 or 3 objects, with few of them involving more than 4 objects.
In contrast, we observe a relatively large number of entity nodes with 4 outgoing edges, and we believe this can be attributed to the bias caused by the few-shot examples provided in our query template.



\begin{figure}
    \centering
    \begin{subfigure}{0.36\textwidth}
        \centering
        \includegraphics[width=\textwidth]{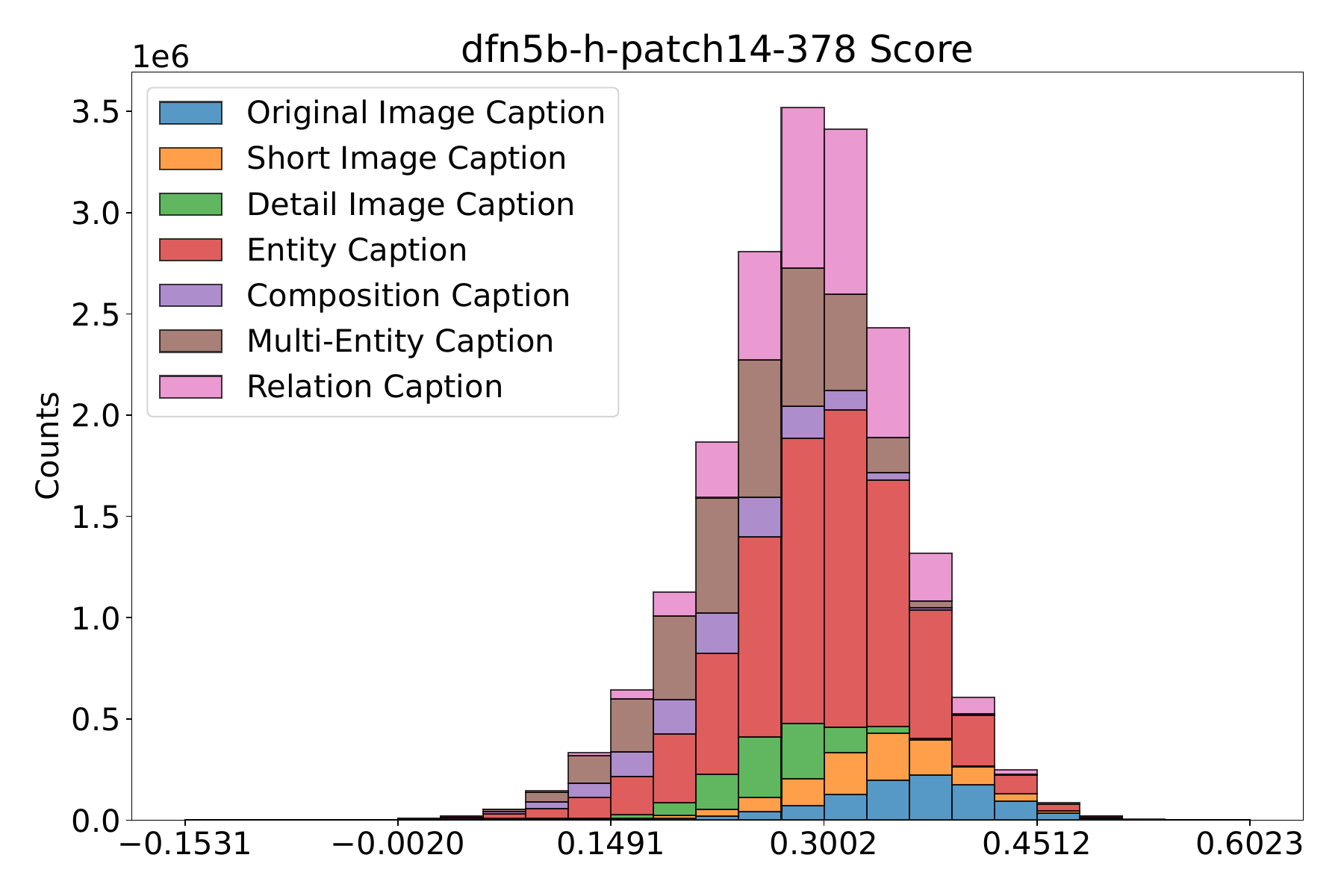}
        \caption{GBC1M}
    \end{subfigure}
    \hspace{2em}
    \begin{subfigure}{0.36\textwidth}
        \centering
        \includegraphics[width=\textwidth]{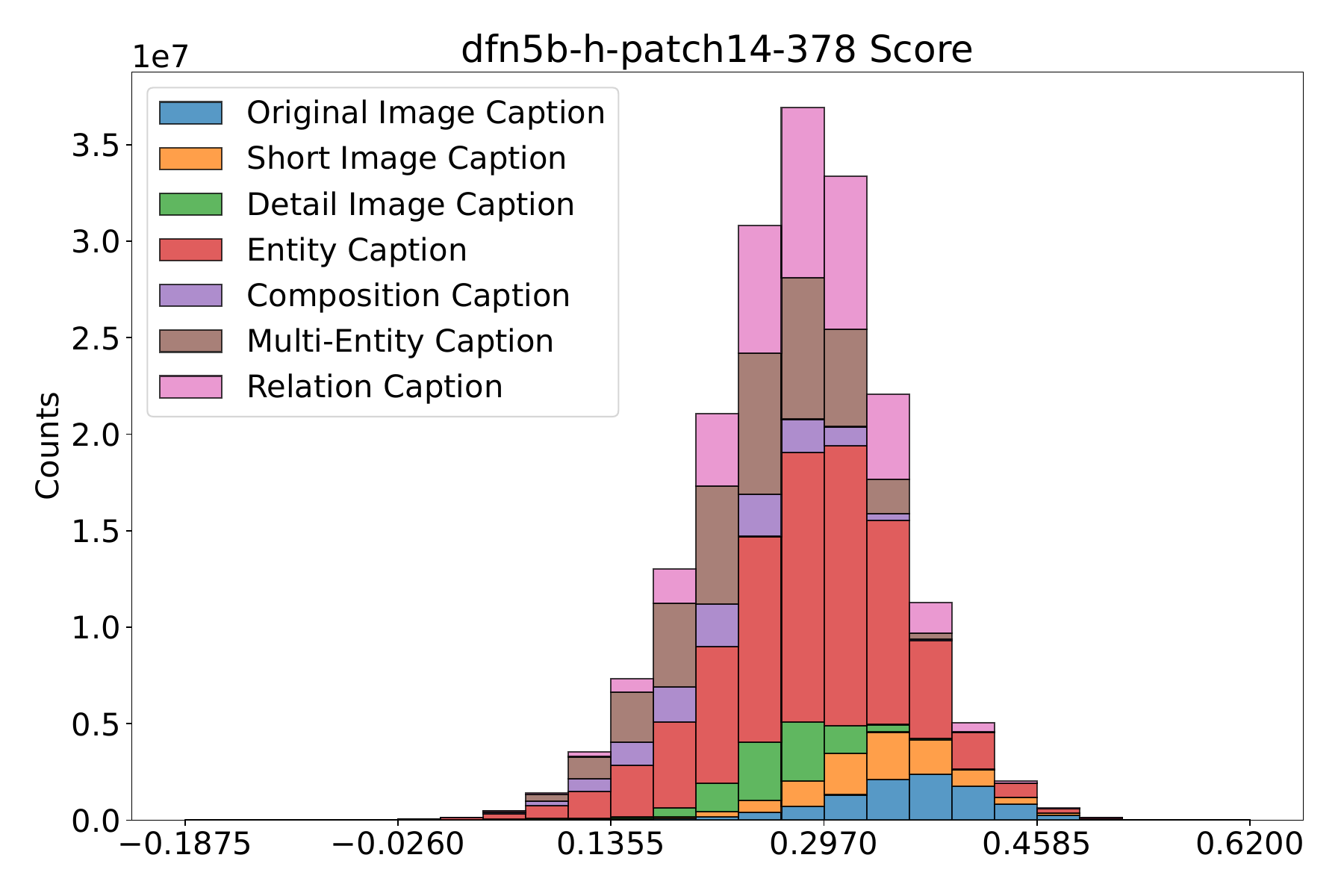}
        \caption{GBC10M}
    \end{subfigure}
    \caption{Distributions of DFN-5B CLIP scores across different types of captions in the GBC1M (left) and GBC10M (right) datasets.}
    \label{fig:gbc1m-10m-clip-score}
\end{figure}

\clearpage

We also provide analysis for the edge labels.
These edge labels should represent the objects that are associated to their respective target vertices.
In particular, during our annotation process, we use these labels as input of the detection model to obtain the bounding boxes of the entity nodes.
In \cref{fig:gbc1m-edge-label-length,fig:gbc10m-edge-label-length}, we plot the distributions of the numbers of words and tokens contained in the edge labels.
As expected, most of the time we use only 1 or 2 words to represent the entities.

\begin{figure}[p]
    \centering
    \includegraphics[width=0.42\textwidth]{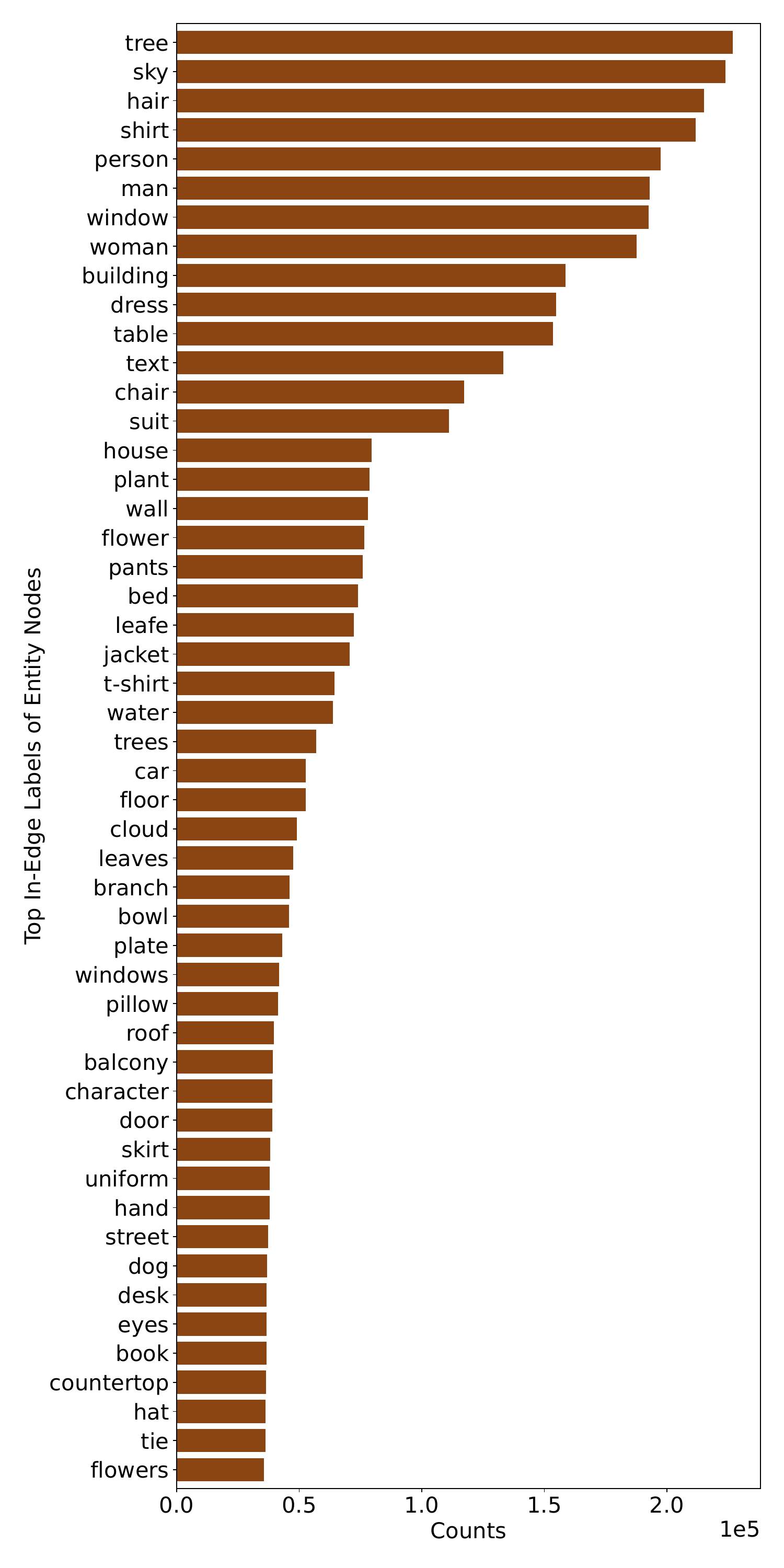}
    \includegraphics[width=0.42\textwidth]{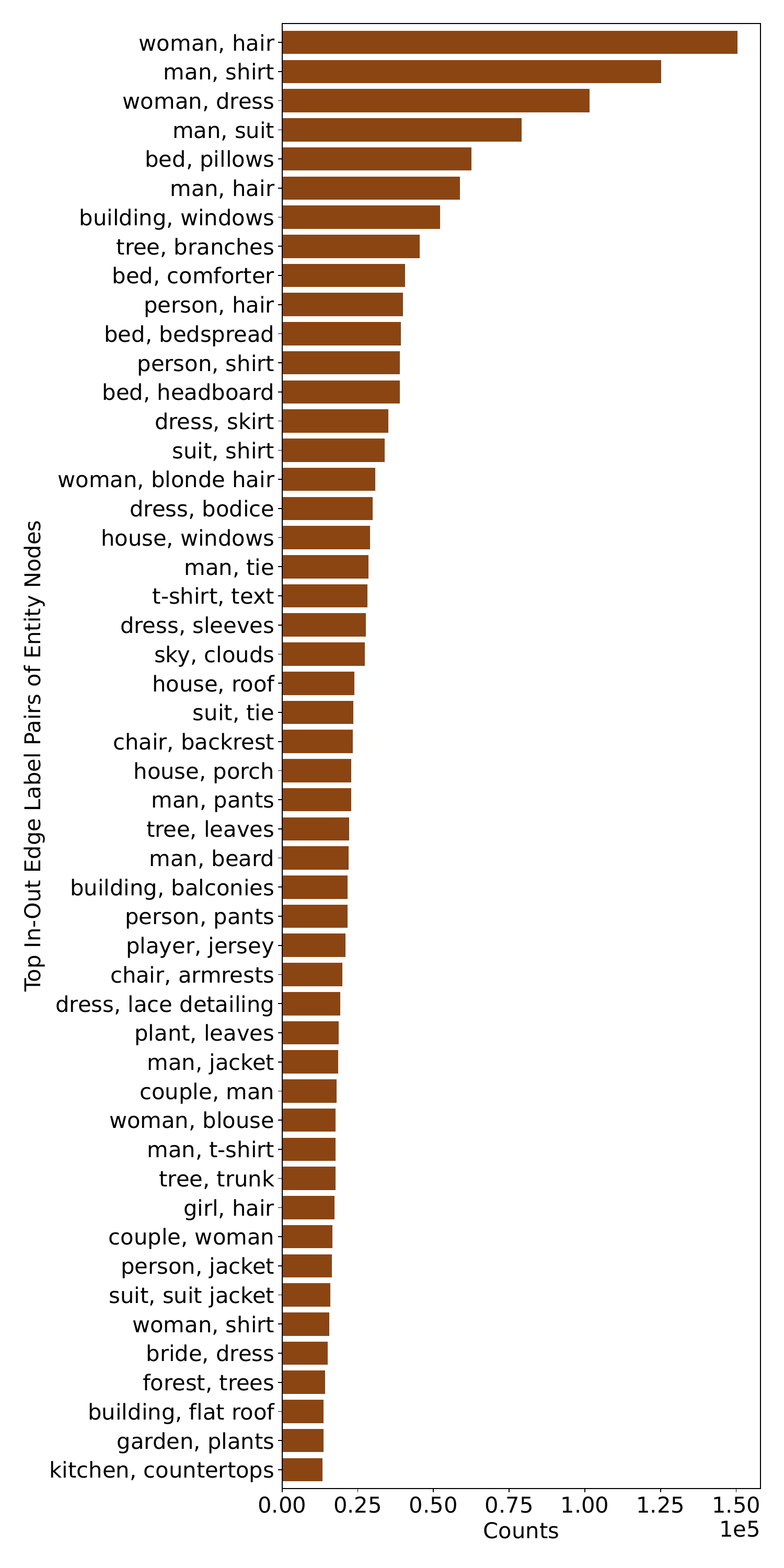}
    \caption{Distributions of the 20 most common in-edge labels and in-/out-edge label pairs at entity nodes in the GBC1M dataset. We remove numbers from the edge labels for the computation of their occurrences in these plots.}
    \label{fig:gbc1m-edge-label-content}
\end{figure}

\begin{figure}
    \centering
    \includegraphics[width=0.42\textwidth]{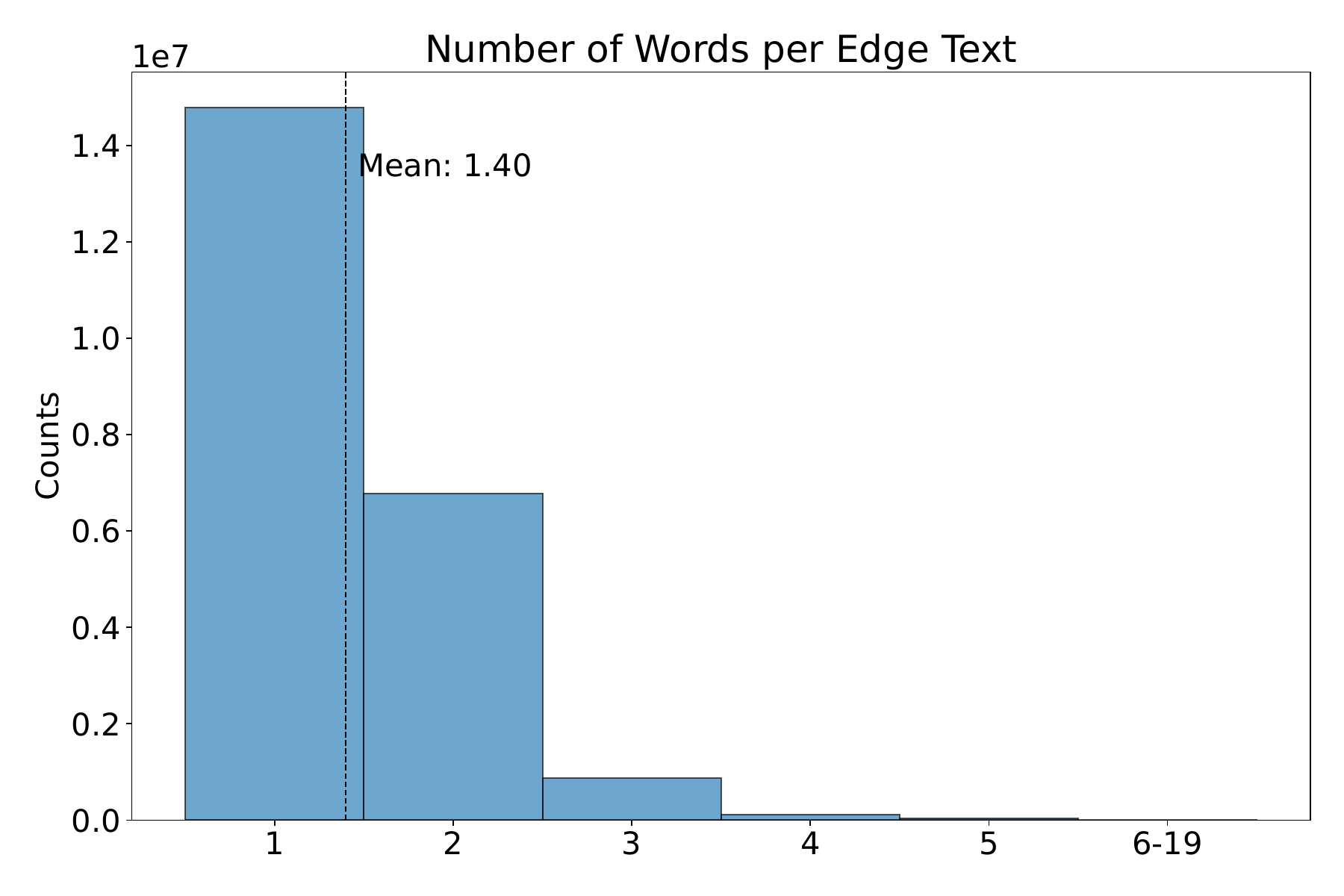}
    \includegraphics[width=0.42\textwidth]{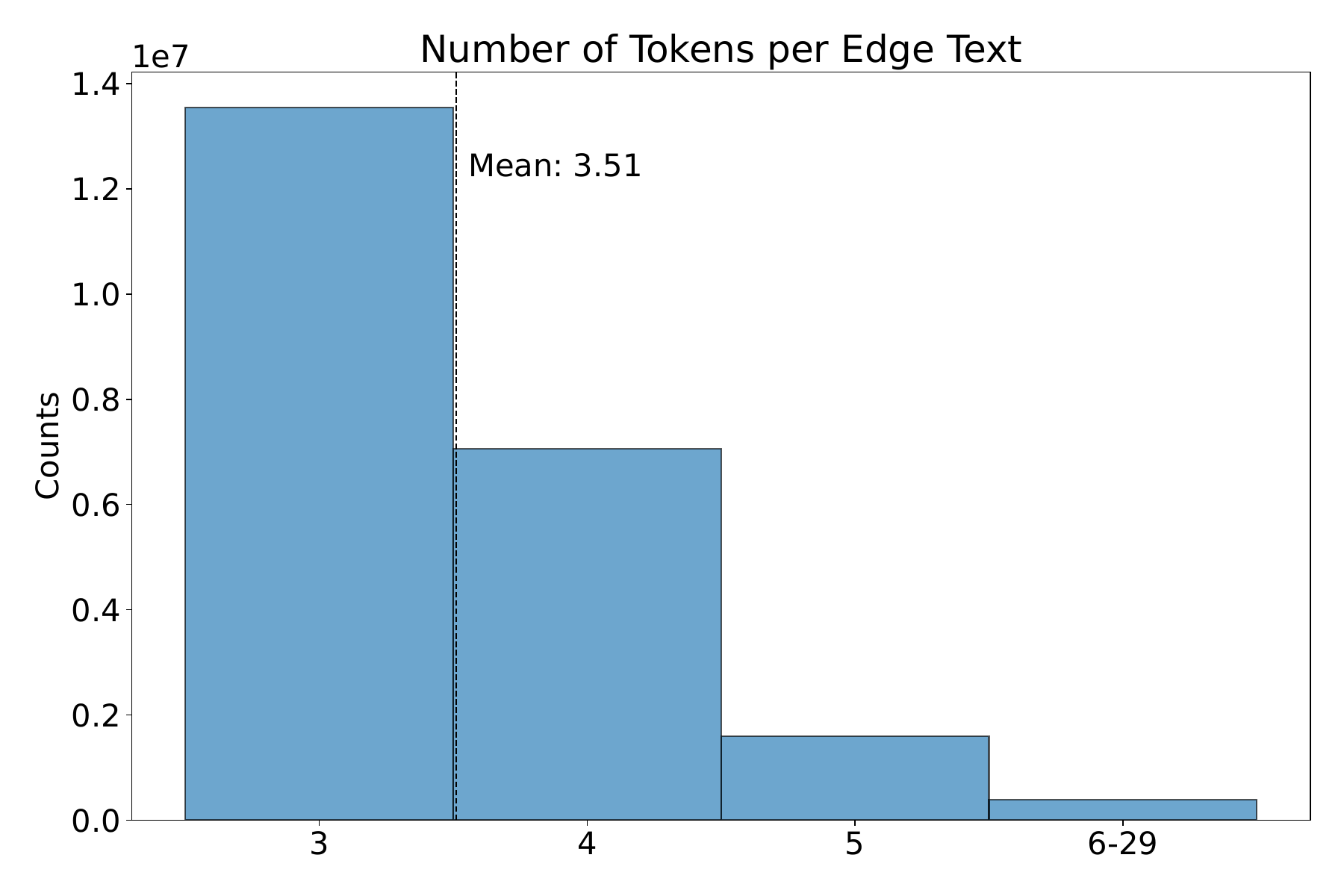}
    \caption{Distributions of numbers of words/tokens in each edge label in the GBC1M dataset.
    To compute the number of tokens we use the standard CLIP tokenizer.}
    \label{fig:gbc1m-edge-label-length}
\end{figure}

\begin{figure}[p]
    \centering
    \includegraphics[width=0.42\textwidth]{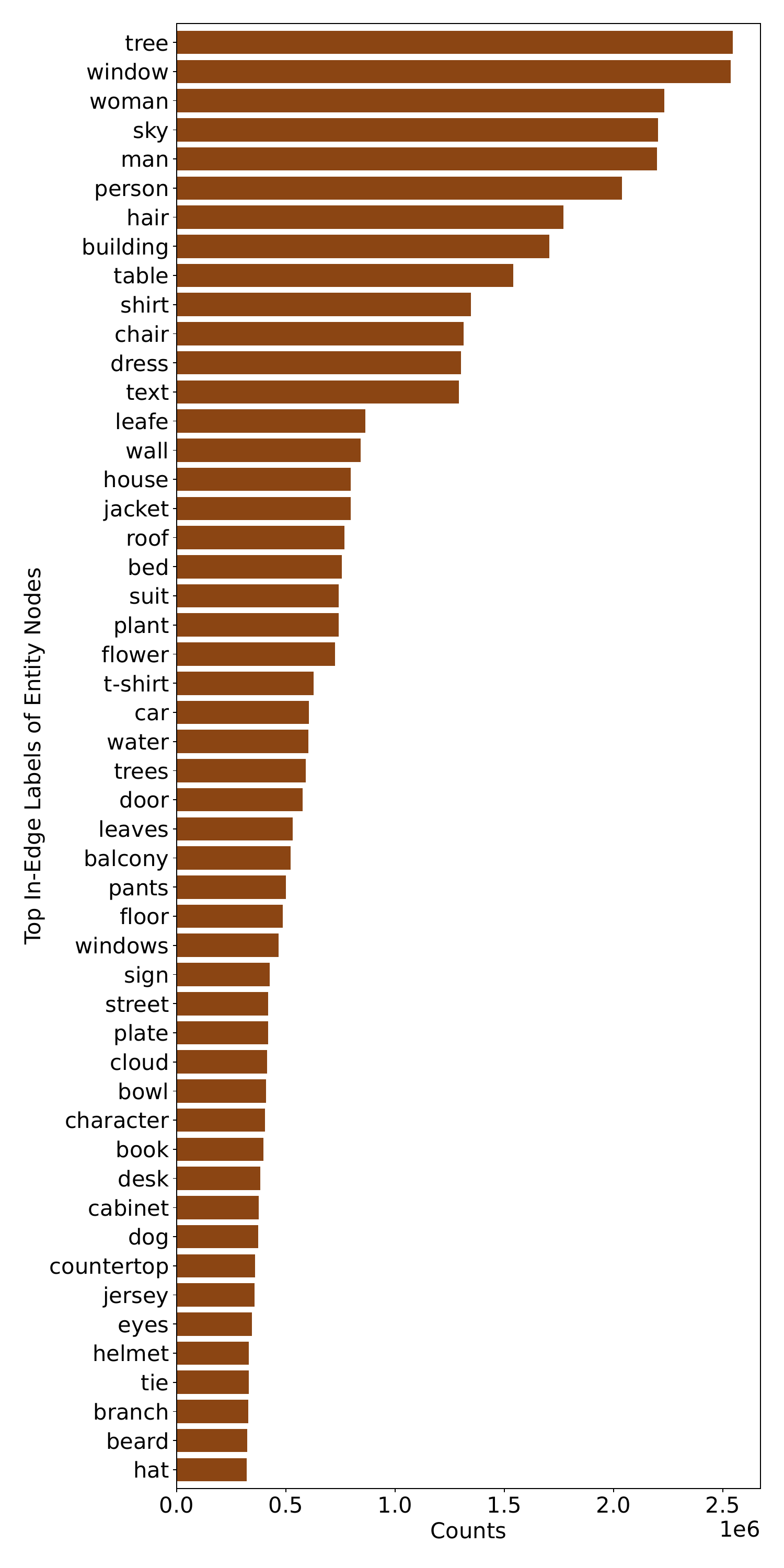}
    \includegraphics[width=0.42\textwidth]{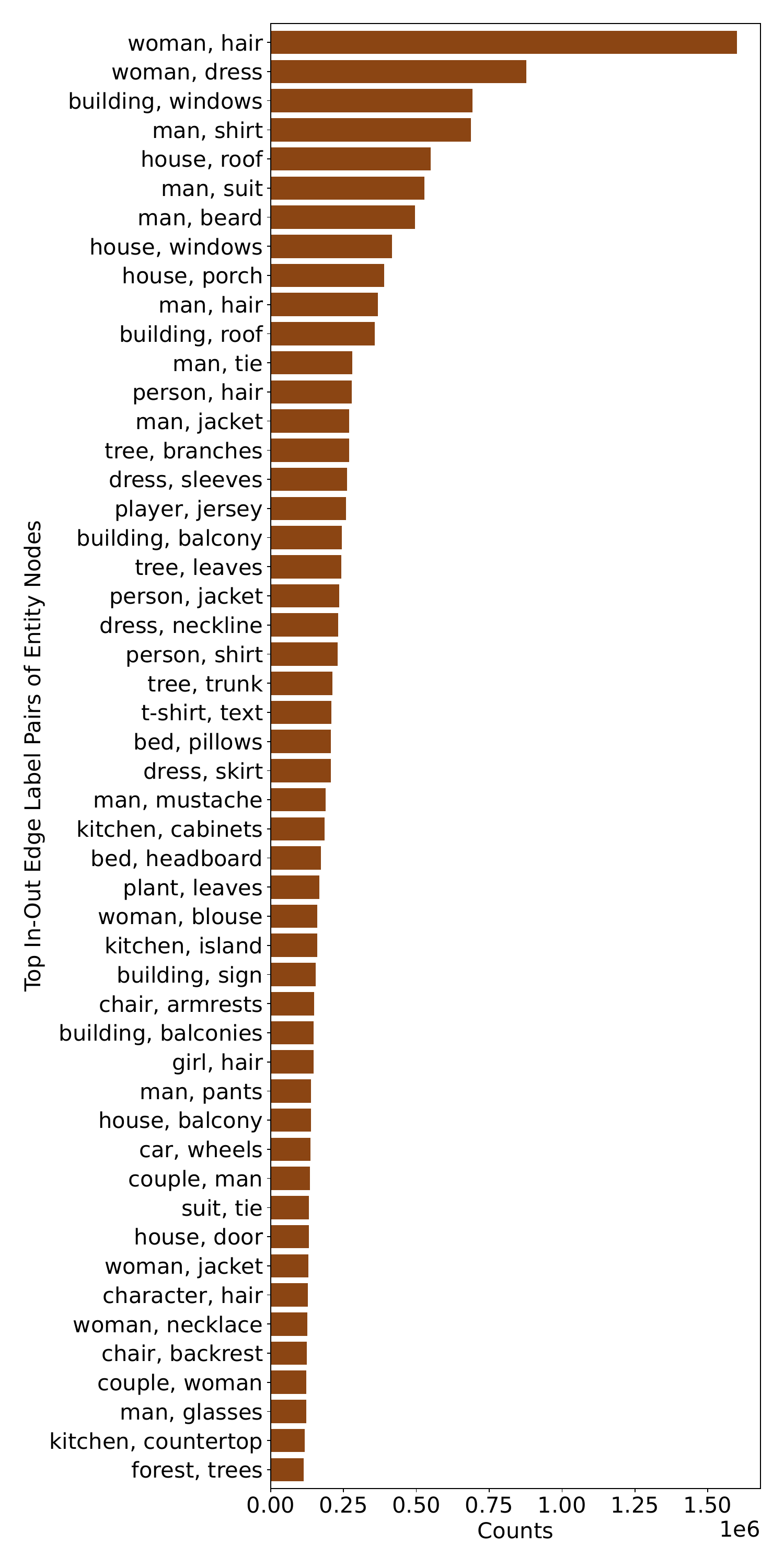}
    \caption{Distributions of the 20 most common in-edge labels and in-/out-edge label pairs at entity nodes in the GBC10M dataset. We remove numbers from the edge labels for the computation of their occurrences in these plots.}
    \label{fig:gbc10m-edge-label-content}
\end{figure}

\begin{figure}
    \centering
    \includegraphics[width=0.42\textwidth]{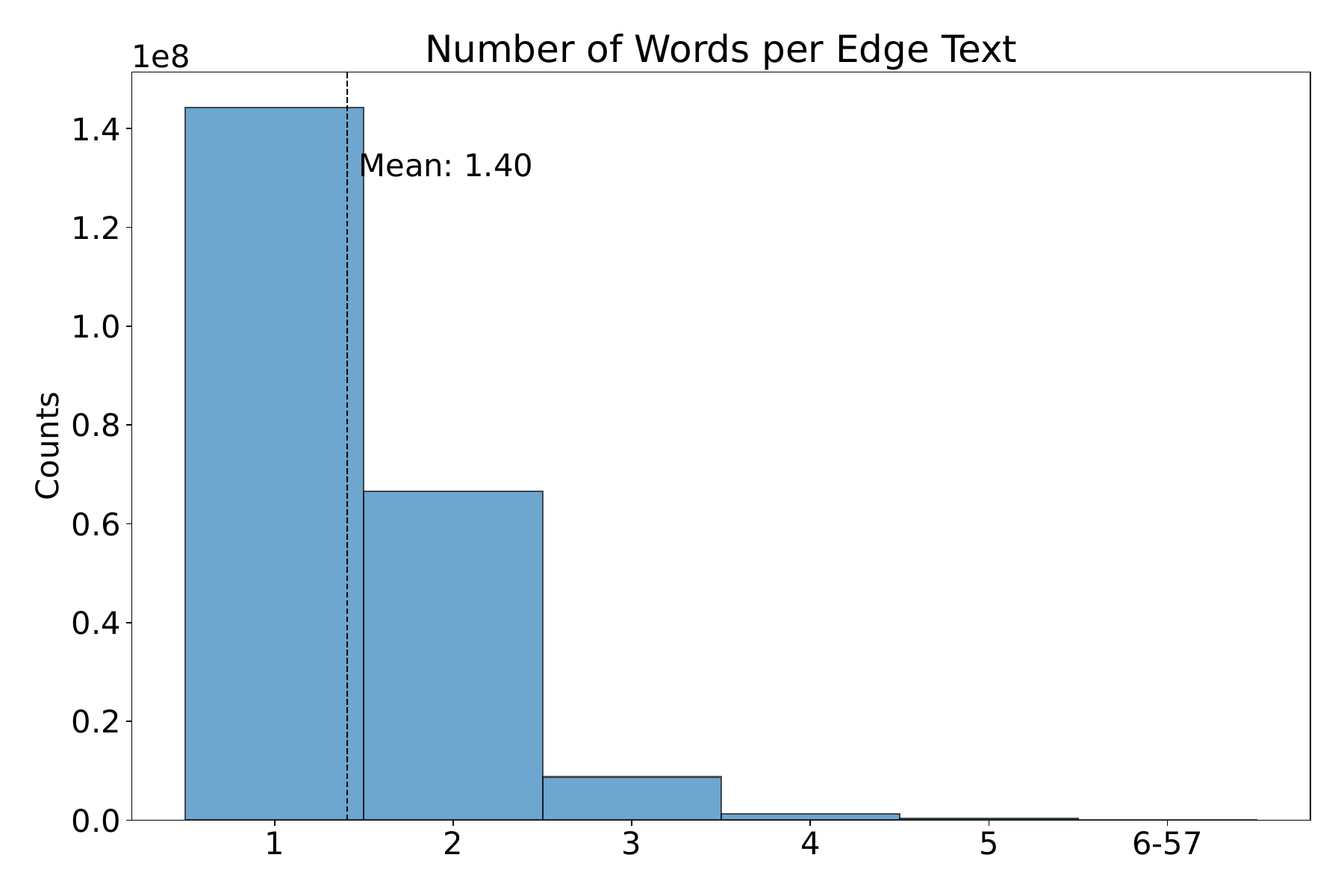}
    \includegraphics[width=0.42\textwidth]{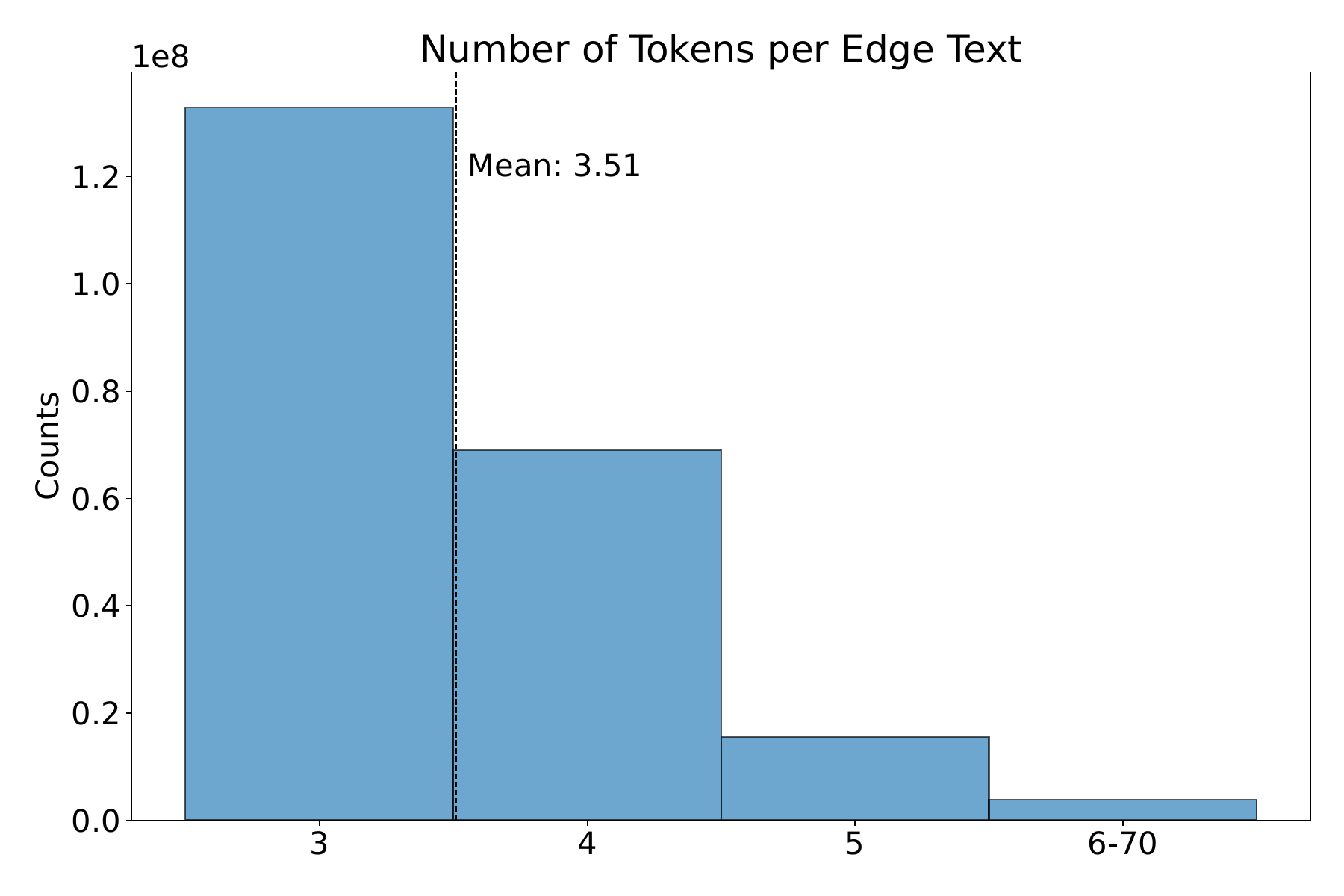}
    \caption{Distributions of numbers of words/tokens in each edge label in the GBC1M dataset.
    To compute the number of tokens we use the standard CLIP tokenizer.}
    \label{fig:gbc10m-edge-label-length}
\end{figure}

We next study the content of these labels. To this end, we plot the distribution of
\emph{(i)}
the 20 most common edge labels at the in-edges of the entity nodes, reflecting the content of these entity nodes, and
\emph{(ii)}
the 20 most common edge label pairs when pairing the in- and out-edges of the entity node, reflecting the situation where we zoom in on an object to further describe a part of it.
The corresponding histograms are presented in \cref{fig:gbc1m-edge-label-content,fig:gbc10m-edge-label-content}.
From these plots, we see that the most common objects from our datasets are ``tree'', ``sky'', ``man'', ``woman'', ``table'', and ``building'', among others.
This distribution aligns well with the ones reported for existing datasets, \cf \cite[Fig. 22]{krishna2017visual} and \cite[Tab. 11]{kuznetsova2020open}. 
Furthermore, while the occurrence of certain labels and label pairs, such as (``woman'', ``hair''), may be influenced by our system prompts, others like (``bed'', ``pillows'') are widely present despite not being included in our prompts.
This suggests potential biases in either the model or the dataset itself.



\begin{table}[t]
    \centering
    \renewcommand{\arraystretch}{1.25}
    \setlength\tabcolsep{0.3em}
    \begin{tabular}{@{\hskip 0.6em}l@{\hskip 1em}l@{\hskip 1.2em}|@{\hskip 0.6em}cccc@{\hskip 0.6em}}
    \toprule
        &
        Caption Type
        &
        \# Captions
        &
        \# Words / Caption
        &
        \# Tokens / Caption
        &
        CLIP score\\
    \midrule
    \multirow{7}{*}{\vspace{-1em}GBC1M}
    & Image Original & \multirow{3}{*}{1,013,592}
    & 17.4
    & 24.5
    & 0.36
    \\
    & Image Short &
    & 28.1
    & 35.3
    & 0.33
    \\
    & Image Detail &
    & 110.3
    & 130.9
    & 0.26
    \\
    \cmidrule{2-6}
    & Entity & 7,512,638
    & 37.5
    & 46.3
    & 0.29
    \\
    \cmidrule{2-6}
    & Composition & 1,117,935
    & 35.8 
    & 44.1
    & 0.23
    \\
    & Multi-Entity & 3,487,562
    & 17.8
    & 23.1
    & 0.25
    \\
    \cmidrule{2-6}
    & Relation & 3,493,543
    & 22.0 
    & 27.2
    & 0.30
    \\
    \midrule
    \multirow{7}{*}{\vspace{-1em}GBC10M}
    & Image Original & \multirow{3}{*}{10,138,757}
    & 17.4
    & 24.6
    & 0.36
    \\
    & Image Short &
    & 28.1
    & 35.3
    & 0.33
    \\
    & Image Detail &
    & 110.3
    & 130.9
    & 0.26
    \\
    \cmidrule{2-6}
    & Entity & 74,354,424
    & 33.9
    & 42.1
    & 0.28
    \\
    \cmidrule{2-6}
    & Composition & 11,621,125
    & 36.2
    & 44.5
    & 0.22
    \\
    & Multi-Entity & 36,359,826
    & 17.9
    & 23.2
    & 0.24
    \\
    \cmidrule{2-6}
    & Relation & 36,606,028
    & 11.5
    & 15.3
    & 0.28
    \\
    \bottomrule
    \end{tabular}
    \vspace{0.75em}
    \caption{Key caption statistics of the GBC1M and GBC10M datasets across different types of captions.
    We use the DFN-5B CLIP model to compute the CLIP scores.
    }
    \label{tab:gbc-dataset-caption-stat}
\vspace{-1em}
\end{table}

\subsubsection{Caption statistics}

\begin{figure}[t]
    \centering
    \includegraphics[width=0.325\textwidth]{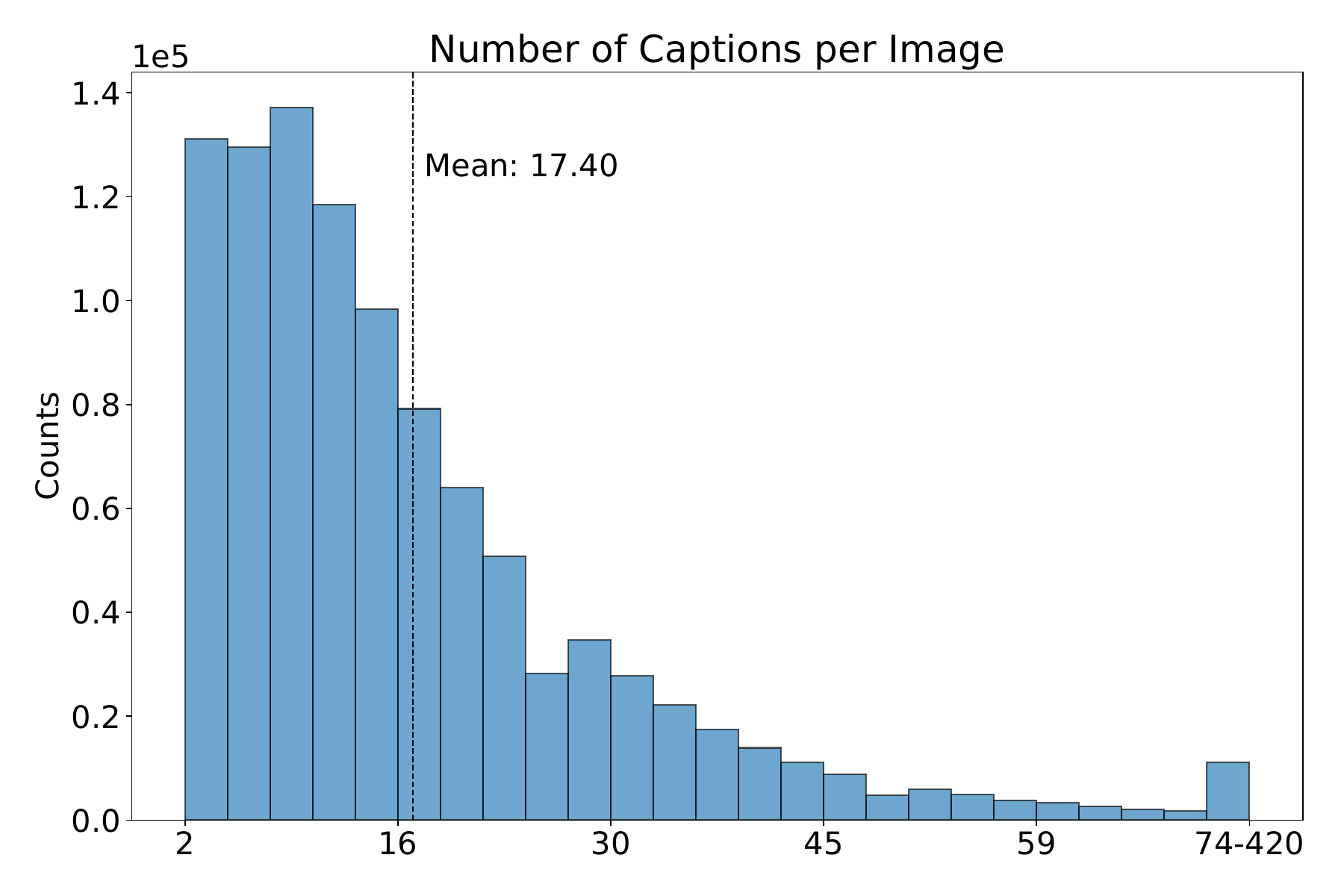}
    \includegraphics[width=0.325\textwidth]{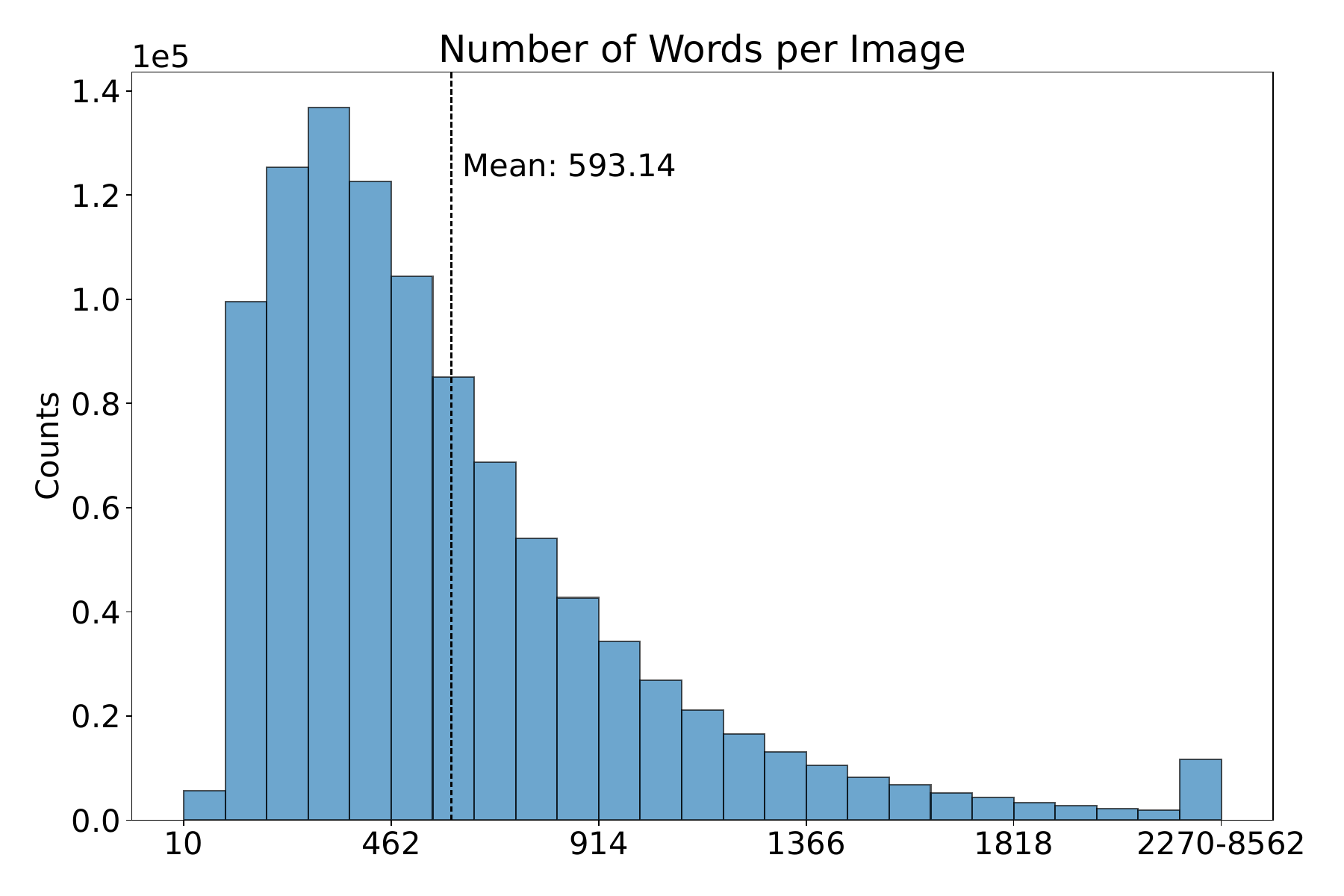}
    \includegraphics[width=0.325\textwidth]{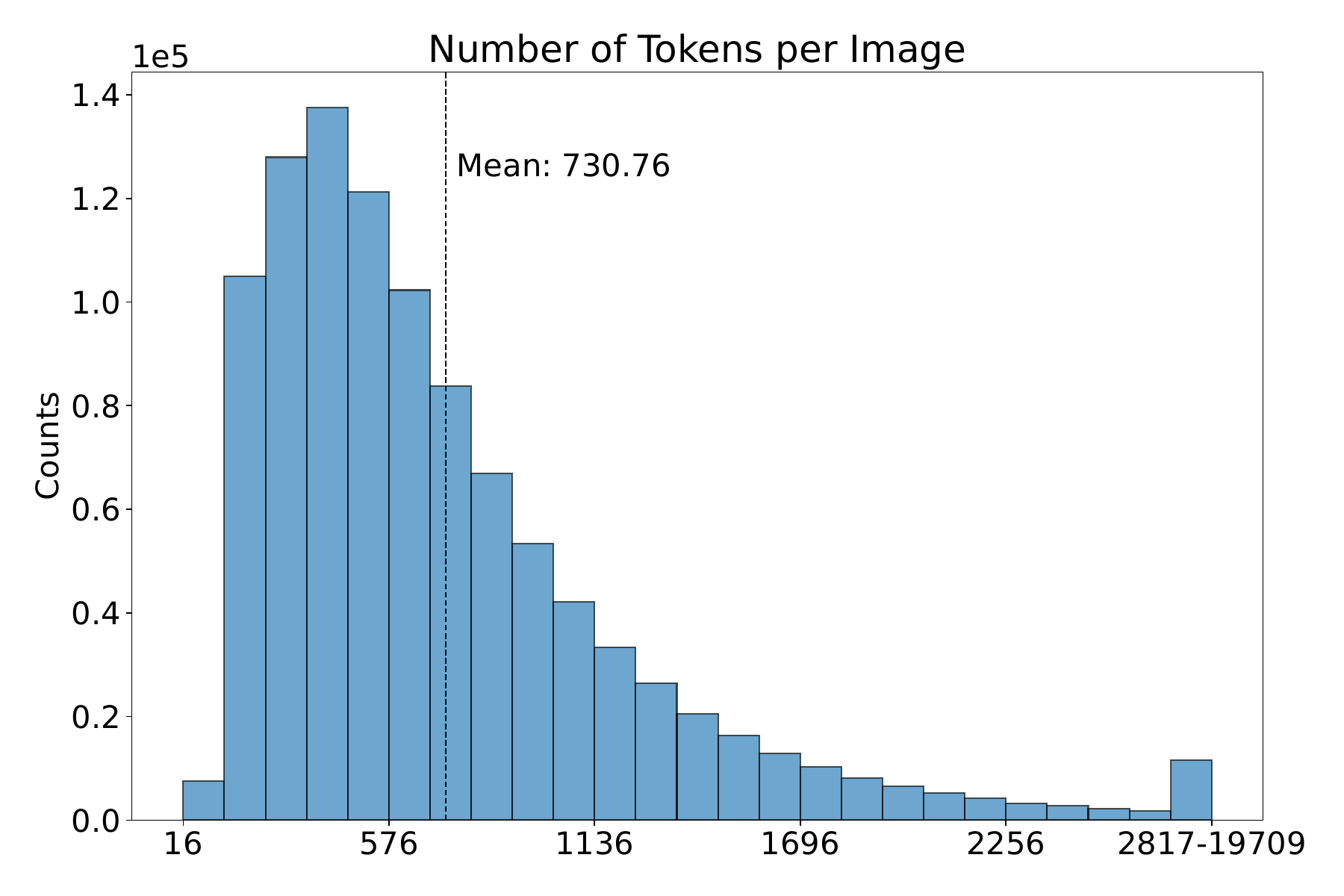}
    \caption{Distributions of numbers of captions, words, and tokens per image in the GBC1M Dataset.}
    \label{fig:gbc1m-caption-number-stat}
\end{figure}

\begin{figure}
    \centering
    \includegraphics[width=0.325\textwidth]{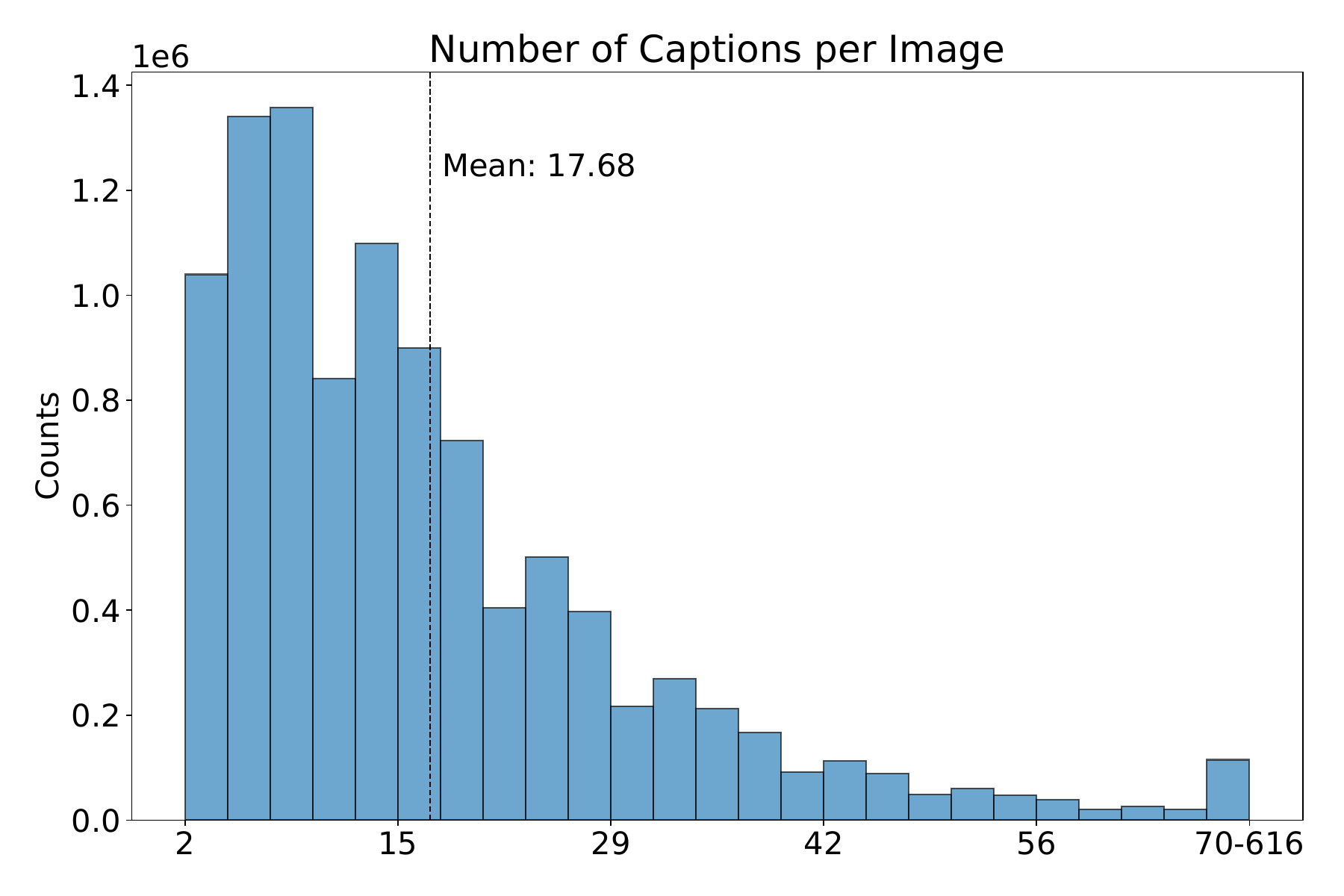}
    \includegraphics[width=0.325\textwidth]{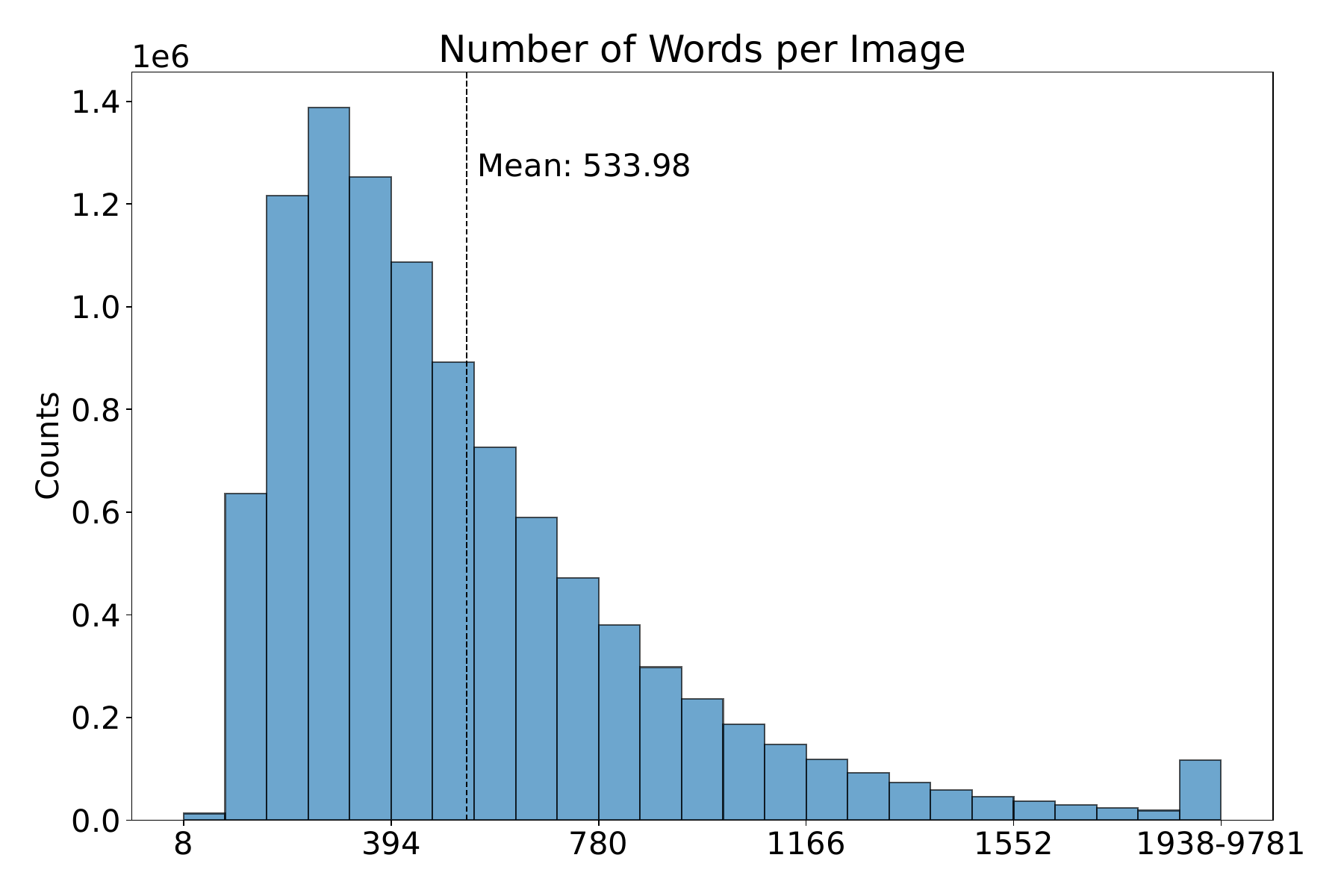}
    \includegraphics[width=0.325\textwidth]{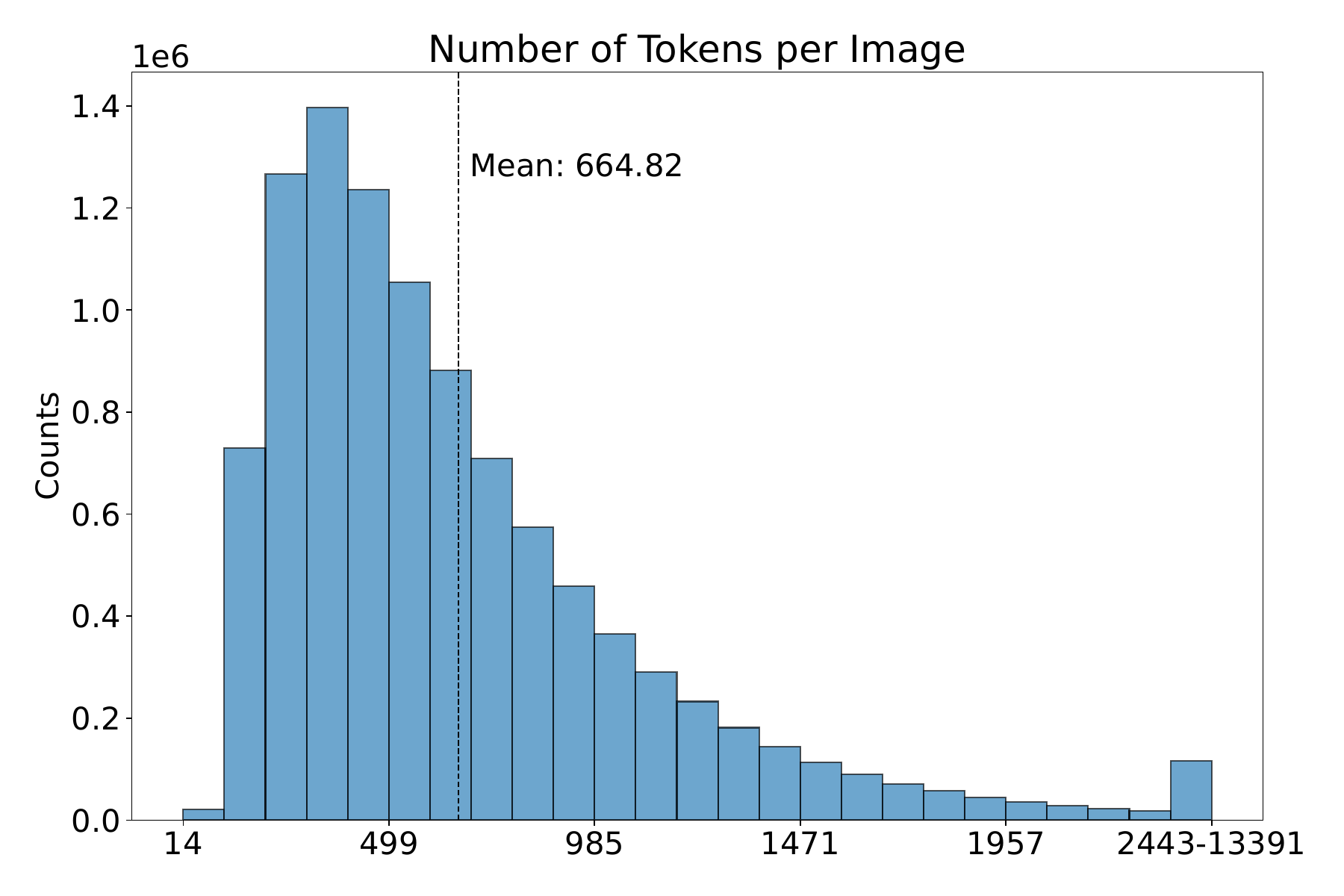}
    \caption{Distributions of numbers of captions, words, and tokens per image in the GBC10M Dataset.}
    \label{fig:gbc10m-caption-number-stat}
\end{figure}

For statistics at the caption level, we first complete \cref{tab:gbc-dataset-stat} 
by providing distribution of CLIP scores on the two datasets in \cref{fig:gbc1m-10m-clip-score}, and distribution of number of captions, words, and tokens per image in \cref{fig:gbc1m-caption-number-stat,fig:gbc10m-caption-number-stat}.
In particular, the significant variation in CLIP score distributions across different caption types motivates our decision to perform CLIP-filtering independently for each type, as mentioned in \cref{subsec:exp-setup}.

Going further, we report the average number of words and tokens per caption across different types of captions in \cref{fig:gbc1m-caption-length}, \ref{fig:gbc10m-caption-length}, and \cref{tab:gbc-dataset-caption-stat}.
We can see that except for the detailed image captions, most captions indeed contain fewer than 77 tokens.
\cref{tab:gbc-dataset-caption-stat} additionally reveals that we have near 2.5 times more region captions (\ie entity and multi-entity captions) than the total of relation and composition captions.
However, as we have seen in \cref{subsec:standard-benchmark} and will further ablate in \cref{apx:caption-type-influence}, these relation and composition captions, unique to our dataset, are crucial for the performance improvement that we observe across different evaluations.

\clearpage

\begin{figure}[p]
    \centering
    \includegraphics[width=0.42\textwidth]{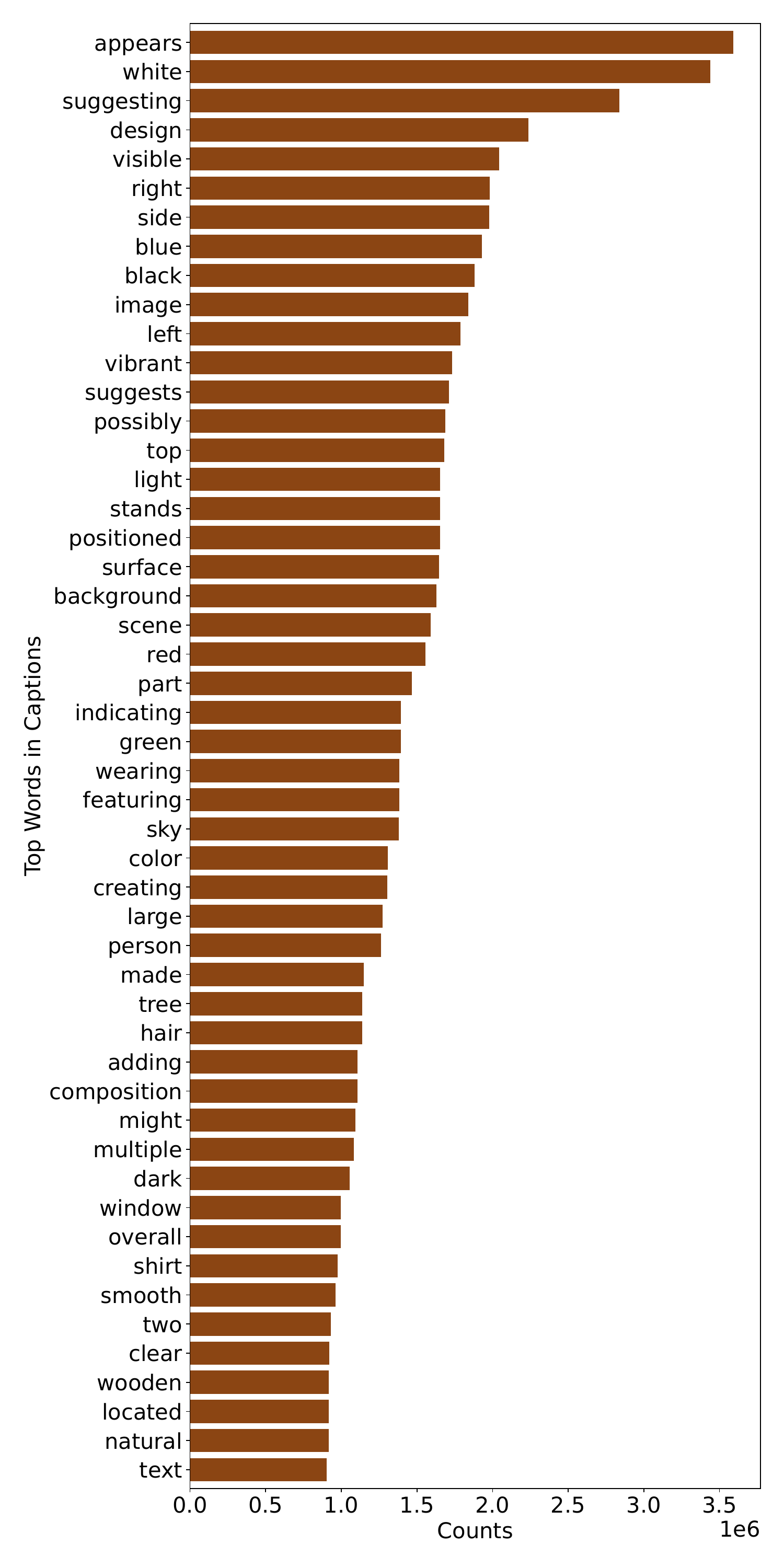}
    \includegraphics[width=0.42\textwidth]{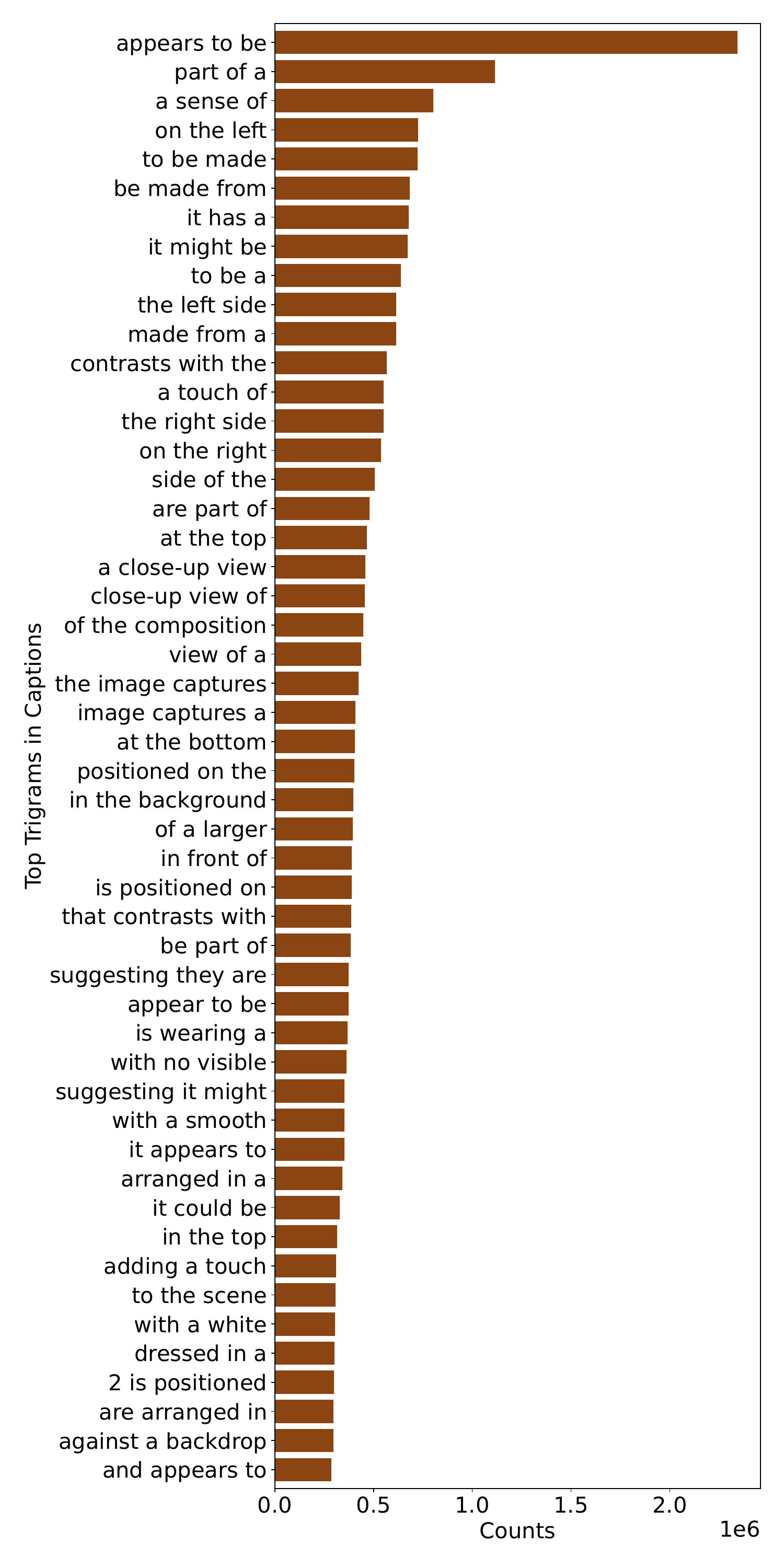}
    \caption{Distributions of the 20 most common words and trigrams that appear in the captions of the GBC1M dataset.}
    \label{fig:gbc1m-caption-content}
\end{figure}

\begin{figure}
    \centering
    \includegraphics[width=0.42\textwidth]{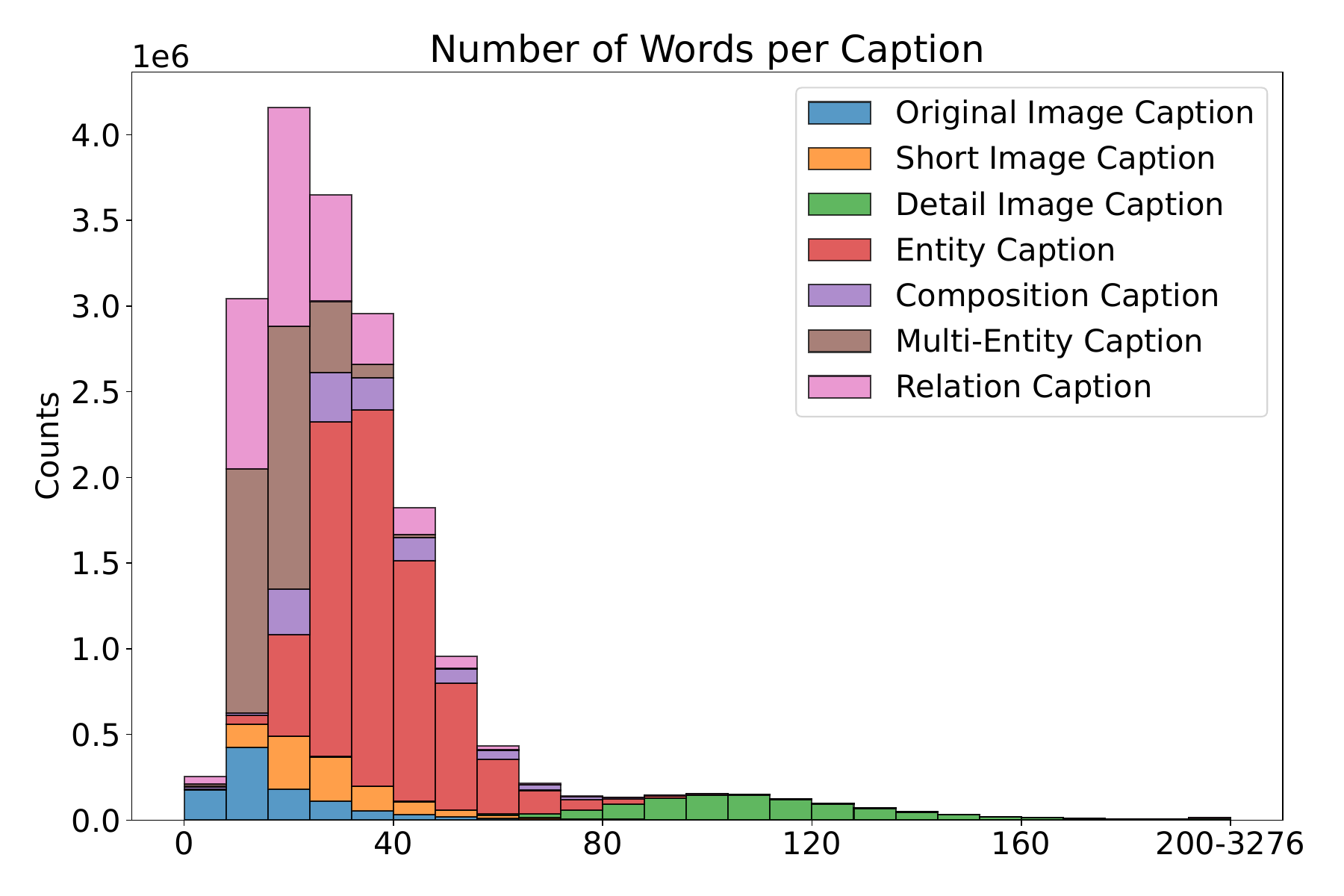}
    \includegraphics[width=0.42\textwidth]{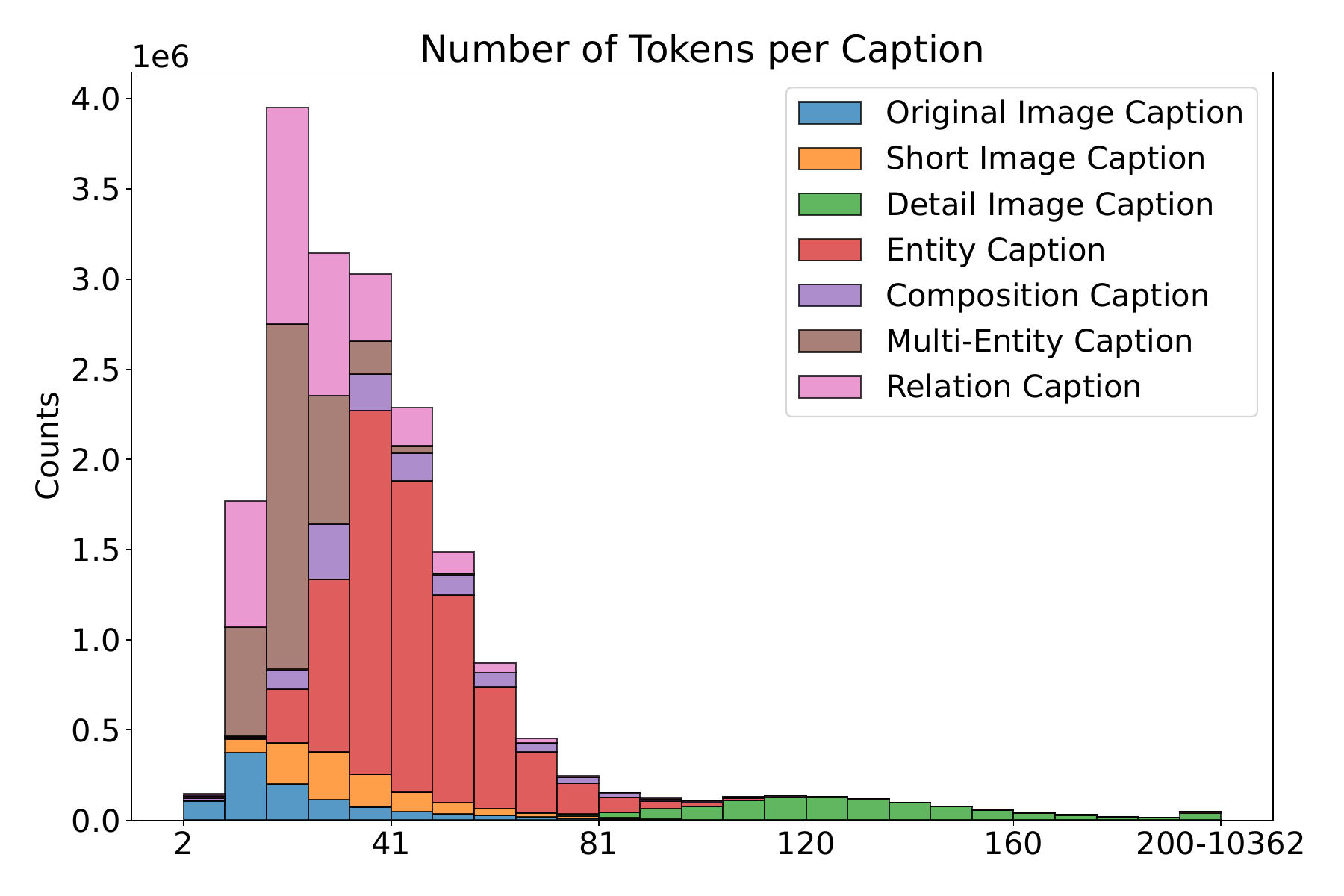}
    \caption{Distributions of numbers of words/tokens across different types of captions in the GBC1M dataset.
    To compute the number of tokens we use the standard CLIP tokenizer.}
    \label{fig:gbc1m-caption-length}
\end{figure}

\begin{figure}[p]
    \centering
    \includegraphics[width=0.42\textwidth]{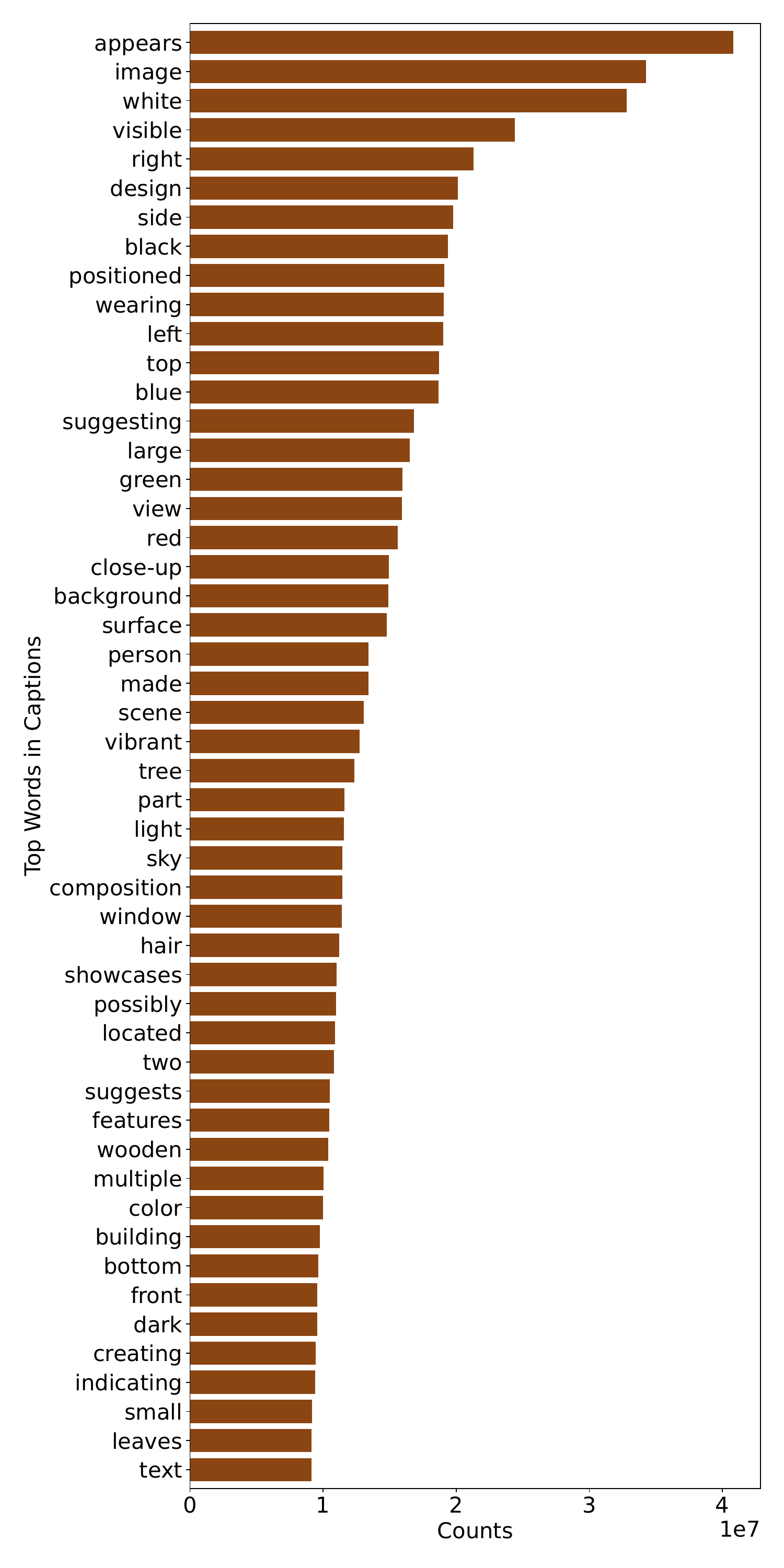}
    \includegraphics[width=0.42\textwidth]{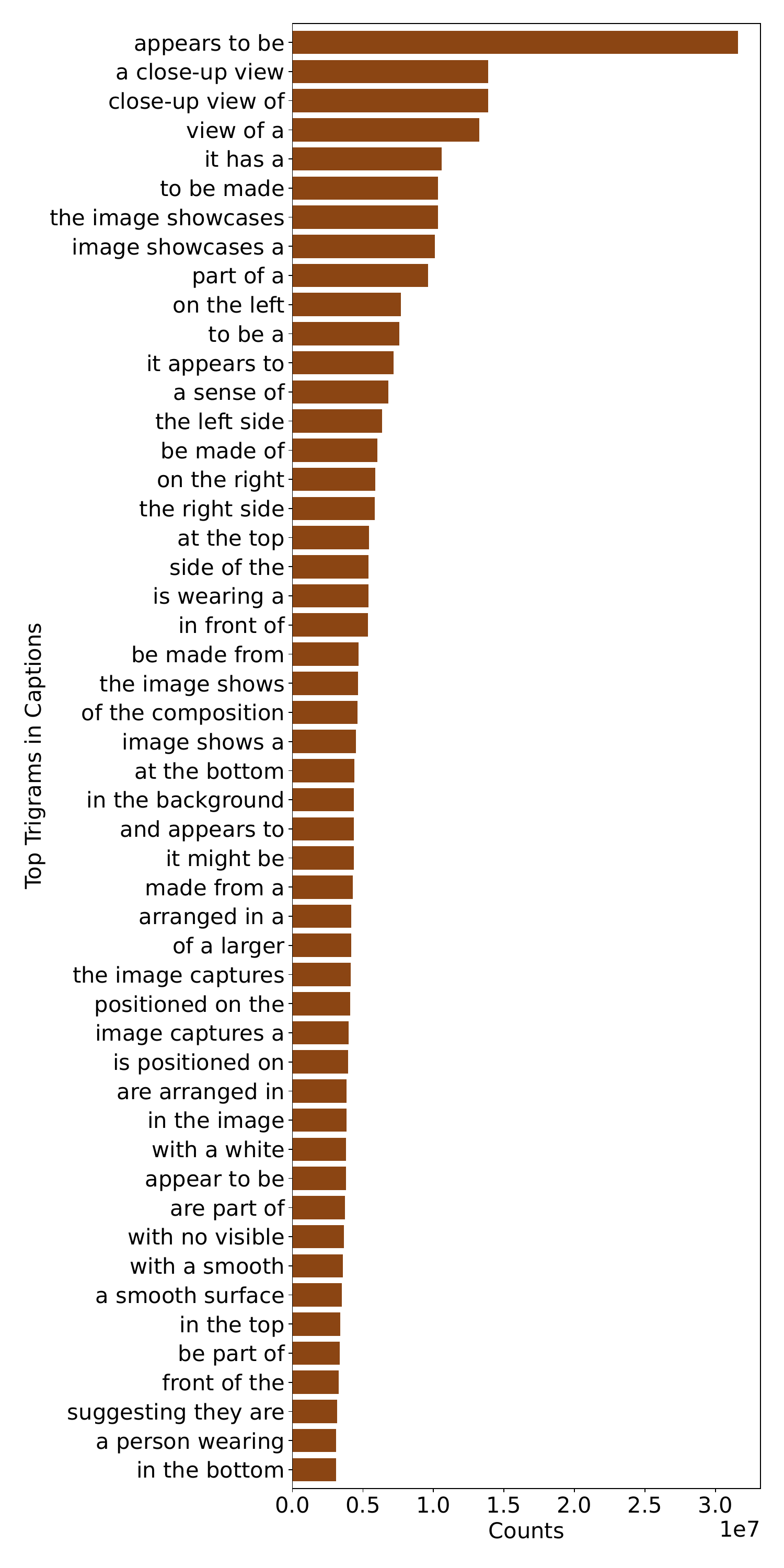}
    \caption{Distributions of the 20 most common words and trigrams that appear in the captions of the GBC10M dataset.}
    \label{fig:gbc10m-caption-content}
\end{figure}

\begin{figure}
    \centering
    \includegraphics[width=0.42\textwidth]{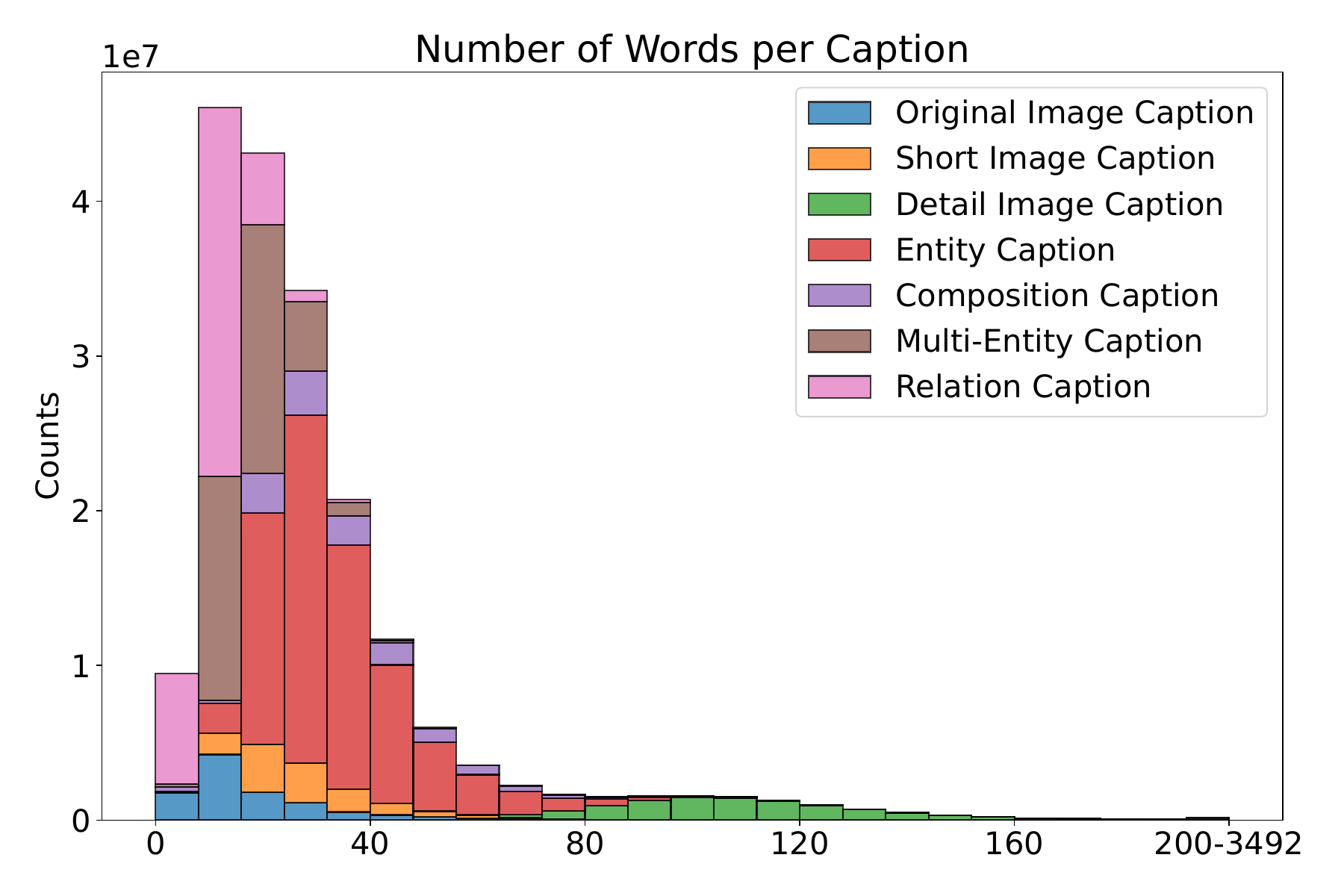}
    \includegraphics[width=0.42\textwidth]{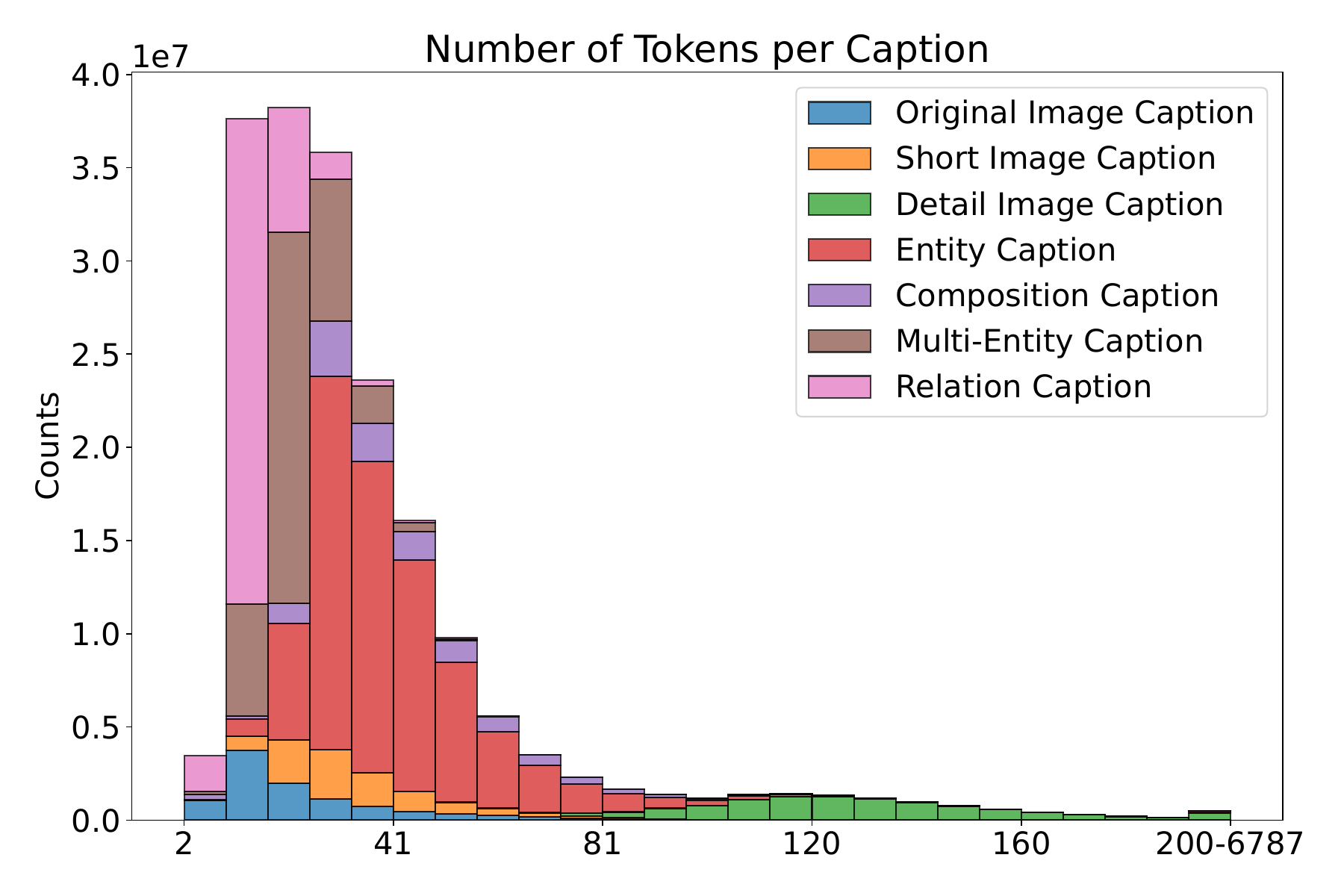}
    \caption{Distributions of numbers of words/tokens across different types of captions in the GBC10M dataset.
    To compute the number of tokens we use the standard CLIP tokenizer.}
    \label{fig:gbc10m-caption-length}
\end{figure}

We conclude this part by showing the distribution of the 20 most common words and trigrams that appear in our captions in \cref{fig:gbc1m-caption-content,fig:gbc10m-caption-content}, with stop words removed when considering the word distributions.
The frequent appearances of colors among the top words again align with the distribution reported in Visual Genome \cite[Fig. 24]{krishna2017visual}.
In addition, phrases like ``appears to be'', ``possibly'', and ``the image captures'' that commonly appear in our data, reflect LLaVA's use of GPT-generated data during instruction tuning.




\subsection{Examples from GBC10M}
\label{apx:dataset-examples}

\begin{figure}
\begin{tcolorbox}[sharp corners,colback=white,colframe=white!20!black]
    \flushleft{\textbf{Flame:}}\\[-0.6em]
    \centering
    \hspace{5.5em}
    \includegraphics[width=0.18\textwidth]{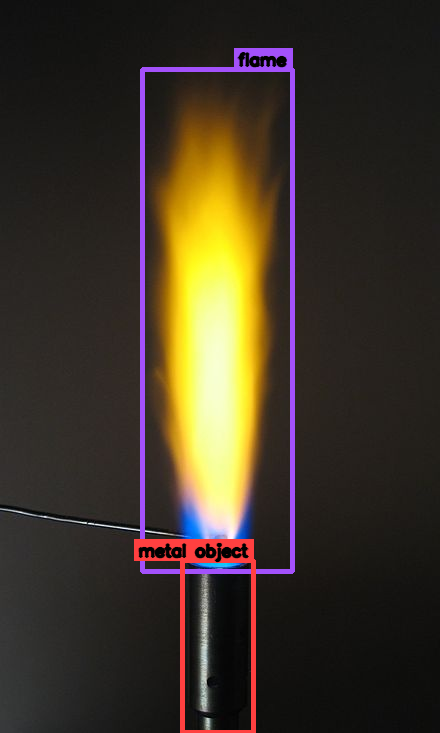}
    \hspace{3em}
    \includegraphics[width=0.35\textwidth]{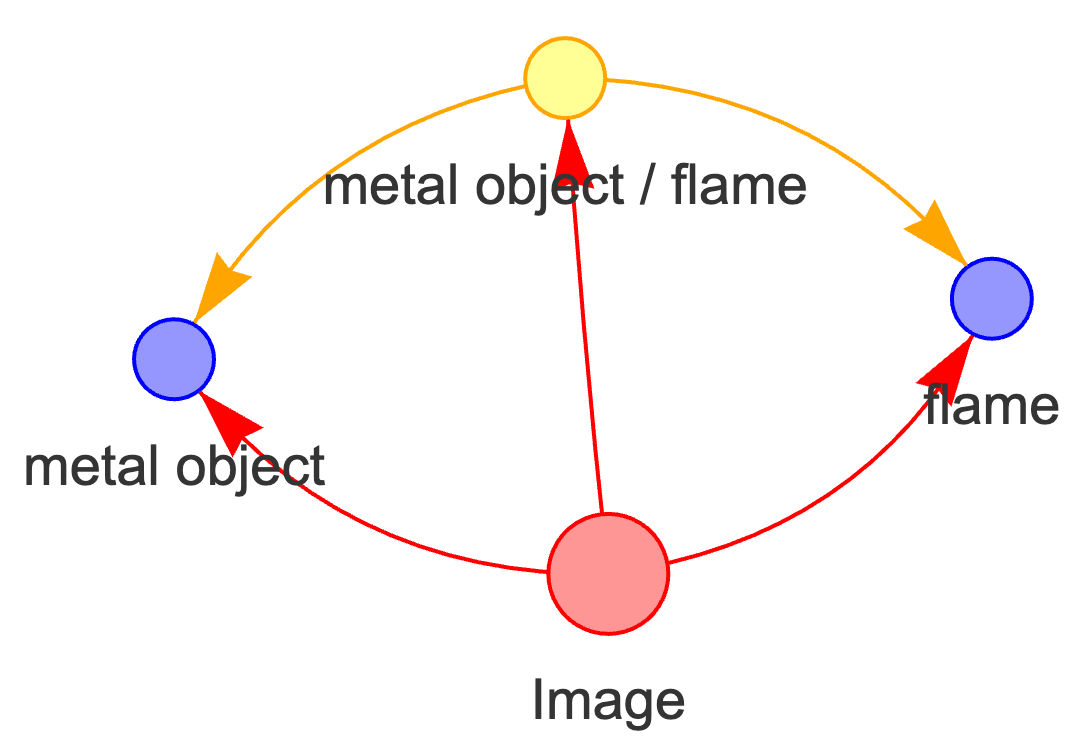}
    \tcbline
    \flushleft{\textbf{Messe:}}\\[-0.7em]
    \includegraphics[width=0.7\textwidth]{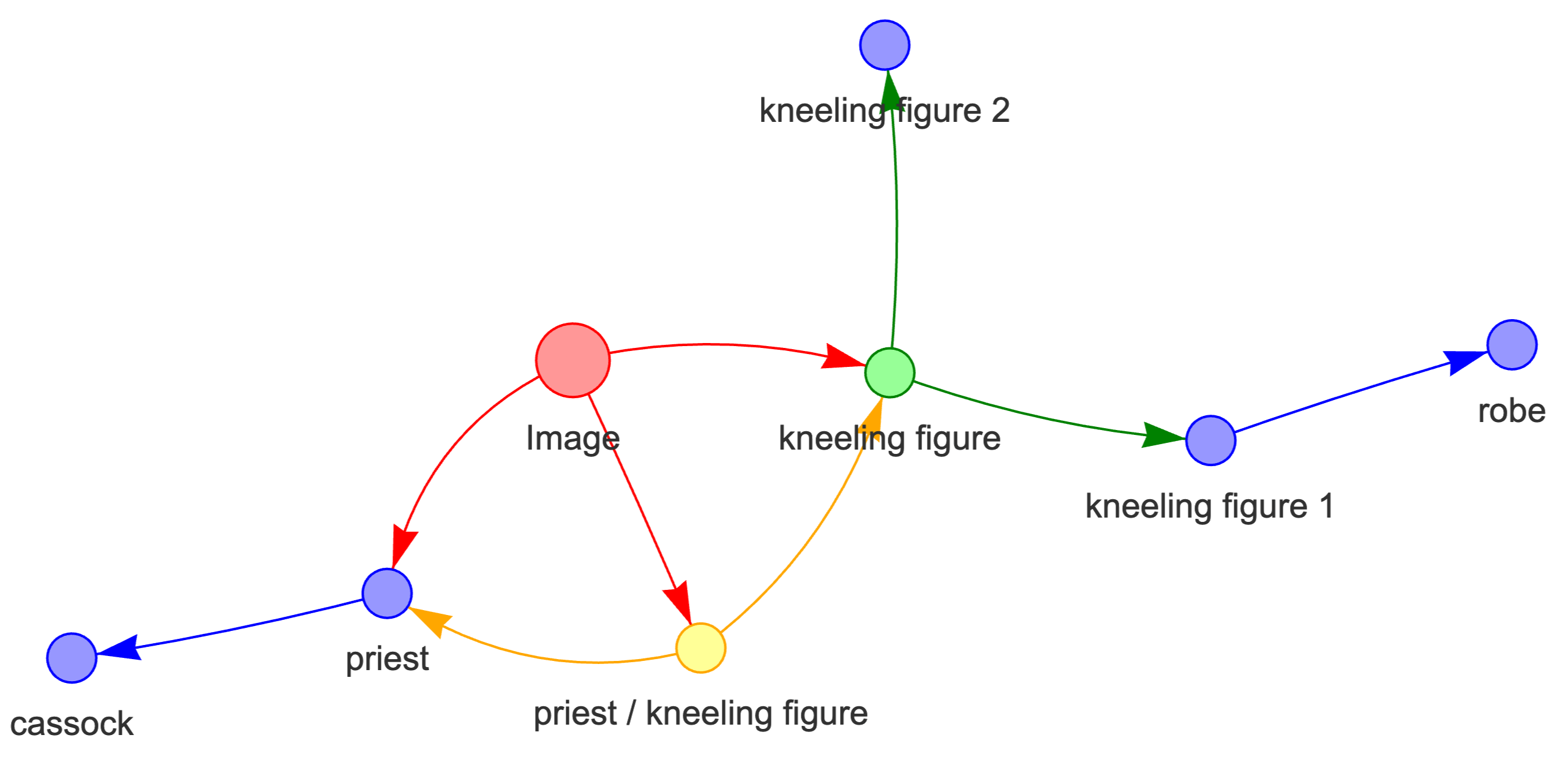}
    \hfill
    \includegraphics[width=0.25\textwidth]{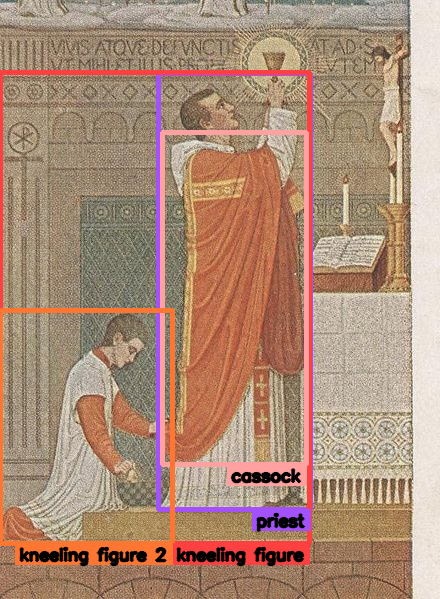}
    \tcbline
    \flushleft{\textbf{Regalia:}}\\[-1.5em]
    \includegraphics[width=0.4\textwidth]{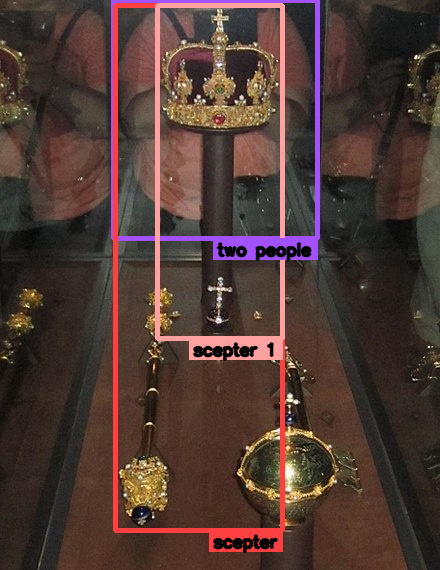}
    \hfill
    \includegraphics[width=0.57\textwidth]{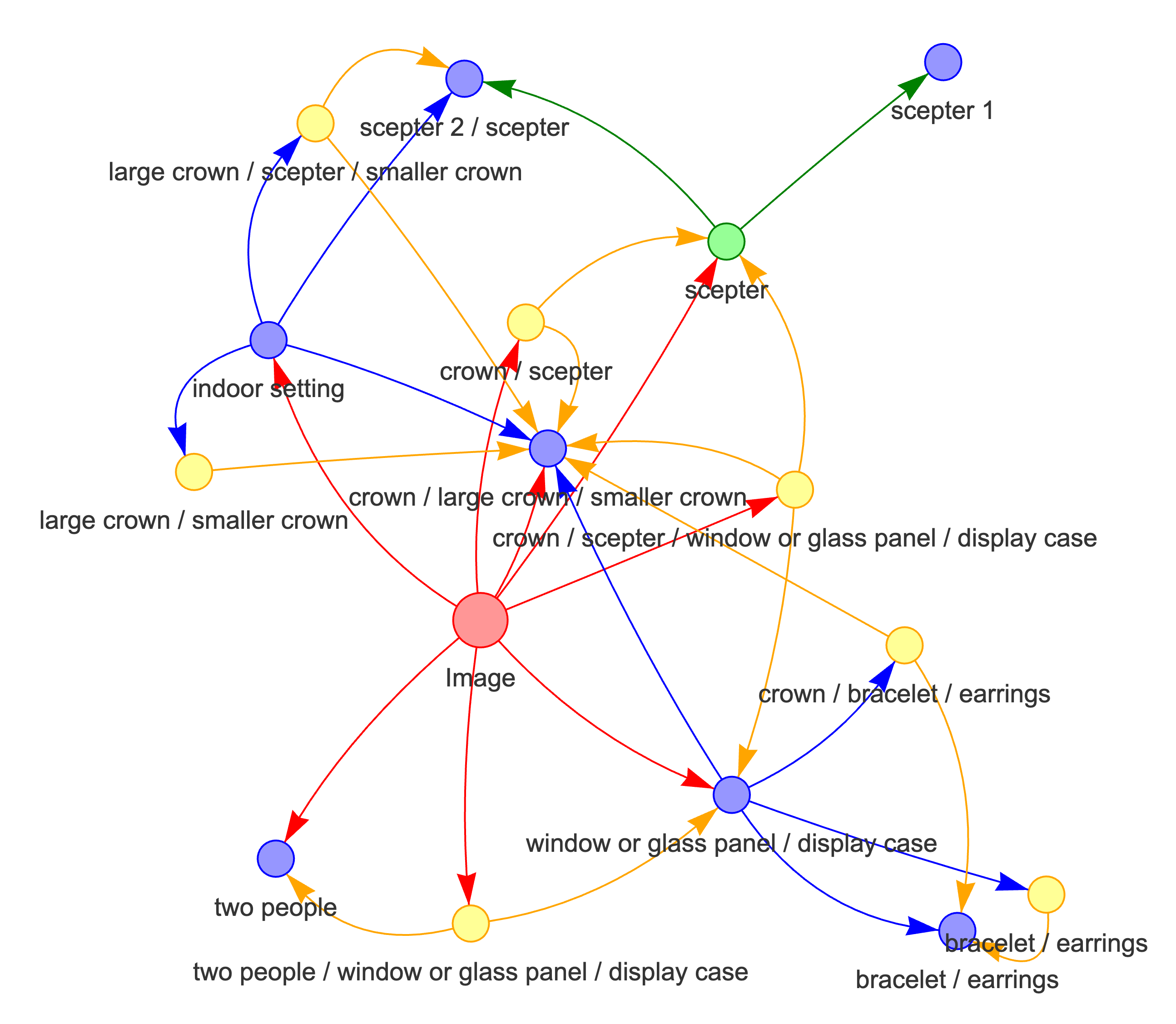}
\end{tcolorbox}
    \caption{Example images and graphs from the GBC10M dataset. For ease of visualization we do not show all the bounding boxes.}
    \label{fig:gbc10m-ex-flame-missa-scepter}
\end{figure}

\begin{figure}[p]
\begin{tcolorbox}[sharp corners,colback=white,colframe=white!20!black]
    \textbf{Elephant:}\\
    \begin{center}
    \includegraphics[width=0.8\textwidth]{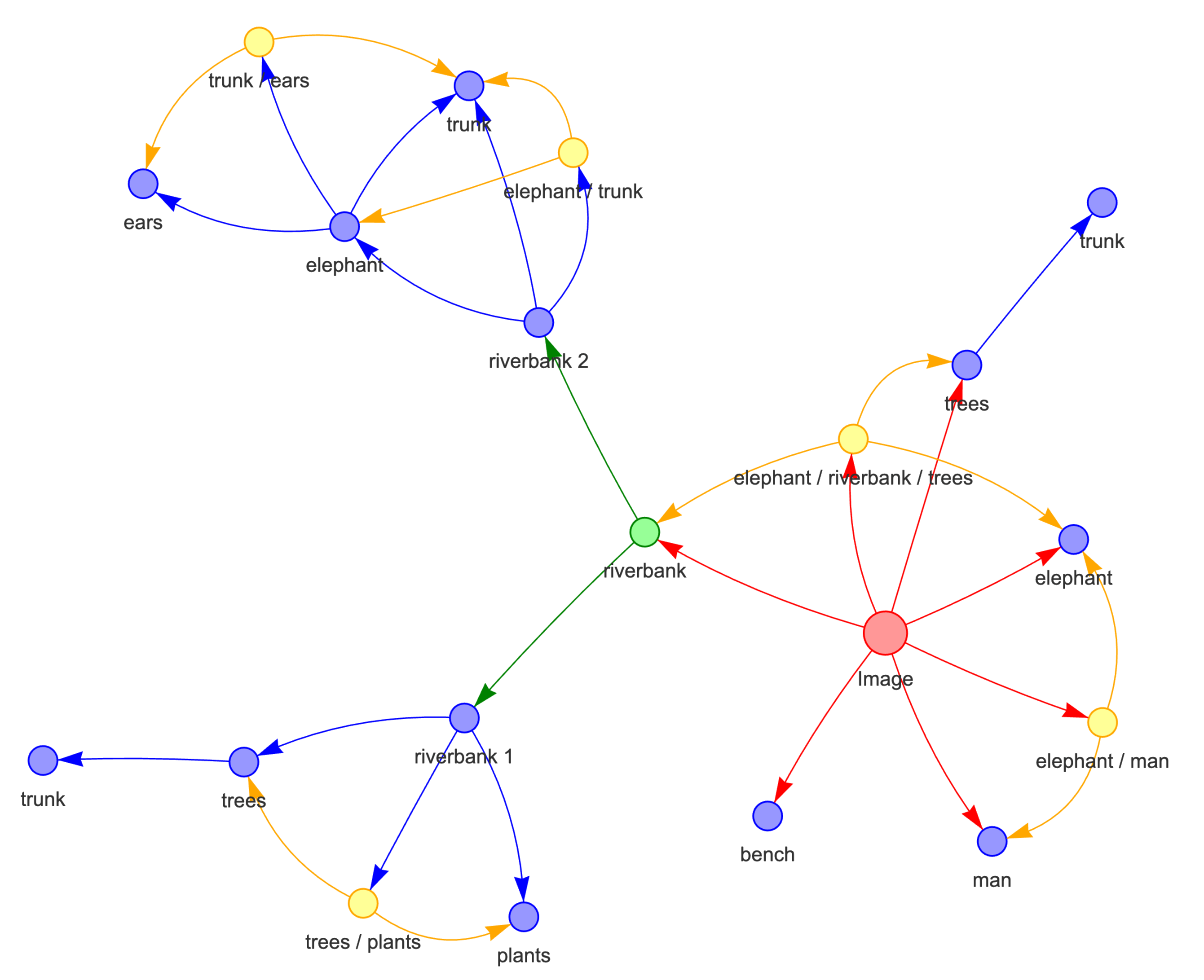}\\[1em]
    \vfill
    \includegraphics[width=0.8\textwidth]{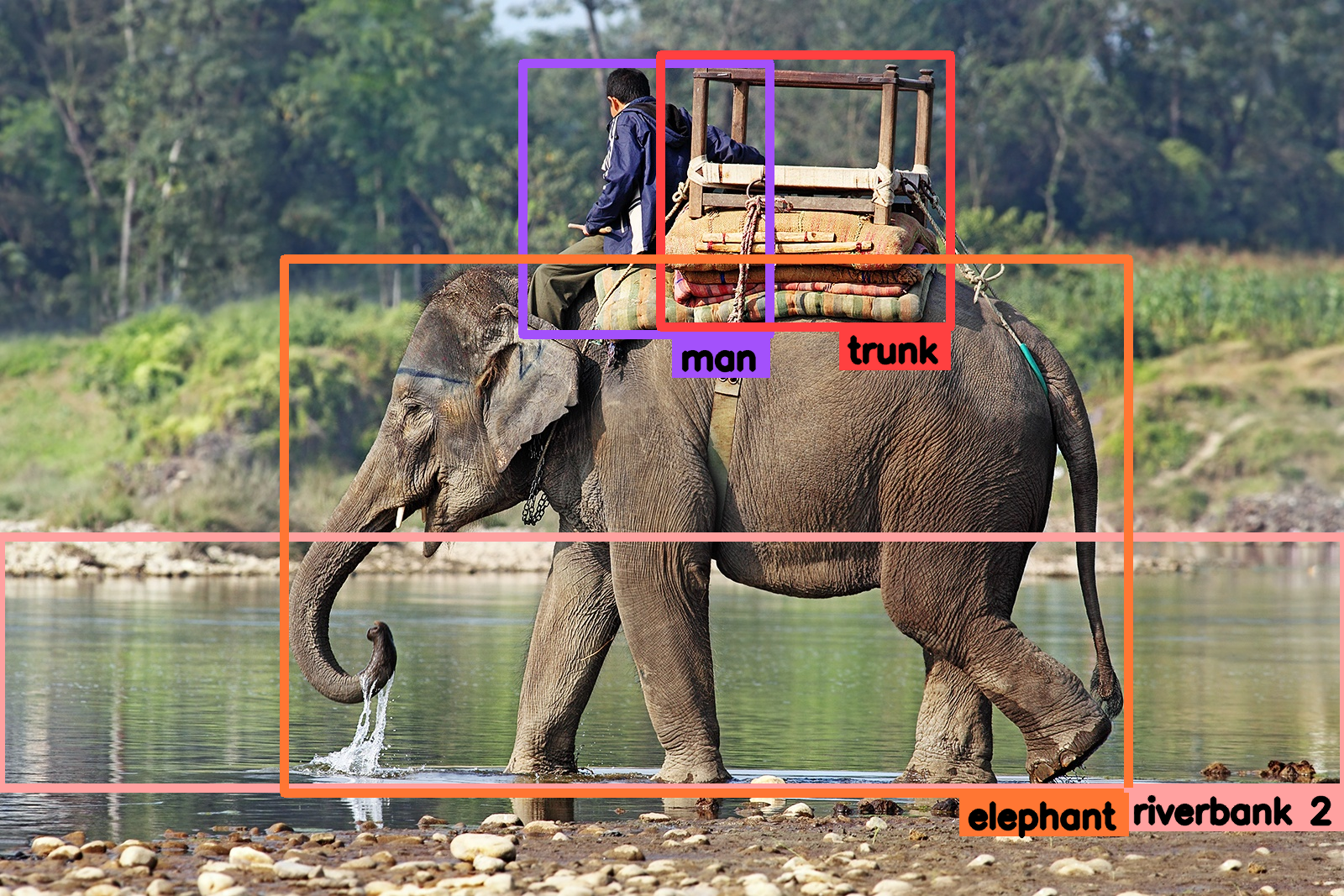}
    \end{center}
\end{tcolorbox}
    \caption{An example image with its graph from the GBC10M dataset.
    For ease of visualization do not show all the bounding boxes.}
    \label{fig:gbc10m-ex-elephant}
\end{figure}

As a complement to the dataset statistics presented in the previous section, we showcase a few illustrative examples from GBC10M in Figures~\ref{fig:gbc10m-ex-flame-missa-scepter} and~\ref{fig:gbc10m-ex-elephant}.
These examples demonstrate the varying levels of graph complexity across our dataset.
The number of nodes varies from just a few (first example in Figure~\ref{fig:gbc10m-ex-flame-missa-scepter}) to over 10 (third example in Figure~\ref{fig:gbc10m-ex-flame-missa-scepter} and the example in Figure~\ref{fig:gbc10m-ex-elephant}).
In most cases, this complexity aligns with the visual complexity of the corresponding image.

On the other hand, these examples also reveal limitations arising from the object detection models used. 
For instance, in the Messe example from \cref{fig:gbc10m-ex-flame-missa-scepter}, the detection model incorrectly identifies a standing priest as a ``kneeling figure''. 
Similarly, in \cref{fig:gbc10m-ex-elephant}, two of the three nodes labeled ``trunk'' are derived from tree nodes and erroneously associated with the elephant’s trunk or other non-trunk objects on the elephant.
These limitations become particularly severe in the Regalia example of \cref{fig:gbc10m-ex-flame-missa-scepter}, where the presence of more specific objects like crowns, scepters, bracelets, and earrings leads to frequent confusion by the object detection model.

Next, we focus on the captions associated with these images.
A subset of these captions is presented in \cref{tab:captions,tab:captions-more}.
We observe that hallucination is particularly important for detailed captions.
These erroneous descriptions can then be inherited by the shorter captions derived from them.
We also note there are situations where the model describes an object that actually does not exist in the corresponding region of the image, such as the caption for ``scepter 1'' in the Regalia example.
As we can see from the figure, in the corresponding bounding box, there is no scepter visible but only a crown on a wooden base.

In spite of these inaccuracies in object detection and captioning, the overall graph structure and captions still align well with the images. 
On top of this, the granularity of our descriptions significantly enhances the descriptive power of our dataset, allowing for a more nuanced understanding of the visual content. 

\begin{table}[p]
    \renewcommand{\arraystretch}{1.35}
    \centering
    \begin{NiceTabular}{c|p{0.8\textwidth}}
    \toprule
    \multirow{9}{*}{\makecell{Short\\ Captions}}
    &
    \textbf{Flame [\cref{fig:gbc10m-ex-flame-missa-scepter}].} A \empha{flame} with yellow base and \hallucination{blue peak} emerges from a \empha{metal object} against a dark background.
    \\
    &
    \textbf{Messe [\cref{fig:gbc10m-ex-flame-missa-scepter}].} A \empha{priest} holds a chalice aloft while another figure \empha{kneeling figure} kneels on the floor, set against a backdrop of architectural details and ornamentation within what appears to be a religious setting. 
    \\
    &
    \textbf{Regalia [\cref{fig:gbc10m-ex-flame-missa-scepter}].}
    \hallucination{\empha{Two people}} stand behind a \empha{display case} containing a \empha{crown}, \empha{scepter} with a blue gem, and a golden orb with \hallucination{a red gem}, all \hallucination{under natural light from a \empha{window or glass panel}} within an \empha{indoor setting}.
    \\
    &
    \textbf{Elephant [\cref{fig:gbc10m-ex-elephant}].}
    A \empha{man} rides \hallucination{atop a \empha{bench}} strapped on a \empha{elephant} drinking from a \empha{riverbank}, surrounded by \empha{trees} under a clear sky. 
    \\
    \midrule
    \multirow{22}{*}{\makecell{\hphantom{--}Detailed\hphantom{--}\\ Captions}}
    & 
    \textbf{Flame [\cref{fig:gbc10m-ex-flame-missa-scepter}].}
    The image captures a close-up view of a \hallucination{blue \empha{flame}} emanating from a small \empha{metal object}, which appears to be a lighter or torch. The \empha{flame} has a vibrant yellow hue at its base, transitioning to \hallucination{a bright blue at its peak.} The \empha{flame}'s shape is irregular with wisps extending outward from its core, suggesting it's in motion or has been recently ignited. The \empha{metal object} has a cylindrical shape with a pointed tip from which the \empha{flame} emerges. The background is dark, providing a stark contrast that accentuates the \empha{flame}'s colors and form.
    \\
    &
    \textbf{Messe [\cref{fig:gbc10m-ex-flame-missa-scepter}].} The image portrays a religious scene set within what appears to be a church or chapel. At the center of the composition stands a \empha{priest}, dressed in traditional religious attire with a red robe and a white cowl. He holds a golden chalice aloft with both hands, suggesting he may be performing a sacrament or ritual. \hallucination{To his right}, another figure, possibly another \empha{priest} or religious figure, kneels on the floor, seemingly in prayer or reverence. The background features ornate architectural details, including arches and intricate patterns on the walls, indicative of Gothic or similar architectural styles. The overall atmosphere suggests a solemn or sacred moment within a religious ceremony or service.
    \\
    &
    \textbf{Regalia [\cref{fig:gbc10m-ex-flame-missa-scepter}].}
    The image captures a scene where \hallucination{two individuals} are standing behind a \empha{display case} containing various items. The \empha{display case} houses a collection of ornate jewelry pieces, including a \empha{crown} with intricate detailing, a \empha{scepter} with a blue gem at its top, and a golden orb with \hallucination{a red gem}.
    The individuals are dressed in pink shirts and are positioned behind the \empha{display case}, which has a reflective surface. The background suggests an \empha{indoor setting} with \hallucination{a \empha{window or glass panel} allowing natural light to illuminate the scene.}
    \\
    &
    \textbf{Elephant [\cref{fig:gbc10m-ex-elephant}].}
    The image captures a serene scene at a \empha{riverbank} where a \empha{man} is riding on the back of a large \empha{elephant}. The \empha{elephant}, with its majestic gray skin, is partially submerged in the water, drinking from it. The \empha{man}, dressed in casual attire, \hallucination{sits comfortably on a wooden \empha{bench}} strapped securely on the \empha{elephant}'s back. The \empha{bench} is adorned with colorful cushions for added comfort during the ride.
    The backdrop features lush greenery with \empha{trees} lining the \empha{riverbank}, adding to the tranquil atmosphere of the scene. The overall setting suggests a peaceful interaction between humans and nature.
    \\
    \bottomrule
    \end{NiceTabular}
    \vspace{0.75em}
    \caption{Corresponding synthetic detailed and short captions for the examples shown in \cref{fig:gbc10m-ex-flame-missa-scepter,fig:gbc10m-ex-elephant}.
    We highlight the objects described in the children nodes in dark blue and mark some erroneous descriptions in italic.
    }
    \label{tab:captions}
\end{table}

\begin{table}[p]
    \renewcommand{\arraystretch}{1.35}
    \centering
    \begin{NiceTabular}{c|p{0.8\textwidth}}
    \toprule
    \multirow{9}{*}{\makecell{Entity\\ Captions}}
    & 
    \textbf{Flame---flame [\cref{fig:gbc10m-ex-flame-missa-scepter}].}
    A bright yellow flame emanates from a blue lighter. The flame is intense and vibrant, with a distinct orange hue at the center. It appears to be burning steadily, casting a warm glow.
    \\
    &
    \textbf{Messe---priest [\cref{fig:gbc10m-ex-flame-missa-scepter}].}
    A priest dressed in traditional religious attire, which includes a red robe with gold trim, a white \empha{cassock}, and a red sash. \hallucination{He holds a cross in his right hand.}
    \\
    &
    \textbf{Regalia---scepter 1 [\cref{fig:gbc10m-ex-flame-missa-scepter}].}
    \hallucination{A gold scepter with a cross atop it. The scepter has intricate designs and patterns throughout its length.}
    \\
    &
    \textbf{Elephant---elephant [\cref{fig:gbc10m-ex-elephant}].}
    A large elephant with a long trunk is seen walking through a body of water. Its skin appears rough and wrinkled, typical of elephants. The elephant has small tusks and large ears that are characteristic of this species.
    \\
    \midrule
    \multirow{5}{*}{\makecell{Relation\\ Captions}}
    &
    \textbf{Flame---flame/metal object [\cref{fig:gbc10m-ex-flame-missa-scepter}].} The \empha{flame} is positioned above the \empha{metal object}.
    \\
    &
    \textbf{Messe---priest/kneeling figure [\cref{fig:gbc10m-ex-flame-missa-scepter}].} The \empha{priest} is standing in front of the \empha{kneeling figure}.
    \\
    &
    \textbf{Regalia---crown/scepter [\cref{fig:gbc10m-ex-flame-missa-scepter}].} The \empha{scepter} is positioned \hallucination{next to} the \empha{crown}.
    \\
    &
    \textbf{Elephant---elephant/riverbank/trees [\cref{fig:gbc10m-ex-elephant}].}
    The \empha{elephant} is standing near the \empha{riverbank} with \empha{trees} in the background.
    \\
    \midrule
    \multirow{7}{*}{\makecell{Composition\\ Captions}}
    &
    \textbf{Messe---kneeling figure [\cref{fig:gbc10m-ex-flame-missa-scepter}].} \empha{Kneeling figure 1}, positioned at the top right, appears to be in a state of prayer or reverence, while \empha{kneeling figure 2}, located at the bottom left, seems to be in a similar posture, possibly indicating a shared moment of devotion or reflection.
    \\
    &
    \textbf{Regalia---scepter [\cref{fig:gbc10m-ex-flame-missa-scepter}].} 
    \empha{Scepter 2}, which is in the bottom left corner, has a golden handle with a blue gemstone at its center, while \hallucination{\empha{scepter 1}}, positioned above \empha{scepter 2}, features a \hallucination{golden handle} with a red gemstone at its center. Both scepters are ornate, with intricate designs and a regal appearance.
    \\
    &
    \textbf{Elephant---riverbank [\cref{fig:gbc10m-ex-elephant}].}
    \empha{Riverbank 1} is positioned above \empha{Riverbank 2}, with \empha{Riverbank 2} located at the bottom of the composition.
    \\
    \midrule
    \multirow{5}{*}{\makecell{Multi-Entity\\ Captions}}
    &
    \textbf{Messe---kneeling figure [\cref{fig:gbc10m-ex-flame-missa-scepter}].} Both figures are depicted in a posture commonly associated with prayer or worship, suggesting a religious or spiritual context for their actions. 
    \\
    &
    \textbf{Regalia---scepter [\cref{fig:gbc10m-ex-flame-missa-scepter}].}
    The gemstones in their handles add a touch of elegance and value to each scepter.
    \\
    &
    \textbf{Elephant---riverbank [\cref{fig:gbc10m-ex-elephant}].}
    They are situated near a body of water, which suggests a peaceful, natural setting.
    \\
    \bottomrule
    \end{NiceTabular}
    \vspace{0.75em}
    \caption{Some example relational and region captions for the examples shown in \cref{fig:gbc10m-ex-flame-missa-scepter,fig:gbc10m-ex-elephant}.
    We highlight the objects described in the children nodes in dark blue and mark some erroneous descriptions in italic.
    }
    \label{tab:captions-more}
\end{table}

\FloatBarrier

\newpage
\section{Algorithm details}
\label{apx:algo}
This appendix provides missing details about our architectures, CLIP training objective, and sampling algorithms.

\subsection{Structure-aware hierarchical attention}
\label{apx:SAHA}

We propose a simple text encoder architecture to incorporate structural information encoded in GBC graph along with node captions. 
Specifically, we present \ac{SAHA} block which treats each caption as an individual sample, and introduces an additional cross-attention layer that enforces the captions to attend to their children.

\begin{figure*}[t]
    \centering
    \includegraphics[width=\textwidth]{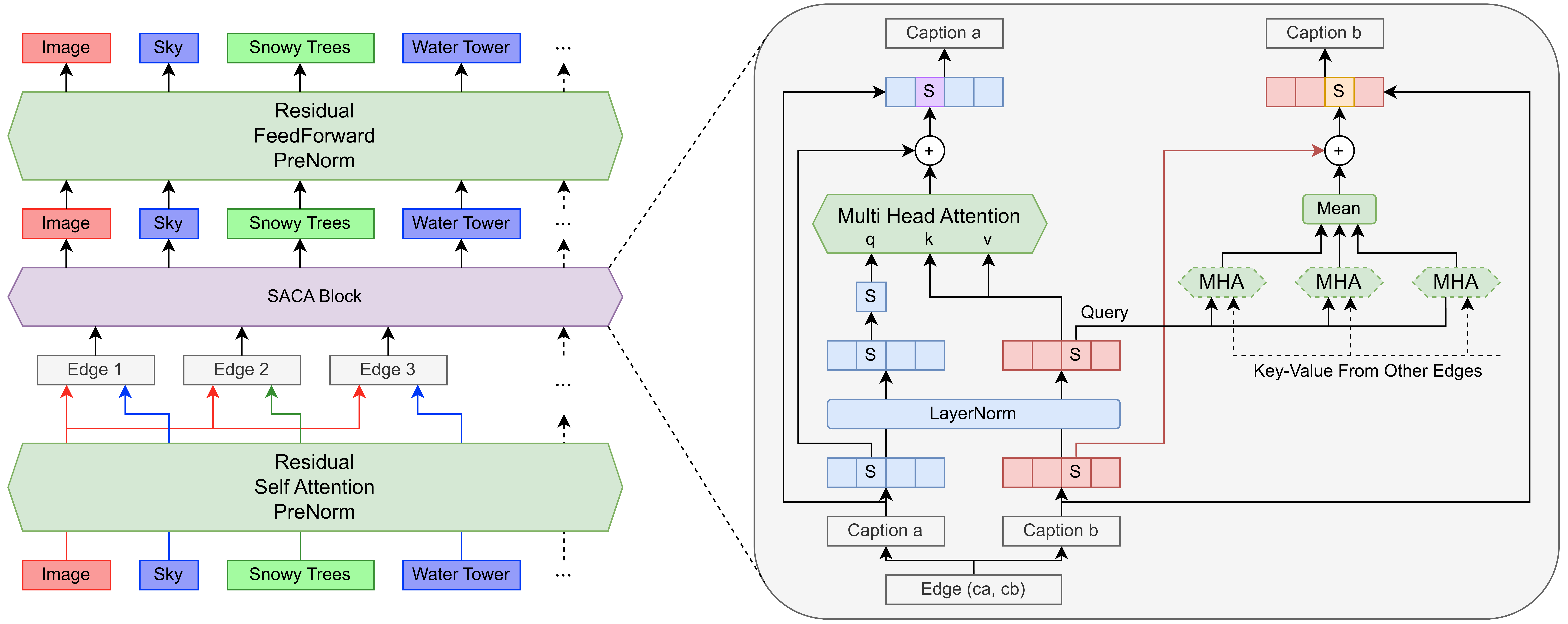}
    \caption{Illustration of the proposed SAHA block when applied to the graph shown in \cref{fig:GBC-ex-main}. For the sake of simplicity, we assume here there is only one caption per node.}
    \label{fig:SACA-arch-main}
\end{figure*}

Formally,
we consider a caption graph $\vcap[\graph][\captions] = (\captions, \vcap[\edges][\captions])$ with vertices $\captions=\Union_{\vertex\in\vertices}\vv[\captions]$ and edges $\vcap[\edges][\captions]\subseteq\captions\times\captions$ such that $(\capn, \capnalt)\in\vcap[\edges][\captions]$ if and only if $\capn\in\vv[\captions][\vertexalt]$, $\capnalt\in\vv[\captions][\vertex]$, $\edge=(\vertexalt, \vertex)\in\edges$, and the label $\ve[\labeling][\edge]$ is included within the caption $\capn$.
In words, each vertex in the graph represents a caption from a node of the original graph and there is an edge from one caption to another only if the second caption describes part of the first caption.
After tokenization of the captions, we can map the edge labels to a set of token positions of the source caption, which we write as $\ve[\positionids][\edge]$.
Then, the target caption \emph{annotates} the source caption via the tokens at positions $\ve[\positionids][\edge]$.
Therefore, we can simply consider cross-attention with queries from these tokens and keys and values from the target caption.
We illustrate this idea in \cref{fig:SACA-arch-main}, where we zoom in on the additional cross-attention layer (SACA) on the right side of the figure. 

To express this via mathematical formula, we denote by $\vcap[\nhdgraph][\capn]$ the children of caption $\capn$ in caption graph $\vcap[\graph][\capn]$
and write the features of $\capn$ in the input of our \ac{SACA} layer as $\vcap[\features]=[\vcapc[\feature][\capn][1], \ldots, \vcapc[\feature][\capn][\vcap[\ntokens][\capn]]]$.
Then, the SACA layer maps each feature vector $\vcapc[\feature][\capn][\tokenpos]$ to
\begin{equation}
    \label{eq:SACA}
    \saca(\vcapc[\feature][\capn][\tokenpos]) = \frac{\sum_{\capnalt\in\vcap[\nhdgraph][\capn]} \one_{\tokenpos\in\ve[\positionids][(\capn,\capnalt)]}
    \mha(\vcapc[\feature][\capn][\tokenpos], \vcap[\features][\capnalt], \vcap[\features][\capnalt])}
    {\min(1, \sum_{\capnalt\in\vcap[\nhdgraph][\capn]} \one_{\tokenpos\in\ve[\positionids][(\capn,\capnalt)]})},
\end{equation}
where $\mha$ implements the standard multi-head attention mechanism.
Note that we average across the results from all the relevant captions that describe this token, as we show in the figure.

Our text encoder then stacks a number of \ac{SAHA} blocks, effectively interleaving the vanilla self-attention layers that process, \emph{local, intra-caption} information, with the \ac{SACA} layers that process \emph{global, inter-caption} information in a structure-aware manner.
Furthermore, the model acts as a classic text encoder in the absence of edges in the graph, \ie when $\vcap[\edges][\captions]=\emptyset$.

As a side note, we highlight that with SAHA, information is only propagated from each node to its direct parent within a block.
Consequently, the number of blocks must exceed the depth of the $\vcap[\graph][\captions]$ to ensure that information reaches the root node from all levels of the graph.

\vspace{0.25em}
\textbf{Complexity analysis\afterhead}\quad
To estimate the computational complexity of our approach, we assume that the captions have a fixed sequence length $\ntokens$.
Then, implementing SACA using masked cross-attentions between captions leads to a total complexity of $\bigoh(|\vcap[\edges][\captions]|\ntokens^2)$. 
Additionally, we must account for the self-attention operations, resulting in a combined complexity of $\bigoh(|\captions|\ntokens^2 + |\vcap[\edges][\captions]|\ntokens^2)$.
Provided that many of the graphs in our dataset are trees, we have $|\vcap[\edges][\captions]| = |\captions|-1$ and the complexity simplifies to $\bigoh(|\captions|\ntokens^2)$. In contrast, a naive approach that performs self-attention on the concatenated set of all captions would result in a significantly higher complexity of $\bigoh(|\captions|^2\ntokens^2)$.

\subsection{Multi-positive contrastive loss}
\label{apx:objective}
To pair multiple positive captions to an image, we extend standard contrastive loss~\cite{radford2021learning} into multiple-positive contrastive loss, as also considered in prior studies~\cite{doveh2023dense, fan2023improving}. Specifically, consider a batch of $N$ images $\{I_i\}_{i=1}^N$, where each image $I_i$ is associated with $M_i$ captions $\{T_{i,j}\}_{j=1}^{M_i}$, we utilize the following loss function to account for multiple positive texts per image:
\begin{align}
    \mathcal{L}_{\mathrm{I}} = -\frac{1}{Z} \sum_{i=1}^N \sum_{j=1}^{M_i} \log \frac{S(I_i, T_{i,j})}{ S(I_i, T_{i,j}) + \sum_{k=1, k \neq i}^{N}\sum_{l=1}^{M_k} S(I_i, T_{k,l})},
\end{align}
where $S(I, T) = \exp(\cos(I, T) / \tau)$, $\tau$ is a learnable temperature parameter, and $Z = \sum_{i=1}^N M_i$ is a normalizer. On the other hand, each caption still only has one paired image. Therefore, we use the standard contrastive loss on for text-to-image alignment:
\begin{align}
    \mathcal{L}_{\mathrm{T}} = -\frac{1}{Z} \sum_{i=1}^N \sum_{j=1}^{M_i} \log \frac{S(I_i, T_{i,j})}{ \sum_{k=1}^{N} S(I_k, T_{i,j})}.
\end{align}

\subsection{Sampling algorithm for GBC-to-image generation}
\label{apx:gbc-sampling}

In \cref{sec:exp-t2i}, we did not provide details on how the negative prompts are handled and how the images are generated when only graph and text information from GBC are available.
We explain them below.

\textbf{Negative GBC prompts\afterhead}\quad
Given a base negative prompt $\capn^{neg}$, we create a negative GBC prompt for each positive GBC prompt by preserving the same graph structure and bounding boxes, while replacing the caption of each node with $\capn^{neg}$.
For each node, we then prepend the base negative prompt with the concatenated labels of the outgoing edges $\edge_1, \ldots, \edge_k$ from that node, resulting in "$\ve[\labeling][\edge_1], \ldots, \ve[\labeling][\edge_k], \capn^{neg}$".
During sampling, we apply the same cross-attention masks based on bounding boxes and graph structure for the negative GBC prompt. 
Putting together, the above ensures that we prevent $\ve[\labeling][\edge]$ from appearing outside the region associated with the vertex that edge $\edge$ points to using the negative prompt, as mentioned in \cref{sec:exp-t2i}.

\begin{figure*}[t]
    \centering
    \includegraphics[width=0.96\textwidth]{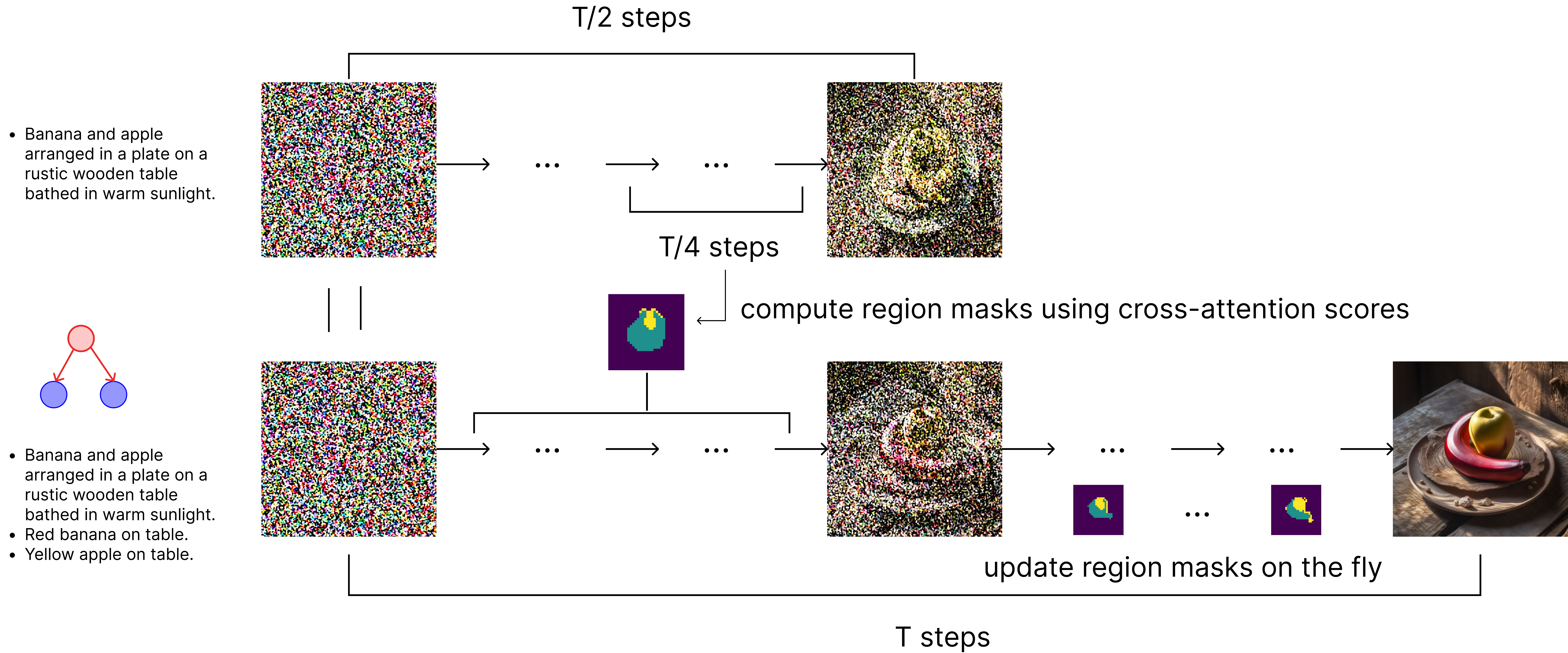}
    \caption{
    To sample images from GBC without bounding box information, we infer a segmentation map that maps image patches to graph vertices using cross-attention scores.
    We first run the sampling process for half of the total number of time steps to get the initial map.
    We then restart the process with the provided map, with the map being updated dynamically during the second half of the sampling process.
    }
    \label{fig:GBC-diff-graph}
\vspace{0em}
\end{figure*}

\textbf{Sampling with only graph and text from GBC\afterhead}\quad
Our overall procedure for sampling images from GBC without bounding box information is illustrated in \cref{fig:GBC-diff-graph}.
We restrict ourselves to star graphs, and split the images into non-overlapping segments, each corresponding to a distinct node, based on cross-attention scores.
Concretely, for each set of tokens $\ve[\positionids][\edge]$ within the global prompt that maps to a child caption, we store the cross attention scores $\ve[\attnscore][\edge]\in\R^{(\width\times\height)}$ that averages across all the tokens in $\ve[\positionids][\edge]$ and across all the cross-attention layers of the same query dimension $(\width\times \height)$.
We focus especially on the lower-resolution cross-attention maps---for SDXL with an output resolution of $1024\times1024$, this corresponds to $\width=\height=32$.

At each sampling step, we then get a set of cross-attention maps $\ve[\attnscore][\edge_1], \ldots, \ve[\attnscore][\edge_{\ncap}]$, where $\ncap$ is the number of leaves in the star graph.
For each of these maps $\ve[\attnscore][\edge]$, we regard it as an image of size $\width\times\height$ and apply
Felzenszwalb's graph-based image segmentation algorithm~\cite{felzenszwalb2004efficient} with the default hyperparameters of skimage~\cite{van2014scikit} to get the image segments $\mathcal{\segment}^{\edge}=\segment^{\edge}_{1}, \ldots, \segment^{\edge}_{m_\edge}$.
We then combine the segmentation maps to form $\segment_1, \ldots, \segment_m$, where two image patches belong to the same segment if and only if they appear together in the same segment in each of the segmentation maps $\mathcal{\segment}^{\edge_1}, \ldots, \mathcal{\segment}^{\edge_{\ncap}}$.
Finally, we merge the segments to get $\alt{\segment}_1, \ldots, \alt{\segment}_{\ncap+1}$ by assigning each segment $\segment_{i}$ to one of the $\ncap+1$ vertices.
This is achieved by using Otsu's method~\cite{otsu1975threshold} to determine a threshold for each leaf, based on the average cross-attention scores of each segment across both image patches and relevant text tokens.
If a segment is initially assigned to multiple leaves, we retain only the assignment to the leaf with the highest normalized score (all average scores are normalized to the range $[0, 1]$). 
Segments unassigned to any leaf are instead assigned to the root node.

The aforementioned segmentation provides a cross-attention mask for each time step, which we apply to both the lower-resolution ($32 \times 32$) and higher-resolution ($64 \times 64$) cross-attention layers within the UNet (assuming an output resolution of $1024 \times 1024$).
To maintain stability in the mask application, we further employ an EMA version of the masks with a momentum of $0.9$.
To generate the initial mask, we first run the sampling process using only the global prompt for $\nSteps/2$ steps, with the EMA mask calculated from step $\nSteps/4$ to $\nSteps/2$.
We then restart the process with the same initial noise, applying the obtained EMA mask during the first half of sampling.
During the second half, we update the EMA mask dynamically and apply it to the UNet at each step.

\section{Experimental details for CLIP training}
\label{apx:exp-setup}
This appendix provides additional details on the experiments presented in \cref{sec:exp}.

\subsection{Data filtering}

For the computation of CLIP score, we split any caption that contains more than 77 tokens into individual sentences, compute the score for each of these sentences, and compute the average of these scores.
Then, we start by filtering out images whose short synthetic captions have CLIP scores that are lower than the 5\% quantile.
After this, we consider three filtering strategies depending on the annotation formats.

\vspace{0.25em}
\textbf{Long caption\afterhead}\quad
In this case, we just further filter out a portion of original captions and long captions with the lowest CLIP scores (by considering the 5\% quantiles from the non-filtered dataset).

\vspace{0.25em}
\textbf{GBC\afterhead}\quad
Naive CLIP filtering and tokenizer truncation could break the graph structure as some of the edge labels would not appear in the captions of its source node anymore after these operations.
We address this issue by filtering out the captions following the reverse of a topological ordering of the graph, drop a node along with its in edges when all its captions and children get filtered, and otherwise, if necessary, add \emph{bag-of-words} captions that collects edge labels from the remaining out edges of a node to ensure all these labels still appear in some captions of this node.
Moreover, we split the captions whose length are longer than 77 tokens into concatenations of sentences that fit within this limit, and drop any caption which contains sentences that are of more than 77 tokens.

\vspace{0.25em}
\textbf{Short and Region\afterhead}\quad
We follow the strategy mentioned in GBC, but use selected types of captions.
Moreover, bag-of-words captions are not used.

We remark that the filtering procedure is only applied to the training set, and \emph{not} the GBC test set.

\subsection{Dynamic batch size}

Given the varying sizes of our graph, setting a fixed number of images per batch could result in out-of-memory errors unless we opt for a conservatively small batch size.
To overcome this challenge, we implement a dynamic batching strategy for the setups where the number of captions per image is in principle unbounded.
This encompasses notably region, GBC-captions, and GBC-graph.
With this strategy, we ensure that the number of captions, and, in the case of GBC-graph, the number of edges, that are included in each batch do not exceed a certain limit.
In this regard, the batch size that we report in \cref{sec:exp} is actually just an upper bound on the number of images included in each batch.
More specifically, we set this limit based on the number of average captions/edges per image in the filtered dataset.
For example, for GBC-captions and GBC-graph we have in average 17.61 captions per graph.
We thus set the limit on caption number to $18\times 64 = 1152$ on each GPU (as mentioned in Appx.~\ref{apx:exp-computation-cost}, we use 64 GPUs for most of our experiments, which gives a batch size of $64$ per GPU).

\subsection{Hyperparameters for CLIP training}
We used a consistent set of hyperparameters for all model training runs, as detailed in Table~\ref{tab:hparam-clip}.
The sole exception is training with original CC12M captions, where we used a larger batch size of 8,192 to ensure the model sees a comparable number of texts as during training with both short synthetic and original captions.
For this specific setup with the larger batch size, we reported evaluation results from the EMA checkpoint at the end of epoch 15, for it achieving the best performance among the evaluated checkpoints.
For GBC-graph, we drop the edges with probability $0.5$ so that the model also learns how to match images with short captions.

\begin{table}[t]
    \centering
    \small
    \begin{NiceTabular}{l|cccccc}[colortbl-like]
    \toprule
        &
        Short
        &
        Long
        &
        Region
        &
        GBC-captions
        &
        GBC-concat
        &
        GBC-graph
        \\
    \midrule
        Training time (hr) & 22.7 & 19.7 & 29.9 & 38.9 & 20.3 & 42.0 \\
    \bottomrule
    \end{NiceTabular}
    \vspace{0.25em}
    \caption{CLIP model training time for 45,000 iterations with different annotation formats.
    }
    \label{tab:clip-compute}
\end{table}

\begin{table}[t]
    \small
    \centering
    \begin{minipage}{0.47\linewidth}
\centering
\scalebox{0.95}{
\begin{tabular}{l|c}
\toprule
Hyperparameters & Values \\
\midrule
Data augmentation & RRC  \\
Crop size & 224$\times$224  \\
Train iterations & 45k \\
Global batch size & 4,096 \\
Optimizer & AdamW \\
Min / max learning rate & \{1e-6, 1e-3\} \\
LR. decay schedule type & Cosine \\
Warmup iterations & 1,000 \\
Weight decay rate & 0.05 \\
EMA factor & 0.9995 \\
\bottomrule
\end{tabular}
}
\vspace{0.25em}
\caption{Hyperparameters for CLIP model training. RRC stands for \texttt{RandomResizedCrop}}
\label{tab:hparam-clip}
    \end{minipage}
    \hfill 
\begin{minipage}{0.47\linewidth}
\centering
\scalebox{0.95}{
\begin{tabular}{l|c}
\toprule
 Hyperparameters & Values \\
\midrule
Data augmentation & RRC  \\
Crop size & 512$\times$512  \\
Train iterations & 160k \\
Global batch size & 16 \\
Optimizer & AdamW \\
Peak learning rate &
\makecell{[5e-4, 2e-4,\\ 1e-4, 7e-5, 5e-5]} \\
LR. decay schedule type & Polynomial \\
Warmup iterations & 1,500 \\
Weight decay rate & 0.01 \\
\bottomrule
\end{tabular}
}
\vspace{0.25em}
\caption{Training hyperparameters for semantic segmentation experiments on ADE20k. RRC stands for \texttt{RandomResizedCrop}.}
\label{tab:hyperparams_segm}
    \end{minipage}
\end{table}

\subsection{Computation cost}
\label{apx:exp-computation-cost}
We train all CLIP models on A100-80G GPUs. As training with different annotation formats requires varying size of GPU memory, we use different total numbers of GPUs to ensure the same batch size. Specifically, we utilize 16 GPUs for training with \textit{Short} captions, and utilize 64 GPUs for training with all other annotation formats. We list the corresponding  time required for training with different annotation formats in \cref{tab:clip-compute}.

\subsection{Evaluation details}

Our evaluation uses the validation set of ImageNet-1k~\cite{russakovsky2015imagenet} and the test sets of Flickr30k~\cite{plummer2015flickr30k} and MS-COCO~\cite{lin2014microsoft}.
For SugarCrepe~\cite{hsieh2024sugarcrepe} we report the average performance across all variants.
As for ShareGPT4V~\cite{zhou2019semantic}, we use a subset of size $15,295$ from ShareGPT4V-cap100k.
These images were also used for LLaVA training.\footnote{\url{https://huggingface.co/datasets/liuhaotian/LLaVA-Pretrain} ~~Accessed: 2024-05-14}
When each image is paired with multiple captions, we only select one of them.
The evaluation setups with ADE20K and DCI are more involved, as we explain below.

\vspace{0.25em} 
\textbf{Evaluation on ADE20K\afterhead}\quad
We evaluate the quality of CLIP models' image encoder for dense prediction tasks like image segmentation by performing full finetuning on ADE20k~\cite{zhou2019semantic} dataset. We follow the same setup as described in~\cite{fastvit, clipft2024} where we use a ViTDet style feature pyramid network with UperNet~\cite{upernet} head. All models were trained using the MMSegmentation library~\cite{mmseg2020}. We sweep through peak learning rate for all the results reported in the paper and the ranges are listed in~\Cref{tab:hyperparams_segm}.

\vspace{0.25em}
\textbf{Evaluation on DCI\afterhead}\quad
We perform text-to-image and image-to-text evaluations on DCI~\cite{urbanek2023picture} using either long captions or concatenated captions.
The long captions are marked as \texttt{extra\_caption} in the released DCI dataset.
We filter out samples with empty long captions, resulting in a subset of 7,602 images for evaluation with long captions.
Regarding evaluation with concatenated captions, we leverage the full set of 7,805 images.
We retain masks containing summary captions (these are masks with bounding boxes larger than $224\times 224$). 
If the human-annotated caption contains fewer than 77 tokens and is longer than the first summary caption, we use it.
Otherwise, we use the first summary caption.
For concatenation,  we follow the \ac{BFS} order based on the provided tree structure between the masks.

\section{Additional results and experiments for CLIP training}
\label{apx:exp-add}
In this appendix, we present additional experiments that we performed but did not present in the main paper due to space constraints.

\subsection{Retrieval with multiple captions using maximum CLIP score}
An alternative to the mean CLIP score we considered in \cref{tab:gbc-eval} is to take the maximum, for which we report the results in \cref{tab:gbc-eval-clip-max}.
Compared to taking the average, using the maximum is more robust to low CLIP scores, and thus gives better results when the model is not trained to match the image with all local captions, as with Short, Long caption, and GBC-concat.
\footnote{In GBC-graph, the entire graph is encoded, so we do pair the image with all the captions during training.}
Nonetheless, despite these differences, the overall retrieval performance still significantly lags behind that achieved using a single caption.

\begin{table*}[t]
    \small
    \centering
    \begin{minipage}{0.33\linewidth}
    \centering
    \begin{NiceTabular}{l|cc}[colortbl-like]
    \toprule
    Annotation & T2I & I2T \\
    \midrule
    Short               & 71.5 & 78.1 \\
    Long                & 72.9 & 80.1 \\[0.05em]
    \vphantom{\rule{0pt}{1em}}%
    \rowcolor{TableRowHighlight}GBC-captions        & 86.4 & 87.3 \\
    \rowcolor{TableRowHighlight}GBC-concat          & 73.3 & 79.7 \\
    \rowcolor{TableRowHighlight}GBC-graph           & 85.2 & 85.9 \\
    \bottomrule
    \end{NiceTabular}
    \vspace{0.25em}
    \caption{Image and text retrieval performance on GBC test when using max CLIP score over all the captions.}
    \label{tab:gbc-eval-clip-max}
    \end{minipage}
    \hfill 
    \begin{minipage}{0.63\linewidth}
    \centering
    \begin{NiceTabular}{l|cc|cc|cc}[colortbl-like]
    \toprule
        \multirow{2}{*}[-0.25em]{Annotation}
        &
        \multicolumn{2}{c}{DCI-Long}
        &
        \multicolumn{2}{c}{DCI-concat}
        &
        \multicolumn{2}{c}{ShareGPT4V-15k}
        \\
        \cmidrule(lr){2-3}
        \cmidrule(lr){4-5}
        \cmidrule(lr){6-7}
        & T2I
        & I2T
        & T2I
        & I2T
        & T2I
        & I2T
        \\
    \midrule
        Long & 53.6 & 53.3 & 63.3 & 65.8 & \textbf{93.4} & \textbf{93.9} \\[0.05em]
        \vphantom{\rule{0pt}{1em}}%
        \rowcolor{TableRowHighlight}GBC-captions & 42.5 & 43.3 & 64.3 & 63.8 & 78.7 & 82.1 \\
        \rowcolor{TableRowHighlight}GBC-concat & 51.4 & 52.3 & \textbf{69.0} & \textbf{70.8} & 89.5 & 91.4  \\
    \bottomrule
    \end{NiceTabular}
    \vspace{0.25em}
    \caption{Image and text retrieval performance on DCI and a 15k subset of ShareGPT4V-cap100k of our models trained on longer captions. We also include GBC-captions that by design can only handle short captions as a baseline for comparison.
    }
    \label{tab:dci-eval}
    \end{minipage}
\vspace{-1em}
\end{table*}

\subsection{Retrieval with long captions} 
To complement the results presented in \cref{tab:standard-eval},
we evaluate the retrieval performance of our extended context models on datasets with richer annotations.
We focus specifically on ShareGPT4V~\cite{chen2023sharegpt4v}, which offers GPT-style detailed captions closely resembling those obtained from LLaVA, and DCI~\cite{urbanek2023picture}, containing human-annotated detailed and region captions.
The latter allows us to perform retrieval using either detailed captions or concatenated short captions, as we did in Section~\ref{subsec:gbc-test}.
Our results shown in \cref{tab:dci-eval} demonstrates that the close caption distribution with ShareGPT4V effectively enables strong retrieval results for models trained on our long captions.
However, potentially due to the distribution shift, all the models perform badly on DCI retrieval with long captions.
In this setup, using concatenated captions for training and retrieval significantly outperformed other baselines, indicating the broader benefit of the concatenation approach.

\begin{table}[t]
    \centering
    \small
    \renewcommand{\arraystretch}{1.1}
    \setlength\tabcolsep{0.5em}
    \begin{NiceTabular}{l|ccc|cccccc}[colortbl-like]
    \toprule
    \multirow{2}{*}[-0.85em]{Annotation}
    &
    \multicolumn{3}{c}{Hyperparameter}
    &
    \multicolumn{6}{c}{Evaluation results}
    \\[0.3em]
    \cmidrule(lr){2-4}
    \cmidrule(lr){5-10}
        &
        Epoch
        &
        Batch size
        &
        \# Tokens
        &
        ImageNet
        &
        Flickr
        &
        COCO
        &
        \makecell[c]{Share-\\GPT4V}
        &
        \makecell[c]{DCI-\\concat}
        &
        \makecell[c]{GBC\\test}
        \\
    \midrule
    \multirow{5}{*}[-0.4em]{Short}
    & \multirow{2}{*}[-0.25em]{10} & \multirow{2}{*}[-0.25em]{4,096} & 77
    & 38.8 & 64.8 & 38.7 & 79.1 & 57.5 & 87.8
    \\
    \cmidrule(lr){4-4}
    &  &  & 512 
    & 39.0 & 64.7 & \underline{39.3} & \underline{86.7} & 56.4 & \underline{89.7}
    \\
    \cmidrule(lr){2-4}
    & 10 & & 
    & 33.2 & 59.1 & 34.7 & 74.0 & 52.3 & 83.4
    \\
    & 28 & 16,384 & 77
    & \underline{40.0} & 67.4 & 38.7 & 80.6 & \underline{58.1} & 88.6 
    \\
    & 40 &  &
    & 39.0 & 65.7 & 37.3 & 79.9 & 57.8 & 88.6
    \\[0.1em]
    \specialrule{0.05pt}{0.05pt}{-0.05pt}
    \vphantom{\rule{0pt}{1.1em}}%
    \rowcolor{TableRowHighlight}GBC-captions
    & & &
    & \textbf{40.8} & \textbf{70.0} & \textbf{43.0} & 80.4 & 64.1 & 91.2
    \\
    \rowcolor{TableRowHighlight}GBC-concat
    & 10 & 4,096 & 77
    & 39.0 & 66.1 & 40.0 & \textbf{90.5} & \textbf{69.9} & 94.8
    \\ 
    \rowcolor{TableRowHighlight}GBC-graph 
    & & &
    & 38.4 & \underline{67.5} & 40.6 & 78.0 & 61.3 & \textbf{96.0}
    \\
    \bottomrule
    \end{NiceTabular}
    \vspace{0.25em}
    \caption{Comparative performance across various benchmarks when we perform CLIP training on short captions with different hyperparameters.
    For ease of reference, we also include the results from the methods that use GBC annotations.
    We report the average image and text Recall@1 for all retrieval benchmarks.
    Specifically, as explained in \cref{subsec:gbc-test}, we perform retrieval using various annotation formats for GBC test.
    We thus report here the average of the \emph{highest} image and text Recall@1 scores.
    The number of iterations is consistently set at 45,000, corresponding to 20 epochs with a batch size of 4,096 and 76 epochs with a batch size of 16,384.
    }
    \label{tab:eval-short}
\vspace{-1em}
\end{table}

\subsection{Matching compute resource for training with short captions}

All our models presented in \cref{sec:exp} used 8 nodes for training, except for the models trained on short captions, which only used 2 nodes.
This raises the question of whether the performance gap could be bridged by providing more computational resources to this setup.
To address this, we specifically considered two modifications that would naturally necessitate using more nodes for training with short captions: \emph{(i)} extending the context length to 512, as done for training with Long and GBC-concat captions, and \emph{(ii)} using a batch size that is four times larger, \ie a batch size of 16,384 instead of 4,096.
All other hyperparameters remained unchanged.
We then trained the models on 8 nodes, each with 8 GPUs, as in the other setups, which resulted in training times of 18 and 48 hours for the two modifications respectively.
The evaluation results are presented in \cref{tab:eval-short}.

\paragraph{Training with extended context length.}
Provided that the models are only trained with short captions, we do not expect any tangible benefit from extending the context length.
Yet, surprisingly, while this is indeed the case for classic benchmarks such as ImageNet, Flickr, and COCO, we do observe a significant performance boost on ShareGPT4V retrieval, suggesting that the longer context length is still beneficial for retrieval with long caption even though the model is not explicitly trained for this task.
On the other hand, we do not observe any benefit when evaluated using concatenated caption from DCI.
Finally, we also get a slight performance improvement on GBC test, and it turns out this improved performance is achieved by performing retrieval using the long caption.
This is in line with the performance gain that we observe for the ShareGPT4V benchmark.

\paragraph{Training with larger batch size.}
More interestingly, CLIP is known to perform better when trained with a large batch size, so 
we might be able to bridge the performance gap by simply including more images and captions in each batch.
To enable a fair comparison for this setup, we report evaluation results from three checkpoints at varying training stages in \cref{tab:eval-short}. These checkpoints are chosen to align with key training milestones.
\begin{itemize}
\item \textbf{Number of images seen:} We consider the EMA checkpoint at the end of epoch 10 to align the number of images seen.
\item \textbf{Number of iterations:} We include the EMA checkpoint at the end of epoch 40 to compare models at a fixed number of training iterations.
\item \textbf{Best performing checkpoint:} Additionally, we report results for the EMA checkpoint at the end of epoch 28, as it gives the best performance among all evaluated checkpoints (see Figure~\ref{fig:eval-ema}).
\end{itemize}

As we can see from the table, while the use of a larger batch size indeed leads to better performance on ImageNet and Flickr, the results still lag behind those achieved with GBC-captions. This discrepancy underscores the \textbf{importance of including multiple captions per image to enhance performance}.

\subsection{The importance of multi-positive contrastive loss}

\begin{table}[t]
    \centering
    \small
    \begin{NiceTabular}{l|ccccc}[colortbl-like]
    \toprule
        Annotation
        &
        ImageNet
        &
        Flickr-1k
        &
        MSCOCO-5k
        &
        SugarCrepe
        &
        Average Drop
        \\
    \midrule
        Short & 
        38.8 $\rightarrow$ 35.2 &
        64.8 $\rightarrow$ 61.0 & 38.7 $\rightarrow$ 36.7 &  76.0 $\rightarrow$ 74.4 & -2.75 \\
        Long & 
        39.6 $\rightarrow$ 30.5 &
        65.8 $\rightarrow$ 56.8 & 40.1 $\rightarrow$ 33.5 &  77.0 $\rightarrow$ 74.0 & -6.93 
        \\[0.05em]
        \vphantom{\rule{0pt}{1em}}%
        \rowcolor{TableRowHighlight}GBC-captions 
        & 40.8 $\rightarrow$ 31.9
        & 70.0 $\rightarrow$ 58.3 & 43.0 $\rightarrow$ 33.2 &  76.7 $\rightarrow$ 73.3 & -8.45\\
    \bottomrule
    \end{NiceTabular}
    \vspace{0.25em}
    \caption{Performance degradation across different annotation types when switching from multi-positive contrastive loss to standard contrastive loss with randomly sampled positive captions.
    For Flickr-1k and MSCOCO-5k we report the average image and text Recall@1.
    }
    \label{tab:eval-no-sampled}
\end{table}

We next look into the influence of the objective function when an image is paired with multiple captions.
Instead of employing the multi-positive contrastive loss introduced in \cref{apx:objective}, we can use a standard contrastive loss with a single randomly sampled caption paired with each image.
\cref{tab:eval-no-sampled} presents the evaluation results for both the models trained with the original objective (left side of the arrow), and this new, sampled, objective (right side of the arrow).

The table clearly shows a performance decline across all the considered annotation formats and benchmarks when sampling is applied, as also observed by \citet{doveh2023dense} and \citet{fan2023improving}.
The performance drop is particularly important when the captions vary significantly (e.g., long versus short captions, or image versus region captions), and when many captions are involved.
More surprisingly, this alternative loss does not lead to improvement but rather to performance degradation when we increase the number of captions paired with each image.
We conjecture this is because the additional captions that we consider here are less relevant for these specific benchmarks, leading to a worse performance when they are forced to be treated as positive in the sampled objective.

Overall, these results confirm \textbf{the importance of our multi-positive contrastive loss in leveraging the presence of multiple captions for an image.}

\subsection{Impact of caption type on CLIP training}
\label{apx:caption-type-influence}

\begin{table}[t]
    \centering
    \small
    \begin{NiceTabular}{l|cc|cc|cc|c|c|c}[colortbl-like]
    \toprule
        \multirow{2}{*}[-0.25em]{Annotation}
        &
        \multicolumn{2}{c|}{Flickr-1k}
        &
        \multicolumn{2}{c|}{MSCOCO-5k}
        &
        \multicolumn{2}{c|}{DCI-concat}
        &
        \multirow{2}{*}[-0.25em]{ImageNet}
        &
        \multirow{2}{*}[-0.25em]{SugarCrepe}
        &
        \multirow{2}{*}[-0.25em]{ADE20K}
        \\
        \cmidrule(lr){2-3}
        \cmidrule(lr){4-5}
        \cmidrule(lr){6-7}
        & T2I
        & I2T
        & T2I
        & I2T
        & T2I
        & I2T
        & 
        &
        & %
        \\
    \midrule
        Short 
        & 56.3 & 73.2 & 30.7 & 46.7 
        & 57.5 & 57.5
        & 38.8 & 76.0 & 42.0
        \\
        Region 
        & 58.3 & \underline{76.6} & 31.5 & 49.1 
        & 61.8 & 61.5
        & 38.5 & 75.6 & 43.5 
        \\[0.05em]
        \vphantom{\rule{0pt}{1em}}%
        \rowcolor{TableRowHighlight}GBC-relation
        & \underline{60.4} & 76.5 & \textbf{34.8} & \textbf{52.5}
        & 62.0 & 61.4 
        & \textbf{41.5} & \underline{76.4} & \underline{44.5} 
        \\
        \rowcolor{TableRowHighlight}GBC-captions & \textbf{60.6} & \textbf{79.3} & \underline{34.1} & \underline{51.9}
        & \textbf{64.1} & \textbf{63.4}
        & \underline{40.8} & \textbf{76.7} & \textbf{45.0}
        \\
    \bottomrule
    \end{NiceTabular}
    \vspace{0.25em}
    \caption{Comparative performance on various existing benchmarks when trained using different subsets of GBC-captions.
    }
    \label{tab:caption-type-influence}
\vspace{-1em}
\end{table}

To further highlight the value of relation and composition captions from GBC, we trained a CLIP model using only these captions alongside short image captions.
As shown in the third row of \cref{tab:caption-type-influence}, these captions, despite being more than twice as scarce as region captions, not only provided a larger performance gain than using only region captions, but sometimes even enabled the model to achieve comparable or better performance than using all captions combined.
This underscores \textbf{the significant benefit of the relational captions from GBC datasets}.

Looking closely, we note that region captions primarily benefit retrieval and dense prediction tasks, while relation and composition captions improve performance across the board. 
While using all captions remains the best approach for most benchmarks, the marginal improvement from region captions hints at the potential for more efficient training with these captions through alternative training objectives.

\subsection{Impact of the underlying graph on retrieval}

\begin{table}[t]
    \centering
    \small
    \renewcommand{\arraystretch}{1.1}
    \setlength\tabcolsep{0.7em}
    \begin{NiceTabular}{l|cc|cccccccccc}[colortbl-like]
    \toprule
        \multirow{3}{*}[-0.24em]{Annotation}
        &
        \multicolumn{2}{c}{Short}
        &
        \multicolumn{6}{c}{GBC-graph}
        &
        \multicolumn{2}{c}{Star graph}
        &
        \multicolumn{2}{c}{Line graph}
        \\
        \cmidrule(lr){2-3}
        \cmidrule(lr){4-9}
        \cmidrule(lr){10-11}
        \cmidrule(lr){12-13}
        & \multirow{2}{*}[-0.24em]{T2I}
        & \multirow{2}{*}[-0.24em]{I2T}
        & \multicolumn{2}{c}{Groundtruth}
        & \multicolumn{2}{c|}{Last token}
        & \multicolumn{6}{c}{Random token}
        \\
        & 
        & 
        & T2I
        & I2T
        & T2I
        & \multicolumn{1}{c|}{I2T}
        & T2I
        & I2T
        & T2I
        & I2T
        & T2I
        & I2T
        \\
    \midrule
        \rowcolor{TableRowHighlight}
        \vphantom{\rule{0pt}{1.05em}}
        GBC-graph  & 84.8 & 85.7 
        & \textbf{95.9} & \textbf{96.1}
        & 93.1 & 93.6
        & \underline{95.2} & \underline{95.7}
        & \underline{94.7} & \underline{95.0}
        & 92.2 & 92.4
        \\[0.1em]
    \bottomrule
    \end{NiceTabular}
    \vspace{0.25em}
    \caption{Image and text retrieval performance on the GBC test set when the model is trained using GBC-graph and evaluated across various underlying graph structures.}
    \label{tab:gbc-wrong-graph}
\end{table}

In this part, we investigate how much GBC-graph relies on the underlying graph structure for retrieval.
For this, we probe the performance of our model when the graph is modified either in the mapped tokens or in the connectivity patterns.
In terms of the mapped tokens, we consider
\begin{itemize}
    \item Last token: For any edge from a caption $\capn$ to another caption $\capnalt$, we mapped the information of $\capnalt$ to the last token before the summary token in $\capn$.\footnote{We also experimented with mapping the information to the summary token but this completely destroys the performance.}
    \item Random token: For each edge, we randomly map the information to one token in the source caption.
\end{itemize}
As for the connectivity pattern, we investigate
\begin{itemize}
    \item Star graph: All the captions are mapped to the short image synthetic caption.
    \item Line graph: We map each caption to its next caption in a list (ordered as in GBC-concat following the BFS order), with the short image synthetic caption being the first in the list.
\end{itemize}

The results are shown in \cref{tab:gbc-wrong-graph}.
Since random-token mapping consistently leads to better result than last-token mapping, we only report results for this in the case of star graph and line graph.
First of all, we observe that no matter which graph is given, we always achieve better performance than retrieval with only short caption, suggesting that the model is always able to exploit the additional captions to some extent.
Furthermore, employing random-token mapping, whether with the groundtruth graph topology or the star graph, yields performance that closely matches that of using the groundtruth graph with correct mapping (interestingly, when using star graph the performance is also very close to that obtained with GBC-concat, see \cref{tab:gbc-eval}).
This suggests that the specific retrieval task we are examining is not highly dependent on the provided mapping and topology.
However, we do believe the mapping and topology could play a significant role in other tasks or when more fine-grained distinctions between images are necessary.

\subsection{Evaluating at non-EMA checkpoints}

\begin{table}[t]
    \centering
    \small
    \begin{NiceTabular}{l|cccccc}[colortbl-like]
    \toprule
        Annotation
        &
        ImageNet
        &
        Flickr-1k
        &
        MSCOCO-5k
        &
        SugarCrepe
        &
        ShareGPT4V-15k
        &
        GBC test
        \\
    \midrule
        CC12M & 37.1 & 50.5 & 27.9 & 39.4 & 47.2 &  49.5 \\
    \midrule
        Short & 37.5 & 62.0 & 36.1 & 74.5 & 77.3 & 86.9 \\
        Long & 38.5 & 64.5 & 38.4 & 75.8 & \textbf{93.5} & 95.5
        \\
        Region & \underline{40.3} & \underline{68.6} & \underline{40.8} & \textbf{76.1} & 78.9 & 91.8
        \\[0.05em]
        \vphantom{\rule{0pt}{1em}}%
        \rowcolor{TableRowHighlight}GBC-captions
        & \textbf{41.3}
        & \textbf{70.6}
        & \textbf{43.1}
        & \textbf{76.4}
        & 80.1 & 91.9 \\
        \rowcolor{TableRowHighlight} GBC-concat
        & 38.2 & 63.4 & 37.1 & 74.9 & \underline{89.8} & \textbf{96.1} \\
        \rowcolor{TableRowHighlight} GBC-graph &
        39.7 & 68.1
        & \underline{40.8} & 75.3 & 78.2 & \textbf{96.2} \\
    \bottomrule
    \end{NiceTabular}
    \vspace{0.25em}
    \caption{Comparative performance on various existing benchmarks when trained using different annotation schemes.
    Unlike the other tables that report performance for EMA checkpoints, this table presents the performance at the final non-EMA checkpoints obtained from the end of training.
    }
    \label{tab:eval-non-ema}
\vspace{-1.5em}
\end{table}

\begin{figure}[p]
    \centering
    \includegraphics[width=0.42\textwidth]{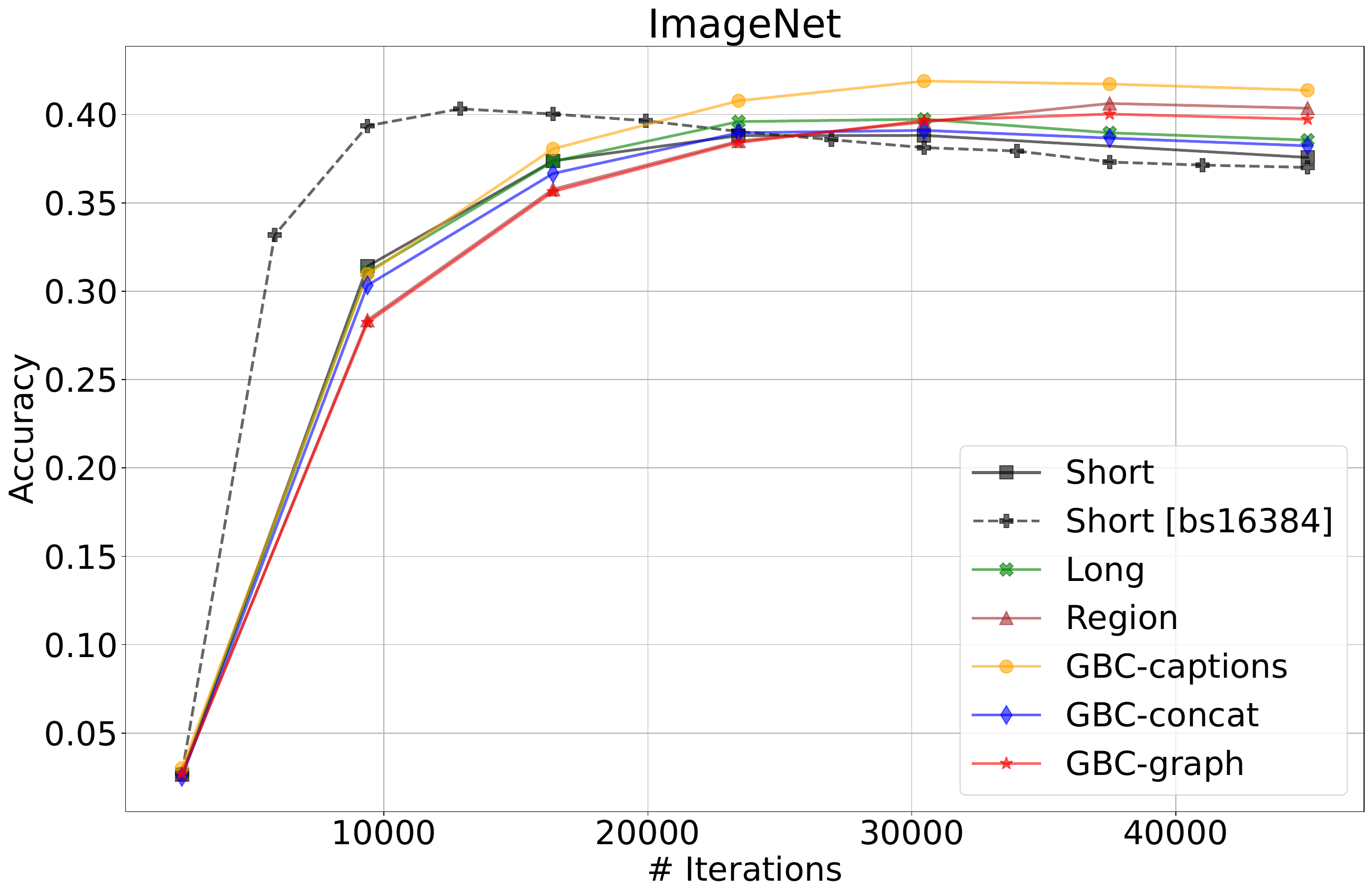}
    \includegraphics[width=0.42\textwidth]{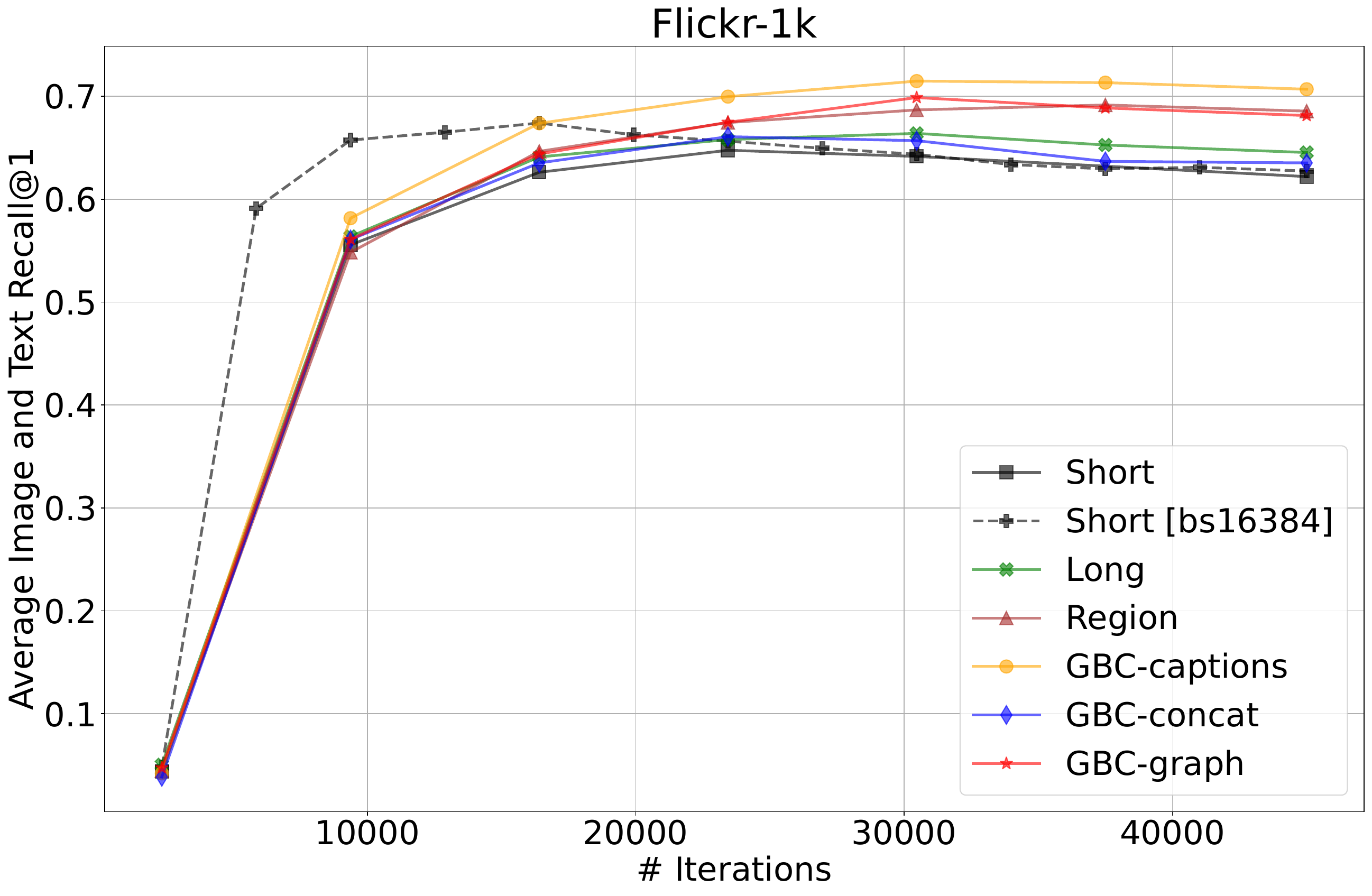}
    \includegraphics[width=0.42\textwidth]{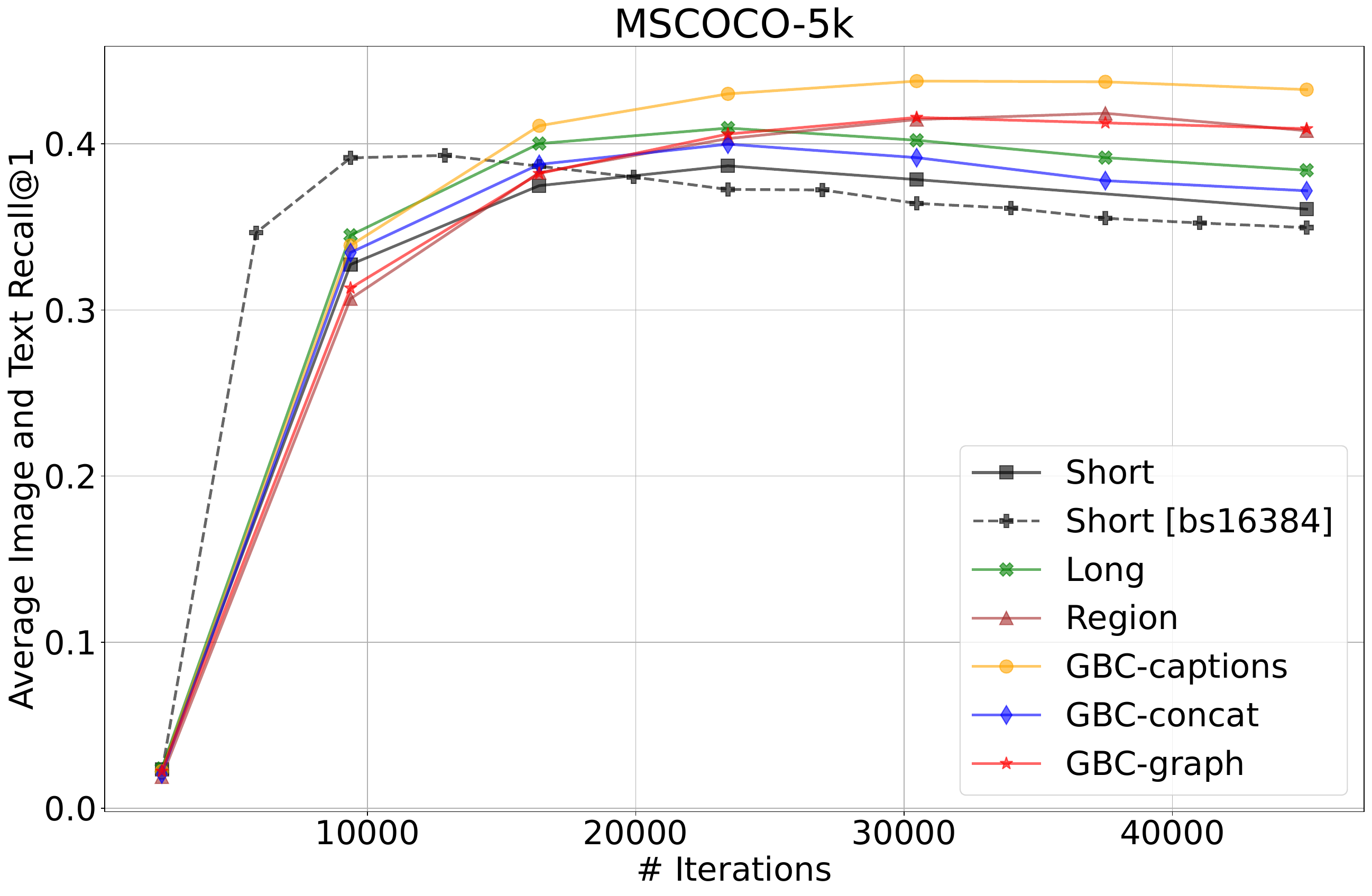}
    \includegraphics[width=0.42\textwidth]{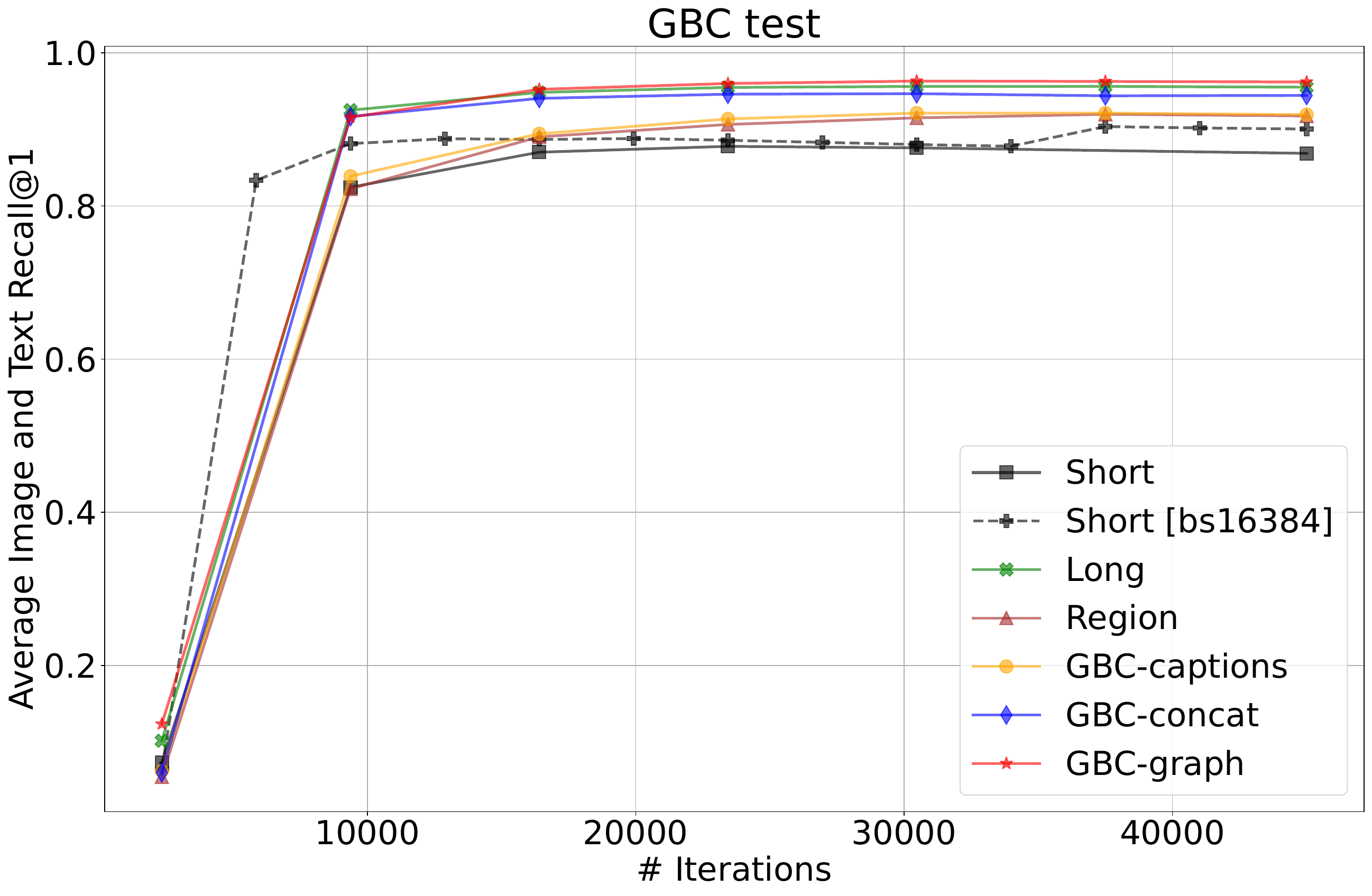}
    \caption{Benchmark performances on ImageNet, Flickr-1k, MSCOCO-5k, and GBC test for EMA checkpoints of models trained with different annotations / hyperparameters.
    For GBC test we use different formats for retrieval at test time and average the highest scores that are respectively obtained for text-to-image and image-to-text retrievals.
    }
    \label{fig:eval-ema}
\end{figure}

\begin{figure}
    \centering
    \includegraphics[width=0.42\textwidth]{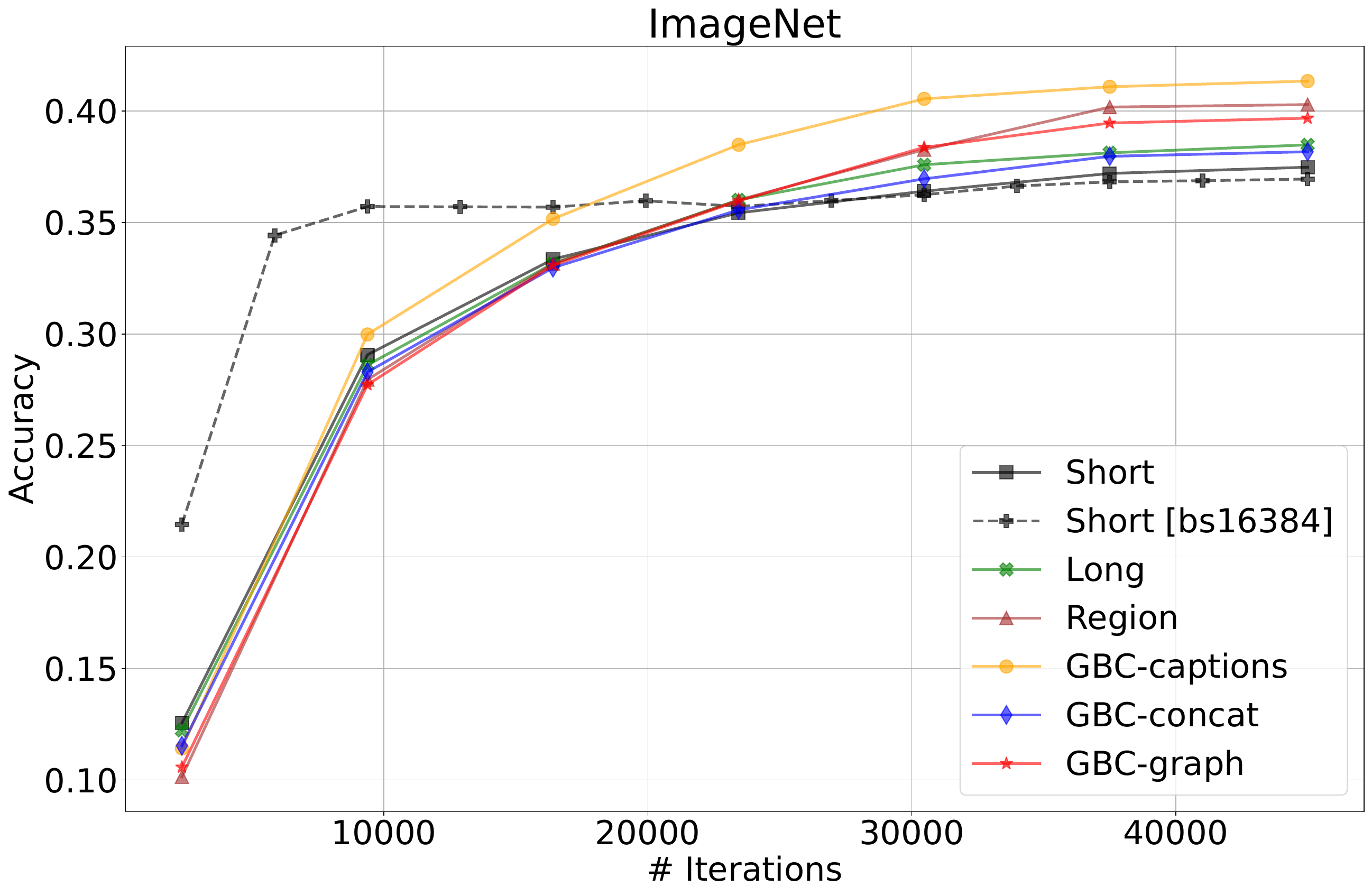}
    \includegraphics[width=0.42\textwidth]{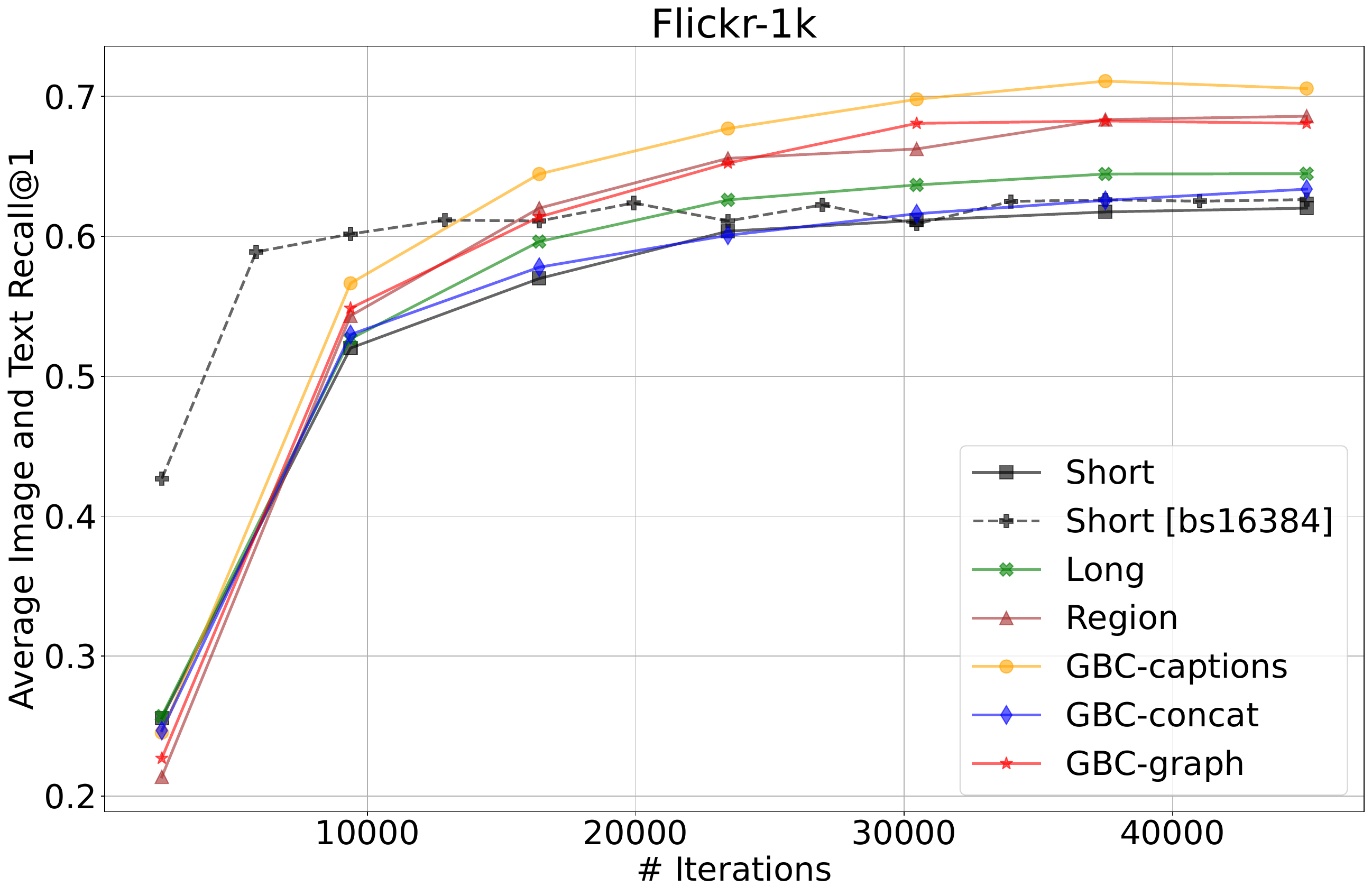}
    \includegraphics[width=0.42\textwidth]{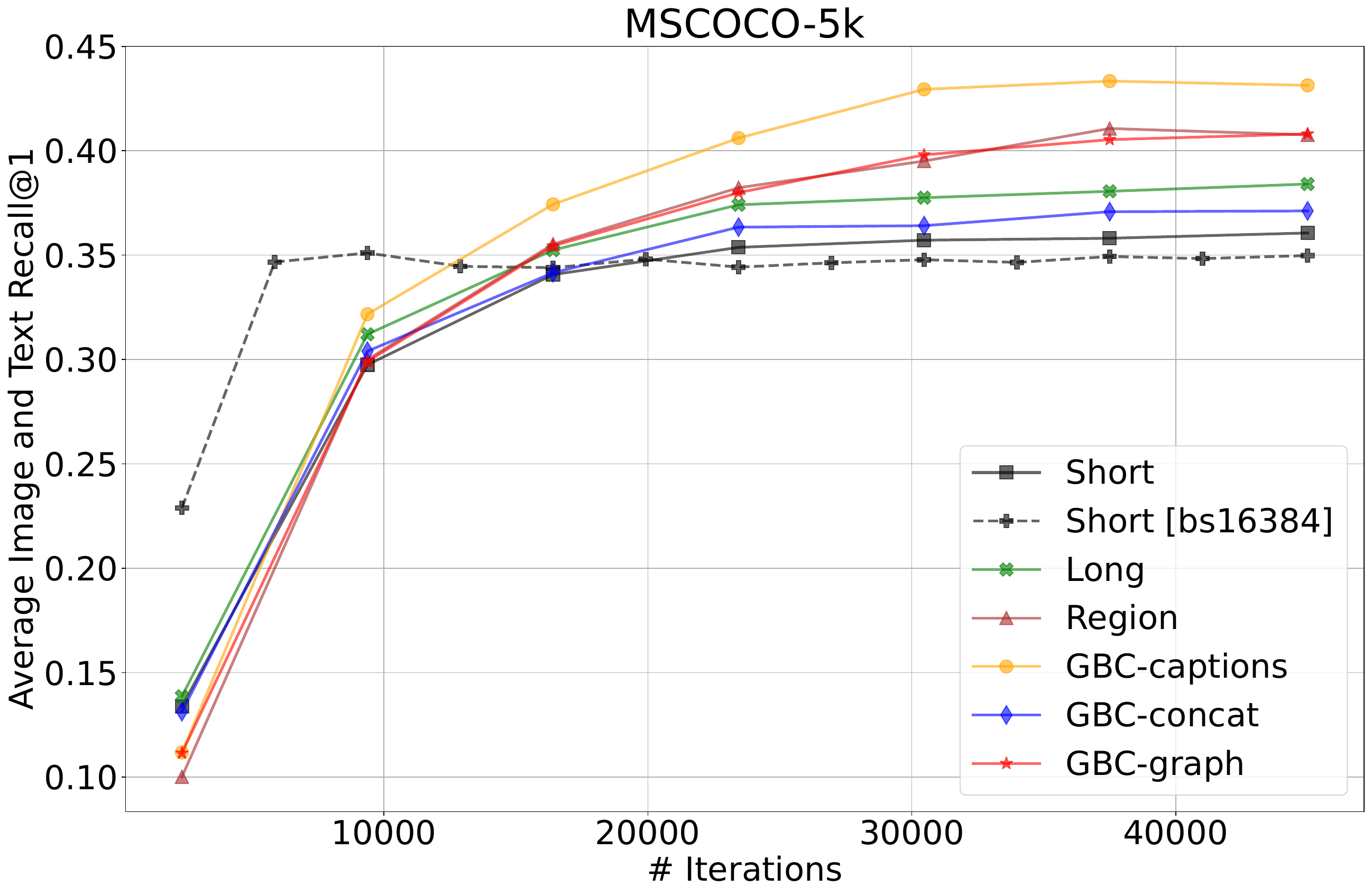}
    \includegraphics[width=0.42\textwidth]{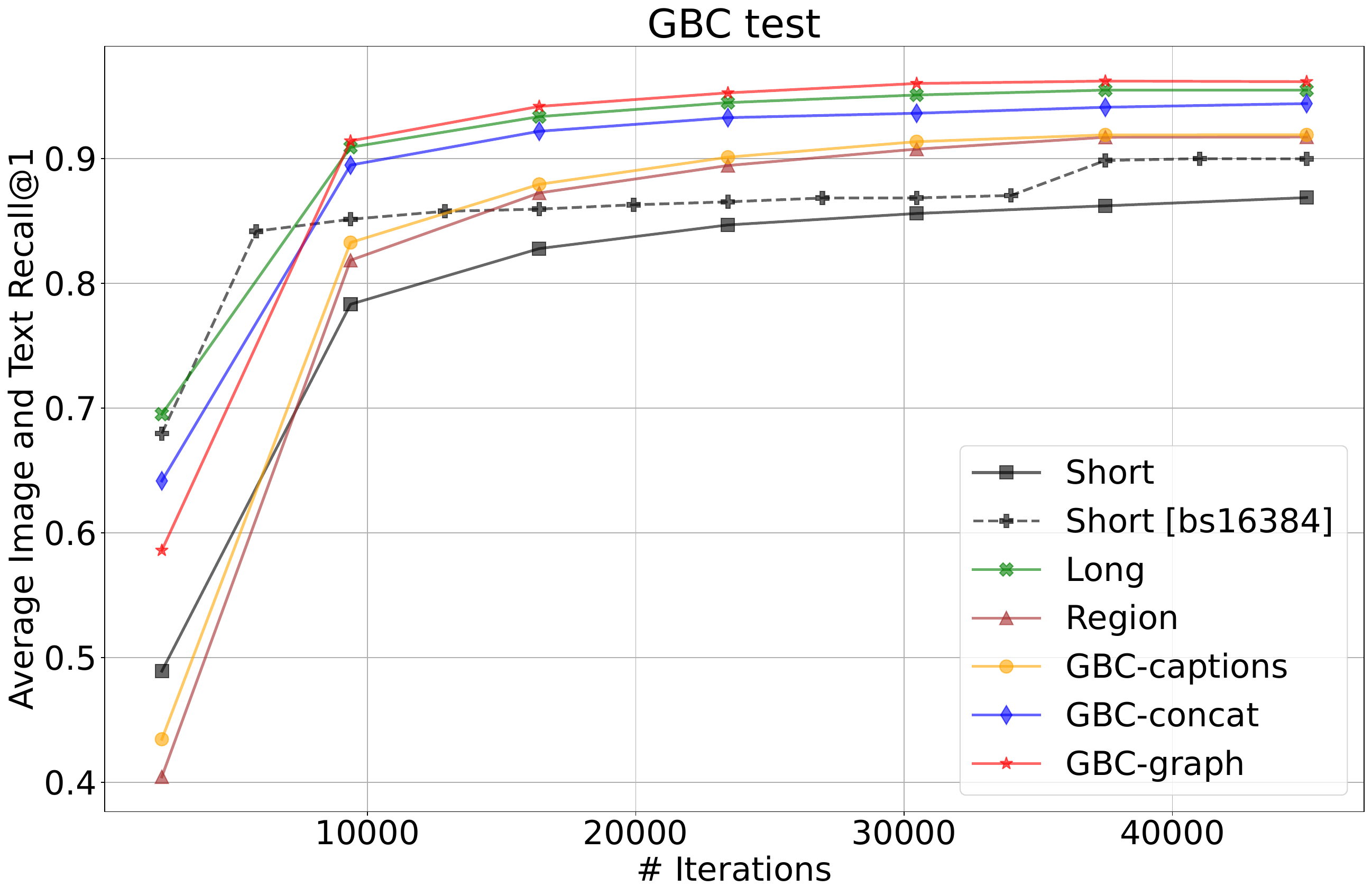}
    \caption{Benchmark performances on ImageNet, Flickr-1k, MSCOCO-5k, and GBC test for \emph{non}-EMA checkpoints of models trained with different annotations / hyperparameters.
    For GBC test we use different formats for retrieval at test time and average the highest scores that are respectively obtained for text-to-image and image-to-text retrievals.
    }
    \label{fig:eval-non-ema}
\end{figure}

For the sake of completeness, we also perform evaluation on the non-EMA checkpoints, with results shown in \cref{tab:eval-non-ema,fig:eval-non-ema}.
Comparing \cref{fig:eval-ema} with \cref{fig:eval-non-ema}, we see that while EMA checkpoints may experience a drop in performance during later training stages, non-EMA checkpoints typically exhibit best performance at the final training checkpoint.
Consequently, our evaluations in \cref{tab:eval-non-ema} are based on these last checkpoints.
From the evaluation results, we observe a similar trend in the performance comparison of annotation formats with non-EMA checkpoints as with EMA ones, confirming the validity of our previous claims.
Finally, we also note that the use of larger batch size when training with short captions is only beneficial when we consider EMA checkpoints.


\section{Experimental details for text-to-image generation}
\label{apx:exp-setup-t2i}
This appendix provides additional details on the experiments presented in \cref{sec:exp}.

\subsection{Text-to-GBC}

We provide below detailed information on data processing and model training for our GBC prompt generation model.

\paragraph{Data format and dataset.} 
As illustrated in \cref{fig:Text-to-GBC}, each node in our GBC graph is encoded in plain text as follows

\begin{verbatim}
Node #<id> <name>
type: <type>
is_leave: <True|False>
desc: <description>
parents: #<parent_id>(<parent_name>: <name>)
bbox: <bbox>
\end{verbatim}

\noindent
The elements of the above format are defined as:
\begin{itemize}
    \item \texttt{<id>} is a unique numerical identifier for the node.
    \item \texttt{<name>} is the node's unique string identifier as assigned in our dataset.
    \item \texttt{<type>} indicates the node type (image, entity, composition, or relation).
    \item \texttt{<is\_leave>} specifies whether the node is a leaf node.
    \item \texttt{<description>} contains the textual description of the node content. For image node we use the short caption while for the remaining we use the first caption stored in the node.
    \item \texttt{<parent\_id>} and \texttt{<parent\_name>} identify the parent node(s).
    \item \texttt{<bbox>} represents the bounding box coordinates.
\end{itemize}

\vspace{1em}
Our training dataset is derived from GBC10M by processing each graph $G=(V,E)$ into a sequence of node descriptions encoded in the aforementioned format.
Moreover, we ensure that this node sequence forms a topological order of the original graph so that the hierarchical structure of the graph is respected.
The numerical identifier \texttt{<id>} of each node naturally corresponds to its position in this sequence.
We also remove all relation nodes and composition captions to align with our GBC-to-image generation pipeline.

\paragraph{Training Configuration.}
Our text-to-GBC model is built upon TIPO-200M~\cite{tipo2024yeh}. This lightweight architecture ensures efficient processing while maintaining high-quality graph generation capabilities.
We train the model using the scheduler-free AdamW optimizer~\cite{loshchilov2018decoupled,defazio2024road} for one epoch, which amounts to approximately 20,000 steps with a fixed global batch size of 512.
The learning rate is set to 5e-5, with a warmup period of 100 steps and a weight decay of 0.1. We use $\beta_1=0.9$, $\beta_2=0.99$ for AdamW, and limit the context length to a maximum of 4,096 tokens.

\subsection{GBC-to-image}

For sampling from SDXL, we use Euler sampler with $\nSteps=24$ sampling steps and a cfg scale of 6.
All prompts are either truncated or padded to 77 tokens.
The base negative prompt is set to "low quality, worst quality".

Regarding \cref{fig:GBC-diff-graph},
we observe that encoding prompts independently performs better for the first example, while encoding them with contextual information from parent prompts works better for the second example.
Accordingly, we adopt the respective strategy for these two examples when only prompt and graph information is provided.

\pagebreak
\section{Additional results for text-to-image generation}
\label{apx:exp-add-t2i}
In this appendix, we present additional results for text-to-image generation experiments.

\subsection{GBC-to-image}

\begin{figure}
    \centering
    \begin{subfigure}{\textwidth}
    \centering
        \includegraphics[width=0.9\textwidth]{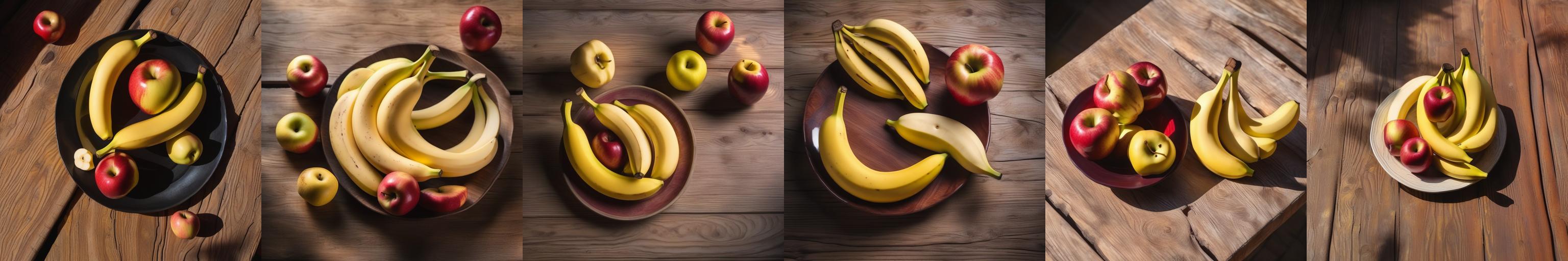}
        \vspace{.2em}
        \caption{Generated with concatenation of text prompts.}
    \end{subfigure}
    \\[.6em]
    \begin{subfigure}{\textwidth}
    \centering
        \includegraphics[width=0.9\textwidth]{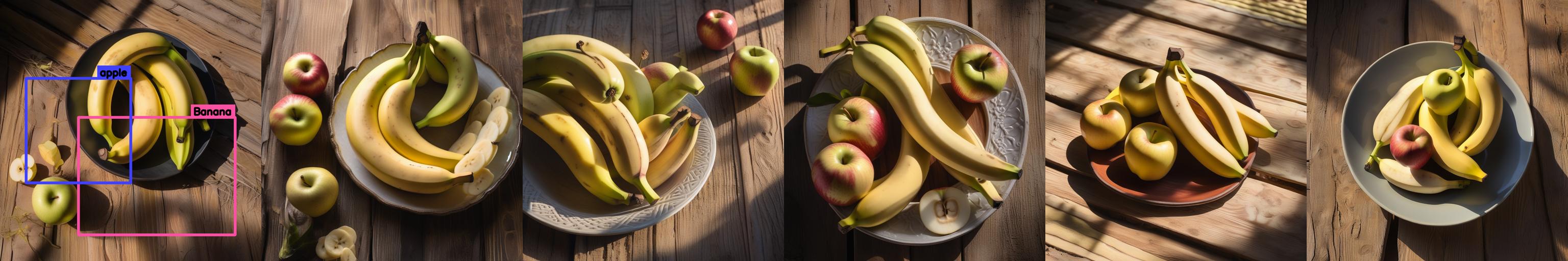}
        \vspace{.2em}
        \caption{Generated with text prompts and bounding box information.}
    \end{subfigure}
    \\[.6em]
    \begin{subfigure}{\textwidth}
    \centering
        \includegraphics[width=0.9\textwidth]{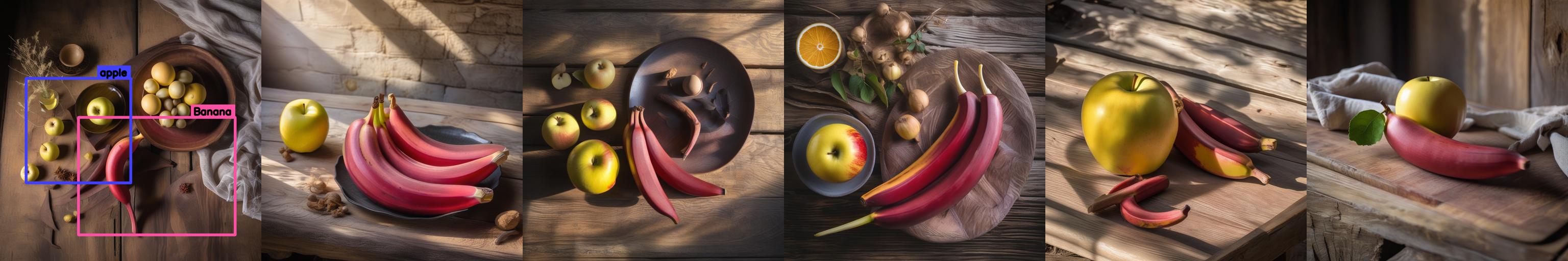}
        \vspace{.2em}
        \caption{Generated with text prompts, graph, and bounding box information. Prompts are encoded independently.}
    \end{subfigure}
    \\[.6em]
    \begin{subfigure}{\textwidth}
    \centering
        \includegraphics[width=0.9\textwidth]{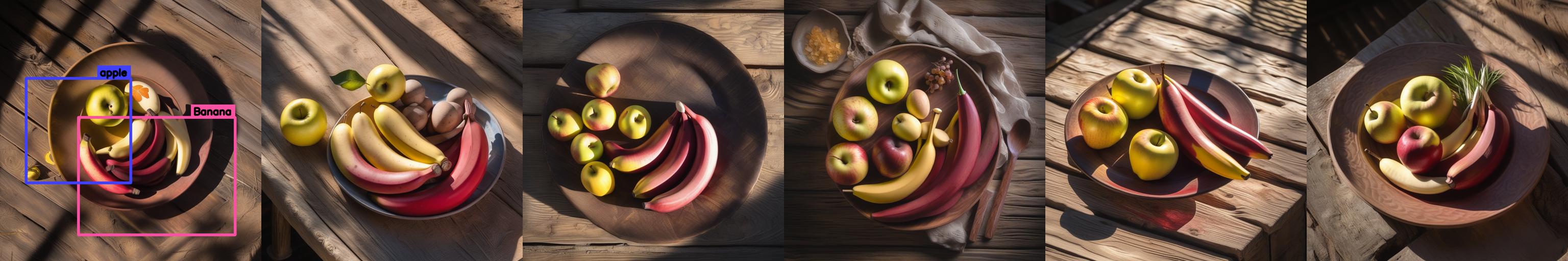}
        \vspace{.2em}
        \caption{Generated with text prompts, graph, and bounding box information. Prompts are encoded with parent prompts as context information.}
    \end{subfigure}
    \\[.6em]
    \begin{subfigure}{\textwidth}
    \centering
        \includegraphics[width=0.9\textwidth]{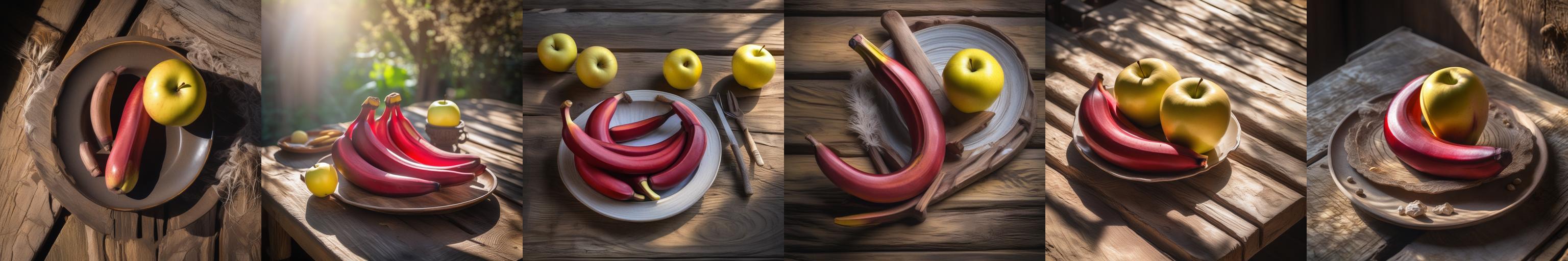}
        \vspace{.2em}
        \caption{Generated with text prompts and  graph information. Prompts are encoded independently.}
    \end{subfigure}
    \caption{Non cherry-picked generations for the first example presented in \cref{fig:GBC-diff-main}.}
    \label{fig:banana-apple-gen}
\end{figure}

\begin{figure}
    \centering
    \begin{subfigure}{0.9\textwidth}
    \centering
        \includegraphics[width=\textwidth]{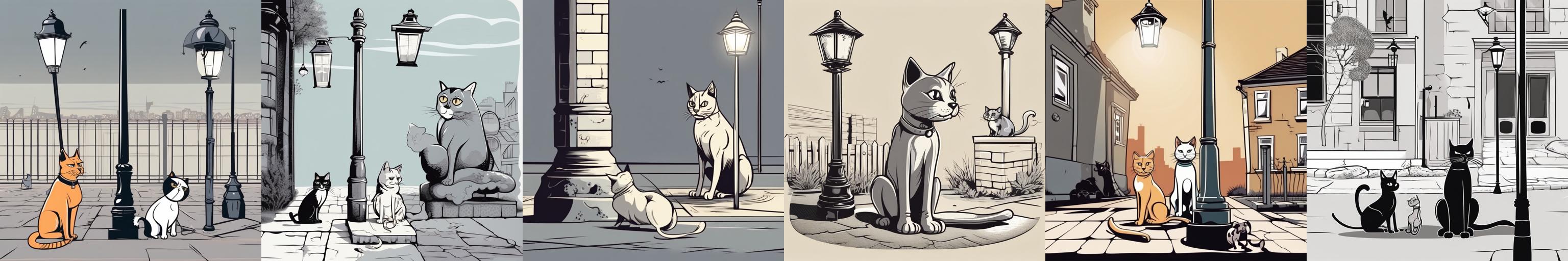}
        \\[.2em]
        \caption{Generated with concatenation of text prompts.}
    \end{subfigure}
    \\[.6em]
    \begin{subfigure}{0.9\textwidth}
    \centering
        \includegraphics[width=\textwidth]{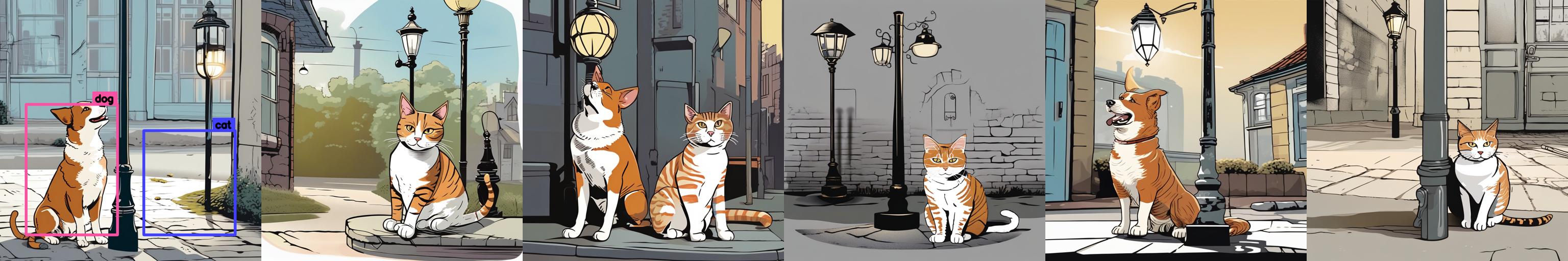}
        \\[.2em]
        \caption{Generated with text and image prompts and bounding box information.}
    \end{subfigure}
    \\[.6em]
    \begin{subfigure}{0.9\textwidth}
    \centering
        \includegraphics[width=\textwidth]{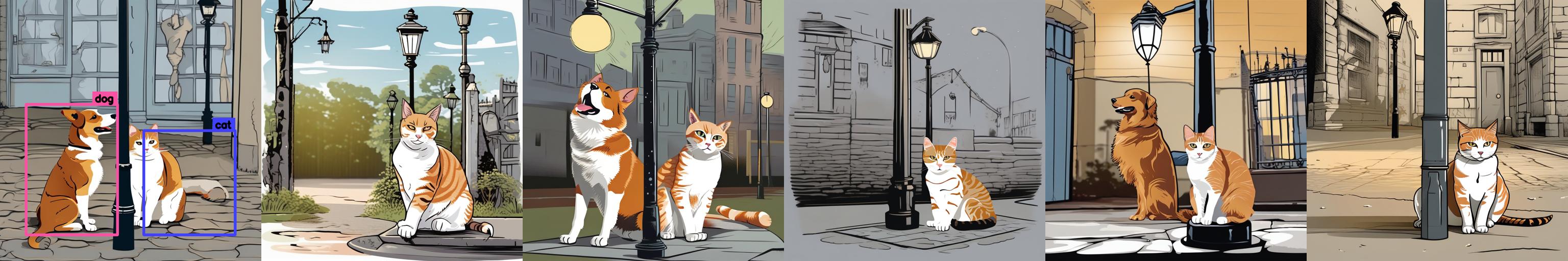}
        \\[.2em]
        \caption{Generated with text and image prompts, graph, and bounding box information. Prompts are encoded independently.}
    \end{subfigure}
    \\[.6em]
    \begin{subfigure}{0.9\textwidth}
    \centering
        \includegraphics[width=\textwidth]{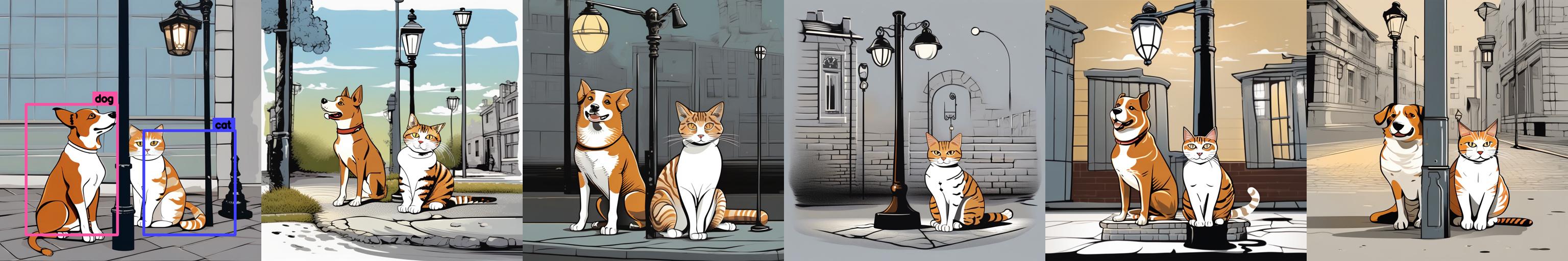}
        \\[.2em]
        \caption{Generated with text and image prompts, graph, and bounding box information. Prompts are encoded with parent prompts as context information.}
    \end{subfigure}
    \\[.6em]
    \begin{subfigure}{0.9\textwidth}
    \centering      \includegraphics[width=\textwidth]{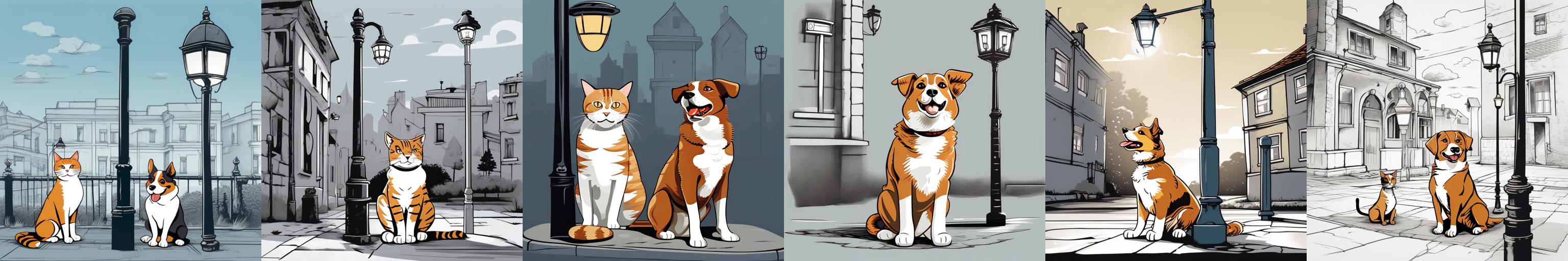}
        \\[.2em]
        \caption{Generated with text and image prompts and  graph information. Prompts are encoded with parent prompts as context information.}
    \end{subfigure}
    \caption{Non cherry-picked generations for the second example presented in \cref{fig:GBC-diff-main}.}
    \label{fig:corgi-cat-gen}
\end{figure}

\begin{figure}
    \centering
    \begin{subfigure}{\textwidth}
    \centering
        \includegraphics[width=0.9\textwidth]{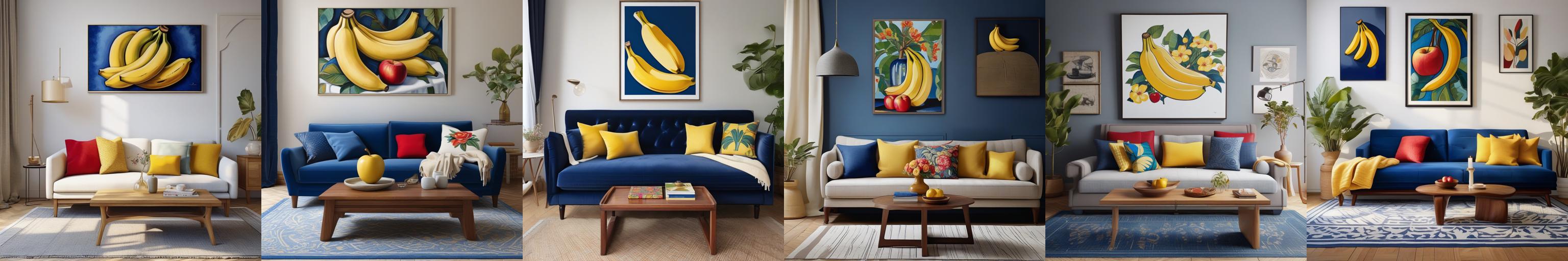}
        \vspace{.2em}
        \caption{Generated with concatenation of text prompts.}
    \end{subfigure}
    \\[.6em]
    \begin{subfigure}{\textwidth}
    \centering
        \includegraphics[width=0.9\textwidth]{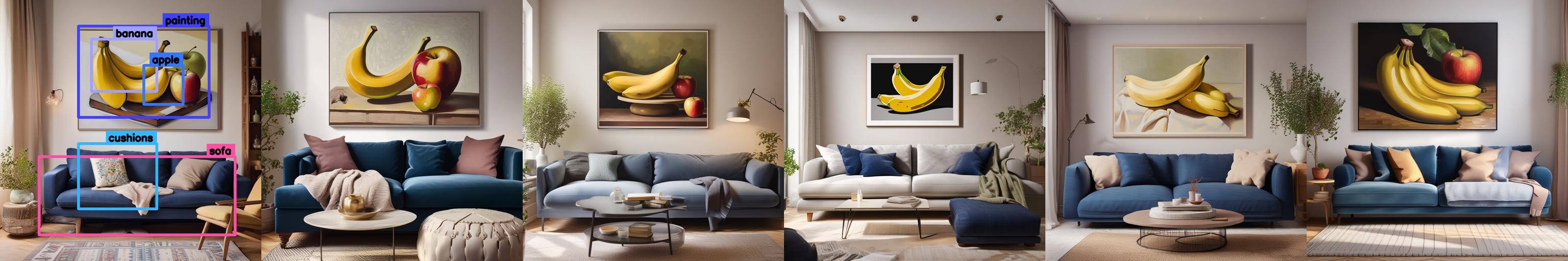}
        \vspace{.2em}
        \caption{Generated with text prompts and bounding box information.}
    \end{subfigure}
    \\[.6em]
    \begin{subfigure}{\textwidth}
    \centering
        \includegraphics[width=0.9\textwidth]{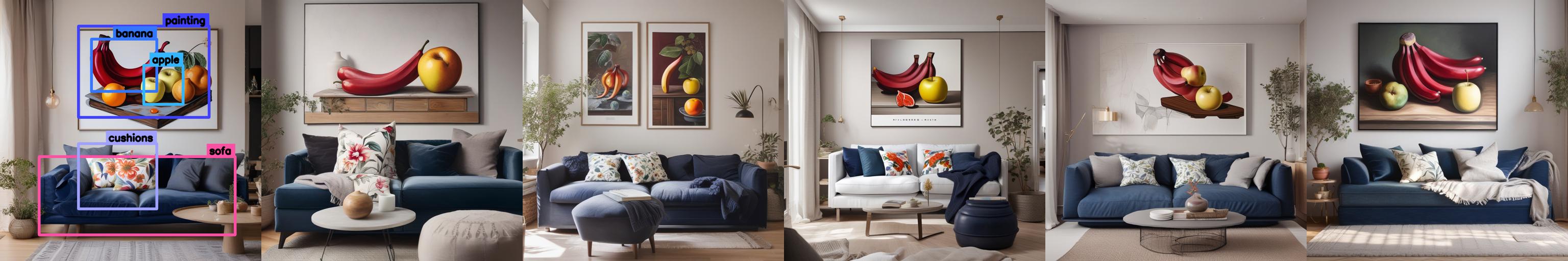}
        \vspace{.2em}
        \caption{Generated with text prompts, graph, and bounding box information. Prompts are encoded independently.}
    \end{subfigure}
    \\[.6em]
    \begin{subfigure}{\textwidth}
    \centering
        \includegraphics[width=0.9\textwidth]{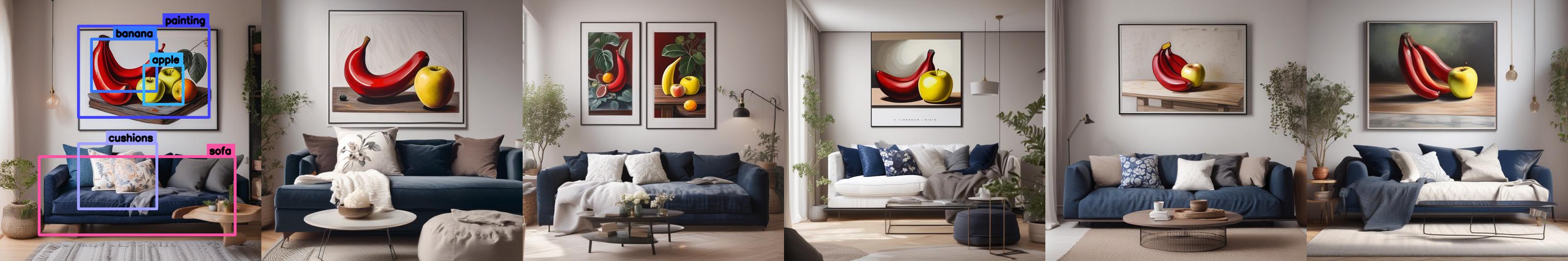}
        \vspace{.2em}
        \caption{Generated with text prompts, graph, and bounding box information. Prompts are encoded with parent prompts as context information.}
    \end{subfigure}
    \caption{Non cherry-picked generations for the third example presented in \cref{fig:GBC-diff-main}.}
    \label{fig:living-room-gen}
\end{figure}

Unbiased generations from different methods for the prompts considered in \cref{fig:GBC-diff-main} are shown in \cref{fig:banana-apple-gen,fig:corgi-cat-gen,fig:living-room-gen}.
These results confirm our observations in \cref{sec:exp-t2i}.

In \cref{fig:banana-apple-gen}, we additionally observe that restricting banana and apple from appearing outside the bounding boxes could lead to the appearance of ambiguous or incomplete objects.
These artifacts seem to be the result of the model attempting to generate apples or bananas but being constrained by the bounding box, causing the object to morph into unnatural or distorted forms.
Given how different text tokens and image patches exchange information in text encoder and in the self-attention and convolutional layers of UNet, this leakage of information seems unavoidable with our current approach.

In \cref{fig:corgi-cat-gen}, we see that vanilla SDXL struggles to generate a cat and a dog with prompt concatenation.
Similarly, with the naive approach that only leverages bounding box information with forward cross-attention control, the model still fails to generate one of the two animals with high probability.

\subsection{Text-to-image with GBC as middleware}

\begin{figure}[p]
    \centering
    \includegraphics[width=\linewidth]{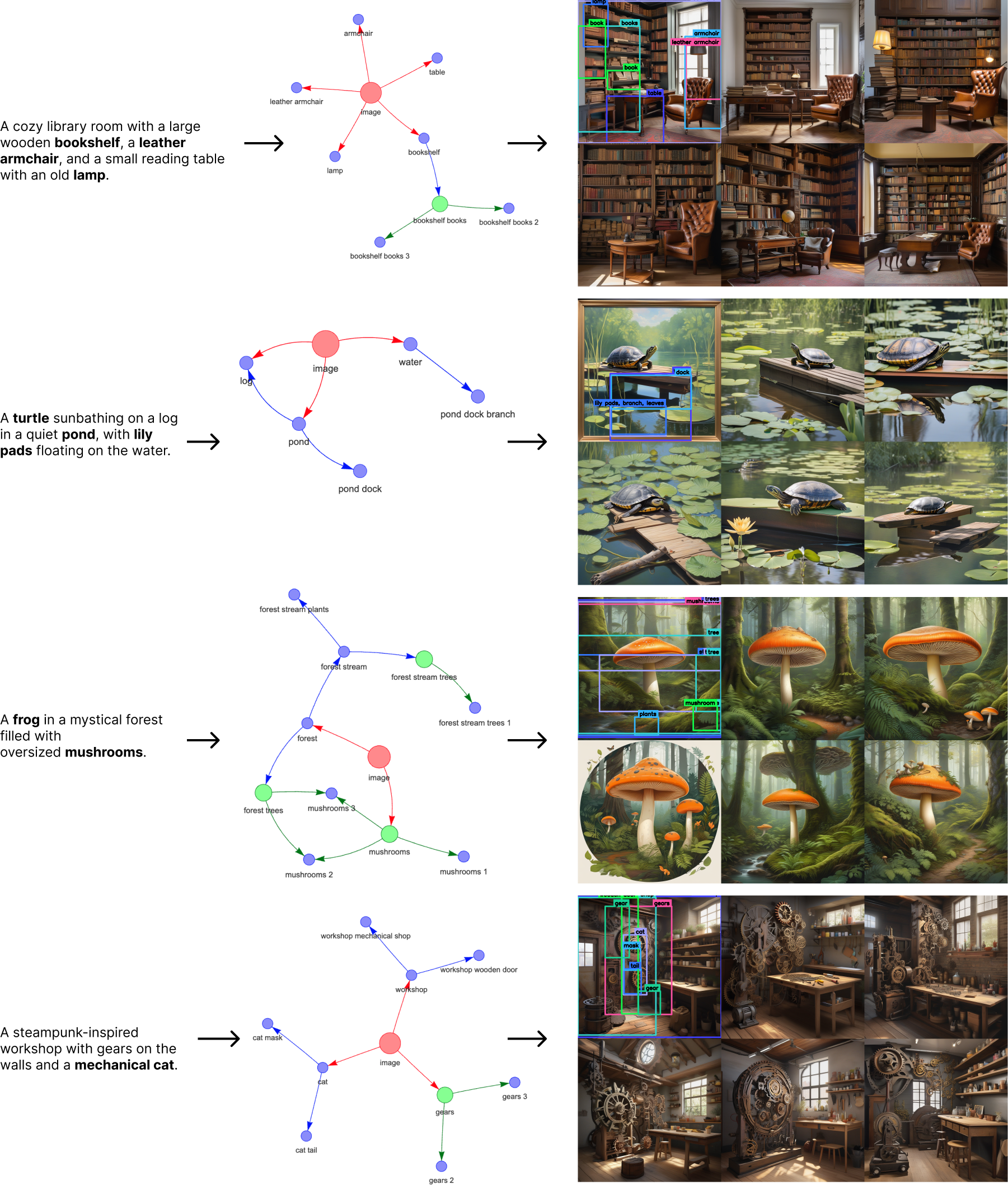}
    \vspace{-0.4em}
    \caption{Example generations with GBC as middleware, presented without cherry picking.
    Our prompt generation model can generate complex graphs when used naively.
    This, however, cannot be handled by our image generation algorithm, leading to unsatisfying results.
    For better visualization, bounding boxes are shown for only one of the $6$ generated images.
    Some generated prompts are provided in \cref{tab:gbcpromptgen}.}
    \label{fig:t2gbc2i-complex}
\end{figure}

We next study the possibility of combining our text-to-GBC and GBC-to-image pipeline.
For GBC-to-image, we use the algorithm that exploits all information from GBC and encode each prompt with parent prompts as contextual information.
In \cref{fig:t2gbc2i-complex}, we first show that our prompt generation model is able to generate complex graphs from a simple prompt.
However, we find the our image generation algorithm fail to generate images that adhere to such complex GBC prompt for the following reasons.
\begin{itemize}
    \item The generated GBC is not perfect, potentially due the small size of our prompt generation model.
    While each node's description is correct for their corresponding object, inconsistencies can arise between the descriptions of different nodes.
    For instance, in the last example of \cref{fig:t2gbc2i-complex}, a cat is described simply as an animal rather than as a mechanical cat.
    Moreover, the generated bounding boxes can also contradict the descriptions.
    We believe the latter could be improved by encoding bounding boxes differently, as for example done in \cite{omost}.
    \item
    The image generation algorithm struggles with bounding boxes that have large overlapping areas when they are on disjoint paths of the graph.
    This limitation arises because the algorithm relies on the graph hierarchy to manage overlapping bounding boxes and determine their priority.
    As far as we are aware, no existing training-free approach can effectively handle highly complex overlapping bounding boxes of this kind.
    \item 
    Not all important objects from the seed prompt are guaranteed to receive additional descriptions from our prompt generation model.
    Combined with the previously mentioned limitation, this can result in certain objects failing to appear in the generated image.
    For example, in the third example of \cref{fig:t2gbc2i-complex}, the frog is absent from the final images.
\end{itemize}

\vspace{0.5em}

\begin{figure}[p]
    \centering
    \includegraphics[width=\linewidth]{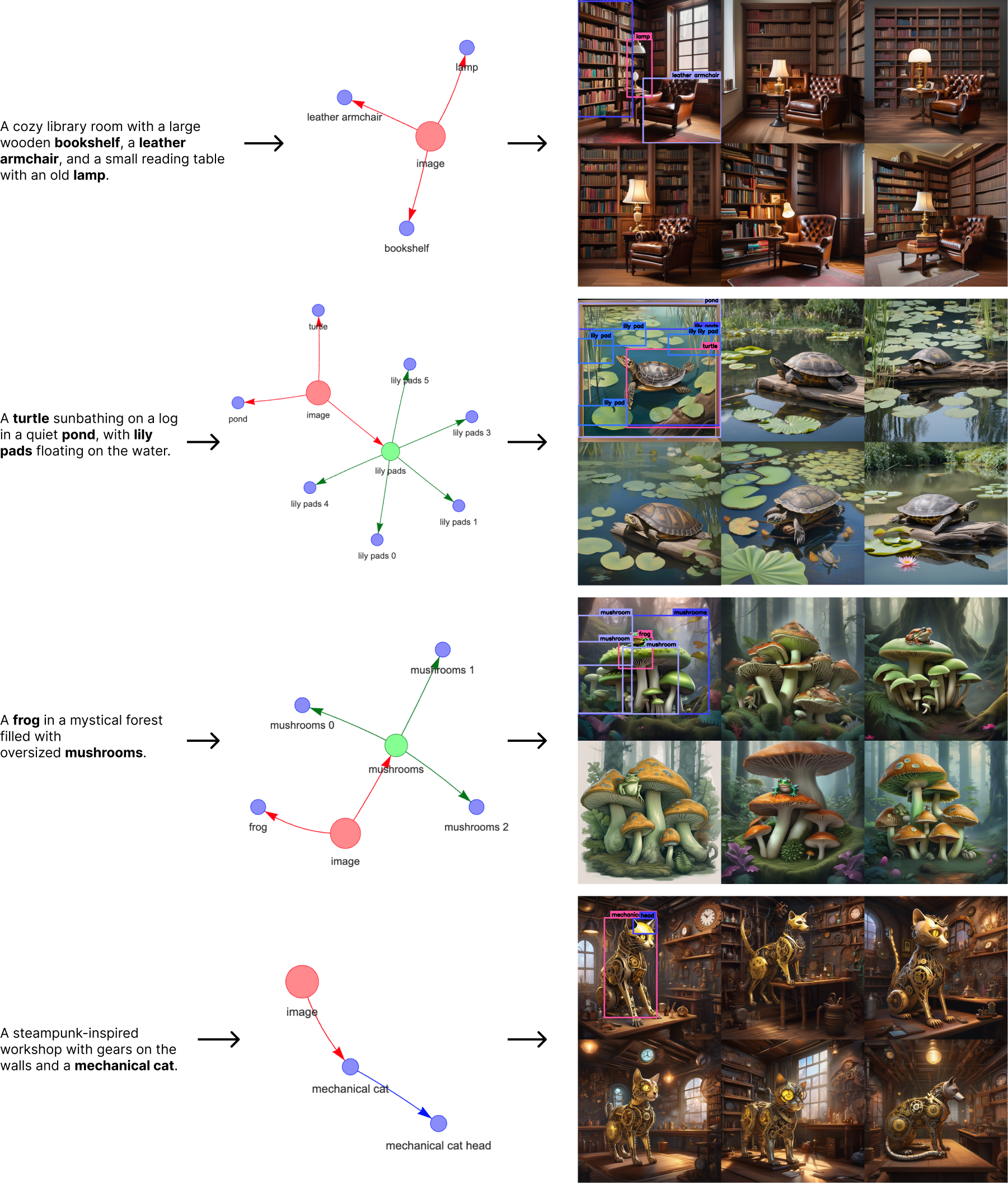}
    \vspace{-0.4em}
    \caption{Example generations with GBC as middleware, presented without cherry picking.
    Here, we manipulate the generation process to start from a certain object (lamp, turtle, frog, or mechanical cat).
    For better visualization, bounding boxes are shown for only one of the $6$ generated images.
    Some generated prompts are provided in \cref{tab:gbcpromptgen}.}
    \label{fig:t2gbc2i-simple}
\end{figure}

To circumvent the above limitations while still benefiting from the advantages of using GBC as middleware, we restrict the process to simpler graphs and adjust the prompt generation to ensure the model explicitly produces detailed descriptions for a select number of key objects.
The results, shown in \cref{fig:t2gbc2i-simple}, demonstrate that the generated images align more closely with both the seed prompt and the intermediate GBC.
However, some failure cases remain, such as the absence of a frog in two images for the third example or the generation of a mechanical dog instead of a mechanical cat in two images for the forth example. 

\begin{table}[p]
    \renewcommand{\arraystretch}{1.35}
    \centering
    \begin{NiceTabular}{c|p{0.8\textwidth}}
    \toprule
    \multirow{18}{*}{\makecell{Example\\ Generated\\ Prompts for\\ \cref{fig:t2gbc2i-complex}}}
    &
    \textbf{Lamp.} A vintage table lamp with a yellow shade. The lamp has a sturdy base with intricate detailing. It features a classic design with a metal frame and a ribbed texture.
    \\
    &
    \textbf{Bookshelf.} A large wooden bookshelf filled with various \empha{books}. The bookshelf has multiple shelves, each holding an assortment of \empha{books} with different sizes and colors. Some \empha{books} are stacked horizontally while others are arranged vertically.
    \\
    &
    \textbf{Pond.} A serene pond with a wooden \empha{dock} extending into it. The water is calm, reflecting the surrounding greenery. A few lily pads are scattered across the surface, adding to the tranquil atmosphere.
    \\
    &
    \textbf{Dock.}
    A wooden dock extends into a body of water. The dock has a sturdy structure with visible planks and supports. A small boat is docked at one end, with a single oar resting against it.
    \\
    &
    \textbf{Mushrooms.}
    The mushrooms are likely part of a larger forest ecosystem, contributing to the overall environment.
    \\
    &
    \textbf{Steam.}
    The image showcases a serene stream meandering through a lush forest. The stream is flanked by verdant greenery, with various \empha{plants} and \empha{trees} lining its path. The water appears calm and clear, reflecting the surrounding foliage. The stream's banks are adorned with fallen leaves, hinting at the season being autumn.
    \\
    &
    \textbf{Cat.}
    The image showcases a stylized depiction of a cat. It is characterized by a sleek body with smooth muscles and a distinctive \empha{mask}-like pattern on its head. The cat's ears are pointed upwards, and it has a small, round \empha{tail}. 
    Its eyes are closed, giving it a serene appearance.
    \\
    &
    \textbf{Wooden door.}
    A wooden door with a visible grain pattern and a slightly worn appearance. It has a rectangular shape with a central panel flanked by two smaller panels. A small window or peephole is located near the top center.
    \\
    \midrule
    \multirow{11}{*}{\makecell{Example\\ Generated\\ Prompts for\\ \cref{fig:t2gbc2i-simple}}}
    &
    \textbf{Lamp.} A vintage table lamp with a gold base and a white lampshade. The lamp has a classic design with a curved neck and a floral pattern at the base.
    \\
    &
    \textbf{Leather armchair.}
    A vintage leather armchair with a rich brown color. The chair has a plush seat cushion and armrests. It features a tufted backrest with decorative buttons, adding to its classic design.
    \\
    &
    \textbf{Turtle.}
    A large turtle with a rough, textured shell is seen resting on a log. Its head is slightly raised, and it appears calm and relaxed.
    \\
    &
    \textbf{Frog.}
    A frog with vibrant green skin sits perched on a mushroom. Its large eyes are wide open, and it appears to be looking directly at the camera.
    \\
    &
    \textbf{Mechanical cat.}
    A mechanical cat with intricate gears and cogs visible on its body. The cat has a sleek design with a streamlined \empha{head} and tail. Its eyes are glowing with a yellowish hue.
    \\
    &
    \textbf{Head.}
    The image showcases a close-up view of a head, which appears to be that of a mechanical or robotic robot. It has a sleek, metallic finish with a shiny surface that reflects light. The head features two eyes that are glowing with a yellowish hue, giving it a somewhat eerie or futuristic appearance. There are no visible mouth or nose, and no other distinguishable facial features are apparent.
    \\
    \bottomrule
    \end{NiceTabular}
    \vspace{0.75em}
    \caption{Examples of generated prompts that we obtain in \cref{fig:t2gbc2i-complex,fig:t2gbc2i-simple} with our prompt generation model.
    We highlight the objects described in the children nodes in dark blue.
    }
    \label{tab:gbcpromptgen}
\end{table}

\paragraph{GBC as middleware enhances image diversity.}

\begin{figure}
    \centering
    \includegraphics[width=0.8\textwidth]{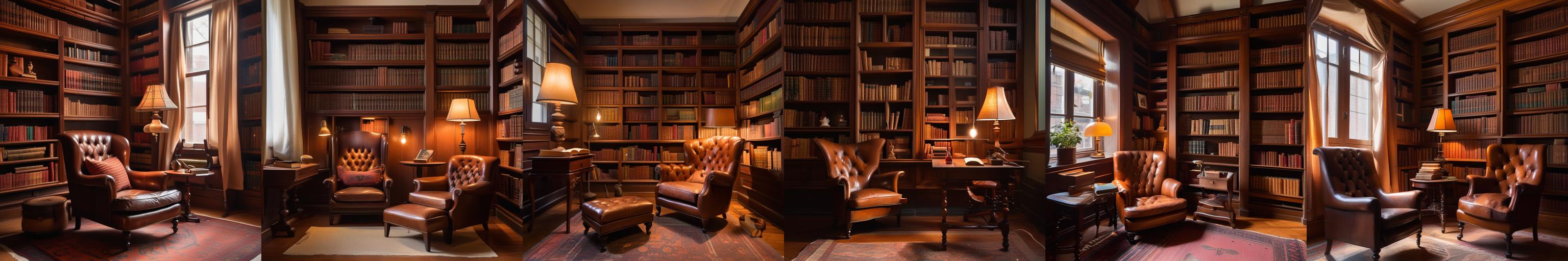}
    \vspace{-0.05em}
    \includegraphics[width=0.8\textwidth]{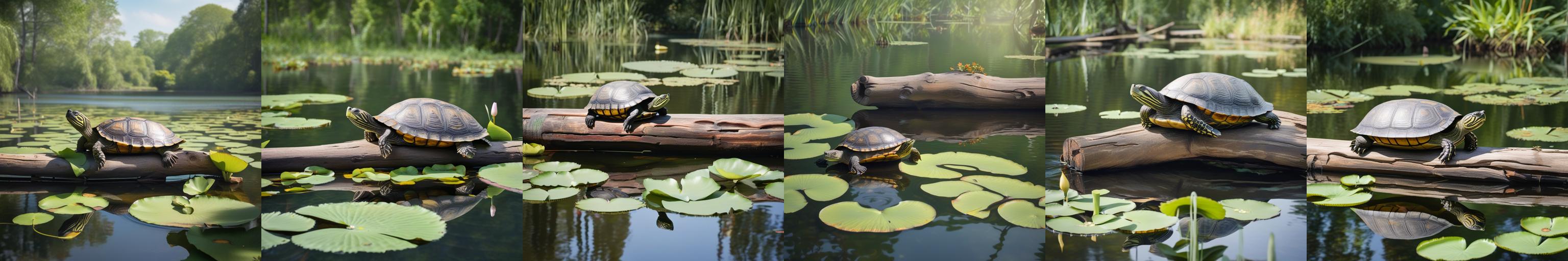}
    \vspace{-0.05em}
    \includegraphics[width=0.8\textwidth]{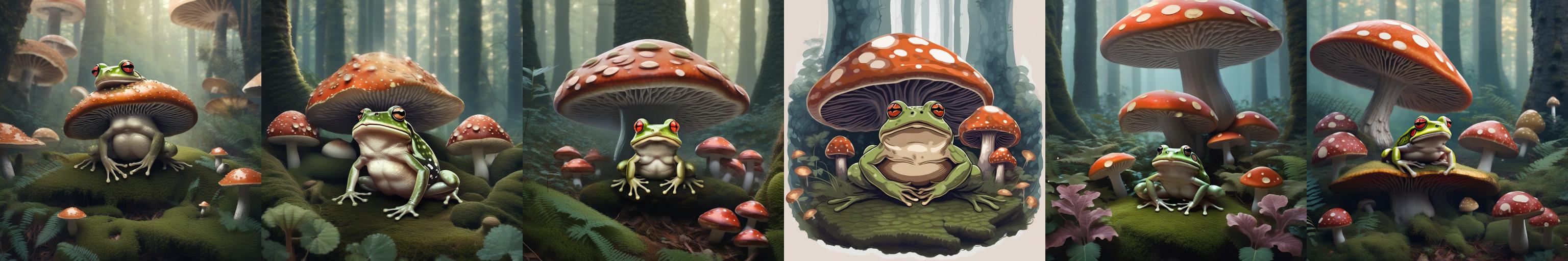}
    \vspace{-0.05em}
    \includegraphics[width=0.8\textwidth]{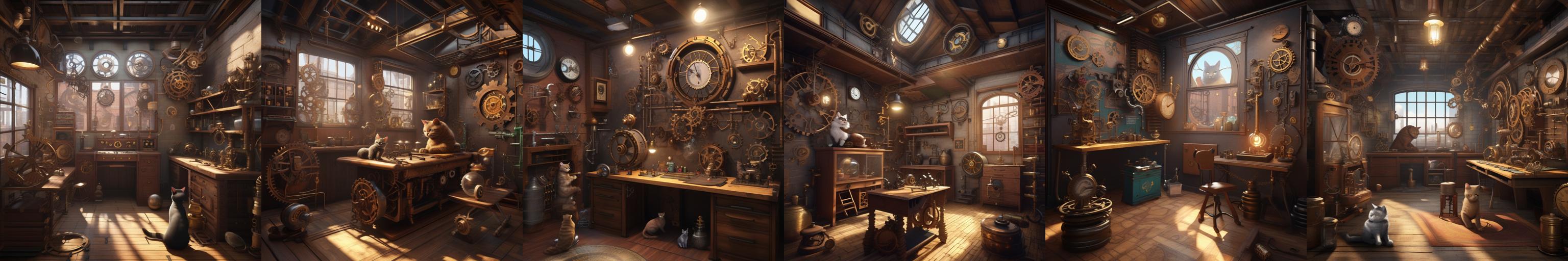}
    \caption{Example generations from the seed prompts in \cref{fig:t2gbc2i-complex,fig:t2gbc2i-simple} when SDXL is used naively with Euler sampling.}
    \label{fig:nogbc-gen}
\end{figure}

\begin{figure}
    \centering
    \includegraphics[width=0.8\textwidth]{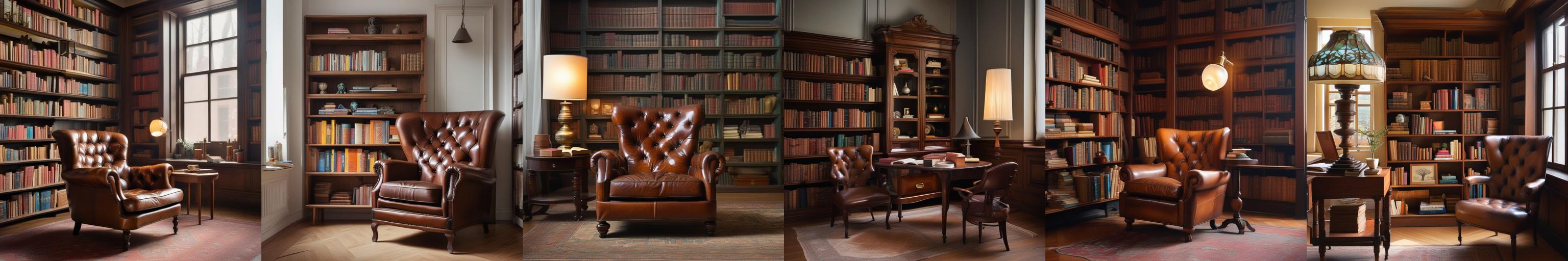}
    \vspace{-0.05em}
    \includegraphics[width=0.8\textwidth]{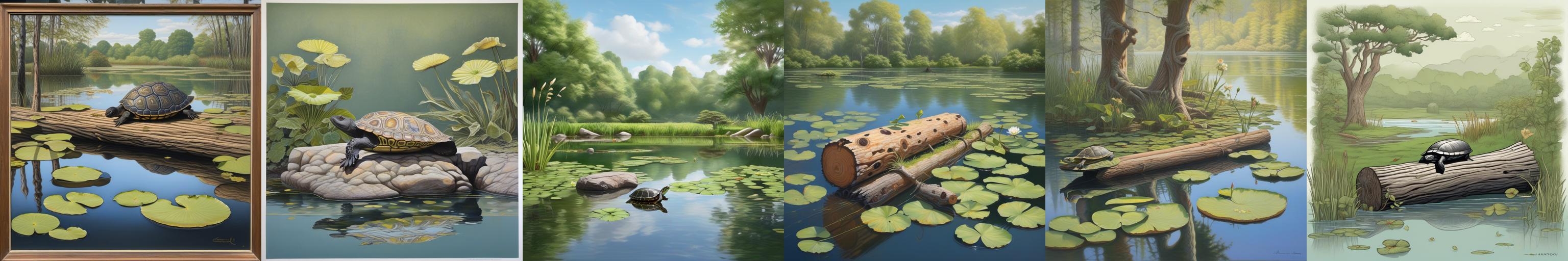}
    \vspace{-0.05em}
    \includegraphics[width=0.8\textwidth]{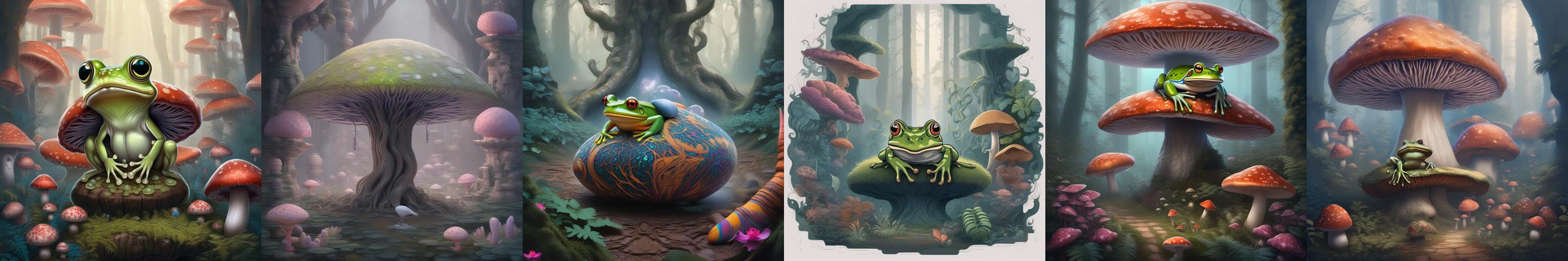}
    \vspace{-0.05em}
    \includegraphics[width=0.8\textwidth]{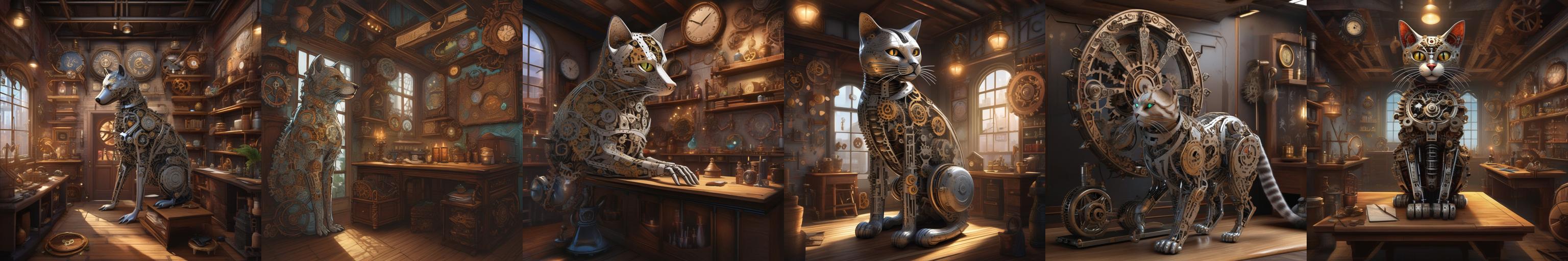}
    \caption{Example generations from the seed prompts in \cref{fig:t2gbc2i-complex,fig:t2gbc2i-simple} when GBC is used as middleware with the manipulation introduced in \cref{fig:t2gbc2i-simple}.}
    \label{fig:t2gbc2i-gen}
\end{figure}

It has been reported that naively prompting the text-to-image models could often result in images with low diversity from a single prompt
\cite{kirchhof2024sparse,kynkaanniemi2024applying}.
Using additional middleware is an effective way to address this issue as it enables the model to generate a diverse set of intermediate representations from the same seed prompt \cite{gu2024kaleido,tipo2024yeh}.
We verify that GBC as middleware can indeed also improve the diversity of generated images in 
\cref{fig:nogbc-gen,fig:t2gbc2i-gen}.
On the downside, as noted previously, this approach may slightly compromise prompt adherence.
Interestingly, for the fourth example, SDXL fails to generate a mechanical cat using only the plain text prompt.
In contrast, employing GBC that provides additional descriptions of the mechanical cat as middleware successfully ensures its inclusion in the generated images. 

\FloatBarrier
\pagebreak
\section{Image attributions}

All the images that we show in this paper come from Wikimedia Commons.
We provide in \cref{tab:images-urls} the exact source urls and license for each of the images.
The urls to the CC BY-SA 2.0, CC BY-SA 3.0, and GFDL 1.2 licenses are respectively \url{https://creativecommons.org/licenses/by-sa/2.0}, \url{https://creativecommons.org/licenses/by-sa/3.0/}, and \url{https://www.gnu.org/licenses/old-licenses/fdl-1.2.txt}.

\vspace{2.5em}
\begin{table}[ht]
    \centering
    \renewcommand{\arraystretch}{1.45}
    {
    \begin{tabular}{lp{8cm}l}
        \toprule
        \textbf{Image} & \textbf{Source URL} & \textbf{License} \\
        \midrule
        \cref{fig:GBC-ex-main} & \url{https://commons.wikimedia.org/wiki/File:Tartu_raudteejaama_veetorn,_2010.JPG} & 
        CC BY-SA 3.0
        \\
        \hline
        \cref{fig:dataset-query} & \url{https://commons.wikimedia.org/wiki/File:Eiffel_Tower_from_north_Avenue_de_New_York,_Aug_2010.jpg} & CC BY-SA 3.0 \\
        \hline
        \cref{fig:GBC-diff-main}
        \\[0.2em]
        Corgi
        &  \url{https://commons.wikimedia.org/wiki/File:Fawn_and_white_Welsh_Corgi_puppy_standing_on_rear_legs_and_sticking_out_the_tongue_(cropped).jpg}
        &
        Public domain
        \\[-0.1em]
        Cat
        &
        \url{https://fr.m.wikipedia.org/wiki/Fichier:Orange_tabby_cat_sitting_on_fallen_leaves-Hisashi-01A.jpg}
        &
        CC BY-SA 2.0
        \\
        \hline
        \cref{fig:gbc10m-ex-flame-missa-scepter} & &
        \\[0.2em]
        Flame
        & \url{https://commons.wikimedia.org/wiki/File:Flametest--Na.swn.jpg} & CC BY-SA 3.0 \\[-0.1em]
        Messe
        & \url{https://commons.wikimedia.org/wiki/File:Messe_mit_Wandlungskerze_Beuron.jpg} & Public domain
        \\[-0.1em]
        Regalia
        & \url{https://commons.wikimedia.org/wiki/File:Crown,_sceptre,_orb_\%26_key_of_the_King_of_Sweden_2014.jpg}
        & Public domain 
        \\
        \hline
        \cref{fig:gbc10m-ex-elephant} & \url{https://commons.wikimedia.org/wiki/File:Indian-Elephant-444.jpg} & GFDL 1.2 
        \\
        \bottomrule
    \end{tabular}
    }
    \vspace{0.75em}
    \caption{Source URLs and licenses of the images shown in this paper.}
    \label{tab:images-urls}
\end{table}

\end{document}